%% file: main.tex
\documentclass[11pt]{article}
\pdfoutput=1

\input{math_commands.tex}

\input{style.tex}

\title{Pointwise Generalization in Deep Neural Networks}

\author{%
  Shaojie Li \hspace{3em}%
  Yunbei Xu\thanks{In keeping with standard practice in mathematics and theory, authors are listed alphabetically. Yunbei Xu (\texttt{yunbei@nus.edu.sg}) is the corresponding author.}%
  \\
  National University of Singapore
  \\
  \texttt{\{li\_sj,yunbei\}@nus.edu.sg}
}

\date{}

\begin{document}

\maketitle

\begin{abstract}
We address the fundamental question of why deep neural networks generalize by establishing a pointwise generalization theory for fully connected networks. This framework resolves long-standing barriers to characterizing the rich nonlinear feature-learning regime and builds a new statistical foundation for representation learning. For each trained model, we characterize the hypothesis via a pointwise Riemannian Dimension, derived from the eigenvalues of the learned feature representations across layers. This establishes a principled framework for deriving hypothesis-dependent, representation-aware generalization bounds. These  bounds offer a systematic upgrade over approaches based on model size, products of norms, and infinite-width linearizations, yielding guarantees that are orders of magnitude tighter in both theory and experiment. Analytically, we identify the structural properties and mathematical principles that explain the tractability of deep networks. Empirically, the pointwise Riemannian Dimension exhibits substantial feature compression, decreases with increased over-parameterization, and captures the implicit bias of optimizers. Taken together, our results indicate that deep networks are mathematically tractable in practical regimes and that their generalization is sharply explained by pointwise, feature-spectrum-aware complexity.
\end{abstract}

\tableofcontents

\section{Introduction}
Deep learning has ushered in a new era of AI, delivering striking generalization across scientific tasks \citep{lecun2015deep,bengio2013representation}. Yet, a fundamental paradox remains: while classical theory predicts severe overfitting for massive models, practice exhibits strong generalization. This gap has fueled a prevailing view that neural networks are opaque “black boxes” resistant to principled explanation \citep{goodfellow2016deep}. We narrow this gap by addressing the generalization problem for the canonical fully connected Deep Neural Networks (DNN). We demonstrate that, under verifiable spectral conditions on the {\it learned feature representations}, deep neural networks fall into a tractable regime with tight generalization guarantees. Crucially, these conditions impose no constraints on parameter count or weight sparsity, standing in sharp contrast to prior statistical conventions based on pure weight-space compression. Methodologically, our characterization leverages a pointwise generalization paradigm that fundamentally transcends classical uniform-convergence and covering-number approaches. We believe this is an especially notable conclusion: deep neural networks can admit sharp, hypothesis-dependent finite-sample guarantees with a degree of tractability often thought possible only for linearized models, {\it throughout the rich nonlinear feature-learning regime}, without relying on infinite-width limits, linearized approximations around initialization, or exponential dependence on key problem parameters. We hope that this framework helps demystify generalization and provides a systematic methodology for analyzing nonlinear, overparameterized models in representation learning.

We study standard fully connected (feed-forward) networks on a dataset
\(X=[x_1,\dots,x_n]\in\mathbb{R}^{d_{0}\times n}\), where each column is one input example.
The network has widths $ d_1,\dots, d_L$, and weight matrices
\(W_l\in\mathbb{R}^{d_l\times d_{l-1}}\) for \(l=1,\dots,L\).
We define the \emph{feature matrix} at layer \(l\) by the recursion 
\begin{align}\label{eq: intro DNN}
F_l(W,X) := \sigma_l\!\bigl(W_l\,F_{l-1}(W,X)\bigr)\in\mathbb{R}^{d_l\times n},
\qquad l=1,\dots,L,
\end{align}
where $F_0 := X$ and the nonlinear activation \(\sigma_l\) acts columnwise. Each \emph{column} of \(F_l\) is the feature vector of
one data point at layer \(l\); each \emph{row} of $F_l$ is the activation of one neuron across the dataset.

Our focus is the \emph{generalization gap}—the difference between test and training loss
at the learned weights \(W\).
Informally—up to universal constants, optimistic logarithmic factors\footnote{Throughout the paper, the phrase “optimistic logarithmic factors” is used in the following  precise sense illustrated by Example~\ref{example optimistic logarithmic}: under common rapid spectral-decay or explicit low-rank assumptions on learned-feature compression, the relevant scaling is logarithmic. Whenever absolute logarithmic terms appear, they are stated explicitly.}, and reasonable simplification (made precise in Theorem \ref{thm dnn} and Theorem~\ref{thm empirical dnn feature iso} with discussion on  the feasibility of these
simplifications)—we prove that this gap is
controlled by the \emph{effective dimension} of the learned features: uniformly over every $W\in \bR^{\sum_{l} d_l\cdot d_{l-1}}$,
\begin{align}\label{eq: intro bound}
\mathcal{L}_{\mathrm{test}}(W)-\mathcal{L}_{\mathrm{train}}(W)
\;\lsim\;
\sqrt{\frac{1}{n}\,\sum_{l=1}^{L}
\bigl(d_l+d_{l-1}\bigr)\;
d_{\mathrm{eff}}\!\Bigl(F_{l-1}(W,X)\,F_{l-1}(W,X)^{\!\top}\Bigr)}.
\end{align}
Here \(d_{\mathrm{eff}}(\cdot)\) denotes the (layerwise) \emph{effective dimension}—a smoothed, spectrum-aware notion of rank—of the feature Gram matrix \(F_{l-1}(W,X)F_{l-1}(W,X)^{\!\top}\), i.e., the number of meaningful directions the feature data actually occupies at that layer. Intuitively, each layer contributes a term proportional to its size \((d_l+d_{l-1})\) multiplied by how many directions its features \(F_{l-1}(W,X)\) truly use, \(d_{\mathrm{eff}}\). 
When features are correlated, low rank, or exhibit a rapidly decaying spectrum (a few large eigenvalues dominating many small ones), \(d_{\mathrm{eff}}\) is small, so the bound remains tight even for very wide/deep networks. Such “feature compression” are widely observed in modern deep learning \citep{huh2021low,wang2025understanding,parker2023neural}. Strikingly, in our experiments, increasing overparameterization often induces pronounced \emph{feature-rank compression}: the bound \eqref{eq: intro bound} decreases as model size grows (Section~\ref{section: experiments}); for example, in ResNet trained on CIFAR-10, a majority of layers compress to (near-)zero effective rank.

Inequality \eqref{eq: intro bound} yields a strong \emph{uniform, hypothesis- and data-dependent} guarantee, which we term \emph{pointwise generalization}.  It tracks how features evolve across layers of the \emph{trained} model and explains overparameterization in practice. Moreover, the right-hand side of \eqref{eq: intro bound} can be used directly as a \emph{regularizer}, leading to algorithms that adapt to the effective ranks around a benchmark $W^\star$ (Section~\ref{subsec implicit bias}).
The spectrum-aware effective-dimension notion  we adopt is standard and minimax-sharp in linear and kernel settings \citep{even2021concentration}. In contrast, existing bounds either (i) rely on infinite–width linearizations (the NTK line of work,
e.g., \cite{jacot2018neural}), (ii) blow up exponentially with products of norms
(e.g., \cite{neyshabur2017pac,bartlett2017spectrally}), or (iii) scale with model size (e.g., VC dimension
\citep{bartlett2019nearly}). Our bounds avoid these pathologies, providing a pointwise, spectrum–aware account grounded in unifying structural principles, clear links to prior frameworks, and qualified tightness discussions. By confronting the long-standing challenge of obtaining sharp, \emph{nonlinear} guarantees for representation learning beyond uniform convergence—emphasized in \cite{bartlett2021deep,  zhang2021understanding, neyshabur2017exploring, nagarajan2019uniform, wilson2025deep}—we aim to show that generalization in deep neural networks can be mathematically tractable in the above sense. 

\paragraph{Fully observable bounds and an open problem.}
On the technical issue of full observability from the empirical sample and learned features, it is important to clarify the role of the ghost sample and the unknown population distribution in our analysis. We first prove an unconditional compressed pointwise theorem under a mixed empirical-population feature metric. We then provide a fully observed-sample theorem that isolates the additional finite-resolution subspace observability needed to compute the same compressed spectrum from the training sample alone. Verifying this condition from intrinsic feature regularity is a substantive open problem, rather than a routine technical step. Algorithm design offers another potential route to resolving the observability problem, for example through sample splitting and cross-validation.

\subsection{Organization and Contributions}
The paper is organized into three parts:  (i) a pointwise generalization framework (Section \ref{sec pointwise generalization});  (ii) structural principles of deep networks (Sections \ref{sec non-perturbative} to \ref{sec generalization DNN}); and (iii) empirical validation (Section \ref{section: experiments}). Related work appears in Appendix \ref{subsec related works}, and all proofs are in Appendices \ref{appendix pointwise}, \ref{appendix DNN}, \ref{sec geometry algebra},  \ref{appendix generalization}. Below we summarize the main novelties in each part.

\paragraph{Pointwise Generalization and Finite-Scale Geometry.}
We develop a pointwise framework that analyzes the \emph{trained} hypothesis and yields generalization bounds with (qualified) matching upper and lower rates via a finite-scale notion of \emph{pointwise dimension}. This fundamentally upgrades generic chaining and all covering–number approaches by assigning each hypothesis its {\it own complexity} that directly controls its error. The bounds can also be read as an optimally tuned PAC–Bayes objective specialized to deterministic predictors. Taken together, this framework reframes generalization as a study of pointwise geometry governed by dimension reduction at finite  precision, clarifying  why nonlinear models can generalize even in the absence of uniform convergence.

\paragraph{Structural Principles and Tight Bounds for Neural Networks.}
We develop a \emph{non-perturbative} analysis based on exact telescoping identities (rather than Taylor linearizations) that preserves the finite-scale geometry of deep networks. This yields our first structural principle: \emph{cross-layer correlations factor through the feature matrices, approximately preserving a pointwise-linear structure} (Section~\ref{subsec Feature Expansion}). Next, we show that bounding the pointwise dimension reduces, on each local chart, to the gold-standard \emph{effective dimension}, and we lift this to a global statement by constructing an \emph{ellipsoidal} covering of the Grassmannian of subspaces. This extension—beyond classical differential-geometric/Lie-algebraic treatments—establishes our second structural principle: \emph{the complexity of the global atlas (covering reference eigenspaces) is commensurate with that of the local charts}. Building on these principles, we introduce the \emph{Riemannian Dimension}—a spectrum-aware, pointwise effective complexity—which governs generalization at the trained model and yields tight, analyzable bounds (Section~\ref{subsec Inter-layer Decomposition}). We provide explicit generalization guarantees for deep neural networks and argue that they are tight in a qualified sense. We establish a principled \emph{exponential} improvement over spectral-norm bounds, provide an explicit separation, and discuss implications of our pointwise generalization theory for algorithm design (Section~\ref{sec generalization DNN}).

\paragraph{Unconditional compression, observability, and an open problem.}
The paper deliberately separates two statements that are often conflated.  Theorem~\ref{thm dnn} is unconditional and already gives a compressed pointwise bound through the mixed empirical-population Riemannian Dimension (Section \ref{subsection gen bound of dnn}).  Theorem~\ref{thm empirical dnn feature iso} gives the fully observable upgrade: under a finite-resolution subspace isomorphism (Section~\ref{subsec fully observable}), it replaces the population feature contribution by the observed spectrum. This is a sufficient finite-resolution condition in our derivation. We further pose its verification from subspace-level sub-Gaussian or small-ball feature regularity as an open question, identifying a precise route toward fully observed, pointwise feature-compressed bounds.

\paragraph{Empirical Findings and Evidences.}  The experiments are designed to systematically examine three central questions in modern deep learning: (i) why does overparameterization often improve generalization? (ii) how does feature learning evolve during training? and (iii) what implicit regularization is encoded by the baseline optimizer? Across the experimental results, we observe that (i) the overparameterization impressively leads to  decreasing Riemannian Dimension; (ii) feature learning compresses the effective ranks of learned features during the training; and (iii) stochastic gradient descent (SGD) with momentum implicitly regularizes the Riemannian Dimension.

\section{The Nature of Pointwise Generalization}
\label{sec pointwise generalization}
In this section, we develop our pointwise framework for generalization analysis, which introduces a tight tool--pointwise  dimension--to characterize  generalization. We illustrate its advancement to existing methodologies  and bring new understandings on the nature of generalization.
\subsection{Pointwise Dimension Strengthens PAC-Bayes and Generic Chaining}
Let \(\mathcal F\) be a hypothesis class, let \(z\in\mathcal Z\) be a random data (e.g., input-label pair $z=(x,y)$), and let \(\ell:\mathcal F\times\mathcal Z\to\mathbb R\) be a real-valued loss. Denote by \(\mathbb P\) the (unknown) population distribution, and by \(\mathbb P_n\) the empirical distribution associated with an i.i.d. sample \(S=\{Z_i\}_{i=1}^n\sim\mathbb P^{\otimes n}\). For any integrable function \(g:\mathcal Z\to\mathbb R\),
define the population and empirical averaging operators by
\[
\bP g := \mathbb{E}_{z\sim \bP}[g(z)],\qquad
\bP_n g := \frac{1}{n}\sum_{i=1}^n g(z_i).
\]
For convenience, we will often write \(g(z)\) inside these operators (e.g., \(g(z)=\ell(f;z)\)). Our goal is to control the \emph{generalization gap} $(\mathbb{P}-\mathbb{P}_n )\ell(f;z)$ in the following {\it pointwise generalization} manner: for $\delta\in(0,1)$, with probability at least $1-\delta$, uniformly over every $f\in\mathcal{F}$, 
\begin{align}\label{eq: generalization gap}
(\mathbb{P}-\mathbb{P}_n )\ell(f;z)
\;:=\;
\mathbb{E}_{z\sim\mathbb{P}}\bigl[\ell(f;z)\bigr]
\;-\;
\frac1n\sum_{i=1}^n \ell(f;z_i)\le\;
C\sqrt{\frac{d(f)+\,\log\frac{1}{\delta}}{n}
},
\end{align}
where $d(f)$ is a hypothesis-dependent complexity measure that aims to characterize the intrinsic complexity of every {\it trained} hypothesis $f$, different from class-wide, uniformly defined complexity measures. In Appendix \ref{subsec uniform pointwise convergence}, we state necessary and sufficient conditions for pointwise generalization through the ``uniform pointwise convergence'' principle proposed in \cite{xu2020towards, xu2025towards}. This stands in opposition to the uniform-convergence paradigm \citep{vapnik2013nature}, which seeks a single worst-case bound on $\sup_{f\in\mathcal F}(\mathbb P-\mathbb P_n)\ell(f;z)$ rather than the precise, hypothesis-dependent (pointwise) control we pursue here.
The failure of class-wide uniform convergence to meaningfully capture generalization in deep networks—in sharp contrast to its success in classical linear and kernel models—motivates a fundamental departure toward tight, hypothesis-dependent complexity measures that can distinguish generalizable solutions from arbitrary interpolants \citep{zhang2021understanding,neyshabur2017exploring,belkin2019reconciling,nagarajan2019uniform,wilson2025deep}.

In the spirit of \eqref{eq: generalization gap}, we introduce the central notion of this section, the \emph{pointwise dimension}: a finite-scale analogue of asymptotic notions from fractal geometry \citep{falconer1997techniques}, a  small-ball characterization to rate distortion and PAC-Bayesian optimization \citep{dembo2002source,zhang2006information}, and a measure of the local complexity encoded by the Fernique-Talagrand integral in generic chaining  \citep{fernique1975regularite,talagrand2014upper}. In addition, \citet{maurer2010majorizing,chu2023majorizing} provide direct precedents for our pointwise-dimension viewpoint by establishing simultaneous envelopes for stochastic processes in terms of point-dependent variants of the Fernique-Talagrand integral. We formalize the underlying ball-mass complexity as a finite-scale analogue of the canonical notion of pointwise dimension in general metric spaces and derive simple, direct generalization bounds in terms of this quantity.
 Throughout the paper, “metric” \(\varrho\) means a \emph{semi-metric}: all metric axioms hold except that $\varrho(f_1,f_2)=0$ need not imply $f_1=f_2$. 
\begin{definition}[Pointwise Dimension; cf.\ \citet{falconer1997techniques}] \label{def pointwise dimension}Given a function class $\cF$, a metric $\varrho$ on $\cF$, and a prior $\pi$ over $\cF$, the \emph{pointwise (or local) dimension} at $f$ with scale $\varepsilon$ is defined as the log inverse density of the $\varepsilon-$ ball $B_{\varrho}(f,\varepsilon)=\{f'\in\cF: \varrho(f,f')\leq \varepsilon \}$ centered at $f$:
\begin{align}\label{eq: pointwise dimension}
    \log\frac{1}{\pi(B_{\varrho}(f,\varepsilon))}.
\end{align}
\end{definition}

We first state the one-shot bound suggested by PAC-Bayes optimization, and then the multiscale bound obtained by generic chaining. Write $\mathbb P_S=\mathbb P_n$ and define the mixed empirical-population metrics
\begin{align}
\varrho^{(1)}_{(S,\mathbb P),\ell}(f,g)
&:=(\mathbb P_S+\mathbb P)|\ell(f;z)-\ell(g;z)|,\nonumber\\
\varrho_{(S,\mathbb P),\ell}(f,g)
&:=\sqrt{(\mathbb P_S+\mathbb P)(\ell(f;z)-\ell(g;z))^2}.
\label{eq empirical population metrics}
\end{align}
For the one-shot bound, we can use the smaller gap distance
\begin{align}
\varrho^{\mathrm{gap}}_{(S,\bP),\ell}(f,g)
&:=\big|(\bP-\bPn)(\ell(f;z)-\ell(g;z))\big|,
\label{eq gap distance}\\
\varrho^{\mathrm{gap}}_{(S,\bP),\ell}(f,g)
&\le\varrho^{(1)}_{(S,\bP),\ell}(f,g)
\le\sqrt2\,\varrho_{(S,\bP),\ell}(f,g).
\label{eq gap metric comparison}
\end{align}

This is the smallest symmetric distance controlling differences of generalization gaps. It depends on the unknown population gaps; the mixed $L_1$ distance gives a useful relaxation in terms of loss differences, while the mixed $L_2$ metric is used by chaining and the subsequent DNN analysis.

\begin{theorem}[One-Shot Pointwise Generalization Bound]\label{thm one shot}
Let $\ell(f;z)\in[0,1]$. For any prior $\pi$ on $\mathcal F$ independent of $S$ and any $\delta\in(0,1)$, with probability at least $1-\delta$, uniformly over every $f\in\mathcal F$,
\begin{align}\label{eq: one shot}
(\mathbb P-\mathbb P_n)\ell(f;z)
\le C\left(
\inf_{0\le\varepsilon\le2}\left\{
\underbrace{\varepsilon}_{\text{bias (approximate $f$)}}+
\underbrace{\sqrt{\frac{\log\frac1{\pi\Big(B_{\varrho^{\mathrm{gap}}_{(S,\mathbb P),\ell}}(f,\varepsilon)\Big)}}{n}}}_{\text{variance (PAC-Bayes term)}}
\right\}
+\sqrt{\frac{\log(2\log(2n)/\delta)}{n}}\right),
\end{align}
where $C>0$ is an absolute constant.
\end{theorem}
Here, the bias parameter $\varepsilon$ specifies the one-shot approximation resolution in the posterior optimization problem underlying this closed-form objective and can be optimized pointwise for each realized hypothesis $f$. Theorem~\ref{thm PACBayes-uniform-upper} in Appendix~\ref{subsec PAC-Bayes} provides a direct PAC-Bayes proof of this theorem. Theorem~\ref{thm lower bound} establishes the worst-case optimality of its complexity term.

The concept of pointwise dimension and Theorems~\ref{thm one shot} and~\ref{thm generic chaining} strengthen several established generalization methodologies, including PAC-Bayesian analysis, Kolmogorov complexity, generic chaining, and covering numbers. We elaborate below. The short, direct proof based on PAC-Bayes optimization originally led us to identify pointwise dimension as a quantity governing generalization. The chaining approach extends this core idea and requires developing several new technical ideas and arguments.
\paragraph{Theorem~\ref{thm one shot} sharpens PAC-Bayes posterior optimization.}
A key novelty is that, by measuring the prior mass accumulated in a \emph{localized metric ball} centered at \(f\), rather than the prior mass assigned to the singleton \(\{f\}\), the pointwise dimension framework naturally accommodates general uncountable classes \citep{zhang2006information,dembo2002source}. This circumvents the discreteness limitations inherent in hypothesis-singleton bounds corresponding to singleton localization, such as Occam-type or Kolmogorov-complexity bounds (e.g., \cite{lotfi2022pac,sutskever2023observation}), thereby yielding strictly stronger guarantees; interested readers are referred to \cite{lutz2016note} for related connections between pointwise dimension and algorithmic complexity.  Additionally, our perspective brings the best possible PAC-Bayesian mechanism under the selected metric: generalization is recast as a bias–variance tradeoff optimized over a user–chosen posterior, applies to \emph{deterministic} hypotheses, and shows that the pointwise dimension optimally governs the complexity; see Appendix~\ref{subsec PAC-Bayes} for this perspective. This clarifies and strengthens earlier PAC-Bayes optimization approaches, which typically restrict the posterior to a tractable Gaussian family in the Euclidean weight space, rather than optimizing over arbitrary posteriors on the nonlinear hypothesis class, in order to obtain computable and nonvacuous bounds for random outputs \citep{hinton1993keeping, dziugaite2017computing} and deterministic hypotheses  \citep{neyshabur2017exploring,neyshabur2017pac}; see the end of Section~\ref{subsec finite scale} for more details.

In addition, the scope of Theorems~\ref{thm one shot} and~\ref{thm generic chaining} goes well beyond traditional local Rademacher complexity analysis. The key distinction is that our approach localizes the dimension factor—that is, the model complexity itself—through the ``right’’ notion of pointwise dimension, rather than merely localizing a norm radius as in standard localization arguments. While the latter is sufficient for obtaining fast and adaptive rates, it does not capture the overparameterization phenomena studied in this paper.  See the framework in \cite{xu2020towards,xu2025towards}, where the current work is positioned as showing that pointwise dimension is the central localized complexity for this purpose.

\begin{theorem}[Multiscale Pointwise  Generalization Bound]\label{thm generic chaining}
Let $\ell(f;z)\in[0,1]$. For any prior $\pi$ on $\mathcal F$ independent of $S$ and any $\delta\in(0,1)$, with probability at least $1-\delta$, uniformly over every $f\in\mathcal F$,
\begin{align*}
(\mathbb P-\mathbb P_n)\ell(f;z)
\le C\left(
\inf_{0\le\alpha\le\sqrt2}\left\{
\alpha+\frac1{\sqrt n}\int_\alpha^{\sqrt2}
\sqrt{\log\frac1{\pi(B_{\varrho_{(S,\mathbb P),\ell}}(f,\varepsilon))}}\,d\varepsilon
\right\}
+\sqrt{\frac{\log(2\log(2n)/\delta)}{n}}\right),
\end{align*}
where $C>0$ is an absolute constant and $\varrho_{(S,\mathbb P),\ell}$ is defined in \eqref{eq empirical population metrics}.
\end{theorem}
The proof introduces a ghost sample and then applies Lemma~\ref{lemma ghost population averaging} to obtain the displayed empirical-population bound. A sharper version retaining the conditional ghost expectation appears in Lemma~\ref{lemma conditional ghost}.
The main challenge for empirical processes is to establish a multiscale bound governed by the complexity of the realized hypothesis, integrating over scales $\varepsilon$ rather than using a single resolution, in the spirit of earlier Gaussian-process bounds \citep{maurer2010majorizing,chu2023majorizing}. Here, the prior $\pi$ must be selected independently of the sample, in contrast to class-uniform tools such as generic chaining and empirical covering that rely on sample-dependent measures or covers. The truncation level $\alpha$ can be optimized pointwise for each realized hypothesis $f$, a feature that is also new to the chaining and covering literature. We retain truncation for technical convenience, although it plays a secondary role in the theory of the multiscale integral.

\paragraph{Theorem~\ref{thm generic chaining} upgrades generic chaining and covering numbers to a pointwise form.}
Building on the core idea behind the PAC-Bayes optimization bound in Theorem~\ref{thm one shot}, we develop new techniques to obtain \emph{pointwise} upgrades of of classical generic chaining bounds based on the Fernique-Talagrand integral \citep{fernique1975regularite,talagrand1987regularity,talagrand2014upper,block2021majorizing}.  The resulting \emph{pointwise} generalization bounds can be strictly stronger than entropy-integral bounds based on \emph{uniform} covering numbers
(e.g., Dudley’s integral), whose integrand takes a supremum over the entire class $\mathcal F$ rather than localizing at the realized hypothesis. In particular, \eqref{eq: sandwitch fractional cover} in Appendix~\ref{appendix ambient equivalence} demonstrates that the {\it class-wide} fractional covering number
\begin{align}\label{eq: fractional covering number}
\inf_{\pi\in\Delta(\cF)}\ \sup_{f\in\mathcal{F}}  \frac{1}{\pi\!(B_{\varrho}(f,\varepsilon))}
\end{align}
is (up to absolute constants) equivalent to the \emph{canonical covering number} of $\mathcal{F}$ with metric $\varrho$ at scale~$\varepsilon$.
Consequently, Theorem~\ref{thm generic chaining} goes beyond classical covering analyses by
(i) recasting covering-number complexity as the inverse-prior-density objective \eqref{eq: fractional covering number}, and
(ii) localizing this complexity \emph{pointwise} in $f$.
Through the results and tools developed in this paper, we advocate the “prior-density + localization” (i.e., pointwise dimension) viewpoint as a new paradigm for statistical complexity analysis—one that is markedly more flexible than the classical covering-number approaches prevalent in statistics and  machine learning. 

\begin{corollary}[Common One-Shot $L_2$ Relaxation]\label{coro one shot L2}
Each of Theorems~\ref{thm one shot} and~\ref{thm generic chaining} implies the following bound: under their common hypotheses, with probability at least $1-\delta$, uniformly over every $f\in\mathcal F$,
\begin{align}\label{eq one shot L2}
(\mathbb P-\mathbb P_n)\ell(f;z)
\le C\left(
\inf_{0\le\varepsilon\le\sqrt2}\left\{
\varepsilon+\sqrt{\frac{\log\frac1{\pi\big(B_{\varrho_{(S,\mathbb P),\ell}}(f,\varepsilon)\big)}}{n}}
\right\}
+\sqrt{\frac{\log(2\log(2n)/\delta)}{n}}\right).
\end{align}
\end{corollary}
From Theorem~\ref{thm one shot}, use \eqref{eq gap metric comparison} and the inclusion $B_{\varrho_{(S,\mathbb P),\ell}}(f,\varepsilon)\subseteq B_{\varrho^{\mathrm{gap}}_{(S,\mathbb P),\ell}}(f,\sqrt2\varepsilon)$. From Theorem~\ref{thm generic chaining}, use that its integrand is nonincreasing, so the integral from $\alpha$ to $\sqrt2$ is bounded by $\sqrt2$ times the integrand's value at $\alpha$. Our pointwise optimization allows us to set the truncation level $\alpha$ equal to the chosen approximation resolution $\varepsilon$. In either case, enlarge the absolute constant $C$ as needed.

\paragraph{The gap-distance, $L_1$, and $L_2$ bounds.}
Theorem~\ref{thm one shot} also holds with $\varrho^{\mathrm{gap}}_{(S,\mathbb P),\ell}$ replaced by $\varrho^{(1)}_{(S,\mathbb P),\ell}$. By \eqref{eq gap metric comparison}, the gap-distance bound is no larger than this $L_1$ relaxation, which in turn sharpens the $L_2$ one-shot bound up to absolute constants. At a fixed $L_2$ metric, chaining improves the one-shot relaxation, but there is no general dominance order between the $L_1$ one-shot bound and the $L_2$ chaining bound: the former exploits $L_1$ approximation, whereas the latter exploits all scales. The DNN analysis below uses $L_2$ geometry to express complexity through feature spectra; the experiments evaluate its one-shot relaxation using the observed feature geometry, subject to the observability distinction in Section~\ref{subsec fully observable}.

The multiscale improvement can be substantial in the same $L_2$ metric. For example, a pointwise dimension estimate of order $d(f)\varepsilon^{-2}$ gives a truncated-integral bound of order $\sqrt{d(f)/n}$ up to logarithmic factors, whereas the corresponding one-shot estimate is of order $(d(f)/n)^{1/4}$.

\paragraph{}

The next theorem distinguishes worst-case sharpness from an instancewise characterization.
Let $A_\pi(f)$ and $C_\pi(f)$ denote the infima in Theorems~\ref{thm one shot} and~\ref{thm generic chaining}, respectively, excluding their common confidence remainder and outer constants.

\begin{theorem}[Lower Bounds and Scope of Tightness]\label{thm lower bound}
Let $\ell(f;z)\in[0,1]$ and $n\ge2$. There are positive absolute constants $c,C$ with the following properties.

\textbf{\emph{One-shot gap-distance lower bound.}}
For every sample and every prior $\pi$, we have
\begin{equation}\label{eq gap objective oscillation}
\sup_{f\in\cF}|(\bP-\bPn)\ell(f;z)|\geq \frac{1}{2}\sup_{f\in\cF}A_\pi(f).
\end{equation}

\textbf{\emph{Chaining integral lower bound.}} With
$\varrho_{n,\ell}(f,g)^2=\bPn(\ell(f;z)-\ell(g;z))^2$,
\begin{align*}
\bE\sup_{f\in\cF}|(\bP-\bPn)\ell(f;z)|
\ge{}\frac{c}{\sqrt{n\log n}}\bE\inf_{\pi\in\Delta(\cF)}\sup_{f\in\cF}
\int_0^1\sqrt{\log\frac1{\pi(B_{\varrho_{n,\ell}}(f,\varepsilon))}}\,d\varepsilon-\frac{C\sup_{f\in\cF}\bP\ell(f;z)}{\sqrt{n\log n}}.
\end{align*}
The absolute value may be omitted if the loss class is closed under complementation.

\textbf{\emph{Both upper bounds can be sharp.}} For every pair of integers $2\le d\le n$, there is a finite loss class and a distribution for which, under the uniform prior $U$, each of $\bE\sup_f|(\bP-\bPn)\ell(f;z)|$ and $\bE\sup_f A_U(f)$ lies between $c\sqrt{d/n}$ and $C\sqrt{d/n}$, and
\begin{equation}\label{eq mixed worst case lower}
c\sqrt{\frac dn}
\le\bE\inf_\pi\sup_f C_\pi(f)
\le\bE\sup_f C_U(f)
\le C\sqrt{\frac dn}.
\end{equation}
The empirical and mixed chaining functionals are also equivalent in expectation on this family, both with and without pointwise optimized truncation.

\textbf{\emph{Chaining need not be sharp on every problem.}} For every $n\ge2$ and $0<p<1$, there is another finite loss class and distribution for which each of $\bE\sup_f|(\bP-\bPn)\ell(f;z)|$ and $\bE\sup_f A_U(f)$ lies between $cp$ and $Cp$, but
\begin{equation}\label{eq rare chaining separation}
\frac{\sqrt p}{2}
\le\bE\inf_\pi\sup_f C_\pi(f)
\le\bE\sup_f C_U(f)
\le\sqrt{2p}.
\end{equation}
In both constructions, $U$ is independent of the sample. The chaining lower bounds hold even when the infimum over priors is taken after observing $S$.
\end{theorem}

Theorem~\ref{thm lower bound} establishes worst-case sharpness of the complexity terms in our one-shot and multiscale pointwise generalization bounds (Theorems~\ref{thm one shot} and~\ref{thm generic chaining}). For the one-shot bound, the deterministic comparison \eqref{eq gap objective oscillation} unconditionally controls the class-wide complexity by twice the largest absolute generalization gap for every loss class and distribution, although it uses the unknown population gaps. For the multiscale bound, sharpness is established in the weaker existential sense through matching examples.

The worst-case statements are  analogous in spirit to the minimax optimality in frequentist decision theory \citep{wald1945minimax}. It does not preclude sharper guarantees for a fixed hypothesis $f$. A matching pointwise lower bound, without the outer supremum over $f\in\cF$, cannot hold in general: a hypothesis may have zero generalization gap while its complexity term remains positive. Appendix~\ref{appendix mixed lower bound} provides the proofs and  constructions illustrating both sharpness and limitations.
These statements also highlight the importance of metric choice for both pointwise generalization and uniform convergence.

We defer technical innovations and connections to existing methodologies to the Appendix—most notably the unified pointwise–generalization framework of \citet{xu2020towards,xu2025towards} which we build upon (Appendix~\ref{subsec uniform pointwise convergence}), and the alternative PAC–Bayesian perspective (Appendix~\ref{subsec PAC-Bayes}). We conclude that the proposed \emph{pointwise dimension} yields hypothesis-specific generalization guarantees whose complexity terms can be sharp in the worst case.

\subsection{Necessity of Finite-Scale Pointwise Geometry and Structural Analysis}
\label{subsec finite scale}

The transition from uniform convergence to the ``prior-density + localization'' (pointwise dimension) perspective offers a fundamental tightening over standard covering number approaches. However, translating this theoretical advantage into a practical framework for deep learning requires addressing two distinct challenges. First, we will distinguish the \emph{geometric nature} of generalization from classical infinitesimal geometry: relevance lies not in the limit $\varepsilon \to 0$, but motivates a new program of finite-scale geometric analysis. Second, we must overcome the \emph{computational intractability} of evaluating the pointwise dimension directly, which necessitates a dedicated structural analysis for deep neural networks.

\paragraph{Asymptotic vs.\ Finite-Scale Dimension.}
Although powerful in mathematics, standard differential-geometric tools (e.g., pointwise metrics and subspace angles) have not been systematically used in generalization theory, largely because they define dimension in  infinitesimal notions. For instance, the \emph{asymptotic pointwise dimension}—central to fractal and Riemannian geometry \citep{falconer1997techniques, jost2008riemannian} and used to characterize Hausdorff and packing dimensions (e.g., Theorem~3 of \cite{lutz2016note})—is defined via a limit:
\[
 \lim_{\varepsilon\to 0}\frac{\log \pi(B_{\varrho}(f,\varepsilon))}{\log \varepsilon }.
\]
We argue that generalization is distinct from, and in some ways more challenging than, infinitesimal geometry: the nature of generalization in deep models lies in reducing geometric dimension at a \emph{finite scale} of precision for each hypothesis. Crucially, the pointwise dimension $\log\!\frac{1}{\pi\!(B_{\varrho}(f,\varepsilon))}$ is monotonic: it naturally decreases as the resolution $\varepsilon$ increases. Therefore, a finite-scale analysis reveals significant dimension reduction that infinitesimal analysis misses. In our one-shot bound \eqref{eq: one shot}, the objective is to identify the optimal finite scale $\varepsilon^\star$ where the trade-off between precision and pointwise complexity is minimized. At this scale, the effective dimension can be orders of magnitude smaller than the asymptotic dimension, explaining the tractability of overparameterized models. To the best of our knowledge, this distinction is novel; prior uses of geometric dimension in generalization (e.g., \cite{birdal2021intrinsic}) have largely emphasized globally uniform and infinitesimal notions. And the Neural Tangent Kernel (NTK) \citep{jacot2018neural} and Gaussian-process \citep{lee2018deep} viewpoints are valid only in an infinitesimal neighborhood of initialization (equivalently, in the infinite-width regime). A precise account of deep-model generalization thus calls for a shift from infinitesimal calculus to finite-scale, pointwise geometry.

\paragraph{Computation and the Necessity of Structural Analysis.}
Although worst-case sharp, Theorems~\ref{thm one shot} and \ref{thm generic chaining}
—like many abstract bounds (PAC-Bayes, mutual-information, generic chaining)—is generally not computationally tractable on its own; practical use requires adapting it to the function class at hand and introducing suitable relaxations. If we denote an effective dimension by $d(f)=\log \frac{1}{\pi(B_{\varrho}(f,\varepsilon^\star))}$ ($\varepsilon^\star$ tuned in  one-shot optimization), a brute-force Monte Carlo estimator using i.i.d.\ draws $f'\sim\pi$ would require on the order of $e^{d(f)}$ samples to obtain a single hit $f'\in B_{\varrho}(f,\varepsilon)$ with constant probability. For high-dimensional deep networks, where $d(f)$ should be moderate to large, this is computationally prohibitive.

This intractability helps explain why much of the PAC-Bayes literature turns to tractable weight-space surrogates that are effectively linearized: instead of optimizing over arbitrary posteriors on the nonlinear hypothesis class \(\mathcal F\), one restricts to Gaussian priors and posteriors over the weights \(W\in\mathbb R^p\), typically with isotropic or fixed covariance structure. This restriction yields closed-form KL terms and computable objectives \citep{hinton1993keeping,dziugaite2017computing}; see Sections~3 and~6 of \cite{dziugaite2017computing} for representative formulations. However, this strategy implicitly imposes a uniform linearization that discards the distinctive {\it pointwise} geometry of deep networks, effectively flattening a curved manifold.  To retain the sharpness of pointwise dimension without incurring the simulation barrier, we therefore avoid black-box sampling and instead develop explicit {\it structural principles} of deep networks that allow analytic control of the pointwise dimension—yielding generalization guarantees that are both theoretically rigorous and practically computable.

\section{Deep Neural Networks and Pointwise Riemannian Dimension}\label{sec non-perturbative}
In this section we develop a systematic pointwise dimension analysis for deep neural networks, focusing on the empirical DNN geometry induced by the observed sample \(S\); the separate reduction from the mixed empirical-population metric in Theorem~\ref{thm generic chaining} to this observed-sample complexity is handled in Section~\ref{sec generalization DNN}.
Section~\ref{subsec nn setup} formalizes the standard fully connected architecture and notation.
Section~\ref{subsec Feature Expansion} introduces a non-perturbative calculus (avoiding infinitesimal Taylor expansions) to analyze finite-scale behavior.
Section~\ref{subsec Inter-layer Decomposition} introduces a hierarchical covering scheme—our key technical innovation—that overcomes the well-known linear/kernel bottleneck in classical statistical learning and enables a principled treatment of genuinely nonlinear models.

\subsection{Neural Network Setup}\label{subsec nn setup}
We consider fully connected (feed-forward) networks that map an input
\(x\in\mathbb{R}^{d_0}\) to an output \(f_L(W,x)\in\mathbb{R}^{d_L}\).
The architecture is specified by widths \(d_0,\dots,d_L\) and weight matrices
\(W=\{W_1,\dots,W_L\}\) with \(W_l\in\mathbb{R}^{d_l\times d_{l-1}}\) for \(l=1,\dots,L\).
Let \(\sigma_1,\dots,\sigma_L\) be nonlinear activations (e.g., ReLU), acting componentwise on column vectors, and each \(\sigma_l:\mathbb{R}^{d_l}\to\mathbb{R}^{d_l}\) is assumed
\(1\)-Lipschitz. The network’s forward map is the composition
\[
f_L(W,x)\;:=\;\sigma_L\!\Big(W_L\,\sigma_{L-1}\big(W_{L-1}\,\cdots\,\sigma_1(W_1 x)\big)\Big).
\]
Let \(X=[x_1,\dots,x_n]\in\mathbb{R}^{d_0\times n}\) collect the \(n\) training inputs as columns.
For each layer \(l\in\{1,\dots,L\}\), define the depth-\(l\) map and the corresponding \emph{feature matrix}
\[
f_l(W,x)\;:=\;\sigma_l\!\Big(W_l\,\sigma_{l-1}\big(W_{l-1}\,\cdots\,\sigma_1(W_1 x)\big)\Big), F_l(W,X)\;:=\;\big[f_l(W,x_1)\ \cdots\ f_l(W,x_n)\big]\in\mathbb{R}^{d_l\times n}.
\]
Equivalently (full, non-recursive form consistent with \eqref{eq: intro DNN}),
\[
F_l(W,X)\;=\;\sigma_l\!\Big(W_l\,\sigma_{l-1}\big(W_{l-1}\,\cdots\,\sigma_1(W_1 X)\big)\Big),
\]
where for a matrix \(A=[a_1,\dots,a_n]\) we write
\(\sigma_l(A):=[\,\sigma_l(a_1),\dots,\sigma_l(a_n)\,]\).
Thus \(F_L(W,X)\) collects the network outputs on the dataset \(X\).

We denote \(\|\cdot\|_{\tF}\) for the Frobenius norm, \(\|\cdot\|_{\op}\) for the spectral norm, and $\|\cdot\|_2$ for the Euclidean norm on vectors. We abbreviate norm balls by $B_{\tF}(R)$, $B_{\op}(R)$, and $B_{2}(R)$ (all centered at $0$; with radius $R$).
The empirical \(L_2(\mathbb{P}_n)\) distance between two hypotheses \(W,W'\) is (a $1/\sqrt{n}$ scaling is used to keep consistency with Section~\ref{sec pointwise generalization})
\[
\varrho_n(W,W')\;:=\;\sqrt{\,\big\|F_L(W,X)-F_L(W',X)\big\|_{\tF}^2/n}\,.
\]
The function-level empirical metric and generalization statements in Section~\ref{sec pointwise generalization}
for the loss \(x\mapsto \ell(f_L(W,x),y)\) at data–label pairs \(z=(x,y)\)
specialize, on the dataset \(X\), to the metric $\varrho_n$ defined above. We assume the loss $\ell(\cdot,y)$ is $\beta$-Lipschitz in its first argument with respect to $f_L(W,x)$. This bridges the loss-induced metric on $\mathcal F$, studied in Section~\ref{sec pointwise generalization}, with the weight-space metric used here.
\subsection{Non-Perturbative Expansion and Layer-wise Correlation}\label{subsec Feature Expansion}
Throughout, our finite-scale analysis relies on \emph{non-perturbative} expansions. Borrowing terminology from theoretical physics, “non-perturbative” here means we avoid Taylor/derivative expansions and instead use exact, telescoping algebraic identities that hold at finite scale. For example, 
\[
W'_2W'_1-W_2W_1=W'_2(W'_1-W_1)+(W'_2-W_2)W_1,
\qquad
{\Sigma'}^{-1}-\Sigma^{-1}={\Sigma'}^{-1}(\Sigma-\Sigma')\Sigma^{-1},
\]
with analogous decompositions used throughout. This viewpoint preserves the full finite-scale geometry of deep networks, rather than linearizing around an infinitesimal neighborhood.

To present our non-perturbative expansion for DNN, we define {\it local Lipschitz constant} as follows.

\begin{definition}[Local Lipschitz Constant]\label{def local Lipschitz constant}
For each $l=1,\cdots, L$, we define $M_{l\rightarrow L}(W,\varepsilon)$ as the (outer) local Lipschitz constant, which characterizes the sensitivity of the layer $L$ output, $F_L$, to variations in layer $l$'s output, within a neighborhood around $F_l$. Formally, we assume  that for every $W'\in B_{\varrho_n}(W,\varepsilon)$
\begin{align*}
   || F_L(F_l(W',X),\{W'_{i}\}_{i=l+1}^L)-F_L(F_l(W,X),\{W'_{i}\}_{i=l+1}^L)||_{\tF}\leq M_{l\rightarrow L}(W,\varepsilon)||F_l(W',X)-F_l(W,X)||_{\tF}.
\end{align*}
\end{definition}
Local Lipschitz constants are typically much smaller than products of spectral norms and can be
computed by formal–verification toolchains \citep{shi2022efficiently}. In our bounds, these constants enter only through \emph{optimistic logarithmic factors} and therefore do not affect the leading rates; see Example~\ref{example optimistic logarithmic} for the precise sense in which the relevant effective dimension is logarithmically  sensitive to such multiplicative scale factors.  We propose a telescoping decomposition to replace conventional Taylor expansion, where in each summand the only difference lies in $W_l'$ and $W_l$.
\begin{align}  \label{eq: telescoping equality}
&F_L(W',X)-F_L(W,X)\nonumber\\=&\sum_{l=1}^L[\underbrace{\sigma_L(W'_L\cdots W'_{l+1}}_{\textup{controlled by} M_{l\rightarrow L}}\underbrace{\sigma_l}_{\textup{by} 1}(W'_l\underbrace{F_{l-1}(W,X)}_{\textup{learned feature}}))-\sigma_L(W'_L\cdots W'_{l+1}\sigma_l(W_l\underbrace{F_{l-1}(W,X)}_{\textup{learned feature}}))].
\end{align}
Note that this is a {\it non-perturbative} expansion that holds unconditionally and does not rely on infinitesimal approximation, and crucially keeps the {\it learned} feature $F_{l-1}(W,X)$ at the {\it trained} weight $W$. From this decomposition and applying basic inequalities, we have the following key lemma. 
    \begin{lemma}[Non-Perturbative Feature Expansion]\label{lemma non-perturbative} For all $W'\in B_{\varrho_n}(W, \varepsilon)$,
\begin{align*}
    ||F_L(W',X)-F_L(W,X)||^2_{\tF}\leq \sum_{l=1}^L L \cdot {M_{l\rightarrow L}[W,\varepsilon]^2}\cdot ||(W'_l-W_l)F_{l-1}(W,X)||_{\tF}^2.
\end{align*}
The lemma captures the first structural principle of fully connected DNN:
\emph{cross-layer correlations mostly pass through the feature matrices, preserving an approximate pointwise linear structure}.
\end{lemma}

Since enlarging the metric  only shrinks metric balls and hence
\emph{increases} the pointwise dimension \eqref{eq: pointwise dimension} we analyze in Section \ref{sec pointwise generalization} (formalized as Lemma \ref{lemma simple metric domination}; metric domination lemma), it suffices to analyze
pointwise dimension under the \emph{pointwise ellipsoidal metric} that appears on the right-hand side of
Lemma~\ref{lemma non-perturbative}. Concretely, 
$F_{l-1}(W,X)\,F_{l-1}(W,X)^{\!\top}$,
the feature Gram matrix from layer \(l{-}1\),  faithfully encodes the spectral information
induced by the network–data geometry at layer $l$. Working with the corresponding pointwise ellipsoidal metric yields sharp,
\emph{pointwise, spectrum-aware} bounds with the desired properties for deep networks, and underpins
our tractability results (with the structural principles and technical innovations to be developed in the next subsection).
\subsection{Hierarchical Covering from Local Chart to Global Atlas}\label{subsec Inter-layer Decomposition}
Lemma~\ref{lemma non-perturbative} suggests that the following \emph{pointwise ellipsoidal metric}
dominates $n\cdot\varrho_n$ at every $W$ (here, NP stands for ``non-perturbative''):
\begin{align}\label{eq: metric tensor DNN}
&G_{\textup{NP}}(W)
      =\textup{blockdiag}\left(\cdots, LM^2_{l\rightarrow L}(W,\varepsilon) \cdot F_{l-1}(W,X)F^\top_{l-1}(W,X)\otimes  I_{d_{l}},\cdots\right)\nonumber\\
      &\varrho_{G_{\textup{NP}}(W)}(W,W')^2=\textup{vec}(W'-W)^\top G_{\textup{NP}}(W)\textup{vec}(W'-W).
\end{align}
We are therefore interested in bounding the enlarged pointwise dimension under the
pointwise ellipsoidal metric $\varrho_{G_{\textup{NP}}(W)}$:
\[\log\frac{1}{\pi(B_{\varrho_n}(f(W,\cdot),\varepsilon))}\leq  \log\frac{1}{\pi({B_{\varrho_{G_{\textup{NP}}(W)}}(W,\sqrt{n}\varepsilon)})}.\] This section offers a deep dive past classical effective dimension, shifting to hierarchical covering and a global geometric analysis.

\subsubsection{Gold Standard:  Effective Dimension} Classical studies of static ellipsoidal metrics suggest that if $\pi$ is chosen to be uniformly constrained on the top-$r$ eigenspace of a PSD matrix $G(W)$, and the vectorized weights $W \in \mathbb{R}^p$ are restricted to the Euclidean ball
$B_2(R) := \{w \in \mathbb{R}^p : \|w\|_2 \le R\}$, then one can achieve a tight effective dimension as follows:
define the {\it effective rank}
\begin{align}\label{eq: eff rank}
    r_{\textup{eff}}(G(W), R, \varepsilon) := \max \{ k : \lambda_k(G(W)) R^2 \ge n\varepsilon^2/2 \},
\end{align}
where the eigenvalues \(\{\lambda_k(G(W))\}\) are ordered nonincreasingly; and define the spectrum-aware {\it effective dimension}
\begin{align}\label{eq: effective dimension}
    d_\eff(G(W), R,\varepsilon):=\frac{1}{2} \sum_{k=1}^{r_{\eff}(G(W), R, \varepsilon)} \log\left( \frac{8 R^2 \lambda_k(G(W))}{n\varepsilon^2} \right).
\end{align}
This definition serves as a gold standard for static ellipsoidal metrics and is asymptotically tight, as established by the covering number of the unit ball with ellipsoids in Section 3.3 of \cite{even2021concentration}. For this static ellipsoidal metric, the same volume-ratio calculation also controls, up to absolute constants,
the local prior mass of ellipsoidal balls under the uniform measure on the top-$r$ eigenspace. Thus \(d_{\mathrm{eff}}\) is also the gold standard for
pointwise dimension. For brevity, we write $r$ for $r_\eff(G(W),R,\varepsilon)$, and denote by $\cV\subseteq\bR^p$ the $r$-dimensional subspace corresponding to the top-$r_\eff$ eigenspace of $G(W)$.

\begin{example}[Logarithmic Sensitivity to Scale]\label{example optimistic logarithmic}
    It is useful to record two elementary regimes in which the dependence of \eqref{eq: effective dimension} on the radius $R$, the precision $\varepsilon$, and any multiplicative rescaling of the metric tensor is only logarithmic, in the optimistic sense used throughout this paper.  Let
\[
    \widetilde G(W):=aG(W),\qquad a>0,
\]
so that $a$ may represent, for instance, the layerwise multiplier $L M_{l\to L}^2(W,\varepsilon)$ appearing in the neural-network metric tensor.

\paragraph{Exponential spectral decay.}  Suppose that for some $\lambda_0>0$ and $\gamma>0$,
\[
    \lambda_k(G(W))\le \lambda_0 e^{-\gamma(k-1)},\qquad k\ge 1.
\]
Then by the definition of effective rank $
    r_{\textup{eff}}(\widetilde G(W), R, \varepsilon)  = \max \{ k : a\lambda_k(G(W)) R^2 \ge n\varepsilon^2/2 \}$,
\[
    r_{\eff}(\widetilde G(W),R,\varepsilon)
    \le
    1+\frac{1}{\gamma}  \left[\log\!\left(\frac{2a\lambda_0R^2}{n\varepsilon^2}\right)\right]_+,
    \qquad [t]_+:=\max\{t,0\}.
\]
Moreover, if the decay is exact, $\lambda_k(G(W))=\lambda_0 e^{-\gamma(k-1)}$, and $r=r_{\eff}(\widetilde G(W),R,\varepsilon)$, then
\[
    d_{\eff}(\widetilde G(W),R,\varepsilon)
    =\frac{1}{2} \sum_{k=1}^{r} \log\left( \frac{8 R^2a\lambda_0 e^{-\gamma(k-1)}}{n\varepsilon^2} \right)=\frac{r}{2}\log\!\left(\frac{8a\lambda_0R^2}{n\varepsilon^2}\right)
      -\frac{\gamma}{4}r(r-1).
\]
In particular, in the general upper-bound case, combining the two bounds above yields 
\[
    d_{\eff}(\widetilde G(W),R,\varepsilon)
    \le
    C_\gamma   \left[\log\!\left(e+\frac{a\lambda_0R^2}{n\varepsilon^2}\right)\right]^2,
\]
where $C_\gamma$ depends only on the exponential-decay rate. 
 Thus even a large multiplicative factor $a$ changes the effective dimension only through a squared logarithm.
\paragraph{Strict low rank.}  Suppose instead that $\mathrm{rank}(G(W))\le q$ and $\lambda_1(G(W))\le \lambda_0$.  Then, for the rescaled metric $\widetilde G(W)=aG(W)$,
\[
    r_{\eff}(\widetilde G(W),R,\varepsilon)\le q,
\]
and hence
\[
    d_{\eff}(\widetilde G(W),R,\varepsilon)=\frac{1}{2} \sum_{k=1}^{r_{\eff}} \log\left( \frac{8 R^2 \lambda_k(\widetilde G(W))}{n\varepsilon^2} \right) \le \frac{r_{\eff}}{2}  \log\left(e+ \frac{8a R^2 \lambda_0}{n\varepsilon^2} \right)
    \le
    \frac{q}{2}   \log\!\left(e+\frac{8a\lambda_0R^2}{n\varepsilon^2}\right).
\]
In this strict feature-compression regime, the leading dependence is the intrinsic rank $q$, not the ambient dimension $p$, while $R$, $\varepsilon^{-1}$, and the metric multiplier $a$ enter only through a single logarithm.  This is the precise sense in which the effective-dimension calculus demotes products of norms and local-Lipschitz multipliers from leading-order complexity to logarithmic scale factors whenever the learned feature spectrum is compressed.
\end{example}

\subsubsection{Key Challenge: Prior Independence from $W$.}
However, the main challenge is that the prior $\pi$ must be chosen independently of the training data. This means that the construction of $\pi$ cannot rely on knowledge of the learned
weights $W$, including their top-$r_\eff$ eigenspace, yet still capture the underlying geometric structure. The next lemma extends classical results on static ellipsoidal metrics by showing that a uniform prior over a reference subspace $\bar\cV$ suffices to bound the pointwise dimension for all $W$ whose top-$r$ eigenspace of $G(W)$ can be approximated by $\bar\cV$.

\begin{lemma}[Pointwise Dimension via Reference Subspace]\label{lemma approximate subspace}
Consider the weight space $B_{2}(R)\subset \bR^p$ for vectorized weights, and a pointwise ellipsoidal metric defined via PSD $G(W)$.  
Let $\bar\cV\subseteq\bR^p$ be a fixed $r$-dimensional subspace.  
Define the prior $\pi_{\bar{\cV}}=\textup{Unif}\!\bigl(B_{2}(1.58R)\cap\bar{\cV}\bigr)$.  
Then, uniformly over all $(W,\varepsilon)$ such that the top-$r$ eigenspace $\cV$ of $G(W)$ can be approximated by $\bar\cV$ to precision
\begin{align}\label{eq: precision main}
\varrho_{\proj,G(W)}(\cV,\bar\cV)
:=\bigl\|\,G(W)^{1/2}\bigl(\cP_{\cV}-\cP_{\bar\cV}\bigr)\,\bigr\|_{\op}
\;\leq\; \tfrac{\sqrt{n}\varepsilon}{4R},
\end{align}
we have
\[
\log \frac{1}{\pi_{\bar{\cV}}(B_{\varrho_{G(W)}}(W,\sqrt{n}\varepsilon))}
\;\le\;
\frac{1}{2} \sum_{k=1}^{r_{\eff}(G(W),R,\varepsilon)} 
\log\!\left( \frac{40 R^2 \lambda_k(G(W))}{n\varepsilon^2} \right)
\;\leq\; d_\eff(G(W),\sqrt{5}R,\varepsilon).
\]
\end{lemma}
In \eqref{eq: precision main}, $\cP_\cV$ denotes the orthogonal projector onto the subspace $\cV$, and $\varrho_{\proj,G(W)}$ thus defines an ellipsoidal projection metric between subspaces. Further details are provided in the appendix.

\subsubsection{Hierarchical covering (mixture prior over subspaces).}
We introduce a hierarchical covering framework that pushes statistical learning beyond classical linear and kernel paradigms, providing a principled toolkit for genuinely nonlinear models—one of the central innovations of this work.  It operates on two levels: a bottom-level local-chart covering that captures spectrum-aware behavior within a fixed subspace, and a top-level global geometric analysis over the Grassmannian.

(i) For each reference subspace $\bar{\mathcal V}$, placing a uniform prior on $\bar{\mathcal V}$ yields a tight pointwise-dimension bound for all ``local'' weights $W$ whose top$-r$ eigenspace of $G(W)$ is well approximated by $\bar{\mathcal V}$ (see Lemma~\ref{lemma approximate subspace}).

(ii) At the top level, we place a prior over reference subspaces $\bar{\mathcal V}$ and average the local priors, producing a data-independent prior and the final bound. 

\begin{figure}[t]
    \centering
    \includegraphics[width=0.97\linewidth]{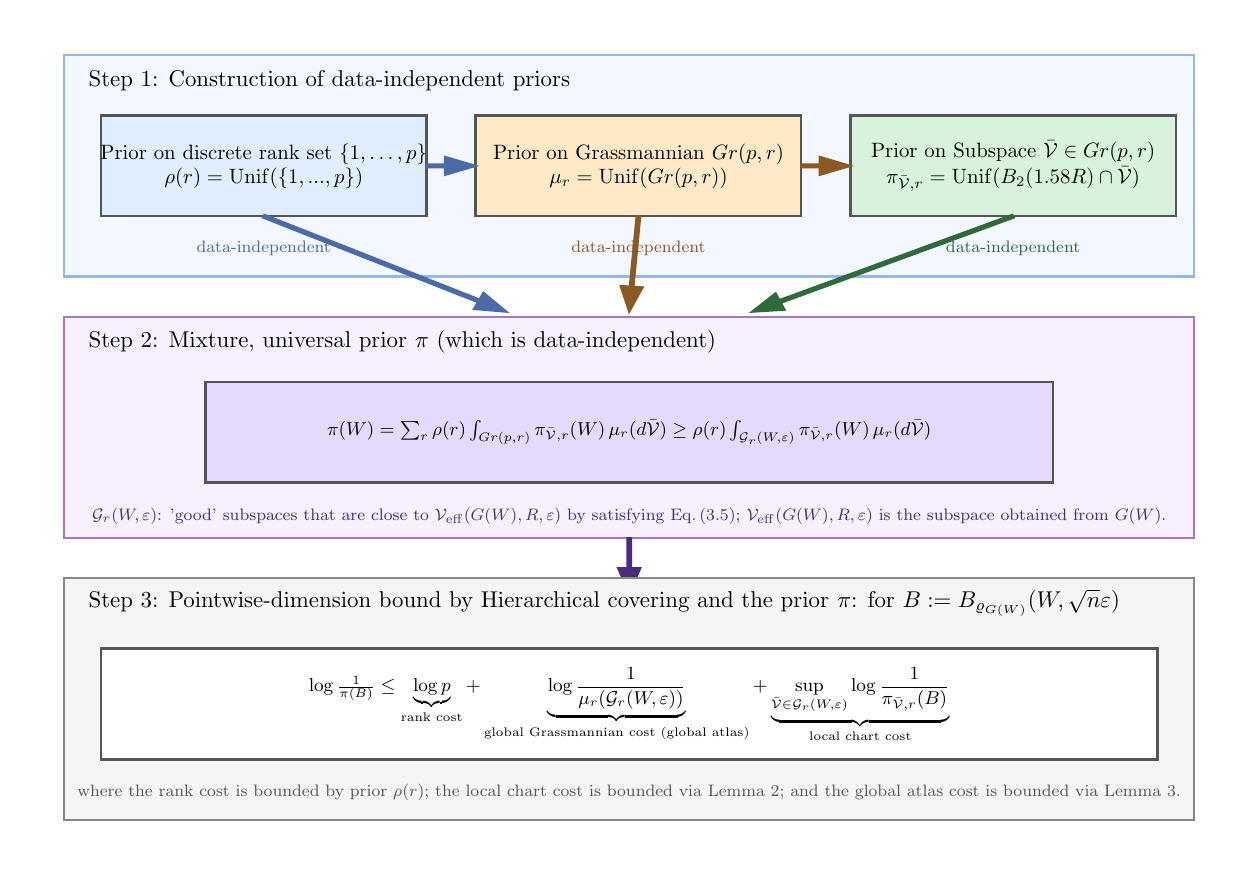}
    \caption{Hierarchical construction of the data-independent prior $\pi$ and its role in the pointwise-dimension bound (one single-layer case).}
    \label{fig:prior-diagram}
\end{figure}

By combining these two levels of priors, we obtain a pointwise dimension bound using a prior $\pi$ that is completely blind to the choice of $W$. To formalize this, we introduce a top-level distribution $\mu$ over the Grassmannian
\begin{align*}
    \textup{Gr}(p,r)\;:=\;\bigl\{\text{$r$–dimensional linear subspaces of }\bR^p\bigr\}
\end{align*}the collection of all $r$-dimensional subspaces, and define
\[
  \pi(W)
  \;=\;
  \sum_{\mathcal V}\mu(\mathcal V)\,\pi_{\mathcal V}(W).
\]
We refer to this two-stage construction as the hierarchical covering argument. Under the resulting prior $\pi$, the following bound holds uniformly over all (vectorized) $W \in B_{2}(R)$, the pointwise dimension $\log\frac{1}{\pi(B_{\varrho_{G(W)}}(W,\sqrt{n}\varepsilon))}$ is bounded by two parts:
\begin{align}\label{eq: hierarchical covering}
   \underbrace{\log\frac{1}{\mu(B_{\varrho_{\proj,G(W)}}(\cV,\sqrt{n\varepsilon/4R}))}}_{\textup{covering  Grassmannian (global atlas) }}+\underbrace{\sup_{\bar{\cV}\in B_{\varrho_{\proj,G(W)}}(\cV,\sqrt{n\varepsilon/4R})}\log\frac{1}{\pi_{\bar\cV}(B_{\varrho_{G(W)}}(W,\sqrt{n}\varepsilon))}}_{\textup{covering local charts}},
\end{align}
In differential–geometric terms, our argument has two components. 
\begin{itemize}
\item \emph{Local (chart) analysis:} fixing a reference subspace $\bar\cV$, we use effective dimension as the gold standard to determine the metric entropy of the corresponding local chart. 
\item \emph{Global (atlas) covering:} we cover the Grassmannian by such reference subspaces, i.e., we bound the metric entropy of the global atlas and account for the cost of transitioning across local charts. 
\end{itemize}
Lemma~\ref{lemma approximate subspace} controls the local part, while the following new result (Lemma \ref{lemma ellipsoidal Grassmannian}) on the \emph{ellipsoidal} covering of the Grassmannian controls the global part.\footnote{Since the effective rank \(r\) of $\bar\cV$ can take any value in \(\{1,\ldots,p\}\), the top-level Grassmannian covering must range over all \(\mathrm{Gr}(p,r)\). This adds only a negligible \(O(\log p)\) overhead to the global-level cost. Accordingly, we construct a data-independent prior in a three-level prior hierarchy: (``global-$r$'') choose the rank \(r\)  (paying the \(\log p\) overhead),  (``global-$\bar\cV$'') choose a reference subspace \(\bar{\mathcal V}\in \mathrm{Gr}(p,r)\), and (``local'') sample within \(\bar{\mathcal V}\) using the local chart prior; see Figure~\ref{fig:prior-diagram} for an illustration. For conceptual clarity, Lemma~\ref{lemma ellipsoidal Grassmannian} focuses on the Grassmannian covering cost at a fixed rank \(r\); and we defer the layer–specific specialization (to each \(d_{l-1}\!\times d_{l-1}\) feature Gram–matrix block) to the calculation in \eqref{eq: global cost}.} \begin{lemma}[Ellipsoidal Covering of the Grassmannian manifold]\label{lemma ellipsoidal Grassmannian} Consider the Grassmannian $\Gr(d,r)$. For uniform prior $\mu=\textup{Unif}(\Gr(d,r))$, we have that for every $\cV\in \Gr(d,r)$,  every $\varepsilon>0$ and every PSD matrix $\Sigma$ with eigenvalues $\lambda_1\geq\cdots\geq \lambda_{d}$, we have the pointwise dimension bound
\begin{align*}
   \log\frac{1}{\mu(B_{\varrho_{\textup{proj},{{\Sigma}}}}(\cV, \varepsilon))}\leq \frac{d-r}{2}\sum_{k=1}^{r}\log \frac{C\max\{\lambda_k,\varepsilon^2\}}{\varepsilon^2}+ \frac{r}{2}\sum_{k=1}^{d-r}\log \frac{C\max\{\lambda_k,\varepsilon^2\}}{\varepsilon^2},
\end{align*}
  where $C>0$ is an absolute constant.  
\end{lemma}
The result above is mathematically significant in its own right. It extends the classical
metric‐entropy (covering number) theory for the Grassmannian—where log covering number $\asymp r(d-r)\log(C/\varepsilon)$
under the \emph{isotropic} projection metric— to an \emph{ellipsoidal} (anisotropic) metric  that
captures feature– and model–induced geometry. This generalization translates the traditional
differential‐geometric and Lie‐algebraic treatments (see Appendix \ref{sec geometry algebra}) and, we believe, illustrates a two–way exchange:
deep mathematical structure is essential to understanding generalization in modern neural networks,
and, conversely, generalization theory can motivate new questions and results in pure mathematics.

Leveraging the block–decomposable structure in \eqref{eq: metric tensor DNN}, the $l$–th block is
\[
G_l(W)
= A_l(W)\;\otimes\; I_{d_l}, \textup{ where } A_l(W)=LM^2_{l\rightarrow L}(W,\varepsilon)\cdot F_{l-1}(W,X)F_{l-1}(W,X)^\top\in \bR^{d_{l-1}\times d_{l-1}}.
\]
Since the Kronecker factor is $I_{d_l}$, the spectrum of $G_l$ consists of the eigenvalues of the
feature Gram matrix $F_{l-1}F_{l-1}^\top$ (scaled by $LM^2_{l\rightarrow L}$), each repeated $d_l$
times.
For readability, in the following display we suppress the radius and precision
parameters in $d_{\mathrm{eff}}$; the precise convention is given in
Theorem~\ref{thm matrix} below.
Consequently, the \emph{local–chart} (within–subspace) covering cost at layer $l$ scales as
\begin{align}\label{eq: local cost}
d_l \cdot d_{\mathrm{eff}}\!\Bigl(LM^2_{l\rightarrow L}(W,\varepsilon)\cdot\;F_{l-1}(W,X)F_{l-1}(W,X)^\top\Bigr),
\end{align}
while the \emph{atlas} (subspace–selection) cost is the Grassmannian term over
\(\Gr\bigl(d_{l-1},\, r_{\mathrm{eff}}[W,l]\bigr)\), where
\(r_{\mathrm{eff}}[W,l]\) is the effective rank of \(A_l(W)\in\bR^{d_{l-1}\times d_{l-1}}\).
By Lemma \ref{lemma ellipsoidal Grassmannian} (and the footnote preceding it), the {\it global-atlas} (choosing-subspace) covering cost at layer $\ell$ scales as 
\begin{align}\label{eq: global cost}
d_{l-1} \cdot d_{\mathrm{eff}}\!\Bigl(LM^2_{l\rightarrow L}(W,\varepsilon)\cdot\;F_{l-1}(W,X)F_{l-1}(W,X)^\top\Bigr) +\log(d_{l-1}).
\end{align}Together, \eqref{eq: local cost} and \eqref{eq: global cost} yield a clean layerwise decomposition: the width \(d_l\) multiplies the spectral
complexity of incoming features (local charts), whereas the input dimension \(d_{l-1}\) governs the
Grassmannian covering (global atlas). This complementary, seemingly magical “duality'' underlies the calculation below. 
\begin{theorem} [Riemannian Dimension for DNN]\label{thm matrix}
Consider the weight space $B_{\tF}(R)$, and a pointwise ellipsoidal metric defined via the ellipsoidal metric   $G_{\textup{NP}}(W)$ defined in \eqref{eq: metric tensor DNN}.  Define the {\it pointwise Riemannian Dimension }
\begin{align*}
d_{\textup{R}}(W,\varepsilon)=\sum_{l=1}^L\Big( \underbrace{d_l\cdot d_{\eff}(A_l(W))}_{\textup{covering local charts}}+\underbrace{d_{l-1}\cdot  d_\eff(A_l(W))}_{\textup{covering global atlas} }+ \underbrace{\log (d_{l-1})}_{\textup{covering discrete }  r_\eff}+\log n\Big),
\end{align*}
where $A_l(W)$ is the feature Gram matrix  $LM^2_{l\rightarrow L}(W,\varepsilon) \cdot F_{l-1}(W,X)F^\top_{l-1}(W,X)$; and $d_\eff(A_l(W))$ is abbreviation of $d_\eff(A_l(W), C\max\{\|W\|_{\tF}, R/2^n\}, \varepsilon)$ with $C>0$ an absolute constant. Then we have the pointwise dimension bound: there exists a prior $\pi$ such that uniformly over all $W\in B_{\tF}(R)$,
\begin{align*}
    \log\frac{1}{\pi(B_{\varrho_n}(f(W,\cdot),\varepsilon))}\leq d_{\textup{R}}(W,\varepsilon).
\end{align*}
\end{theorem}

This concludes our program for fully connected networks: we establish \emph{Riemannian Dimension}
as a principled complexity measure that explains—and sharply bounds—generalization. We summarize
the \emph{second structural principle of fully connected DNN}:
The complexity of the \emph{global atlas} (covering the space of reference top eigenspaces) remains commensurate with the layerwise, spectrum–aware  complexity of covering the \emph{local charts}.   On closer inspection, the effect hinges on the block–decomposable structure in \eqref{eq: metric tensor DNN}. This structure is intrinsic to layered neural networks and typically absent in generic nonlinear models, which helps explain why DNN are particularly amenable to sharp generalization analysis.

 For intuition, we illustrate the construction of the prior $\pi$ in the single-layer case—via the schematic in Figure~\ref{fig:prior-diagram}. From a top-down view, the prior $\pi$ can be generated by first sampling the effective rank $r$, then a subspace $\bar{ \cV}$
 on the Grassmannian, and finally a weight $W$ inside that subspace. The general $L$-layer setting is then obtained by applying the same construction independently to each layer and taking a product measure, which is enabled by the layer-wise decomposable structure of neural networks (a consequence of our non-perturbative analysis).

\section{Generalization Bounds and  Implications}\label{sec generalization DNN}
\subsection{Unconditional Generalization Bound for DNN}\label{subsection gen bound of dnn}

Theorem~\ref{thm generic chaining} is stated in the mixed empirical-population metric. We first apply it to the learned feature geometry, and then give a fully observed-sample bound under a finite-resolution subspace condition. The first bound does not require that condition; the condition is used only to replace population feature information by the observed spectrum.

Write $F_l^S(W):=F_l(W,X_S)$ and define
\[
\Sigma_l^{\mathbb P}(W):=\mathbb E_{X\sim\mathbb P}
 \big[f_l(W,X)f_l(W,X)^\top\big],\qquad
\Gamma_l^{S,\mathbb P}(W):=F_l^S(W)F_l^S(W)^\top+n\Sigma_l^{\mathbb P}(W).
\]
Thus $n^{-1}\Gamma_l^{S,\mathbb P}$ is the feature second-moment matrix for $\mathbb P_S+\mathbb P$. 

Let $d_{\mathrm R}^{S,\mathbb P}(W,\varepsilon)$ be the Riemannian Dimension from Theorem~\ref{thm matrix} with $F_{l-1}F_{l-1}^\top$ replaced by $\Gamma_{l-1}^{S,\mathbb P}(W)$. The outer transfer constants $\bar M_{l\to L}(W,\varepsilon)$ are common bounds validating the non-perturbative expansion in $L_2(\mathbb P_S)$ and $L_2(\mathbb P)$ on the same bounded weight charts used in Theorem~\ref{thm matrix}. Uniform bounds over inputs on these charts are sufficient. These population transfer bounds, like the population feature moments, need not be observable from $S$; the empirical result below uses the same valid bounds.

Combining Theorem~\ref{thm matrix} with Theorem~\ref{thm generic chaining} gives the following result.

\begin{theorem}[Generalization Bound for DNN; mixed empirical-population form]\label{thm dnn}
Assume the finite population feature moments and common transfer bounds specified above. Let the loss $\ell(f(W,x),y)$ be bounded in $[0,1]$ and $\beta$-Lipschitz with respect to $f(W,x)$.  For every $\delta\in(0,1)$, with probability at least $1-\delta$ over the observed sample $S$, uniformly over all $W\in B_{\tF}(R)$,
\begin{align*}
&(\bP-\bPn)\ell(f(W,x),y)\nonumber\\
&\le C_1\left(\beta\inf_{0\le\alpha\le 1/\beta}\left\{\alpha+\frac{1}{\sqrt n}\int_\alpha^{1/\beta}\sqrt{d_{\textup{R}}^{S,\mathbb P}(W,c\varepsilon)}\,d\varepsilon\right\}
+\sqrt{\frac{\log\frac{2\log(2n)}{\delta}}{n}}\right),
\end{align*}
where $c,C_1>0$ are absolute constants and
\begin{align}\label{eq: R D DNN}
d_{\textup{R}}^{S,\mathbb P}(W,\varepsilon)=\frac{1}{2}\sum_{l=1}^L \bigg(&(d_l+d_{l-1}) \sum_{k=1}^{r^{S,\mathbb P}_{\eff}[W, l]}\underbrace{\log\frac{8C_2^2\lambda_k(\Gamma_{l-1}^{S,\mathbb P}(W))}{n\varepsilon^2}}_{\textup{spectrum of inner layers }1{:}l\!-\!1}\nonumber\\
&+(d_l+d_{l-1}) r^{S,\mathbb P}_{\eff}[W, l]\cdot\underbrace{\log\Big(\bar{M}^2_{l\rightarrow L}(W,\varepsilon)  L\max\{||W||_{\tF}^2,R^2/4^n\}\Big)}_{\textup{spectrum of outer layers }l+1:L}+2\log(d_{l-1}n)\bigg),
\end{align}
 with
\[
r^{S,\mathbb P}_{\eff}[W,l]
:=r_{\eff}\!\left(L\bar{M}^2_{l\rightarrow L}(W,\varepsilon)\Gamma_{l-1}^{S,\mathbb P}(W), C_2\max\{||W||_{\tF}, R/2^n\}, \varepsilon\right),
\]
and $C_2>0$ is an absolute constant.
\end{theorem}

\paragraph{Tightness of Each Step and Resulting Theorem.}  We conclude by reviewing our comprehensive theory for generalization in fully connected networks and
justifying the tightness of the resulting bound in Theorem \ref{thm dnn}. 
\begin{enumerate}
    \item \textbf{First}, in Section~\ref{sec pointwise generalization} we develop a framework based on
\emph{pointwise dimension}. 
The upper bounds are complemented by a general empirical-metric lower bound and a worst-case lower bound in the same mixed metric (Theorem~\ref{thm lower bound}), and the framework has a profound connection
to finite-scale geometry—evidence that this is the right organizing principle. 
\item \textbf{Second}, Section~\ref{sec non-perturbative} introduces a \emph{non-perturbative} expansion.
Lemma~\ref{lemma non-perturbative} applies Cauchy-Schwarz layerwise (treating each layer as a block).
While there may be room to improve depth dependence, the telescoping decomposition \eqref{eq: telescoping equality}
 is an exact  \emph{equality}, so the expansion is generally sharp
(and fully avoid linearization). 
\item \textbf{Third}, the hierarchical covering argument shows that the resulting \emph{Riemannian Dimension}
bound matches the gold standard of \emph{effective dimension}. Thus our pointwise, spectrum-aware bounds
achieve the optimal form dictated by static ellipsoid theory, now in  strongly correlated deep networks.
\end{enumerate}

\subsection{Fully Observable Generalization Bound for DNN}\label{subsec fully observable}

Building on Theorem~\ref{thm dnn}, we state a fully empirical bound under a finite-resolution \emph{subspace} condition. We require population-to-empirical comparison only on the observed active eigenspaces and control the population energy on their inactive complement at the same finite resolution. The condition is stated separately on the observed active subspace and its inactive complement at each scale.

For the observed-sample theorem, write
\begin{align}\label{eq empirical observed gram main}
\Gamma_l^S(W):=F_l^S(W)F_l^S(W)^\top,\qquad l=0,\ldots,L-1,
\end{align}
and let $d_{\textup{R}}^S(W,\varepsilon)$ denote the observed-sample Riemannian Dimension obtained from Theorem~\ref{thm matrix} after replacing each $F_{l-1}(W,X)F_{l-1}(W,X)^\top$ by $\Gamma_{l-1}^S(W)$.  Equivalently, it has the same explicit $FF^\top$ eigenvalue form as the mixed quantity in Theorem~\ref{thm dnn}:
\begin{align}\label{eq empirical feature iso RD explicit}
 d_{\textup{R}}^S(W,\varepsilon)
 :=\frac{1}{2}\sum_{l=1}^L \bigg(& (d_l+d_{l-1})
 \sum_{k=1}^{r_{\eff}^S[W,l]}
 \underbrace{\log\frac{8C_2^2\lambda_k\!\big(\Gamma_{l-1}^S(W)\big)}{n\varepsilon^2}}_{\textup{observed spectrum of inner layers }1{:}l\! -\!1}
 \nonumber\\
& +(d_l+d_{l-1})r_{\eff}^S[W,l]
 \cdot \underbrace{\log\!\Big(\bar{M}_{l\to L}^2(W,\varepsilon)\,L\max\{\|W\|_{\tF}^2,R^2/4^n\}\Big)}_{\textup{spectrum of outer layers }l+1{:}L}+2\log(d_{l-1}n)\bigg),
\end{align}
where
\begin{align*}
r_{\eff}^S[W,l]
:=r_{\eff}\!\left(
L\bar{M}_{l\to L}^2(W,\varepsilon)\Gamma_{l-1}^S(W),\,
C_2\max\{\|W\|_{\tF},R/2^n\},\,\varepsilon
\right),
\end{align*}
and $C_2>0$ is an absolute constant.

For later use, define the finite spectral resolution associated with feature layer $j$ by
\begin{align}\label{eq subspace iso threshold main}
R_W:=C_2\max\{\|W\|_{\tF},R/2^n\},\qquad
\vartheta_j(W,\varepsilon)
:=\frac{n\varepsilon^2}{2L\bar M_{j+1\to L}^2(W,\varepsilon)R_W^2},
\qquad j=0,\ldots,L-1 .
\end{align}
The resolution \(\vartheta_j(W,\varepsilon)\) is the finite spectral resolution
that determines which observed feature directions contribute to the effective
rank at the next layer. Indeed, by the definition of \(r_{\mathrm{eff}}^S[W,j+1]\),
\[
\lambda_k\!\left(L\bar M_{j+1\to L}^2(W,\varepsilon)\Gamma_j^S(W)\right)R_W^2
\ge \frac{n\varepsilon^2}{2}
\quad\Longleftrightarrow\quad
\lambda_k(\Gamma_j^S(W))\ge \vartheta_j(W,\varepsilon).
\]
Thus the eigendirections of \(\Gamma_j^S(W)\) with eigenvalues at least
\(\vartheta_j(W,\varepsilon)\) are precisely the observed feature directions
counted by \(r_{\mathrm{eff}}^S[W,j+1]\). Let \(P_j^S(W,\varepsilon)\) be the spectral projector of
\(\Gamma_j^S(W)\) onto the span of those eigendirections; we call its image  the
observed active subspace. Set
\[
    Q_j^S(W,\varepsilon):=I_{d_j}-P_j^S(W,\varepsilon),
\]
whose image is the corresponding inactive complement.

\begin{definition}[Layer-wise finite-resolution subspace isomorphism]\label{def feature isomorphism}
Fix constants $\kappa\ge1$ and $b_{\mathrm{sub}}\ge1$. We say that the observed sample $S$ satisfies the layer-wise finite-resolution population subspace isomorphism if the following two estimates hold simultaneously for every feature layer $j=0,\ldots,L-1$, every center $W\in B_{\tF}(R)$, and every scale $\varepsilon\in(0,1/\beta]$:
\begin{align}
P_j^S(W,\varepsilon)\big(n\Sigma_j^{\mathbb P}(W)\big)P_j^S(W,\varepsilon)
&\preceq
\kappa\,P_j^S(W,\varepsilon)\Gamma_j^S(W)P_j^S(W,\varepsilon),
\label{eq main feature isomorphism active}\\
\big\|Q_j^S(W,\varepsilon)\big(n\Sigma_j^{\mathbb P}(W)\big)Q_j^S(W,\varepsilon)\big\|_{\op}
&\le b_{\mathrm{sub}}\,\vartheta_j(W,\varepsilon).
\label{eq main feature isomorphism inactive}
\end{align}
\end{definition}

The first display is a multiplicative isomorphism only on the observed active subspace.  The second display says that the population features may have energy outside the observed active subspace, but only below the finite resolution already discarded by the effective-rank truncation. This subspace isomorphism implies finite-scale domination of the mixed empirical-population metric by the empirical metric in the pointwise dimension, yielding the following theorem based only on the observed sample. Moreover, the squared ellipsoidal projection metric in \eqref{eq main feature isomorphism inactive} is precisely the metric used to define the finite-scale resolution in \eqref{eq: precision main} of Lemma~\ref{lemma approximate subspace} and Lemma~\ref{lemma ellipsoidal Grassmannian}, at the same order in $\varepsilon$. It provides a sufficient finite-resolution condition for the empirical bound. We also note that $(\kappa,b_{\textup{sub}})$ enter the Riemannian Dimension below only through optimistic logarithmic factors, so their effect can be mild under fast spectral decay.

\begin{theorem}[Fully empirical DNN bound under subspace isomorphism]\label{thm empirical dnn feature iso}
Assume the setting of Theorem~\ref{thm dnn}, with valid $\bar M_{l\to L}$ bounds available from $S$ and the fixed network constraints.  Suppose that, with probability at least $1-\delta_{\mathrm{iso}}$ over $S$, the layer-wise finite-resolution population subspace isomorphism in Definition~\ref{def feature isomorphism} holds.  Then there is a constant $c_{\mathrm{sub}}>0$, depending only on $(\kappa,b_{\mathrm{sub}})$, such that for every $\delta\in(0,1)$, with probability at least $1-\delta-\delta_{\mathrm{iso}}$ over $S$, uniformly over all $W\in B_{\tF}(R)$,
\begin{align}\label{eq empirical feature iso bound main}
&(\bP-\bPn)\ell(f(W,x),y)\nonumber\\
&\le C\Bigg(\beta\inf_{0\le\alpha\le 1/\beta}\left\{\alpha+\frac1{\sqrt n}\int_\alpha^{1/\beta}\sqrt{d_{\textup{R}}^S(W,c_{\mathrm{sub}}\varepsilon)}\,d\varepsilon\right\}
+\sqrt{\frac{\log(2\log(2n)/\delta)}{n}}\Bigg),
\end{align}
where $C>0$ is an absolute constant.  In particular, if \(\kappa,b_{\mathrm{sub}}=O(1)\), then 
\(c_{\mathrm{sub}}\) is an absolute constant. \end{theorem}

\paragraph{Open problem: subspace-level feature regularity.}
The preceding certificate leads to the following concrete open question.
\begin{center}
\begin{minipage}{0.94\linewidth}
\itshape
Can Definition~\ref{def feature isomorphism} be further reduced from a uniform-in-subspace condition to subspace-level tail regularity of the learned features, by leveraging the approximate pointwise linearity and subspace-decomposition structure of deep neural networks?
\end{minipage}
\end{center}
This question isolates the remaining gap between an unconditional mixed empirical-population compressed theorem and a fully observed-sample compressed theorem.  It is not a limitation of the pointwise-compression principle; rather, it is the precise observability problem one must solve to certify, from the training sample alone, the feature subspaces that the unconditional theory already identifies. We defer further discussion of this open problem, including a concrete route based on the two-layer sub-Gaussian/small-ball special case, to Appendix~\ref{subsec feature regularity}.

\paragraph{Interpreting Theorem~\ref{thm empirical dnn feature iso} to the Informal Rate \eqref{eq: intro bound}.}
In the expression for the observed Riemannian Dimension in \eqref{eq empirical feature iso RD explicit}, the quantity $r_{\mathrm{eff}}^S[W,l]$ incorporates local Lipschitz factors.  Specifically, the effective rank is computed for
\[
L\bar{M}_{l\rightarrow L}^2(W,\varepsilon)\,F_{l-1}(W,X_S)F_{l-1}(W,X_S)^\top
\]
rather than $F_{l-1}F_{l-1}^\top$ alone.  When $F_{l-1}F_{l-1}^\top$ has rapidly decaying eigenvalues this dependence is strongly suppressed, and it disappears entirely under strict low rank.  Consequently, under mild low-rank or spectral-decay conditions, the bound aligns with the informal rate \eqref{eq: intro bound}.  For each layer $l$, the first term in \eqref{eq empirical feature iso RD explicit} quantifies the contribution of the inner layers $1{:}(l-1)$ through the observed feature Gram, while the second term captures the influence of the outer layers $(l+1){:}L$ through $\bar M_{l\to L}$.  Together, these terms provide a complete layerwise account of the effective dimension in the informal rate \eqref{eq: intro bound}.

\subsection{Comparison with Norm Bounds, VC, and NTK}\label{subsec comparison}
We compare our generalization bound for fully connected DNN (Theorem~\ref{thm dnn}) with three established lines of work: (i) bounds based on products of spectral norms, (ii) VC–dimension–type capacity bounds, and (iii) Neural Tangent Kernel (NTK) linearizations that are valid only in an infinitesimal neighborhood of initialization. 
Our framework yields {\it exponentially} tighter rates than norm–product bounds, refines VC–type statements into hypothesis– and data–dependent guarantees, and replaces infinitesimal linearization with a finite-scale, non-perturbative analysis that holds simultaneously for every trained hypothesis. For space, we defer the recovery of representative norm bounds to Appendix~\ref{subsec norm bound} and a broader literature review to Appendix~\ref{subsec related works}.

\paragraph{Norm Bounds:}
The trace relaxation of Theorem~\ref{thm dnn} does not use Definition~\ref{def feature isomorphism}. Write $R_n(W):=\max\{\|W\|_{\tF},R/2^n\}$. The inequality $\log x\le\log(1+x)\le x$ gives, for each layer,
\[
\sum_{k=1}^{r_{\mathrm{eff}}^{S,\mathbb P}[W,l]}
\log\!\left(\frac{C\lambda_k(\Gamma_{l-1}^{S,\mathbb P}(W))L\bar M_{l\to L}^2(W,\varepsilon)R_n(W)^2}{n\varepsilon^2}\right)
\le \frac{C\operatorname{Tr}(\Gamma_{l-1}^{S,\mathbb P}(W))L\bar M_{l\to L}^2(W,\varepsilon)R_n(W)^2}{n\varepsilon^2}.
\]
Here
\[
\operatorname{Tr}(\Gamma_{l-1}^{S,\mathbb P}(W))
=\|F_{l-1}^S(W)\|_{\tF}^2+n\mathbb E\|f_{l-1}(W,X)\|_2^2.
\]
When the activations are $1$-Lipschitz and satisfy $\sigma_l(0)=0$, the inner feature norms are bounded by the corresponding products of spectral norms. If the common outer transfer bounds also satisfy $\bar M_{l\to L}(W,\varepsilon)\le C\prod_{i>l}\|W_i\|_{\op}$ on the relevant weight charts, uniformly over the scales used, the rank-free argument in Appendix~\ref{subsec norm bound} yields
\begin{align}\label{eq: norm main}
(\mathbb P-\mathbb P_n)\ell(f(W,x),y)
\le\widetilde O\!\left(
\frac{\beta\|W\|_{\tF}}{n}
\sqrt{\|X_S\|_{\tF}^2+n\mathbb E\|X\|_2^2}
\sqrt{L\sum_{l=1}^L(d_l+d_{l-1})\prod_{i\ne l}\|W_i\|_{\op}^2}
\right).
\end{align}
As in Corollary~\ref{coro worst case}, the display suppresses the confidence term and the stated logarithmic and discretization terms. If $\|X\|_2\le B_x$ almost surely, the input factor is at most $\sqrt{2n}B_x$, giving the usual $n^{-1/2}$ scaling. Retaining the feature spectra and logarithmic ellipsoidal cost can be exponentially sharper than this representative spectral-norm relaxation. Appendix~\ref{subsec norm bound} derives this relaxation directly from Theorem~\ref{thm dnn} and gives an explicit exponential separation.

\paragraph{VC Dimension:} 
Let $L$ be the number of layers and $P=\sum_{l=1}^L d_l d_{l-1}$ be the total number of weights, \citet{bartlett2019nearly} prove a nearly tight VC–dimension bound $\mathrm{VCdim} \le O\, (P L \log P)$, supported by a lower bound $\mathrm{VCdim} \ge \Omega\, (P L \log (P/L))$. This VC dimension bound is roughly equivalent to be $L\sum_{l=1}^L d_ld_{l-1} $.\footnote{The extra factor $L$ beyond parameter count in VCdim is essentially unavoidable: for nonlinear compositional models, VC/packing dimensions depend on the logarithm of a global worst-case Lipschitz constant, and in depth$-L$ networks that constant grows multiplicatively across layers, yielding an additional linear dependence on $L$.} 

Our Riemannian Dimension bound, by contrast, substantially sharpens this rate: it removes the explicit dependence on depth $L$ and replaces the crude width factor with a (layerwise) effective‐rank term.

\paragraph{Neural Tangent Kernel (NTK):} 
Our approach uses an exact, non-perturbative expansion that preserves the finite-scale geometry of deep networks, thereby going beyond NTK-based Taylor linearizations. A major bottleneck of the NTK approach is that the Taylor linearization remains valid only in an infinitesimal neighborhood of initialization, or equivalently in the infinite-width ``lazy'' regime~\citep{jacot2018neural,arora2019exact}. Outside this regime, the NTK approximation typically breaks down, limiting its explanatory power for practical networks.

From a generalization perspective, this initialization-centric and infinitesimal view suppresses the feature learning that drives generalization in modern deep networks, and therefore cannot fully explain their empirical behavior. In contrast, our results provide a finite-scale, pointwise theory that operates directly in practical regimes and explicitly captures feature learning through the spectra of the \emph{learned} feature matrices.

\subsection{Algorithmic Implications and Excess Risk Bound}\label{subsec implicit bias}
\paragraph{Pointwise Dimension as Regularization and Excess Risk Bound.}
 Our bounds imply a natural regularization strategy for algorithm design.  
Given the pointwise generalization inequality \eqref{eq: generalization gap} (e.g., the Riemannian Dimension bound in Theorem \ref{thm dnn}), we consider a regularized ERM objective that explicitly minimizes this complexity measure:
\begin{align}\label{eq: regularized ERM}
\hat{f} = \arg\min_{f\in\cF} \left\{\mathbb{P}_n \ell(f;z) + C\sqrt{\frac{d(f) + \log (2/\delta)}{n}}\right\}.
\end{align}With probability at least $1-\delta$,
its excess risk  is bounded by  (compared to any benchmark $f^\star\in \cF$):
\begin{align}\label{eq: excess risk}
&\mathbb{P}\ell(\hat{f};z)-\mathbb{P}\ell(f^\star;z)\nonumber\\\leq 
    &\inf_{f \in \mathcal{F}}\left\{\mathbb{P}_n\ell(f;z)+ C\sqrt{\frac{d (f)+ \log (2/\delta)}{n}}\right\}  -\mathbb{P}\ell(f^\star;z) \\\le     &(C+\sqrt{1/2})\sqrt{\frac{d (f^\star)+ \log (2/\delta)}{n}};\nonumber
\end{align}
see Appendix~\ref{appendix implicit bias} for full proof.  
Thus we obtain a problem–dependent oracle bound of order
$\sqrt{d(f^\star)/n}$ that adapts to the optimal hypothesis $f^\star$.

\paragraph{From Explicit Regularization to Implicit Bias of Practical Algorithms.} 
Since modern optimizers like SGD routinely drive empirical risk to near-zero, convergence analysis alone offers limited insight into generalization. The central theoretical challenge is therefore not determining \emph{whether} a minimum is reached, but identifying \emph{which} of the infinite interpolating solutions the optimizer selects. Plain ERM is insufficient for this task: without constraints on pointwise dimension, an empirical risk minimizer yields no guarantee of controlled excess risk. In contrast, our RD–regularized objective \eqref{eq: regularized ERM} explicitly enforces the low-complexity structure required for the generalization bound in \eqref{eq: excess risk}. Although this intuition is rooted in the earliest practices of deep learning, our pointwise theory rigorously articulates the underlying mathematical reasoning.

This motivates a concrete agenda for optimization in deep learning: characterize algorithms whose implicit bias drives iterates toward solutions with \emph{low pointwise complexity}, in particular low \emph{Riemannian Dimension (RD)}. Analogous phenomena are well documented in linear and kernel settings: gradient descent converges to max–margin (logistic loss) or minimum–norm (least squares) solutions \citep{soudry2018implicit,gunasekar2018characterizing}, iterate–averaged SGD behaves like ridge regression \citep{neu2018iterate}, and “ridgeless’’ kernel regression can generalize with an optimally zero ridge parameter \citep{liang2020ridgeless}; see \citet{vardi2023implicit} for a survey. Our regularizer in \eqref{eq: regularized ERM}, based on pointwise dimension and, in particular, the RD from Theorem~\ref{thm dnn}, is strictly more informative than any single norm, making it a natural target for such analyses.

Empirically, we say an algorithm exhibits \emph{Riemannian–Dimension implicit bias} if it preferentially returns solutions with small RD despite RD’s large dynamic range; in Section~\ref{subsec sgd implicit} we observe that SGD  indeed finds low-RD solutions.

\section{Experiments}\label{section: experiments}
We evaluate the one-shot $L_2$ relaxation of the Riemannian Dimension bound in Theorem~\ref{thm empirical dnn feature iso} on two standard architectures—Fully Connected Networks (FCNs) and ResNets, using two benchmark datasets—MNIST \citep{lecun1998gradient} and CIFAR-10 \citep{krizhevsky2009learning}, respectively. We consider a $9$-hidden-layer FCN architecture, where, except for the fixed layers, hidden layers share a common width $h$, with $h \in \{2^{6},2^{7},2^{8},2^{9},2^{10},2^{11},2^{12}\}$. Increasing $h$ monotonically enlarges both layer widths and model sizes. We adopt canonical ResNet architectures—ResNet-$\{20,32,44,56,74,110\}$—which differ only in the number of residual blocks per stage while maintaining the same overall architecture (three-stage, basic-block design) as introduced by \citep{he2016deep}. These ResNet architectures provides a clean capacity sweep via depth. In what follows, we organize experiments around the two complementary regimes—width scaling on FCNs and depth scaling on ResNets. 

This design lets us systematically study three central questions in modern deep learning: (i) why does overparameterization often improve generalization? (ii) how does feature learning evolve during training? and (iii) what implicit regularization is encoded by the baseline optimizer? Detailed experimental setups are deferred to Appendix \ref{appendix experimental setup}. The code to reproduce all experiments is available at our GitHub repository \href{https://github.com/Learning-Theory-Enthusiast/Pointwise-Generalization-in-Deep-Neural-Networks}{\textcolor{magenta}{here}}.

In particular, we use the spectral-norm product $\prod_{i>l}\|W_i\|_{\mathrm{op}}$ as a numerical surrogate for the local Lipschitz constant $\bar{M}_{l\to L}(W,\varepsilon)$; state-of-the-art formal-verification toolchains \citep{shi2022efficiently} can compute local Lipschitz constants much more sharply—with well-developed packages and rigorous numerical guarantees—than this spectral-product surrogate, and could therefore further strengthen all our empirical results (an active research area). This substitution gives a certified upper bound when the spectral product dominates the transfer constants throughout the relevant weight charts, as required in Corollary~\ref{coro worst case}; the pointwise product alone does not establish this condition.

\subsection{Riemannian Dimension Explains Overparameterization}
This section studies why does overparameterization—despite exploding model capacity—often improve generalization. We investigate this paradox by tracking our Riemannian Dimension across models with varying parameter counts, asking whether more parameters truly enlarge capacity or instead deduce complexity.

\begin{table}[t]
\centering
\caption{Final‐epoch Metrics of FCNs on MNIST. Column descriptions are as follows: 1) Width$-2^\star$ denotes a hidden width of $h = 2^\star$; 2) Train: training error; 3) Gen: generalization gap, defined as test error minus training error; 4) Spectral Norm: the spectrally normalized margin bound of \citep{bartlett2017spectrally}, a tightness norm–based bound in the literature to our knowledge; we computed it with margin normalization; 5) \# Parameters: parameter counts of the network; 6) VC dimension: we adopt a nearly tight VC–dimension bound from \citep{bartlett2019nearly} and report $P L \log P$ for brevity (see Section \ref{subsec comparison}); 7) R-D: the proposed Riemannian Dimension.}
\label{tab:fcn_mnist_1_transposed}
\begin{tabular}{lcccccc}
\toprule
Model  & Train  & Gen  & Spectral Norm & \# Parameters & VC dimension & R-D \\
\midrule
Width-$2^6$    & 0.0002 & 0.0205 & $3.146 \times 10^{15}$ & $5.961 \times 10^{6}$ & $9.299 \times 10^{8}$ & $6.433 \times 10^{7}$ \\
Width-$2^7$    & 0.0002 & 0.0187 & $2.695 \times 10^{15}$ & $6.167 \times 10^{6}$ & $9.641 \times 10^{8}$ & $6.097 \times 10^{7}$ \\
Width-$2^8$  & 0.0000 & 0.0191 & $2.093 \times 10^{15}$ & $6.726 \times 10^{6}$ & $1.057 \times 10^{9}$ & $5.589 \times 10^{7}$ \\
Width-$2^9$   & 0.0000 & 0.0186 & $2.401 \times 10^{15}$ & $8.434 \times 10^{6}$ & $1.345 \times 10^{9}$ & $5.316 \times 10^{7}$ \\
Width-$2^{10}$ & 0.0000 & 0.0215 & $4.816 \times 10^{15}$ & $1.421 \times 10^{7}$ & $2.340 \times 10^{9}$ & $5.266 \times 10^{7}$ \\
Width-$2^{11}$ & 0.0000 & 0.0160 & $1.001 \times 10^{16}$ & $3.520 \times 10^{7}$ & $6.116 \times 10^{9}$ & $4.972 \times 10^{7}$ \\
Width-$2^{12}$ & 0.0000 & 0.0210 & $1.466 \times 10^{16}$ & $1.149 \times 10^{8}$ & $2.133 \times 10^{10}$ & $4.803 \times 10^{7}$ \\
\bottomrule
\end{tabular}
\end{table}
\begin{table}[t]
\centering
\caption{Final‐Epoch Metrics of ResNets on CIFAR-10}
\label{tab:cifar101}
\begin{tabular}{lccccc}
\toprule
Model & Train Error & Gen Gap & \# Parameters & VC dimension & R-D \\
\midrule
ResNet-20  & 0.0016 & 0.0752 &  $2.690\times 10^{5}$    & $6.727\times 10^{7}$  & $8.801\times 10^{6}$ \\
ResNet-32  & 0.0003 & 0.0695 &  $4.630\times 10^{5}$     & $1.933\times 10^{8}$  & $9.992\times 10^{6}$ \\
ResNet-44  & 0.0001 & 0.0627 &  $6.570\times 10^{5}$     & $3.872\times 10^{8}$  & $6.339\times 10^{6}$ \\
ResNet-56  & 0.0000 & 0.0637 &  $8.510 \times 10^{5}$    & $6.507\times 10^{8}$  & $5.200\times 10^{6}$ \\
ResNet-74  & 0.0000 & 0.0615 & $1.142\times 10^{6}$   & $1.179\times 10^{9}$  & $3.237\times 10^{6}$ \\
ResNet-110 & 0.0000 & 0.0576 &  $1.724\times 10^{6}$   & $2.723\times 10^{9}$  & $2.583\times 10^{6}$ \\
\bottomrule
\end{tabular}
\end{table}
\begin{table}[t]
\centering
\caption{Final‐epoch Effective Ranks for FCNs on MNIST, where Width$-2^\star$ means $h = 2^\star$, and where  for the form A/B, A
represents the effective rank and B represents the original dimension, and where Layer-1 means the input layer.}
\label{tab:fcn_mnist_rank_1}
\begin{tabular}{lccccccc}
\toprule
Metric   & Width-$2^6$ & Width-$2^7$ & Width-$2^8$ & Width-$2^9$  &    Width-$2^{10}$ & Width-$2^{11}$ & Width-$2^{12}$ \\
\midrule
 Layer-1   & 713/763      & 712/763      & 710/763    & 710/763    & 707/763     &   707/763                 &     704/763        \\
Layer-2    & 2048/2048      & 2044/2048      & 2042/2048    & 2048/2048    &  2047/2048    &   2048/2048                &    2048/2048     \\
Layer-3    & 2048/2048       & 2045/2048      & 2037/2048    &2019/2048  & 1925/2048    &   1460/2048                &    1009/2048        \\
Layer-4    & 61/64      & 97/128   & 92/256   & 85/512   & 79/1024   & 79/2048             &   59/4096       \\
Layer-5  & 23/64     & 43/128    & 34/256  & 33/512    & 28/1024  & 26/2048              &    22/4096   \\
Layer-6  & 20/64       & 24/128     & 20/256   &   21/512  &  19/1024  &        18/2048      &   15/4096    \\
Layer-7   & 15/64      & 18/128     & 17/256   & 15/512    & 15/1024  & 14/2048        &    13/4096     \\
Layer-8   & 15/64      & 14/128    & 15/256  & 11/512   & 13/1024 & 13/2048   &   12/4096     \\
Layer-9   & 14/64      & 14/128     & 15/256   & 13/512   & 13/1024   & 12/2048        &    12/4096    \\
Layer-10  & 13/64      & 13/128   & 12/256  & 14/512   & 12/1024   &13/2048   &  14/4096    \\
Total  & 4970     & 5024     & 4994   & 4969    & 4858  & 4390          &   3908    \\
\bottomrule
\end{tabular}
\end{table}

\begin{table}[t]
\centering
\caption{Final‐epoch Effective Ranks for ResNets on CIFAR-10, where for the form A/B, A represents the effective rank and B represents the original dimension, and where Layer-$0\%$ means the input layer.}
\label{tab:cifar10_layer_evolution}
\begin{tabular}{lccccccc}
\toprule
Metric  & ResNet-20 & ResNet-32 & ResNet-44 & ResNet-56 & ResNet-74 & ResNet-110 \\
\midrule
Layer-$0\%$             & 384/3072  & 384/3072  & 17/3072  & 0/3072   & 0/3072  & 0/3072   \\
Layer-$25\%$           & 2048/16384  & 2048/16384   & 7/16384   & 1/16384   & 0/16384   & 0/16384    \\
Layer-$50\%$     & 1024/8192   & 1024/8192   & 1024/8192   & 227/8192  & 0/8192   & 0/8192    \\
Layer-$75\%$     & 512/4096   & 512/4096  & 512/4096  & 512/4096  & 58/4096  & 0/4096    \\
Layer-$100\%$      & 8/64 & 8/64 & 8/64 & 8/64 & 8/64 & 8/64 \\
Total         & 23432 & 37768 & 27564 & 16294 & 11401 & 6925  \\
\bottomrule
\end{tabular}
\end{table}
Final-epoch metrics of FCNs on MNIST and ResNets on CIFAR-10 are reported in Table~\ref{tab:fcn_mnist_1_transposed} and Table \ref{tab:cifar101}, respectively. In these Tables, the train error quickly collapses to zero for sufficiently large models, confirming their expressive capacity. Consistently, the generalization can continue to be improved as parameters increase, especially on ResNets (Table~\ref{tab:cifar101}). This phenomenon means the overfitting does not appear and reflects a paradoxical truth of deep learning: over-parameterization is not a curse, but can benefit the generalization. However, classical complexity measures—e.g., the spectral norm and the VC dimension, often scale exponentially as the parameter count grows. Notably, the spectral norm is about $10^6$ times larger than the VC dimension and seems to perform worse (see Table~\ref{tab:fcn_mnist_1_transposed}). The two measures therefore struggle to explain the generalization of modern overparameterized networks. In contrast, Riemannian Dimension exhibits a consistent downward trend as model size grows—both under width scaling (last column of Table~\ref{tab:fcn_mnist_1_transposed}) and depth scaling (last column of Table~\ref{tab:cifar101}), and it is about $10^3$ times smaller than the VC dimension, suggesting that the effective dimension—not raw parameter count—is the more informative indicator of generalization in deep learning. In summary, increased parameterization is associated with reduced intrinsic model complexity, and Riemannian Dimension captures this phenomenon.
\subsection{Feature Learning Compresses Effective Rank}
\begin{figure*}[t]
  \centering
  \includegraphics[width=0.47\textwidth]{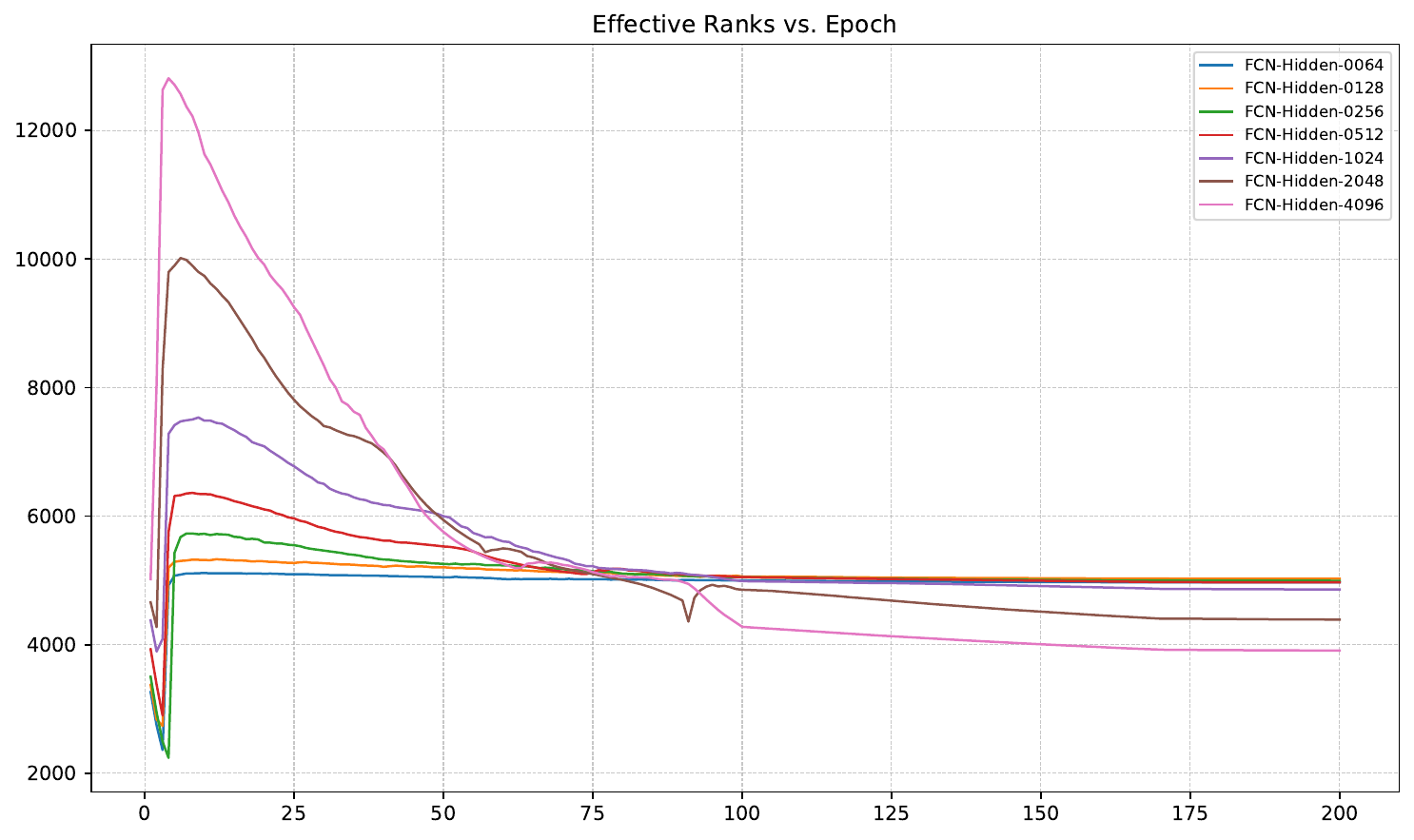}%
   \includegraphics[width=0.47\textwidth]{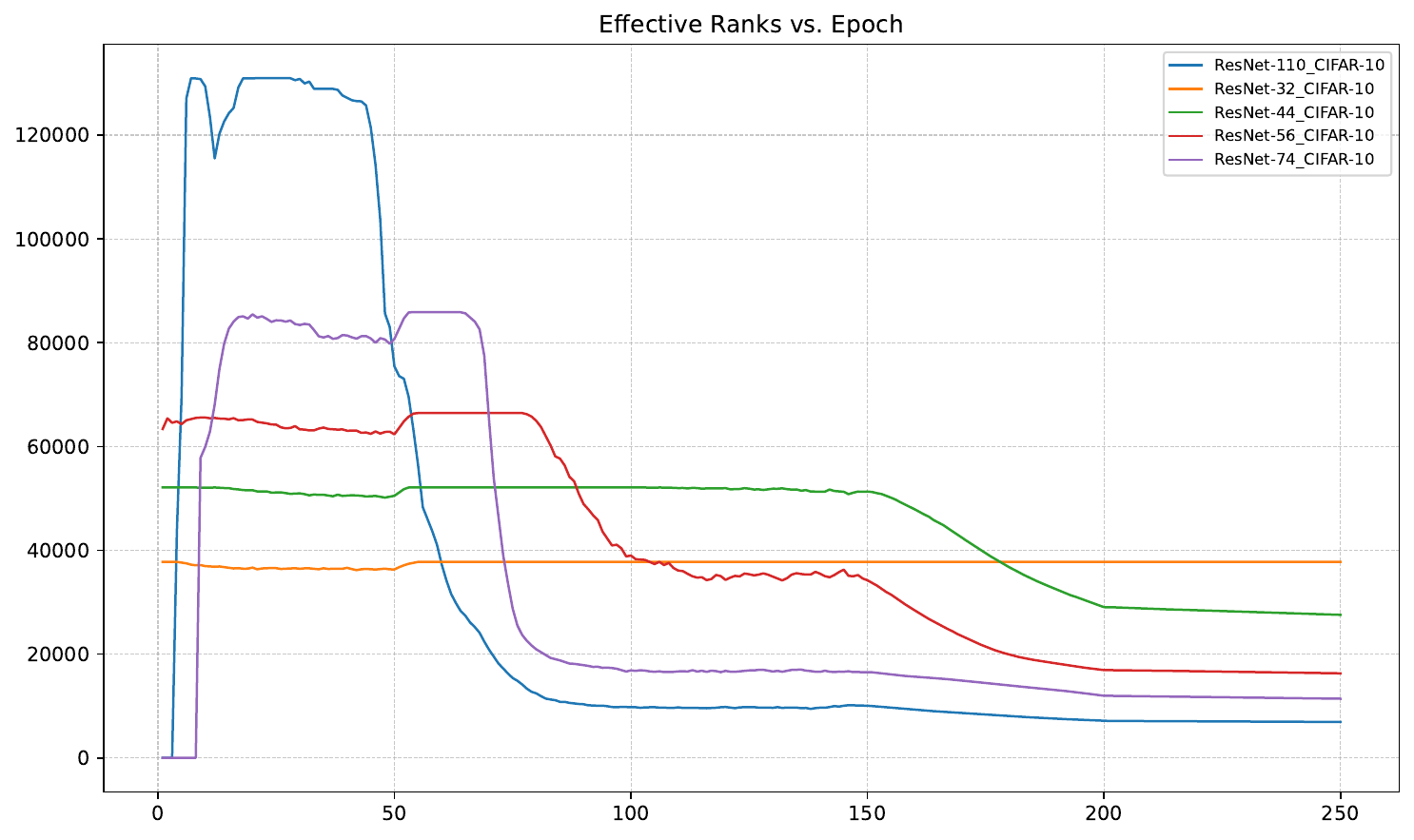}%
    \caption{%
  Effective Rank evolutions of FCNs on MNIST (left) and  ResNets on CIFAR-10 (right) across the training}
  \label{fig:all_in_one_effective_rank}
\end{figure*}

\begin{figure*}[t]
  \centering
  \includegraphics[width=0.47\textwidth]{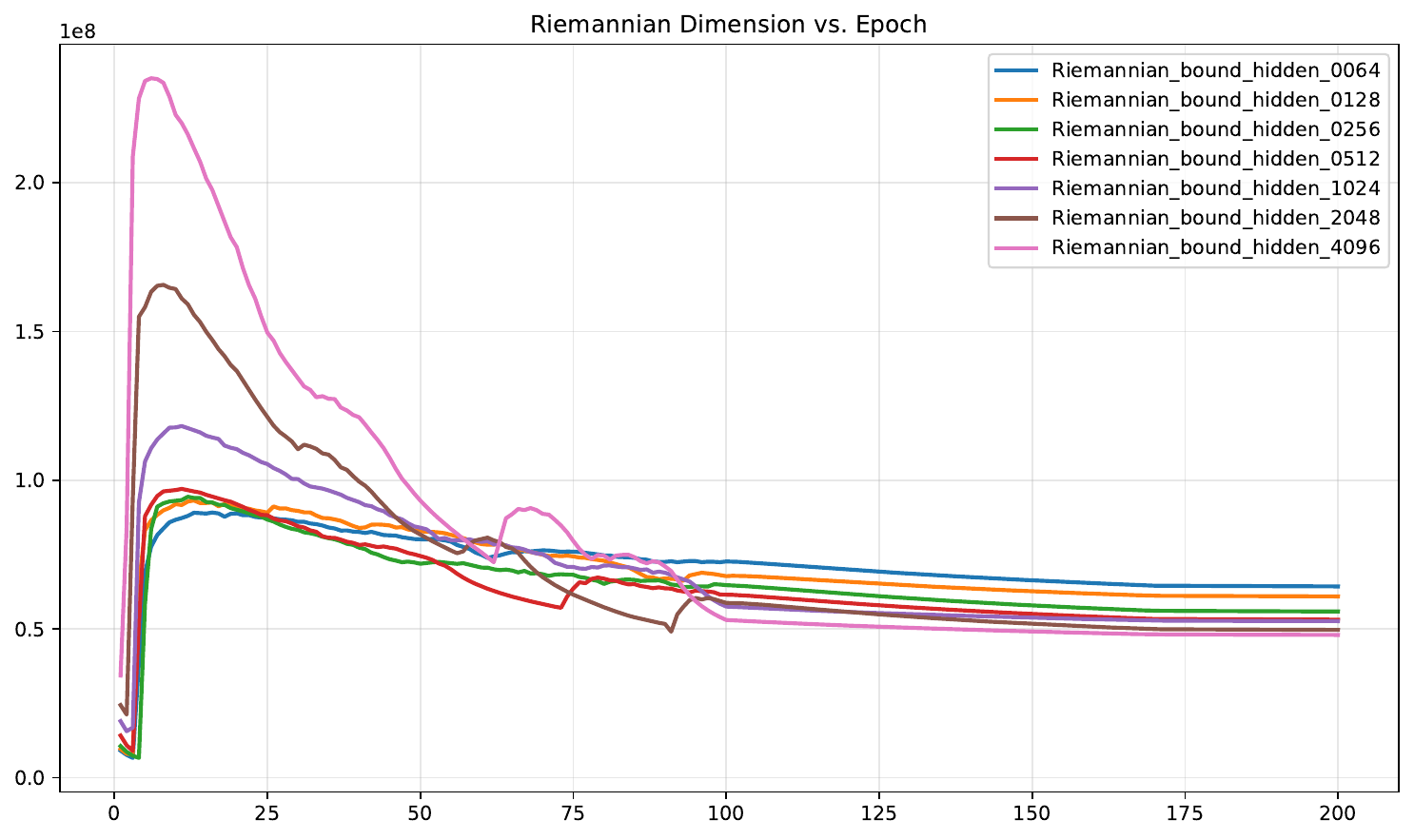}%
    \includegraphics[width=0.47\textwidth]{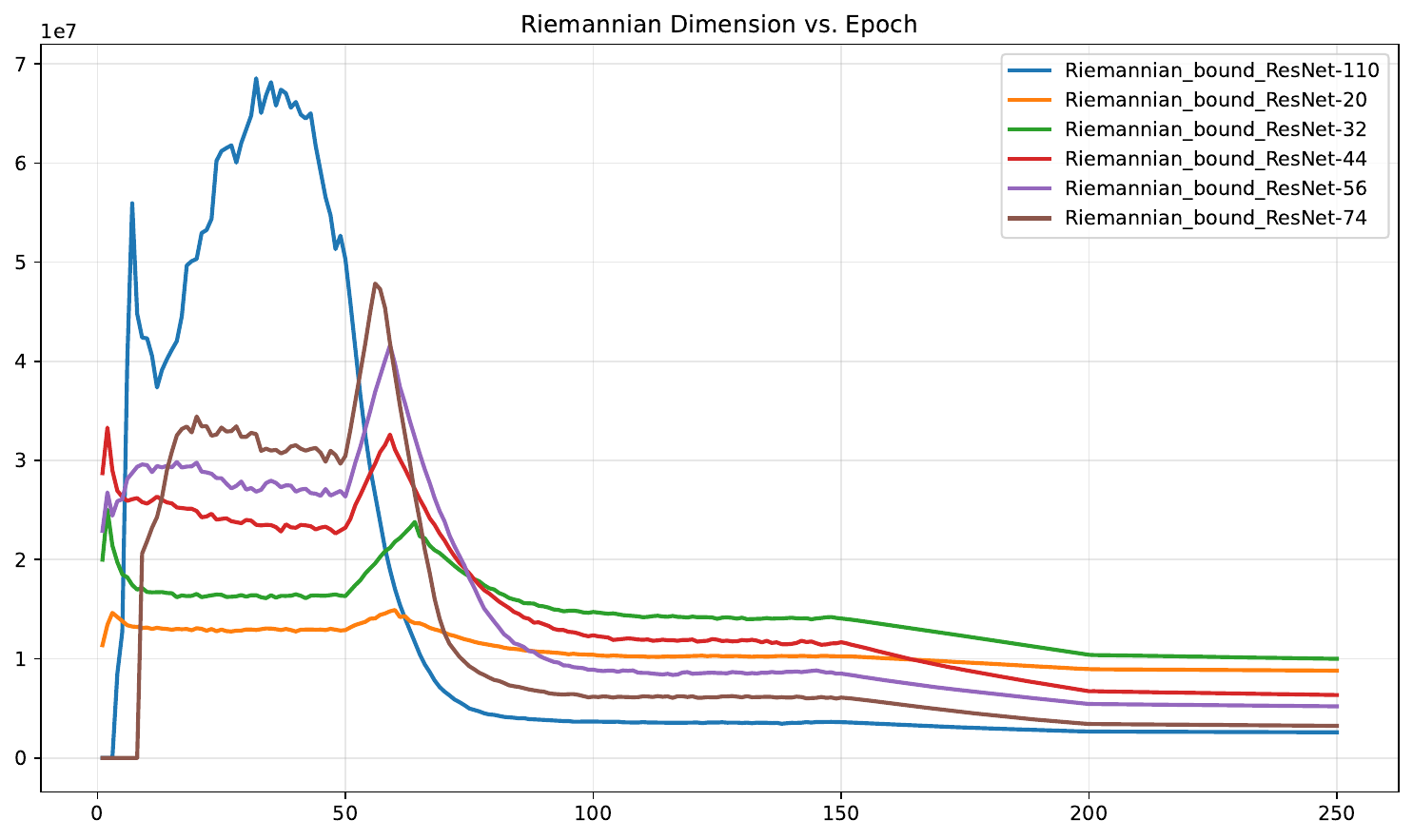}%
    \caption{%
  Riemannian Dimension evolutions of FCNs on MNIST (left) and ResNets on CIFAR-10 (right) across the training
  }
  \label{fig:bound_evolution}
\end{figure*}
We investigate the dynamics of feature learning by monitoring the effective rank of the feature Gram matrices $F_{l-1}F_{l-1}^\top$ scaled by $  L ||W||_{\tF}^2  \prod_{i>l} \|W_i\|_{\textup{op}}^2$ (i.e., $F_{l-1}F_{l-1}^\top \cdot L ||W||_{\tF}^2  \prod_{i>l} \|W_i\|_{\textup{op}}^2$), as dictated by the theory.  We report our empirical results in Tables~\ref{tab:fcn_mnist_rank_1}, \ref{tab:cifar10_layer_evolution} and Figure~\ref{fig:all_in_one_effective_rank}.

Experimental results reveal some clear patterns. First, as training proceeds, the effective ranks of feature Gram matrices decrease sharply after a short transient phase; refer to Figure \ref{fig:all_in_one_effective_rank}. Second, increasing the parameter count, either by width scaling in FCNs or by depth scaling in ResNets, accelerates and strengthens this effective-rank compression; refer to Figure \ref{fig:all_in_one_effective_rank}. Third, on the largest FCN, the degree of effective rank compression can reach as much as $1/300$, which explains the significant improvement of the Riemannian Dimension relative to the VC dimension; refer to Table \ref{tab:fcn_mnist_rank_1}. On the largest ResNet, the effective ranks of the vast majority of layers compress to zero, thereby explaining why deeper networks can exhibit a smaller Riemannian Dimension; refer to Table \ref{tab:cifar10_layer_evolution}. Together, these experimental results suggest that feature learning progressively reduces the intrinsic dimensionality of learned representations, and that overparameterization intensifies this dimension-reduction effect.

\subsection{SGD Finds Low Riemannian Dimension Point}\label{subsec sgd implicit}
Prior work has shown that various norms are implicit bias of optimizers, but typically limited to linear models \citep{vardi2023implicit}. This section studies whether SGD with momentum, in modern deep learning, implicitly regularized the Riemannian Dimension across training dynamics. We examine whether this optimizer preferentially converges to solutions with lower Riemannian Dimension, and the experimental results are presented in Figure  \ref{fig:bound_evolution}. 
 
Empirical results show a repeatable pattern across the architectures: SGD with momentum drives the networks toward solutions with lower intrinsic Riemannian Dimension complexity, after an early transient; refer to Figure  \ref{fig:bound_evolution}. We observe that the Riemannian Dimension drops by orders of magnitude, whereas the VC dimension remains essentially unchanged.
The alignment between optimization dynamics and complexity control supports the view that SGD with momentum implicitly regularizes the Riemannian Dimension. Therefore, optimization is not merely as a mechanism for convergence; it is a primary driver of generalization through its systematic preference for low-complexity solutions. Riemannian Dimension provides a practical and theoretically grounded lens through which the implicit bias of optimizers in machine learning can be quantitatively assessed.

\section{Conclusion}
 We have developed a pointwise, representation-aware foundation for DNN generalization in the nonlinear feature learning regime. Meeting this challenge required several technical innovations: a pointwise generalization framework, a non-perturbative calculus for network mappings, a hierarchical covering scheme, and an ellipsoidal entropy theory for the Grassmannian—yielding structural insights into  cross-weight correlations and global geometric organization. The results strengthen the case that deep-learning generalization admits rigorous explanation and motivate a broader program in finite-scale geometric analysis of strongly correlated learning systems. Our experiments support the theory: the proposed Riemannian Dimension consistently tracks benign overparameterization, feature learning, and the optimizer’s implicit bias. 

Conceptually, the paper also separates two levels of observability.  The unconditional theorem proves compressed pointwise generalization through the mixed empirical-population geometry of the learned features; the fully observable theorem asks for a finite-resolution subspace isomorphism that lets the same geometry be read from the training sample alone.  This distinction is analogous in spirit to observability and uncertainty phenomena in physics, and it gives rise to a concrete open problem on subspace-level feature regularity.  The main message remains unchanged: {\bf pointwise feature compression is intelligence}, with feature compression made rigorous as layerwise learned-feature spectral compression and pointwise made rigorous as hypothesis-dependent finite-scale complexity.

Important directions ahead include integrating the analysis with optimization and algorithmic perspectives, extending it to modern architectures, translating the theory into concrete design principles for new deep-learning systems, and resolving the subspace-level isomorphism question posed in Section~\ref{sec generalization DNN}.

\newpage
\appendix

\section{Related Works and Experimental Setup}

\subsection{Related Works}\label{subsec related works}
Given the breadth of work on generalization and its empirical proxies, the mathematical grounding of our approach, and its conceptual relevance to vision and language practice, we streamline the exposition by concentrating on the most relevant prior results.

\paragraph{Theoretical Generalization Bounds for DNN.}
A substantial line of work anchors generalization bounds to various norms of the network weights, including path norms~\citep{neyshabur2015path}, Frobenius norms~\citep{neyshabur2015norm,golowich2020size}, and spectral norms~\citep{bartlett2017spectrally,neyshabur2017pac,arora2018stronger}. We refer to \citet{neyshabur2017exploring} as an important early contribution to this broader program. While offering conceptual insights, these bounds, often derived from globally uniform complexity measures like covering numbers or Rademacher complexity, frequently suffer from exponential dependencies on depth or layer norms, rendering them vacuous for practical, deep architectures. Compelling empirical evidence \citep{farhang2022investigating,razin2020implicit} further suggests that norm-based bounds alone are insufficient to fully elucidate the generalization phenomenon in deep learning. The kernel perspective \citep{belkin2018understand}, epitomized by NTK theory \citep{jacot2018neural,arora2019exact,golikov2022neural}, yields sharp guarantees by linearizing a network around its initialization—effectively casting training as kernel ridge regression with a fixed kernel. Within this linear/lazy regime, precise calculations explain both double descent \citep{belkin2019reconciling} and benign overfitting \citep{bartlett2020benign}, and an eigenspace-projection viewpoint provides dimension-reduction and feature-compression insights \citep{bartlett2021deep}. Investigations beyond the lazy regime exist, but most analyses either study the
two-layer infinite-width (mean-field) limit (e.g., \citep{mei2018mean,chizat2018global})
or remain in a neighborhood of initialization \citep{woodworth2020kernel}.
While insightful, these settings are idealized and struggle to capture the behavior of finite, deep networks (see Chapter~6 of \citep{misiakiewicz2023six}). More broadly, linear, lazy, or infinite-width approximations fail to reflect the feature learning that arises when parameters move far from initialization and representations evolve. This omission is widely viewed as a central bottleneck in current theory; indeed, the rich, representation-learning regime is often argued to be the key phenomenon distinguishing modern deep learning from long-standing frameworks (see, e.g., \cite{bartlett2021deep, misiakiewicz2023six, radhakrishnan2022mechanism,wilson2025deep}). Building on these directions, we establish—to our knowledge—the first pointwise generalization bounds for nonlinear DNN that are comparable in sharpness to prior linearization results and, crucially, remain valid in the practical feature-learning regime.

\paragraph{Other Theoretical Perspectives of Generalization.} A growing line of work connects generalization to geometric notions of fractal dimension
\citep{birdal2021intrinsic,dupuis2023generalization,simsekli2020hausdorff,andreeva2024topological,camuto2021fractal}, typically through Hausdorff– or Minkowski–type dimensions of optimization trajectories or invariant measures. However, these fractal dimensions are globally uniform, infinitesimal-scale ($\varepsilon \to 0$) notions of complexity. In contrast, our theory is built on a pointwise, finite-scale notion of geometric dimension. Section \ref{subsec finite scale} is precisely devoted to this distinction: we move from globally uniform to pointwise dimension and show that generalization is governed by the finite-scale pointwise dimension rather than its asymptotic limit. 
Several PAC–Bayesian approaches operate directly in parameter space $W$,
endowing the weights with an explicit stochastic model and directly computing the
KL divergence between a hand–designed prior and a posterior over $W$
\citep{hinton1993keeping,dziugaite2017computing,lotfi2022pac,lotfi2024non}; e.g., Gaussian distribution in \citet{dziugaite2017computing}.
These parameter-space bounds are valuable for certifying that certain
trained weight configurations admit nonvacuous PAC-Bayes guarantees, but
they largely treat the network as a black box and do not
directly capture how architecture and feature geometry control
generalization. PAC-Bayes theory has also been used to formalize flatness/sharpness~\citep{neyshabur2017exploring} and spectral-norm bounds~\citep{neyshabur2017pac}, highlighting the central role of this methodology in deep learning generalization and providing strong motivation for pointwise generalization; see the mathematical background below.
Alternative theoretical frameworks include algorithmic stability analyses, which are used primarily for one-hidden-layer networks and connected to the NTK/lazy-training viewpoint \citep{richards2021stability,lei2022stability}; and VC-dimension methods \citep{bartlett2019nearly}, which has been  discussed in  Section~\ref{subsec comparison}.

\paragraph{Pointwise and Non-Perturbative Foundations.}
Our use of ``pointwise'' draws inspiration from several threads that emphasize hypothesis-specific complexity: the asymptotic pointwise dimension in fractal geometry \citep{falconer1997techniques}, PAC-Bayes analyses that tailor complexity to the chosen  posterior or hypothesis \citep{zhang2006information,dziugaite2017computing,neyshabur2017pac}, and the Fernique–Talagrand integral in the majorizing-measure formulation of generic chaining \citep{fernique1975regularite,talagrand1987regularity,talagrand2014upper}. Notably, \citet{maurer2010majorizing,chu2023majorizing} establish simultaneous pointwise envelopes for stochastic processes using variants of the Fernique–Talagrand integral, providing direct precedents for our pointwise-dimension viewpoint. The synthesis of PAC-Bayes bounds with generic chaining dates back to \citet{audibert2003pac,audibert2007combining}, and mutual information based bounds have also been combined with chaining \citep{russo2016controlling,xu2017information,Asadi2018chaining, liu2025lifting}. To the best of our knowledge, this paper is the first work to establish a sharp pointwise bound for deterministic hypotheses in an uncountable class via localization to metric balls, explicitly connecting the result to the formal definition of pointwise dimension. A generic conversion from classical (subset-homogeneous) uniform convergence to pointwise generalization bounds, established in \cite{xu2020towards, xu2025towards}, serves as a guiding principle and plays a central role in our proof of Theorem \ref{thm generic chaining}.
The adjective “non-perturbative,” borrowed from physics \citep{nlab_nonperturbative_qft} and central to the study of strongly correlated systems \citep{nlab_strongly_correlated_system}, underscores that our theory remains valid far beyond infinitesimal neighborhoods of initialization—an essential property for deeply nonlinear, feature-learning DNN.

\paragraph{Connections to Differential Geometry and Lie Algebra.} From a geometric perspective, Hausdorff dimension provides an asymptotic, covering-based notion of capacity (fundamental in geometric measure theory \citep{SimonGMT2018}), while differential and Riemannian geometry \citep{jost2008riemannian} develop the use of local charts and global atlases to analyze non-Euclidean manifolds. Our results motivate viewing generalization as a finite-scale problem in geometric analysis.
The Grassmannian and families of orthogonal subspaces are traditionally studied via Lie groups; using differential-geometric tools, \citet{szarek1997metric,pajor1998metric} established finite-scale isotropic metric-entropy characterizations, which motivate our hierarchical covering viewpoint from local charts to a global atlas and our ellipsoidal entropy framework.

\paragraph{Predictor-Space Flatness and Regularization.}
Flatness has long been used as a geometric intuition for generalization and
regularization \citep{hochreiter1997flat,keskar2017large,foretsharpness}, but
parameter-space formulations are sensitive to reparametrization and scaling
\citep{dinh2017sharp}. The $L_1$ relaxation of the one-shot bound in Theorem~\ref{thm one shot} gives a
coordinate-free version of this idea by defining flatness directly in predictor
space. At scale \(\varepsilon\), the bias term measures loss-level stability over the
metric ball \(B_{\varrho^{(1)}_{(S,\mathbb P),\ell}}(f,\varepsilon)\): nearby predictors have controlled
loss deviation from \(f\). The variance term, namely the pointwise dimension
\(
\log \frac{1}{\pi\Big(B_{\varrho^{(1)}_{(S,\mathbb P),\ell}}(f,\varepsilon)\Big)}
\),
measures the statistical size of the same stable neighborhood through its prior
mass. Thus a favorable predictor is one whose loss is locally stable on a
neighborhood that also has sufficiently large prior mass, and the optimal scale
\(\varepsilon^\star\) balances local approximation against local mass. This also suggests
a regularization principle: prefer predictors with large stable neighborhoods in
predictor space, rather than merely small sharpness in a chosen parameterization.
Since the construction uses loss-metric balls in the function class
\(\mathcal F\), equivalent parameterizations of the same predictor induce the
same ball and the same pointwise dimension. In this sense, \eqref{eq: one shot}
formalizes flatness as predictor-space stability plus local prior mass, avoiding
the reparametrization pathologies of raw parameter-space sharpness. 

\paragraph{Empirical Indicators of Generalization.}
Complementing theory, much research has focused on empirical indicators that explain the generalization of deep learning. Phenomena like \emph{Neural Collapse} \citep{papyan2020prevalence,parker2023neural,kothapalli2022neural} reveals the emergence of low-rank geometric structures in last-layer features. Studies on \emph{Intrinsic Dimension} \citep{li2018measuring,huh2021low} similarly suggest that deeper models exhibit an inductive bias toward low-rank last-layer feature representations. A line of work focuses on \emph{Dynamic NTK variants} \citep{atanasov2021neural,baratin2021implicit,fort2020deep,kopitkov2020neural} or related feature-gradient kernels \citep{radhakrishnan2022mechanism}, where the kernels evolve along optimization trajectories, has  empirically shown that the dynamic kernel evolution is linked to generalization behaviour. Other probes, examining Fisher information \citep{karakida2019universal,jastrzebski2021catastrophic}, Hessian spectral properties \citep{ghorbani2019investigation,rahaman2019spectral}, and output-input Jacobians \citep{novak2018sensitivity}, offer another lens. Collectively, existing empirical probes offer valuable, though often partial, insights—typically from a specific layer perspective, or through a constructed similarity analysis—without a unifying formalism and a theory foundation. Our proposed empirical indicator, rooted in a mathematically sharp theory, resonates with their goals (our theory is in fact supported by many of their experiments) while advancing them. 
It provides a principled, formal measure for studying pointwise generalization in deep neural networks.

\paragraph{Feature Compression in Deep Models for Vision and Language.}
Across vision and language, deep networks exhibit a robust layer–wise compression of representations. In computer vision, \citet{ansuini2019intrinsic} measure intrinsic dimensionality across convolutional layers and find early expansion followed by sharp reduction, with lower late–stage dimensionality correlating with stronger generalization; \citet{feng2022rank} likewise show that feature matrices in CNNs and vision transformers become progressively low–rank with depth, at fixed width, indicating active compression of task–relevant information. Parallel trends appear in NLP: \citet{cai2021isotropy} demonstrate that contextual embeddings (e.g., BERT) occupy narrow, anisotropic cones despite high nominal dimension, and \citet{razzhigaev2024shape} document a two–phase training trajectory—initial expansion, then sustained compression. A complementary line grounded in the Information Bottleneck \citep{tishby2015deep} interprets these findings as the selective removal of task–irrelevant variability: \citet{shwartz2017opening} observe that networks spend most of training compressing internal features toward a prediction–compression trade–off, while \citet{patel2024learning} show gradient descent reduces the local rank of intermediate activations. \citet{balzano2025overview} provide a complementary tutorial on low-rank structures arising during the training and adaptation of large models, emphasizing how gradient-descent dynamics and implicit regularization generate low-rank representations. Taken together, these phenomena motivate our investigation: compression is not merely qualitative, but admits precise, hypothesis–specific complexity that governs generalization. 

\subsection{Experimental Setup}\label{appendix experimental setup}
We introduce detailed experimental setups. We evaluate the one-shot $L_2$ relaxation of the Riemannian Dimension bound derived from Theorem~\ref{thm empirical dnn feature iso} on two standard architectures—Fully Connected Networks (FCNs) and ResNets, using two benchmark datasets—MNIST \citep{lecun1998gradient} and CIFAR-10 \citep{krizhevsky2009learning}, respectively. The architecture of FCNs: we consider a $9$-hidden-layer FCN in which the first two hidden layers have width $2^{11}$ and the remaining seven hidden layers share a common width $h$, with $h \in \{2^{6},2^{7},2^{8},2^{9},2^{10},2^{11},2^{12}\}$. The output layer is a linear classifier mapping to $10$ logits, and we use ReLU as the activation and use PyTorch's default initialization (Kaiming uniform for ReLU). Increasing $h$ monotonically enlarges both layer widths and the total parameter count, yielding a clean capacity sweep at fixed depth.  The architecture of ResNets: we adopt the canonical ResNet architectures, ResNet-20, ResNet-32, ResNet-44, ResNet-56, ResNet-74, and ResNet-110, which differ only in the number of residual blocks per stage while maintaining the same overall architecture (three-stage, basic-block design) as introduced by \citep{he2016deep}. Following the practice of \citep{he2016deep}, we apply BatchNorm and ReLU after each convolution, with shortcut connections added as needed, and a global average pooling layer precedes the final linear classifier. These ResNet architectures provides a clean capacity sweep via depth.

We adopt standard training pipelines widely used in the benchmarks. (1) The training Protocol of FCNs is: SGD with momentum optimizer where momentum  $= 0.9$, learning rate $= 0.01$, and weight decay $= 5 \times 10^{-4}$; $200$ epochs and $128$ batch size;  a step decay at epochs $\{100, 170\}$, where the learning rate is scaled by $\times 0.1$.  (2) The training Protocol of ResNets is: SGD with momentum optimizer where momentum  $= 0.9$, learning rate $= 0.1$, and weight decay $= 5 \times 10^{-4}$;  $250$ epochs and $128$ batch size; a step decay at epochs $\{50, 150, 200\}$, where the learning rate is scaled by $\times 0.1$; Following practical training conditions, we apply standard data augmentation on CIFAR-10: random horizontal flips and $4$-pixel random crops with zero-padding. 

In the experiments of FCNs and ResNets, to enable layerwise analysis of the evolving feature representations and support our computation of Riemannian Dimension, we register forward hooks on all nonlinearity layers. For layers followed by pooling, we replace the last recorded ReLU activation with the corresponding pooled output. We also pre-register the input hook to capture the feature matrix of the data. These hooks ensure precise extraction of nonlinearity activations at each depth throughout training. We set the hyper-parameter $\varepsilon$ via a one–dimensional ternary-search procedure: at the end of each training stage, we perform a $500$-step ternary search for FCNs and a $50$-step ternary search for ResNets over the admissible interval specified by the following finite-resolution search range: $[\sqrt{1/n},\;\max_{l=1,...,L}\sqrt{\frac{2L\lambda_{\max}(F_{l-1}F_{l-1}^\top) \cdot ||W||_{\tF}^2  \prod_{i>l} \|W_i\|_{\op}^2}{n}}]$. The search selects the value of $\varepsilon$ that minimizes the one-shot $L_2$ relaxation of the Riemannian Dimension-based generalization bound in Theorem~\ref{thm empirical dnn feature iso}; it does not directly evaluate the gap-distance balls in \eqref{eq: one shot}. We note that tighter bounds could be achieved with more refined optimization procedures on $\varepsilon$.  For FCNs, we compute full feature gram matrices. While for ResNets, the feature matrix $F$ is formed by flattening each activation map into a vector of dimension $d = C \cdot H \cdot W$, where $C, H, W$ are the channel, height, and width of the feature map respectively. To align with our theory, we simplify ResNets to fully connected (feed-forward) networks when computing our bound; we apply the same simplification to the associated VC-dimension and parameter-count calculations to maintain consistency. To avoid out-of-memory in computing full feature  gram matrices in high-dimensional convolutional layers, we use the standard Gaussian sketching approximation, where each feature  gram matrix uses a Gaussian sketch with parameter $r = \min(8192,   \lfloor d/8 \rfloor )$ \citep{woodruff2014sketching}. By standard subspace-embedding guarantees, such Gaussian sketches preserve Gram quadratic forms—and hence the spectra—of the feature matrices with high probability, introducing only negligible distortion and leaving our conclusions unchanged \citep{woodruff2014sketching}.

\section{Proofs for Pointwise Generalization Framework (Section \ref{sec pointwise generalization})}\label{appendix pointwise}
Much of this section is devoted to a full proof of Theorem~\ref{thm generic chaining} (the integral upper bound).
Conceptually, the pointwise–dimension principle already follows from elementary PAC-Bayes arguments—see
Theorem~\ref{thm PACBayes-uniform-upper} and the subsequent remark in Appendix \ref{subsec PAC-Bayes}. We present the full derivation to make
explicit structural properties (e.g.,  unified blueprint, subset homogeneity, mixed empirical-ghost comparison) that a
rigorous proof requires.
\subsection{The ``Uniform Pointwise Convergence'' Principle}\label{subsec uniform pointwise convergence}
In this section, we present a unified blueprint for establishing pointwise generalization bounds. We state necessary and sufficient conditions for pointwise generalization and show that, when applied carefully, the resulting pointwise bounds are no harder to obtain than classical uniform-convergence guarantees.

We begin by citing a general principle for converting {\it subset-homogeneous} uniform convergence guarantees—i.e., bounds in which the same pointwise
complexity applies for every fixed subset $\cH\subseteq\cF$—into pointwise generalization bounds. This conversion, introduced by the name ``uniform localized convergence'' principle in \citep{xu2020towards} (short conference version) and \cite{xu2025towards} (full journal version), identifies minimal conditions under which
pointwise envelopes can be obtained, up to negligible double-logarithmic
factors, while avoiding unrealistic Bernstein-type and sub-root
assumptions. We state this result as ``uniform pointwise convergence'' principle. This subsection provides the necessary and sufficient
justifications, together with discussions of data dependence.
\begin{lemma}[``Uniform Pointwise Convergence'' Principle] \label{lemma uniformed localized convergence}\textup{\bf(Proposition~1 in \citet{xu2020towards,xu2025towards}).}
For a function class  $\cF$ and   functional $d:\cF\rightarrow [0,R]$, assume there is a function  $\psi(r;\delta)$, which is non-decreasing with respect to $r$, non-increasing with respect to $\delta$, and satisfies that $\forall \delta\in(0,1)$, $\forall r\in[0,R]$, with probability at least $1-\delta$,
\begin{align}\label{eq: surrogate}
\sup_{f\in\cF: d(f)\leq r}(\bP-\bPn)\ell(f;z)\leq \psi(r;\delta).
\end{align}
 Then, given any $\delta\in(0,1)$ and $r_0\in (0,R]$, with probability at least $1-\delta$,  uniformly over all $f\in \cF$,
\begin{align}\label{eq: peeling}
    (\bP-\bPn) \ell(f;z)\leq  \psi\left(\max\{2d(f), r_0\};\delta\left({\log_2\frac{2R}{r_0}}\right)^{-1}\right).
\end{align}
\end{lemma}

This lemma provides a succinct proof that serves as a unifying principle to sharpen classical localization, building on Section 2 of \cite{xu2025towards}.  A key advantage of this framework is its level of abstraction: it establishes subset homogeneity as the necessary and sufficient condition for pointwise generalization when the complexity functional $d(\cdot)$ is data–independent, and likewise when $d(\cdot)$ is swap–invariant and depends on both the observed sample $S=\{z_i\}_{i=1}^n$ and an i.i.d. ghost sample $S'=\{z_i'\}_{i=1}^n$. It also provides a clean treatment of data–dependent functionals and their induced (random) sublevel sets $\{f\in\mathcal F: d(f)\le r)$, as outlined before Section 4  of \cite{xu2025towards}. Crucially, this approach circumvents the circularities that often arise when combining symmetrization with localization or offset arguments.

\subsubsection{Necessary and Sufficient Conditions for Pointwise Generalization} \label{appendix necessary sufficient}
We leverage this ``uniform pointwise convergence'' principle to streamline the derivation of our bounds. Let $d(\cdot)$ denote a pointwise complexity functional, which we categorize into data-independent forms and data-dependent forms. Let $\psi(\cdot;\delta)$ be a non-decreasing function (typically $\psi(r;\delta)\asymp\sqrt{(r+\log (1/\delta))/n}$). We provide a clean characterization of pointwise generalization.

\paragraph{Necessary Condition: Subset Homogeneity.} A valid pointwise generalization guarantee (i.e., \eqref{eq: generalization gap}) necessitates \emph{subset homogeneity}. That is, if the pointwise inequality 
    \begin{align}\label{eq: pointwise psi}
        (\bP-\bPn) \ell(f;z)\leq \psi(d(f);\delta)
    \end{align}
    holds with probability at least $1-\delta$, then \eqref{eq: pointwise psi} must imply that for every fixed (i.e., data-independent) subset $\cH\subseteq\cF$, 
    \begin{align*}
    \sup_{f\in\cH}\bigl(\bP-\bPn\bigr)\,\ell(f;z)
    \;\leq\; \sup_{f\in\cH}\,\psi\!\bigl(d(f);\delta\bigr).
    \end{align*}
    Crucially, the complexity evaluation $d(f)$ must \emph{not} depend on the chosen subset \(\cH\). For instance, for the pointwise dimension $\log \tfrac{1}{\pi(B_{\varrho}(f,\varepsilon))}$, the prior $\pi$ (in particular, its support) should be independent of \(\cH\). This contrasts with classical empirical-process techniques—e.g., naive uses of Rademacher complexity and generic chaining—where the \emph{pre-specified} index sets dictate the proxy $d(\cdot)$ via the chosen Rademacher expectation, admissible tree construction, or prior.
    
    Subset homogeneity is thus the primary eligibility check for any candidate pointwise complexity functional. In Appendix \ref{appendix ambient equivalence}, we complete this check by establishing that the pointwise dimension is \emph{ambiently equivalent}: using a prior \(\pi\in\Delta(\cF)\) or its restriction \(\pi\in\Delta(\cH)\) produces complexities that agree in order (up to absolute constants).

\paragraph{Sufficient Condition: Subset Homogeneity + Data-Independent (or Symmetrized) $d(\cdot)$.}
Assuming the following subset-homogeneity uniform convergence condition:  for every fixed (i.e., data-independent) subset \(\cH\subseteq\cF\) and $\delta\in(0,1)$, with probability at least $1-\delta$,
\begin{align}\label{eq: U1}
    \sup_{f\in\cH}(\bP-\bPn)\ell(f;z)\leq \sup_{f\in\cH}\,\psi\!\bigl(d(f);\delta\bigr).
\end{align}By taking the sublevel set
\[
\cH \;=\; \{f\in\cF:\ d(f)\le r\},
\]
the condition \eqref{eq: U1} (applied to this fixed sublevel set) directly implies the surrogate conditions \eqref{eq: surrogate} in Lemma~\ref{lemma uniformed localized convergence}, and hence the pointwise bound \eqref{eq: peeling}. Thus, subset homogeneity is a necessary and sufficient condition for a data-independent $d(\cdot)$ to imply a pointwise generalization bound.

Likewise, in Appendix~\ref{appendix ghost}, we show that when the complexity $d(\cdot)$ may depend on both the observed sample $S=\{z_i\}_{i=1}^n$ and an i.i.d.\ ghost sample $S'=\{z_i'\}_{i=1}^n$, \emph{provided it is swap–invariant in $(S,S')$ (i.e., invariant under any exchange $z_i\!\leftrightarrow\! z_i')$}, subset homogeneity suffices to yield a pointwise generalization bound via a final swap–symmetrization argument. This establishes Theorem~\ref{thm ghost}: a pointwise generalization result in which the
complexity is evaluated using both the observed sample \(S\) and the ghost sample \(S'\).

\paragraph{Toward pointwise bounds using only observed sample.}
If one seeks bounds that are fully computable from the observed data \(\{z_i\}_{i=1}^n\) alone, without ghost sample or sample splitting, the analysis
is more involved. A practical route is two–step:
(i) first derive a symmetrized pointwise bound using a complexity functional based on \((S,S')\) or \((S,\bP)\)
(which is already valid and sharp);
(ii) then prove or assume an isomorphism between the \(L_2(\mathbb P_S)\)– and \(L_2(\mathbb P)\)–induced
pointwise complexities so as to replace population
or ghost–dependent terms by empirical ones, yielding a fully data–dependent bound.

\subsection{PAC-Bayes Optimization and Pointwise Dimension Bound} \label{subsec PAC-Bayes}
We illustrate why pointwise dimension is a natural consequence of  {\it best} PAC-Bayes optimization.
\begin{lemma}[PAC-Bayes Bound;  \citet{catoni2003pac}; Theorem~2.1, Theorem~2.4 in \citet{alquier2024user}]\label{lemma PAC-Bayes}
Let $\pi$ be a prior on a hypothesis class \(\mathcal{F}\) independent of the data, and let \(\ell\colon\mathcal{F}\times\mathcal{Z}\to[0,1]\) be a bounded loss. Fix confidence \(\delta\in(0,1)\) and sample size \(n\). For each fixed $\eta>0$, with probability at least $1-\delta$ over $S\sim\bP^{\otimes n}$, simultaneously for every posterior $\mu$,
\begin{equation}\label{eq:pac-bayes-fixed-eta}
(\bP-\bPn)\langle\mu,\ell\rangle
\le \frac{\mathrm{KL}(\mu,\pi)+\log(1/\delta)}{\eta n}+\frac{\eta}{8}.
\end{equation}
Define
\[
J_n:=\left\lceil\log_2(4\sqrt n)\right\rceil,
\qquad M_n:=J_n+1.
\]
There is also a single event of probability at least $1-\delta$ on which, simultaneously for every $\mu$,
\begin{equation}\label{eq:pac-bayes-lemma}
(\bP-\bPn)\langle\mu,\ell\rangle
\le \frac54\sqrt{\frac{\mathrm{KL}(\mu,\pi)+\log(M_n/\delta)}{2n}}.
\end{equation}
On this same event, for every $\mu$ and every $\eta>0$,
\begin{equation}\label{eq:pac-bayes-uniform-eta}
(\bP-\bPn)\langle\mu,\ell\rangle
\le \frac54\left\{\frac{\eta}{8}
+\frac{\mathrm{KL}(\mu,\pi)+\log(M_n/\delta)}{\eta n}\right\}.
\end{equation}
\end{lemma}

\paragraph{Proof of the uniform optimization.}
Apply the fixed-$\eta$ bound at confidence $\delta/M_n$ to each
$\eta\in\{4\cdot2^{-j}:0\le j\le J_n\}$ and take a union bound.
Write $K=\mathrm{KL}(\mu,\pi)+\log(M_n/\delta)$.
If $K\le2n$, then $\eta^\star=\sqrt{8K/n}$ lies between the smallest grid point and $4$.
Choose a grid point $\eta\in[\eta^\star,2\eta^\star]$. For $t=\eta/\eta^\star\in[1,2]$,
\[
\frac{\eta}{8}+\frac{K}{\eta n}
=\frac{t+t^{-1}}2\sqrt{\frac K{2n}}
\le\frac54\sqrt{\frac K{2n}}.
\]
If $K>2n$, the same conclusion follows from the bound
$(\bP-\bPn)\langle\mu,\ell\rangle\le1$.
Finally, the arithmetic-geometric mean inequality gives
\eqref{eq:pac-bayes-uniform-eta} for all $\eta>0$ on the same event.
\hfill$\square$

We now use the PAC-Bayes bound (which holds uniformly for every random posterior $\mu$) to approximate a deterministic hypothesis $f$. On the event in \eqref{eq:pac-bayes-uniform-eta}, uniformly over every $\mu\in\Delta(\mathcal F)$, $f\in\mathcal F$, and $\eta>0$,
\begin{align}\label{eq: double uniform}
(\bP-\bPn)\ell(f;z)
&=\langle\mu,(\bP-\bPn)\ell(\cdot;z)\rangle\nonumber\\
&\quad+\langle\mu,(\bP-\bPn)(\ell(f;z)-\ell(\cdot;z))\rangle\nonumber\\
&\le\frac54\left\{\frac\eta8+
\frac{\mathrm{KL}(\mu,\pi)+\log(M_n/\delta)}{\eta n}\right\}
+\langle\mu,\varrho^{\mathrm{gap}}_{(S,\bP),\ell}(\cdot,f)\rangle,
\end{align}
where
\begin{equation}\label{eq: mixed metric}
\varrho^{\mathrm{gap}}_{(S,\bP),\ell}(f',f)
:=\big|(\bP-\bPn)(\ell(f';z)-\ell(f;z))\big|
\le \varrho^{(1)}_{(S,\bP),\ell}(f',f).
\end{equation}
In \eqref{eq: double uniform}, Lemma~\ref{lemma PAC-Bayes} bounds the variance term, and taking the absolute value of the signed gap difference bounds the bias term.
Among symmetric nonnegative costs $d$ satisfying $(\bP-\bPn)(\ell(f;z)-\ell(g;z))\le d(f,g)$ for every ordered pair, $\varrho^{\mathrm{gap}}_{(S,\bP),\ell}$ is pointwise minimal: reversing the pair gives $|(\bP-\bPn)(\ell(f;z)-\ell(g;z))|\le d(f,g)$. This does not establish optimality among all PAC-Bayes bounds; retaining the signed posterior bias can be sharper.

Motivated by the above bias-variance optimization \eqref{eq: double uniform} via PAC-Bayes, for a given prior $\pi$, metric $\varrho$, and confidence $\delta\in(0,1)$ we define the {\it PAC-Bayes optimization objective}
\begin{align}\label{eq: PAC-Bayes optimization objective}
    V(\mu,\eta, f, \varrho):=\underbrace{\frac{\eta}{8}+\frac{\textup{KL}(\mu,\pi)+\log\frac{M_n}{\delta}}{\eta n}}_{\textup{Variance}}+\underbrace{\langle \mu, \varrho(\cdot,f)\rangle}_{\textup{Bias}},
\end{align}
where $\eta>0$, $n$ is the sample size, $\mu$ is a posterior over hypotheses, and $M_n=\left\lceil \log_2(4\sqrt n)\right\rceil+1$ is the grid size for uniform optimization. Since the bias is nonnegative, \eqref{eq: double uniform} implies, on the same event,
\[
(\bP-\bPn)\ell(f;z)
\le\frac54 V\bigl(\mu,\eta,f,\varrho^{\mathrm{gap}}_{(S,\mathbb P),\ell}\bigr)
\quad\text{for all }f\in\mathcal F,\ \mu\in\Delta(\mathcal F),\ \eta>0,
\]
 Here, the “Variance” term arises from a PAC-Bayes bound (Lemma~\ref{lemma PAC-Bayes}) applied to $\mu$, and the ``Bias'' term $\langle \mu,\varrho(\cdot,f)\rangle:=\mathbb{E}_{h\sim\mu}\big[\varrho(h,f)\big]$ measures how well the randomized $\mu$ approximates the target $f$.

\paragraph{Optimizing the Posterior $\mu$ for the Objective \eqref{eq: PAC-Bayes optimization objective}}
The intuitive analysis \eqref{eq: double uniform} explains how the PAC-Bayesian optimization objective naturally bounds the generalization gap.  We now minimize the posterior $\mu$ in \eqref{eq: PAC-Bayes optimization objective}. For fixed $\eta$, \eqref{eq: PAC-Bayes optimization objective} is minimized over $\mu$ by the Gibbs posterior. To obtain a closed-form characterization of the optimized value,
we proceed in two steps: (i) derive an explicit pointwise-dimension upper bound by
taking $\mu$ to be the $\pi-$normalized density on the
metric ball $B_\varrho(f,\varepsilon)$ (Theorem \ref{thm PACBayes-uniform-upper}), and (ii) show that this choice is near-optimal at matched scales
(Lemma~\ref{lemma pacbayes-uniform-opt}).

\subsubsection{Pointwise Dimension Bound via  Metric Ball}
Given any prior $\pi$ on $\cF$ and any $f\in\cF$, take $\mu$ to be the $\pi-$normalized density on the metric ball $B_{\varrho}(f,\varepsilon)$, i.e.,
\begin{equation}\label{eq:uniform-ball}
\mu(A)=\frac{\pi\!\left(A\cap B_{\varrho}(f,\varepsilon)\right)}{\pi\!\left(B_{\varrho}(f,\varepsilon)\right)}
\quad\text{for all measurable }A\subseteq\mathcal F.
\end{equation}
This choice gives an explicit upper bound on the optimized objective; Lemma~\ref{lemma pacbayes-uniform-opt} supplies the corresponding lower bound.

\begin{theorem}[Pointwise Dimension and Pointwise Generalization Upper Bound]\label{thm PACBayes-uniform-upper}
For the PAC-Bayes objective \eqref{eq: PAC-Bayes optimization objective}, let $\mu$ be the $\pi-$normalized density on $B_{\varrho}(f,\varepsilon)$, assuming that this ball has positive prior mass, i.e.
\[
   \frac{d\mu}{d\pi}(h)
   \;=\;
   \begin{cases}
     \displaystyle \frac{1}{\pi\!\bigl(B_{\varrho}(f,\varepsilon)\bigr)}, & h\in B_{\varrho}(f,\varepsilon),\\[6pt]
     0, & h\notin B_{\varrho}(f,\varepsilon).
   \end{cases}
\]
Then, with $\eta^\star=\sqrt{8\bigl(\textup{KL}(\mu,\pi)+\log(M_n/\delta)\bigr)/n}$,
\begin{align}\label{eq:variance-bias-optimized-bound}
  V\bigl(\mu,\eta^\star,f,\varrho\bigr)
  \;\le\;
  \sqrt{\frac{\textup{KL}(\mu,\pi)+\log\frac{\left\lceil\log_2(4\sqrt n)\right\rceil+1}{\delta}}{2n}}
  \;+\;
  \varepsilon
  \;=\;
  \sqrt{\frac{\log\!\frac{1}{\pi(B_{\varrho}(f,\varepsilon))}+\log\frac{\left\lceil\log_2(4\sqrt n)\right\rceil+1}{\delta}}{2n}}
  \;+\;
  \varepsilon.
\end{align}
Combining the upper bound \eqref{eq:variance-bias-optimized-bound} with \eqref{eq: double uniform} yields the pointwise generalization bound: for every $\delta\in(0,1)$, with probability at least $1-\delta$, uniformly over every $f\in\cF$,

\begin{align*}
(\bP-\bPn)\ell(f;z)
\le\frac54\inf_{\varepsilon>0}\left\{
\varepsilon+
\sqrt{\frac{\log\!\frac{1}{\pi\Big(B_{\varrho^{\mathrm{gap}}_{(S,\mathbb P),\ell}}(f,\varepsilon)\Big)}+\log\frac{\left\lceil\log_2(4\sqrt n)\right\rceil+1}{\delta}}{2n}}
\right\}.
\end{align*}
Here $\varrho^{\mathrm{gap}}_{(S,\mathbb P),\ell}$ is the generalization-gap metric in \eqref{eq: mixed metric}. This proves Theorem~\ref{thm one shot}, after enlarging its absolute constant.
\end{theorem}
\paragraph{Remark on the metric choice and formulation.}
The generalization-gap metric gives the sharpest symmetric pointwise bound on the bias, but depends on the unknown gaps themselves. Its mixed $L_1$ relaxation retains the loss geometry, while the mixed $L_2$ metric used by chaining is
\begin{equation}\label{eq: mixed L2}
\varrho_{(S,\mathbb P),\ell}(f',f)
:=\left((\mathbb P_S+\mathbb P)
[\ell(f';\cdot)-\ell(f;\cdot)]^2\right)^{1/2}.
\end{equation}
The inequalities $\varrho^{\mathrm{gap}}_{(S,\mathbb P),\ell}\le\varrho^{(1)}_{(S,\mathbb P),\ell}\le\sqrt2\,\varrho_{(S,\mathbb P),\ell}$ connect these one-shot bounds. Corollary~\ref{coro one shot L2} records the common $L_2$ consequence of the PAC-Bayes and chaining theorems. The  comparison between the multiscale-integral and one-shot objectives is made in that same metric; it does not assert that the mixed $L_2$ integral dominates the mixed $L_1$ one-shot bound. Theorem~\ref{thm lower bound} shows that the one-shot bound based on the gap distance is always worst-case optimal, whereas the chaining bound based on the mixed metric can be worst-case suboptimal.

\paragraph{Proof of Theorem~\ref{thm PACBayes-uniform-upper}:}
For the choice \eqref{eq:uniform-ball},
\begin{align}
  \textup{KL}(\mu,\pi)
  &= \int_{\cF} \log\!\Bigl(\frac{d\mu}{d\pi}(h)\Bigr)\,\mu(dh)
   = \int_{B_{\varrho}(f,\varepsilon)} \log\!\Bigl(\tfrac{1}{\pi(B_{\varrho}(f,\varepsilon))}\Bigr)\,\mu(dh)
   = \log\!\frac{1}{\pi(B_{\varrho}(f,\varepsilon))}. \label{eq:KL-uniform-ball}
\end{align}
Moreover, by construction,
\[
  \langle \mu,\varrho(\cdot,f)\rangle
  = \int_{B_{\varrho}(f,\varepsilon)} \varrho(h,f)\,\mu(dh)
  \le \varepsilon.
\]
Plugging \eqref{eq:KL-uniform-ball} into \eqref{eq: PAC-Bayes optimization objective} and minimizing
$\frac{\eta}{8}+\frac{\textup{KL}(\mu,\pi)+\log\frac{M_n}{\delta}}{\eta n}$ over $\eta>0$ gives
$\sqrt{\frac{\textup{KL}(\mu,\pi)+\log\frac{M_n}{\delta}}{2n}}$, which together with the bias bound
$\langle \mu,\varrho(\cdot,f)\rangle\le \varepsilon$ yields the claimed bound \eqref{eq:variance-bias-optimized-bound}. 

\hfill$\square$

\subsubsection{Lower Bound and Optimality of PAC-Bayes Optimization}
The following lemma compares the ball-posterior upper bound with the optimized Gibbs objective. The localization terms are $a+\varepsilon$ above and $\min\{a,\varepsilon\}$ below, apart from the stated additive term, where $a=\log(1/\pi(B_\varrho(f,\varepsilon)))/(\eta n)$. They are comparable when $a$ and $\varepsilon$ are comparable.
\begin{lemma}[Optimality of Pointwise Dimension in PAC-Bayes Optimization]\label{lemma pacbayes-uniform-opt}
For the PAC-Bayes optimization objective $V(\mu, \eta, f, \varrho)$ defined in \eqref{eq: PAC-Bayes optimization objective}, 
we have that for every $f\in\mathcal F$, $\eta>0$, and $\varepsilon>0$,
\begin{equation}\label{eq:LB-min}
\inf_{\mu }V(\mu, \eta, f, \varrho)\ \ge\ \frac{\eta}{8} +\frac{ \log\frac{\left\lceil \log_2(4\sqrt n)\right\rceil+1}{\delta}}{\eta n}+ 
\min\!\Big\{\frac{1}{\eta n}\log\frac{1}{\pi\!\left(B_{\varrho}(f,\varepsilon)\right)},\ \varepsilon\Big\}
\ -\ \frac{\log 2}{\eta n}.
\end{equation}
Consequently, for every $f\in\mathcal F$, $\eta>0$, and $\varepsilon>0$,
\begin{align}\label{eq:sandwich}
&\frac{\eta}{8} +\frac{ \log\frac{\left\lceil \log_2(4\sqrt n)\right\rceil+1}{\delta}}{\eta n} + 
\min\!\Big\{\frac{\log\frac{1}{\pi\!\left(B_{\varrho}(f,\varepsilon)\right)}}{\eta n},\ \varepsilon\Big\}
\ -\ \frac{\log 2}{\eta n} \nonumber\\ \le &\inf_{\mu }V(\mu, \eta, f, \varrho)\ \nonumber\\
\le &
\frac{\eta}{8} + \frac{\log\frac1{\pi(B_{\varrho}(f,\varepsilon))}\;+\log\frac{\left\lceil \log_2(4\sqrt n)\right\rceil+1}{\delta}}{\eta n} + \varepsilon.
\end{align}
\end{lemma}

\paragraph{Proof of Lemma \ref{lemma pacbayes-uniform-opt}}
The upper bound in \eqref{eq:sandwich} is already proved in Theorem \ref{thm PACBayes-uniform-upper}, so we only need to prove the lower bound \eqref{eq:LB-min}.  The Donsker-Varadhan variational identity states that for any measurable $h$,
\[-\log \int e^{h}\,d\pi\ =\ \inf_{\mu}\Big\{\textup{KL}(\mu,\pi)\ -\ \int h\,d\mu\Big\}.
\]
Apply it with $h= -\eta n \varrho(\cdot,f)$ to obtain
\begin{align*} 
  -\log \int e^{-\eta n \varrho(\cdot,f)}\,d\pi\ = \inf_{\mu}\Big\{\textup{KL}(\mu,\pi)\ +\ \int \eta n \varrho(\cdot,f)\,d\mu\Big\},
\end{align*}
which implies that
\begin{equation}\label{eq:dv}
 \frac{\eta}{8} +\frac{ \log\frac{\left\lceil \log_2(4\sqrt n)\right\rceil+1}{\delta}}{\eta n}-\frac{1}{\eta n}\log \int e^{-\eta n \varrho(\cdot,f)}\,d\pi\ = \inf_{\mu}\Big\{\frac{\eta}{8} + \frac{\textup{KL}\bigl(\mu,\pi\bigr)\;+\log\frac{\left\lceil \log_2(4\sqrt n)\right\rceil+1}{\delta}}{\eta n}+\langle\mu,\varrho(\cdot,f)\rangle\Big\}.
\end{equation}
By splitting the dual integral,
\begin{align*}
\int e^{-\eta n \varrho(\cdot,f)}\,d\pi
&= \int_{B_{\varrho}(f,\varepsilon)}  e^{-\eta n \varrho(\cdot,f)}\,d\pi
  + \int_{{B_{\varrho}(f,\varepsilon)}^c}  e^{-\eta n \varrho(\cdot,f)}\,d\pi \\
&\le \pi\!\left(B_{\varrho}(f,\varepsilon)\right) + e^{-\eta n \varepsilon}\!\left(1-\pi\!\left(B_{\varrho}(f,\varepsilon)\right)\right) \\
&\le \pi\!\left(B_{\varrho}(f,\varepsilon)\right) + e^{-\eta n \varepsilon},
\end{align*}
where ${B_{\varrho}(f,\varepsilon)}^c$ is complement of $B_{\varrho}(f,\varepsilon)$; and we have used $e^{-\eta n \varrho(\cdot,f)}\le 1$ on $B_{\varrho}(f,\varepsilon)$ and
$e^{-\eta n \varrho(\cdot,f)}\le e^{-\eta n \varepsilon}$ on $B_{\varrho}(f,\varepsilon)^c$. 
Hence
\begin{equation} \label{eq:com}
\inf_{\mu} V(\mu, \eta,f, \varrho)\ \ge\ \frac{\eta}{8} +\frac{ \log\frac{\left\lceil \log_2(4\sqrt n)\right\rceil+1}{\delta}}{\eta n}-\frac{1}{\eta n}\log\!\Big(\pi\!\left(B_{\varrho}(f,\varepsilon)\right) + e^{-\eta n \varepsilon}\Big).
\end{equation}
The simplified form \eqref{eq:LB-min} follows from
$a+b\le 2\max\{a,b\}$ or equivalently
$-\log(a+b)\ge -\log 2 + \min\{-\log a,-\log b\}$ on \eqref{eq:com}.
Combining \eqref{eq: PAC-Bayes optimization objective}, \eqref{eq:KL-uniform-ball} and \eqref{eq:LB-min} yields the sandwich \eqref{eq:sandwich}.

\hfill $\square$

\subsection{Subset Homogeneity  of Pointwise Dimension}\label{appendix ambient equivalence}

We show that, for any \(f\in\cH\subseteq\cF\), the pointwise–dimension functional defined with a prior $\pi$ is unchanged in order (up to absolute constants) whether $\pi$ is supported on \(\cH\) or on the ambient class \(\cF\). Hence one may take \(\pi\in\Delta(\cF)\) without restricting it to any particular subset, which suffices to meet the subset–homogeneity condition in Appendix~\ref{appendix necessary sufficient}.
\begin{lemma}[Ambient Equivalence of Pointwise Dimension]\label{lemma PD equivalence}
Let $(\cF,\varrho)$ be a metric space and let $\cH\subseteq \cF$ be a subset. Consider a nearest-point selector $p:\cF\to \cH$
satisfying $\varrho(f,p(f))=\min_{h\in\cH}\varrho(f,h)$ for all $f\in \cF$. For any ambient prior \(\pi\in\Delta(\mathcal F)\), let
\(\pi_{\mathcal H}:=p_{\#}\pi\) be the pushforward measure induced by \(p\), defined by
\[
\pi_{\mathcal H}(A)
:=
\pi\bigl(\{g\in\mathcal F:\ p(g)\in A\}\bigr),
\qquad A\subseteq\mathcal H\ \text{measurable}.
\]
Equivalently, in the discrete case,
$\pi_{\mathcal H}(h)
=
\sum_{g\in\mathcal F:\,p(g)=h}\pi(g),$
that is, \(\pi_{\mathcal H}\) collects all ambient prior mass whose nearest point in
\(\mathcal H\) is \(h\).
Then for every $\varepsilon>0$ we have
\begin{align*}
    \pi_{\cH}(B_{\varrho}(f,2\varepsilon))\geq \pi(B_{\varrho}(f,\varepsilon)),\quad
    \log\frac{1}{\pi_{\cH}(B_{\varrho}(f,2\varepsilon))}\leq \log \frac{1}{\pi(B_{\varrho}(f,\varepsilon))}.
\end{align*}
Consequently, for any ambient prior  $\pi\in\Delta(\cF)$ and any point $f\in\cF$, define the majorizing measure integral
\begin{align*}
I(\pi,f,\varrho)
:=\inf_{\alpha \geq 0}\left\{\alpha+\frac{1}{\sqrt{n}}\int_{\alpha}^{+\infty}\sqrt{\log\frac{1}{\pi\!\big(B_\varrho(f,\varepsilon)\big)}}\,d\varepsilon\right\}.
\end{align*}
Then  we have
\begin{align}\label{eq:ambient-equivalence-bound}
\frac{1}{2}\inf_{\mu\in\Delta(\cH)}\sup_{f\in\cH}I(\mu,f,\varrho)\leq \inf_{\pi\in\Delta(\cF)}\sup_{f\in\cH}I(\pi,f,\varrho)\leq \inf_{\mu\in\Delta(\cH)}\sup_{f\in\cH}I(\mu,f,\varrho). 
\end{align}
\end{lemma}

\paragraph{Proof of Lemma \ref{lemma PD equivalence}:}
The upper bound in \eqref{eq:ambient-equivalence-bound} is immediate since $\Delta(\cH)\subseteq \Delta(\cF)$: taking $\mu$ supported on $\cH$
gives $\inf_{\pi\in\Delta(\cF)}\sup_{f\in\cH}I(\pi,f,\varrho,r)\leq \inf_{\mu\in\Delta(\cH)}\sup_{f\in\cH}I(\mu,f,\varrho,r)$.

For the lower bound in \eqref{eq:ambient-equivalence-bound}, take $\pi_\cH$ to be the pushforward induced by the nearest-point selector.
For any $f\in \cH$ and $\varepsilon>0$, if $f'\in B_\varrho(f,\varepsilon)$ then
\[
\varrho\big(p(f'),f\big)\ \le\ \varrho\big(p(f'),f'\big)+\varrho(f',f)\ =\ \min_{h\in\cH}\varrho(f',h)+\varrho(f',f)\ \le\ 2\varepsilon,
\]
hence $p(f')\in B_\varrho(f,2\varepsilon)$ and
\begin{align}\label{eq: inclusion inequality}
\pi_{\cH}\!\big(B_\varrho(f,2\varepsilon)\big)\ \ge\ \pi\!\big(B_\varrho(f,\varepsilon)\big),\quad  \log\frac{1}{\pi_{\cH}(B_{\varrho}(f,2\varepsilon))}\leq \log \frac{1}{\pi(B_{\varrho}(f,\varepsilon))}.
\end{align}
Therefore,
\begin{align*}
I(\pi,f,\varrho)
\,=&\,\inf_{\alpha \geq 0}\left\{\alpha+\frac{1}{\sqrt{n}}\int_{\alpha}^{+\infty}\sqrt{\log\frac{1}{\pi(B_\varrho(f,\varepsilon))}}\,d\varepsilon\right\}\\
\ \ge\ &\inf_{\alpha \geq 0}\left\{\alpha+\frac{1}{\sqrt{n}}\int_{\alpha}^{+\infty}\sqrt{\log\frac{1}{\pi_{\cH}(B_\varrho(f,2\varepsilon))}}\,d\varepsilon\right\}\\
\,=&\frac{1}{2} \inf_{\alpha \geq 0}\left\{\alpha+\frac{1}{\sqrt{n}}\int_{\alpha}^{+\infty}\sqrt{\log\frac{1}{\pi_{\cH}(B_\varrho(f,\varepsilon))}}\,d\varepsilon\right\}\\
= & \frac{1}{2} I(\pi_{\cH},f,\varrho),
\end{align*}
where the first inequality is by \eqref{eq: inclusion inequality}; the second equality is by the change of variables.
Taking $\sup_{f\in \cH}$ and then $\inf_{\pi\in\Delta(\cF)}$, $\inf_{\mu\in\Delta(\cH)}$ yields
the desired lower bound. 

\hfill$\square$

\paragraph{Relationship to Fractional and Ordinary Covering Numbers}
As with ordinary covering numbers in Definition~\ref{def covering number}, we distinguish between internal and external fractional covering numbers. For $\mathcal H\subseteq\mathcal F$, define
\begin{align*}
\mathrm{N}'(\mathcal H,\varrho,\varepsilon)
:= \inf_{\pi\in\Delta(\mathcal F)}\ \sup_{f\in\mathcal H}
\frac{1}{\pi\big(B_{\varrho}(f,\varepsilon)\big)}.
\end{align*}
This is the \emph{external fractional covering number} relative to the ambient class $\mathcal F$; restricting the infimum to $\Delta(\mathcal H)$ gives the internal version. See \citet[Section~4.1]{chen2024assouad} for the general formulation and its connections to information-theoretic upper and lower bounds. 

Let $\mathrm{N}(\mathcal H,\varrho,\varepsilon)$ denote the internal covering number from Definition~\ref{def covering number}. A standard packing argument gives the following comparison; see \citet[Lemma~14]{chen2024assouad} and, in the batch $L_2(\bPn)$ setting, \citet[Lemma~8]{block2021majorizing}:
\begin{align}\label{eq: sandwitch fractional cover}
\log \mathrm{N}(\mathcal H,\varrho,2\varepsilon)
&\le
\log \mathrm{N}'(\mathcal H,\varrho,\varepsilon)
\notag\\
&=
\inf_{\pi\in\Delta(\mathcal F)}\ \sup_{f\in\mathcal H}
\log\frac{1}{\pi\big(B_{\varrho}(f,\varepsilon)\big)}
\le
\log \mathrm{N}(\mathcal H,\varrho,\varepsilon).
\end{align}
 The cited proofs establish this class-wide comparison through packing, whereas Lemma~\ref{lemma PD equivalence} provides a novel \emph{pointwise} comparison of ball probabilities simultaneously across scales.

Finally, taking the ambient class to be the target class in the comparison above and applying Lemma~\ref{lemma int-ext} shows that the internal fractional covering number in \eqref{eq: fractional covering number} is equivalent to the ordinary external covering number from Definition~\ref{def covering number}, up to a constant rescaling of the radius. Together, these arguments establish the equivalence claimed in the main text and explicitly show that external and internal fractional covering numbers are equivalent up to a constant rescaling of the radius. This comparison is not stated separately by \citet{block2021majorizing,chen2024assouad}.

The equivalence of internal and external ordinary covering numbers established in Lemma~\ref{lemma int-ext} of Appendix~\ref{subsec grassmannian main result}, together with the corresponding equivalence for fractional covering numbers derived above, helps clarify the contribution of our Lemma~\ref{lemma PD equivalence}. The equivalence of internal and external covering numbers suggests that ambient equivalence should also hold for class-wide integral functionals, even when the truncation level is optimized separately for each hypothesis. The novelty of Lemma~\ref{lemma PD equivalence} lies in establishing a \emph{pointwise} comparison that applies to \emph{pointwise} dimensions as well as to these class-wide functionals.
\paragraph{Collapsing the Distinction between Chaining and Generic Chaining.}
A simple illustration of the strength of our pointwise blueprint is the
multi–dimensional setting. Let
$(d^{(1)},\ldots,d^{(k)}):\cF\to(0,R]^k$ be coordinatewise, data-independent complexity functions; and \(\psi\) be coordinatewise
nondecreasing in \((d^{(1)},\ldots,d^{(k)})\) and nonincreasing in $\delta$. 
Our blueprint makes no essential distinction between the two uniform
forms
\begin{align*}
\text{(sup--inside)}\quad
&\sup_{f\in\cH}(\bP-\bPn)\,\ell(f;z)
 \;\le\;
 \psi\!\Bigl(\,\sup_{f\in\cH} d^{(1)}(f),\ldots,\sup_{f\in\cH} d^{(k)}(f);\ \delta\Bigr),\\
\text{(sup--outside)}\quad
&\sup_{f\in\cH}(\bP-\bPn)\,\ell(f;z)
 \;\le\;
 \sup_{f\in\cH}\,
 \psi\!\Bigl(d^{(1)}(f),\ldots,d^{(k)}(f);\ \delta\Bigr),
\end{align*}
in the sense that \emph{either} one leads to the same pointwise
conclusion after peeling.

More precisely, fix a base scale $r_0\in(0,R]$. Then with probability at least $1-\delta$, for every
$f\in\cF$,
\begin{equation}\label{eq: multi-dimensional}
(\bP-\bPn)\,\ell(f;z)
\;\leq\;
\psi\!\left(\left(\cdots,\max\!\left\{\,2 d^{(j)}(f),\,r_0\right\},\cdots\right);
\ \delta\Bigl(\log_2\frac{2R}{r_0}\Bigr)^{-k}\right).
\end{equation}
The most straightforward proof uses essentially the same peeling argument as in
Lemma~\ref{lemma uniformed localized convergence}, with the only change that we
use a grid of size $(\log_2(2R/r_0))^{k}$ (partition each coordinate into
$\log_2(2R/r_0)$ dyadic scales); see the short proof of Proposition~1 in
\cite{xu2025towards}. Alternatively, this can proved by applying Lemma \ref{lemma uniformed localized convergence} for $k$ times, where at each step we remove one dimension functional and divided confidence by $\log_2(2R/r_0)$. Moreover, the multi–dimensional pointwise bound \eqref{eq: multi-dimensional}
shows that its right–hand side, viewed as a \emph{scalar} complexity, yields an
equally tight pointwise bound. Hence the multi–dimensional formulation does not
improve the best-achievable rates beyond a suitably defined one–dimensional complexity (as in
generic chaining).

Conceptually, this shows that the apparent gap between classical chaining
(entropy integral; sup–inside), generic chaining (majorizing measures; sup–outside), and our pointwise
generic–chaining bound (Theorem~\ref{thm generic chaining}) disappears within
the blueprint: each is just a subset–homogeneous uniform statement that implies
the same pointwise bound up to absolute constants and logarithms. 
The peeling argument itself does not compare empirical and mixed metrics. For the worst-case family in Theorem~\ref{thm lower bound}, the corresponding class-wide functionals are equivalent in expectation, including pointwise optimized truncation; Appendix~\ref{appendix mixed lower bound} proves this comparison and explains why it does not extend to arbitrary classes.

\subsection{Multiscale Pointwise Dimension Bound via Ghost Sample}\label{appendix ghost}
In this section, we prove an easier variant of Theorem~\ref{thm generic chaining} that permits \emph{swap-invariant} randomized priors depending on both the observed sample and its ghost counterpart. This setting subsumes—and strengthens—the conditional mutual information (CMI) framework of \citet{steinke2020reasoning}.

Let \(S=(z_1,\dots,z_n)\) and \(S'=(z'_1,\dots,z'_n)\) be two i.i.d.\ samples drawn from \(\mathbb P^{\otimes n}\), independent of each other. For each index \(i\in\{1,\dots,n\}\), define the coordinate–swap map
\[
\tau_i(S,S') \;:=\; \bigl((z_1,\dots,z_{i-1},z'_i,z_{i+1},\dots,z_n),\ (z'_1,\dots,z'_{i-1},z_i,z'_{i+1},\dots,z'_n)\bigr).
\]
A randomized, data-dependent prior is a mapping \(\pi_{(\cdot, \cdot)}:\mathcal Z^{2n}\to\Delta(\mathcal F)\); we write \(\pi_{(S,S')}\in\Delta(\cF)\) for the realized prior over $\cF$ (a distribution on $\cF$ that may depend on $(S,S')$).
We say that \(\pi\) is \emph{swap–invariant} on \((S,S')\), if
\[
\pi_{(S,S')} \;=\; \pi_{\tau_i(S,S')} \qquad\text{for all } i=1,\dots,n \text{ and for }\mathbb P^{\otimes 2n}\text{-a.e.\ }(S,S').
\]
Equivalently, \(\pi\) depends only on the unordered multiset \(\{\{(z_i,z'_i)\}_{i=1}^n\}\) and not on which element of each pair is designated as ``observed'' versus ``ghost.''

\paragraph{Connection to CMI.}
This notion covers the conditional–mutual–information (CMI) framework of \citet{steinke2020reasoning}. In the CMI setup, one draws paired data \(Z=((Z_i^{(0)},Z_i^{(1)}))_{i=1}^n \stackrel{\text{i.i.d.}}{\sim} (\mathbb P\times\mathbb P)^{\otimes n}\) and an independent selector \(U\in\{0,1\}^n\). The training and ghost sets are
\(S_U=(Z_1^{(U_1)},\dots,Z_n^{(U_n)})\) and \(S_{\bar U}=(Z_1^{(1-U_1)},\dots,Z_n^{(1-U_n)})\).
Any prior \(\pi\) that is a function of \(Z\) only (independent of \(U\)) is swap–invariant, since flipping \(U_i\) implements \(\tau_i\).
Conversely, swap–invariance for all \(i\) is equivalent to invariance under all coordinatewise flips of \(U\), hence independence from \(U\).

\paragraph{}
Throughout, let \(S=\{z_i\}_{i=1}^n\) be the observed sample and \(S'=\{z'_i\}_{i=1}^n\) an i.i.d. ghost sample, independent of \(S\).
We write \(\bP_S\) for the empirical measure \(\bP_n\) based on \(S\), and \(\varrho_{S,\ell}\) for the metric \(\varrho_{n,\ell}\) from the main paper.
Let \(\bP_{S'}\) denote the empirical measure based on \(S'\).
For any integrable function \(g:\mathcal Z\!\to\!\mathbb R\) (we write \(g(z)\) when convenient; e.g., \(g(z)=\ell(f;z)\)),
define the empirical averaging operators
\[
\bP_S g := \frac{1}{n}\sum_{i=1}^n g(z_i),\qquad
\bP_{S'} g := \frac{1}{n}\sum_{i=1}^n g(z'_i).
\]
We use the shorthand \((\bP_S \pm \bP_{S'})g := \bP_S g \pm \bP_{S'} g\) for the sum/difference of the two sample–average operators, and the same notation when \(\bP_S \pm \bP_{S'}\) appear inside norms or distances.
\begin{theorem}[Multiscale Pointwise Generalization via Ghost Sample]\label{thm ghost}
  Let $\ell(f;z)\in[0,1]$.  There exists an absolute constant $ C>0$ such that for any swap-invariant prior $\pi_{(\cdot,\cdot)}$ on $(S,S')$, and any $\delta\in(0,1)$,  with probability at least
\(1-\delta\) over \((S,S')\), uniformly in \(f\in\cF\),
\begin{align*}
&(\bP_{S'}-\bP_S)\,\ell(f;z)
\ \\\le\
&C \left(
\inf_{\alpha\ge 0}\!\left\{\alpha+\frac{1}{\sqrt{n}}
\int_{\alpha}^{\sqrt{2}}
\sqrt{\log\frac{1}{\pi_{(S,S')}\!\bigl(B_{\varrho_{(S,S'),\ell}}(f,\varepsilon)\bigr)}}\,d\varepsilon\right\}
+\sqrt{\frac{\log\!\bigl(\log(2n)/\delta\bigr)}{n}}
\right),
\end{align*}
where \(\varrho_{(S,S'),\ell}(f_1,f_2)
= \sqrt{(\mathbb{P}_S+\mathbb{P}_{S'})\bigl(\ell(f_1;z)-\ell(f_2;z)\bigr)^2}\).  
\end{theorem}

\paragraph{Proof of Theorem \ref{thm ghost}:}
The proof of the upper bound in Theorem \ref{thm ghost} consists of three steps: 1. Subset-Homogeneous Uniform Convergence; 2.  Generic Conversion to Pointwise Generalization Bound; 3. High-Probability Symmetrization.

\paragraph{Step 1: Subset-Homogeneous Uniform Convergence.}
Let $S=\{z_i\}_{i=1}^n$ be the observed sample, and $S'=\{z'_i\}_{i=1}^n$ be an i.i.d. ghost sample. We consider the symmetrized loss 
\begin{align}\label{eq: symmetrized loss}
    \tilde{\ell}(f;(z,z'))=\ell(f;z')-\ell(f;z).
\end{align}
Since $\ell(f;z)$ is uniformly bounded in $[0,1]$, $\tilde \ell(f;(z,z'))$ is uniformly bounded in $[-1,1]$. We adopt the notation  
\[
\varrho_{S,\ell}(f_1,f_2)=\varrho_{n,\ell}(f_1,f_2)=\sqrt{\bP_S\big(\ell(f_1;z)-\ell(f_2;z)\big)^2}.
\]
 Furthermore, we define the loss-induced  $L_2$ metrics  $\varrho_{S',\ell}$ and $\varrho_{(S,S'),\ell}$  by
\begin{align*}
     &\varrho_{S',\ell}(f_1,f_2)= \sqrt{\bP_{S'}(\ell(f_1;z)-\ell(f_2;z))^2}, \\&\varrho_{(S,S'),\ell}(f_1,f_2)= \sqrt{\left(\bP_S+\bP_{S'}\right)(\ell(f_1;z)-\ell(f_2;z))^2}.
\end{align*} 
By Minkowski's inequality (see, e.g., \cite{wiki:minkowski-inequality}) and $\sqrt{a}+\sqrt{b}\leq \sqrt{2(a+b)}$, we have 
\begin{align}\label{eq: S S' metrics}
    \sqrt{\frac{1}{n}\sum_{i=1}^n(\tilde\ell(f_1;(z_i,z_i'))-\tilde\ell(f_2;(z_i,z_i')))^2}\leq \varrho_{S,\ell}(f_1,f_2)+\varrho_{S',\ell}(f_1,f_2)\leq \sqrt{2}\varrho_{(S,S'),\ell}(f_1,f_2).
\end{align}

Now applying the truncated integral bound (Lemma \ref{lemma truncated integral}) to the empirical Rademacher complexity: let $\{\xi_i\}_{i=1}^n$ be i.i.d. Rademacher variables, then  conditioned on $(S,S')$,
given any subset $\cH\subseteq\cF$, we have that for all $\delta\in(0,1)$, with probability at least $1-\delta$ (the randomness all comes from $\{\xi_i\}_{i=1}^n$),
\begin{align*}
    &\sup_{f\in\cH}\frac{1}{n}\sum_{i=1}^n \xi_i\tilde\ell(f;(z_i,z_i'))\leq \bE_{\xi}\left[ \sup_{f\in\cH}\frac{1}{n}\sum_{i=1}^n \xi_i\tilde\ell(f;(z_i,z_i'))\right]+\sqrt{\frac{2\log\frac{1}{\delta}}{n}}\\\leq &C_0\inf_{\mu\in\Delta(\cH)}\sup_{f\in\cH}\inf_{0\le\alpha\le2}\left\{\alpha+\frac{1}{\sqrt{n}}\int_\alpha^{2}\sqrt{\log\frac{1}{\mu(B_{\tilde\varrho}(f,\varepsilon))}}d\varepsilon\right\}+\sqrt{\frac{2\log\frac{1}{\delta}}{n}},
\end{align*}
where $C_0>0$ is an absolute constant, and $\tilde\varrho(f_1,f_2):= \sqrt{\frac{1}{n}\sum_{i=1}^n(\tilde\ell(f_1;(z_i,z_i'))-\tilde\ell(f_2;(z_i,z_i')))^2}$. Here, the first inequality is by McDiarmid's inequality (Lemma \ref{lemma Mcdiarmid}); and the second inequality is by Lemma \ref{lemma truncated integral}; and the integral is capped at $2$ because
\[
\sup_{f_1\in\cH,f_2\in\cH}\tilde\varrho(f_1,f_2)\leq \sup_{f_1\in\cH,f_2\in\cH} \sqrt{2}\varrho_{(S,S'),\ell}(f_1,f_2)\leq 2,
\]where we have used  \eqref{eq: S S' metrics}. By the ambient–equivalence of the pointwise–dimension functional (Lemma~\ref{lemma PD equivalence}),
 we have (note that we take the support of $\pi$ to be $\cF$ rather than $\cH$)
\begin{align*}
&\inf_{\mu\in\Delta(\cH)}\sup_{f\in\cH}\inf_{0\le\alpha\le2}
\left\{\alpha+\frac1{\sqrt n}\int_\alpha^2
\sqrt{\log\frac1{\mu(B_{\tilde\varrho}(f,\varepsilon))}}\,d\varepsilon\right\}\\
&\qquad\le2\inf_{\pi\in\Delta(\cF)}\sup_{f\in\cH}\inf_{0\le\alpha\le2}
\left\{\alpha+\frac1{\sqrt n}\int_\alpha^2
\sqrt{\log\frac1{\pi(B_{\tilde\varrho}(f,\varepsilon))}}\,d\varepsilon\right\}.
\end{align*}
By \eqref{eq: S S' metrics} and the fact that pointwise dimension is monotone in the underlying metric (Lemma \ref{lemma simple metric domination}), we have that for any $\pi\in\Delta(\cF)$,
\begin{align*}
   \int_{\alpha}^{2}
\sqrt{\log\frac{1}{\pi\!\bigl(B_{\tilde\varrho}(f,\varepsilon)\bigr)}}\,d\varepsilon
\leq \int_{\alpha}^{2}
\sqrt{\log\frac{1}{\pi\!\bigl(B_{\sqrt{2}\varrho_{(S,S'), \ell}}(f,\varepsilon)\bigr)}}\,d\varepsilon
= \sqrt{2}\int_{\alpha/\sqrt{2}}^{\sqrt{2}}
\sqrt{\log\frac{1}{\pi\!\bigl(B_{\varrho_{(S,S'), \ell}}(f,\varepsilon)\bigr)}}\,d\varepsilon,
\end{align*}
where the equality follows by a change of variables.
Combining the above three inequalities, we prove the following {\it subset-homogeneous} uniform convergence argument when choosing an arbitrary $\pi\in\Delta(\cF)$: conditioned on $(S,S')$, for any  $\cH\subseteq\cF$ and $\delta\in(0,1)$, with probability at least $1-\delta$ (the randomness all comes from $\{\xi_i\}_{i=1}^n$),
\begin{align}\label{eq: subset homogeneous symmetric}
\sup_{f\in\cH}\frac{1}{n}\sum_{i=1}^n \xi_i (\ell(f;z_i')-\ell(f;z_i))\leq
\sup_{f\in\cH}C_1\inf_{\alpha\geq 0}\left\{
\alpha+\frac{1}{\sqrt{n}}\int_{\alpha}^{\sqrt{2}}
\sqrt{\log\frac{1}{\pi\!\bigl(B_{\varrho_{(S,S'),\ell}}(f,\varepsilon)\bigr)}}\,d\varepsilon
\right\}+\sqrt{\frac{2\log\frac{1}{\delta}}{n}},
\end{align}
where $C_1=2\sqrt{2}C_0>0$ is an absolute constant.

Conditioned on $(S,S')$, for fixed $\pi_{(S,S')}\in\Delta(\cF)$ that is independent with $\{\xi_i\}_{i=1}^n$, define 
the pointwise complexity
\begin{align}\label{eq: data independent pointwise complexity}
d_{S,S'}(f)
:=\left(\inf_{\alpha\geq0}\left\{\sqrt{n}\,\alpha+\int_{\alpha}^{\sqrt{2}}
\sqrt{\log\frac{1}{\pi_{(S,S')}\!\bigl(B_{\varrho_{(S,S'),\ell}}(f,\varepsilon)\bigr)}}\,d\varepsilon\right\}\right)^{\!2}.
\end{align}
Then, by \eqref{eq: subset homogeneous symmetric}, conditioned on $(S,S')$, for any  $\cH\subseteq\cF_R$ and any $\delta\in(0,1)$, with probability at least $1-\delta$ (the randomness all comes from $\{\xi_i\}_{i=1}^n$),
\begin{align}\label{eq: step 1 thm 1}
\sup_{f\in\cH}\frac{1}{n}\sum_{i=1}^n \xi_i(\ell(f;z_i')-\ell(f;z_i))
\;\le\;
\,\sup_{f\in\cH} \left(C_1\sqrt{\frac{d_{S,S'}(f)}{n}}
\;+\;\sqrt{\frac{2\log\frac{2}{\delta}}{n}}\right).
\end{align}
As discussed in Appendix \ref{appendix necessary sufficient}, this condition is both necessary and sufficient to establish pointwise convergence when the complexity functional is the $\{\xi_i\}_{i=1}^n$-independent $d_{S,S'}(\cdot)$ when conditioned on $(S,S')$.

\paragraph{Step 2: Generic Conversion to Pointwise Generalization Bound.}

All the analysis in this step is condition on $(S,S')$, thus all the randomness discussed here comes from $\{\xi_i\}_{i=1}^n$. 
For every $r\in[0,2n]$, we take the subset
\begin{align*}
  \cH=  \{f\in\cF: d_{S,S'}(f)\leq r\},
\end{align*}
which, by \eqref{eq: step 1 thm 1}, implies that 
$\forall \delta\in(0,1)$ and $\forall r\in [0, 2n]$, with probability at least $1-\delta$ 
\begin{align}\label{eq: condition surrogate}
    \sup_{f: d_{S,S'}(f)\leq r}\frac{1}{n}\sum_{i=1}^n \xi_i(\ell(f;z_i')-\ell(f;z_i))\leq C_1\sqrt{\frac{r}{n}}+\sqrt{\frac{2\log\frac{2}{\delta}}{n}},
\end{align}
where 
and $C_1$ is an absolute constant. The inequality \eqref{eq: condition surrogate} is precisely the condition \eqref{eq: surrogate} in the generic conversion provided in Lemma \ref{lemma uniformed localized convergence} (here, the expectation (equal to $0$) and the empirical average are taken for $\{\xi_i\}_{i=1}^n$). Thus applying  Lemma \ref{lemma uniformed localized convergence} we have the pointwise generalization bound: conditioned on $(S,S')$, for any $\delta\in(0,1)$, by taking $r_0=1/n$, with probability at least $1-\delta$, uniformly over all $f\in \cF$, 
\begin{align}
    \frac{1}{n}\sum_{i=1}^n \xi_i(\ell(f;z_i')-\ell(f;z_i))\leq &\frac{C_1}{\sqrt{n}}\sqrt{\max\left\{2 d_{S,S'}(f),\frac{1}{n}\right\}}+\sqrt{\frac{2\log\frac{2\log_2(4n^2)}{\delta}}{n}}\nonumber\\
    \leq & C_1\sqrt{\frac{2d_{S,S'}(f)}{n}} +\frac{C_1}{n}+\sqrt{\frac{2\log\frac{4\log_2(2n)}{\delta}}{n}}.\nonumber\\
    \leq &C_2\left(\sqrt{ \frac{d_{S,S'}(f)}{n}}+\sqrt{\frac{\log\frac{2\log(2n)}{\delta}}{n}}\right),\label{eq: pointwise data independent}
\end{align}
where $C_2>0$ is an absolute constant, where the second inequality is because there exists $C_2\geq \sqrt{2}C_1$ such that for all positive integer $n$,
\[
\frac{C_1}{n}+\sqrt{\frac{2\left(\log\frac{1}{\delta}+\log(\log2+\log n)+\log\frac{4}{\log 2}\right)}{n}}\leq C_2\sqrt{\frac{\log\frac{1}{\delta}+\log(2(\log 2+\log n))}{n}}.
\]
Thus we prove the pointwise generalization bound \eqref{eq: pointwise data independent} for the complexity functional $d_{S,S'}(\cdot)$ defined in \eqref{eq: data independent pointwise complexity}, under the randomness of $\{\xi_i\}_{i=1}^n$: for all $\pi\in\Delta(\cF)$, conditioned on $(S,S')$, for any $\delta\in(0,1)$, with probability at least $1-\delta$, uniformly over all $f\in\cF$, 
\begin{align}
    &\frac{1}{n}\sum_{i=1}^n\xi_i(\ell(f;z_i')-\ell(f;z_i))\nonumber\\
    \leq &C_2\left(\inf_{\alpha\geq 0}\left\{\alpha+\frac{1}{\sqrt{n}}\int_{\alpha}^{\sqrt{2}}\sqrt{\log \frac{1}{\pi_{(S,S')}(B_{\varrho_{(S,S'),\ell}}(f,\varepsilon))}}d\varepsilon\right\}+\sqrt{\frac{\log\frac{2\log(2n)}{\delta}}{n}}\right).\label{eq: xi pointwise generalization}
\end{align}

\paragraph{Step 3: High-Probability Symmetrization.} 
Recall that \(S=\{z_i\}_{i=1}^n\) and
\(S'=\{z_i'\}_{i=1}^n\) are i.i.d. samples, independent of each other, and 
\(\{\xi_i\}_{i=1}^n\) are i.i.d.\ Rademacher signs, independent of \((S,S')\).
The mixed (ghost) metric
\[
\varrho_{(S,S'),\ell}(f_1,f_2)
=\sqrt{(\bP_S+\bP_{S'})\bigl(\ell(f_1;z)-\ell(f_2;z)\bigr)^2},
\]
is swap-invariant to the pair  $(z_i,z_i')$ for each $i=1,\cdots,n$. By the definition of swap-invariant prior before Theorem \ref{thm ghost}, the  prior \(\pi_{(S,S')}\in\Delta(\cF)\) is also swap-invariant to the pair  $(z_i,z_i')$.  

Denote the functionals
\begin{align*}
    &X(f;S,S';\delta)\\:=&\frac{1}{n}\sum_{i=1}^n(\ell(f;z_i')-\ell(f;z_i))\\&- C_2\left(\inf_{\alpha\geq 0}\left\{\alpha+\frac{1}{\sqrt{n}}\int_{\alpha}^{\sqrt{2}}\sqrt{\log \frac{1}{\pi_{(S,S')}(B_{\varrho_{(S,S'),\ell}}(f,\varepsilon))}}d\varepsilon\right\}+\sqrt{\frac{\log\frac{2\log(2n)}{\delta}}{n}}\right),
\end{align*}
and 
\begin{align*}
   &Y(f;S,S',\{\xi_i\}_{i=1}^n;\delta)\\:= &\frac{1}{n}\sum_{i=1}^n\xi_i(\ell(f;z_i')-\ell(f;z_i))\\
   &-C_2\left(\inf_{\alpha\geq 0}\left\{\alpha+\frac{1}{\sqrt{n}}\int_{\alpha}^{\sqrt{2}}\sqrt{\log \frac{1}{\pi_{(S,S')}(B_{\varrho_{(S,S'),\ell}}(f,\varepsilon))}}d\varepsilon\right\}+\sqrt{\frac{\log\frac{2\log(2n)}{\delta}}{n}}\right).
\end{align*}

\paragraph{Symmetry Argument.} We write \( \stackrel{d}{=}\) to denote equality in distribution (i.e., the random variables have the same law, equivalently the same cumulative distribution function). For each \(i\in\{1,\dots,n\}\), let \(\tau_i(S,S')\) be the pair
obtained by swapping \(z_i\) and \(z_i'\). Since \(\varrho_{(S,S'),\ell}\) and $\pi_{(S,S')}$ are invariant under
\((S,S')\mapsto \tau_i(S,S')\) and \(\tau_i(S,S')\stackrel{d}{=}(S,S')\), we have, for all \(t\in\mathbb{R}\),
\begin{align*}
&\Pr\!\bigl(Y(f;S,S',\{\xi_i\}_{i=1}^n;\delta)\le t\bigr)\\
= &\tfrac12\,\Pr\!\bigl(Y(f;S,S',\{\xi_i\}_{i=1}^n;\delta)\le t|\xi_i=1\bigr)
+ \tfrac12\,\Pr\!\bigl(Y(f;S,S',\{\xi_i\}_{i=1}^n;\delta)\le t|\xi_i=-1\bigr)\\
= &\Pr\!\bigl(Y(f;S,S',\{\xi_i\}_{i=1}^n;\delta)\le t|\xi_i=1\bigr),
\end{align*}
i.e.,\ \(Y(f;S,S',\{\xi_i\}_{i=1}^n;\delta)\stackrel{d}{=}Y(f;S,S',\{\xi_1,\cdots, \xi_{i-1},1,\xi_{i+1},\cdots, \xi_n\};\delta)\). 
In the second equality, we use the following joint-law symmetry. The map
\[
(S,S',\{\xi_j\}_{j=1}^n)\longmapsto
\bigl(\tau_i(S,S'),\{\xi_1,\ldots,\xi_{i-1},-\xi_i,\xi_{i+1},\ldots,\xi_n\}\bigr)
\]
preserves the joint law of \((S,S',\{\xi_j\}_{j=1}^n)\), because
\((z_i,z_i')\) are i.i.d. and \(\xi_i\) is symmetric. Moreover, by the
swap-invariance of \(\varrho_{(S,S'),\ell}\) and \(\pi_{(S,S')}\), this
transformation leaves the value of the process \(Y\) unchanged. Hence the two distributions coincide, and
\[
\Pr\bigl(Y(f;S,S',\{\xi_i\}_{i=1}^n;\delta)\le t\mid\xi_i=1\bigr)
= \Pr\bigl(Y(f;S,S',\{\xi_i\}_{i=1}^n;\delta)\le t\mid\xi_i=-1\bigr).
\]
Iterating over all indices $i=1,\cdots, n$, we obtain that 
\begin{align*}
   Y(f;S,S',\{\xi_i\}_{i=1}^n;\delta) \stackrel{d}{=}Y(f;S,S',\{1,\cdots, 1\};\delta)=X(f;S,S';\delta).
\end{align*}

By the conclusion \eqref{eq: xi pointwise generalization} in Step~2 and the
tower property, we have
\[
\Pr_{S,S',\xi}\Big(
Y(f;S,S',\{\xi_i\}_{i=1}^n;\delta)\le 0
\text{ for all } f\in\mathcal F
\Big)
\ge 1-\delta .
\]
Moreover, the symmetry argument above gives the process-level equality in
distribution
\[
\{Y(f;S,S',\{\xi_i\}_{i=1}^n;\delta)\}_{f\in\mathcal F}
\stackrel d=
\{X(f;S,S';\delta)\}_{f\in\mathcal F}.
\]
Therefore,
\[
\Pr_{S,S'}\Big(
X(f;S,S';\delta)\le 0
 \text{ for all }  f\in\mathcal F
\Big)
\ge 1-\delta .
\]

Hence, with probability at least $1-\delta$ over the draw of $(S,S')$, we
have, uniformly over all $f\in\mathcal F$,
\begin{align*}
&\frac1n\sum_{i=1}^n\bigl(\ell(f;z_i')-\ell(f;z_i)\bigr)
\\\;\le\;
&C_2\Bigg(
  \inf_{\alpha\ge0}\Big\{\alpha+\frac1{\sqrt n}
   \int_{\alpha}^{\sqrt{2}}\sqrt{\log\frac1{\pi_{(S,S')}(B_{\varrho_{(S,S'),\ell}}(f,\varepsilon))}}
   \,d\varepsilon\Big\}
+\sqrt{\frac{\log(2\log(2n)/\delta)}{n}}
\Bigg),
\end{align*}
where \(\varrho_{(S,S'),\ell}(f_1,f_2)
= \sqrt{(\mathbb{P}_S+\mathbb{P}_{S'})\bigl(\ell(f_1;z)-\ell(f_2;z)\bigr)^2}\), and $C_2>0$ is an absolute constant.

 \hfill$\square$

\subsection{Proof of Theorem~\ref{thm generic chaining}}
We first retain the bound with a conditional ghost expectation as an
intermediate lemma.
\begin{lemma}[Conditional Ghost Bound]\label{lemma conditional ghost}
Let $\ell(f;z)\in[0,1]$, and let $\pi$ be a prior on $\cF$ independent
of both $S$ and $S'$. There is an absolute constant $C>0$ such that,
for every $n\ge1$ and $\delta\in(0,1)$, with probability at least
$1-\delta$ over $S$, uniformly over $f\in\cF$,
\begin{align*}
(\bP-\bP_n)\ell(f;z)
\le C\Bigg(&\mathbb E_{S'}\!\left[
\inf_{0\le\alpha\le\sqrt2}\left\{
\alpha+\frac1{\sqrt n}\int_\alpha^{\sqrt2}
\sqrt{\log\frac1{\pi(B_{\varrho_{(S,S'),\ell}}(f,\varepsilon))}}
\,d\varepsilon\right\}\Bigm|S\right]\\
&+\sqrt{\frac{\log(2\log(2n)/\delta)}{n}}\Bigg),
\end{align*}
where $\varrho_{(S,S'),\ell}$ is the mixed empirical-ghost metric from
Theorem~\ref{thm ghost}.
\end{lemma}
\paragraph{Proof of Lemma~\ref{lemma conditional ghost}.}
We consider data-independent prior $\pi$ that is independent to both $S$ and $S'$.
Define
\[
\Psi_{S,S'}(f)
:=
C_0\inf_{\alpha\ge 0}
\left\{
\alpha
+
\frac{1}{\sqrt n}
\int_{\alpha}^{\sqrt2}
\sqrt{
\log \frac{1}{\pi\!\left(B_{\varrho_{(S,S'),\ell}}(f,\varepsilon)\right)}
}\,d\varepsilon
\right\},
\]
where \(C_0>0\) is the absolute constant from Theorem~\ref{thm ghost}. Finally, set
\[
X(S,S')
:=
\sup_{f\in\cF}
\Bigl(
(\bP_{S'}-\bP_S)\ell(f;z)
-
\Psi_{S,S'}(f)
\Bigr)
\]
and
\[
Z(S)
:=
\sup_{f\in\cF}
\Bigl(
(\bP-\bP_S)\ell(f;z)
-
\E_{S'}[\Psi_{S,S'}(f)\mid S]
\Bigr).
\]

\paragraph{Core Poof Idea.} We briefly overview the essential proof idea. The quantity \(X(S,S')\) is the one-sided \emph{ghost-sample fluctuation} controlled by Theorem~\ref{thm ghost}, while \(Z(S)\) is the one-sided fluctuation we ultimately want. The key point is that, conditional on \(S\), the population mean \(\bP\) is exactly the conditional mean of the ghost empirical mean \(\bP_{S'}\). Therefore \(Z(S)\) is dominated by the conditional expectation of \(X(S,S')\):
\begin{align*}
    Z(S)\leq \mathbb{E}_{S'}[X(S,S')|S].
\end{align*} Since we only require a one-sided upper-tail bound, we may transfer the ghost-sample tail bound to $Z(S)$ by applying Jensen’s inequality to the exponential map, namely, to the moment generating function (MGF). This direct tail-to-MGF argument avoids both the unnecessary intermediate tail-to-expectation conversion and a separate McDiarmid step, which is designed for two-sided bounded-differences control and could lead to a worse-rate bound here unless additional techniques are introduced.

\paragraph{Step 1: Comparison between the Target Process and the Ghost Process.}
For each fixed \(f\in\cF\),
\[
(\bP-\bP_S)\ell(f;z)
=
\E_{S'}\big[(\bP_{S'}-\bP_S)\ell(f;z)\mid S\big].
\]
Hence
\[
(\bP-\bP_S)\ell(f;z)
-
\E_{S'}[\Psi_{S,S'}(f)\mid S]
=
\E_{S'}\big[(\bP_{S'}-\bP_S)\ell(f;z)-\Psi_{S,S'}(f)\mid S\big].
\]
Taking the supremum over \(f\in\cF\) and using the generic comparison
\[
\sup_{f\in\cF}\E_{S'}[A_f\mid S]
\le
\E_{S'}\!\left[\sup_{f\in\cF}A_f\mid S\right],
\]
we obtain
\begin{align}\label{eq: target ghost comparison}
Z(S)
\le
\E_{S'}[X(S,S')\mid S].
\end{align}

\paragraph{Step 2: Ghost-Sample Upper Tail.}
Because the prior \(\pi\) is independent of both \(S\) and \(S'\), Theorem~\ref{thm ghost} yields that for every \(\delta\in(0,1)\),
\[
\Pr\!\left(
X(S,S')
>
C_0\sqrt{\frac{\log(2\log(2n)/\delta)}{n}}
\right)
\le \delta.
\]
Then for every \(t\ge 0\),
\[
\Pr\!\left(X(S,S')-C_0\sqrt{\frac{\log(2\log(2n))}{n}}>t\right)\leq \Pr\!\left(X(S,S')>\sqrt{C_0^2\frac{\log(2\log(2n))}{n}+t^2}\right)
\le
\exp\!\left(-\frac{n t^2}{C_0^2}\right).
\]
Therefore, with
\[
Y:=(X(S,S')-b_n)_+, \quad b_n:=C_0\sqrt{\frac{\log(2\log(2n))}{n}},
\]
we have the sub-Gaussian tail bound
\begin{align}\label{eq: tail upper}
\Pr(Y>t)\le \exp\!\left(-\frac{n t^2}{C_0^2}\right),
\qquad t\ge 0.
\end{align}

\paragraph{Step 3: Exponential-Moment Transfer.}
Let \(\lambda>0\). Since \(x\mapsto e^{\lambda x}\) is convex, Jensen's inequality gives
\[
e^{\lambda Z(S)}
\le
e^{\lambda \E_{S'}[X(S,S')\mid S]}
\le
\E_{S'}\!\left[e^{\lambda X(S,S')}\mid S\right],
\]
where the first inequality is due to \eqref{eq: target ghost comparison}.
Taking expectation over \(S\),
\[
\E\big[e^{\lambda Z(S)}\big]
\le
\E\big[e^{\lambda X(S,S')}\big].
\]
Since \(X(S,S')\le b_n+Y\), it follows that
\[
\E\big[e^{\lambda Z(S)}\big]
\le
e^{\lambda b_n}\E[e^{\lambda Y}].
\]

We now bound the moment generating function of \(Y\). Using the tail-integral formula for a nonnegative random variable,
\begin{align}\label{eq: expectation nonnegative formula}
\E[e^{\lambda Y}]
=
1+\lambda\int_0^\infty e^{\lambda t}\Pr(Y>t)\,dt.
\end{align}
Integrating the tail bound \eqref{eq: tail upper} on \(Y\) to \eqref{eq: expectation nonnegative formula}, we have the moment generating function bound
\[
\E[e^{\lambda Y}]
\le
1+\lambda\int_0^\infty
\exp\!\left(\lambda t-\frac{n t^2}{C_0^2}\right)\,dt.
\]
Completing the square,
\[
\lambda t-\frac{n t^2}{C_0^2}
=
-\frac{n}{C_0^2}
\left(t-\frac{C_0^2\lambda}{2n}\right)^2
+
\frac{C_0^2\lambda^2}{4n},
\]
so
\[
\E[e^{\lambda Y}]
\le
1+\lambda e^{C_0^2\lambda^2/(4n)}
\int_{-\infty}^{\infty}
\exp\!\left(-\frac{n s^2}{C_0^2}\right)\,ds.
\]
The Gaussian integral equals \(C_0\sqrt{\pi/n}\), and therefore
\[
\E[e^{\lambda Y}]
\le
1+
C_1\frac{\lambda}{\sqrt n}
\exp\!\left(\frac{C_0^2\lambda^2}{4n}\right)
\]
for some absolute constant \(C_1>0\). Using the elementary bound
\[
x\le e^{x^2/2},
\qquad x\ge 0,
\]
with \(x=\lambda/\sqrt n\), we obtain
\[
\frac{\lambda}{\sqrt n}
\le
\exp\!\left(\frac{\lambda^2}{2n}\right).
\]
Hence
\[
\E[e^{\lambda Y}]
\le
1+
C_1
\exp\!\left(C_2\frac{\lambda^2}{n}\right)
\le
C_3
\exp\!\left(C_2\frac{\lambda^2}{n}\right)
\]
for some absolute constants \(C_2,C_3>0\). Consequently,
\[
\E\big[e^{\lambda Z(S)}\big]
\le
C_3
\exp\!\left(
\lambda b_n
+
C_2\frac{\lambda^2}{n}
\right).
\]

\paragraph{Step 4: Chernoff Bound to Optimize $\lambda$.}
For any \(t>0\) and any \(\lambda>0\),
\[
\Pr\!\left(Z(S)>b_n+t\right)
\le
e^{-\lambda(b_n+t)}\E[e^{\lambda Z(S)}]
\le
C_3
\exp\!\left(
-\lambda t + C_2\frac{\lambda^2}{n}
\right).
\]
Optimizing at
\[
\lambda=\frac{n t}{2C_2},
\]
we obtain
\[
\Pr\!\left(Z(S)>b_n+t\right)
\le
C_3\exp\!\left(-\frac{n t^2}{4C_2}\right).
\]
Thus, for every \(\delta\in(0,1)\), choosing
\[
t=\sqrt{\frac{4C_2}{n}\log\frac{C_3}{\delta}},
\]
we get
\[
\Pr\!\left(
Z(S)>
b_n+\sqrt{\frac{4C_2}{n}\log\frac{C_3}{\delta}}
\right)
\le
\delta.
\]

Finally, since
\[
b_n=C_0\sqrt{\frac{\log(2\log(2n))}{n}},
\]
there exists an absolute constant \(C>0\) such that
\[
b_n+\sqrt{\frac{4C_2}{n}\log\frac{C_3}{\delta}}
\le
C\sqrt{\frac{\log(2\log(2n)/\delta)}{n}}.
\]
Therefore, with probability at least \(1-\delta\),
\[
Z(S)
=
\sup_{f\in\cF}
\Bigl(
(\bP-\bP_S)\ell_f
-
\E_{S'}[\Psi_{S,S'}(f)\mid S]
\Bigr)
\le
C\sqrt{\frac{\log(2\log(2n)/\delta)}{n}}.
\]
Equivalently, uniformly over all \(f\in\cF\),
\[
(\bP-\bP_n)\ell(f;z)
\le
\E_{S'}\!\left[
C_0\inf_{\alpha\ge 0}
\left\{
\alpha
+
\frac{1}{\sqrt n}
\int_{\alpha}^{\sqrt2}
\sqrt{
\log \frac{1}{\pi\!\left(B_{\varrho_{(S,S'),\ell}}(f,\varepsilon)\right)}
}\,d\varepsilon
\right\}
\Bigm| S
\right]
+
C\sqrt{\frac{\log(2\log(2n)/\delta)}{n}}.
\]
Hence we finish the proof of Lemma~\ref{lemma conditional ghost}.

\hfill$\square$

\begin{lemma}[Averaging a Random Metric]\label{lemma ghost population averaging}
Let $n\ge1$, $D>0$, and let $\varrho_\omega$ be a jointly measurable
random semi-metric on $\cF$ with diameter at most $D$ almost surely.
Fix a prior $\pi$ on $\cF$ independent of $\omega$, and set
\[
\bar\varrho(f,g):=\bigl(\mathbb E_\omega
\varrho_\omega(f,g)^2\bigr)^{1/2}.
\]
For any semi-metric $\varrho$ of diameter at most $D$, define
\[
I_{\pi,\varrho}(f):=
\inf_{0\le\alpha\le D}\left\{
\alpha+\frac1{\sqrt n}\int_\alpha^D
\sqrt{\log\frac1{\pi(B_\varrho(f,r))}}\,dr\right\}.
\]
Then, for every $f\in\cF$,
\begin{equation}\label{eq: ghost population averaging}
\mathbb E_\omega I_{\pi,\varrho_\omega}(f)
\le\sqrt2\,I_{\pi,\bar\varrho}(f)
+D\sqrt{\frac{\log2}{n}}.
\end{equation}
\end{lemma}

\paragraph{Proof.}
Fix $f$ and define the prior quantile of its distance by
\[
Q_\varrho(u):=\inf\{r\in[0,D]:\pi(B_\varrho(f,r))\ge u\},
\qquad 0<u\le1.
\]
The integrand in $I_{\pi,\varrho}(f)$ is nonincreasing in the radius.
Optimizing the cutoff and then integrating its level sets therefore give
\begin{align}\label{eq: integral quantile representation}
I_{\pi,\varrho}(f)
&=\int_0^D\min\!\left\{1,
\sqrt{\frac{\log(1/\pi(B_\varrho(f,r)))}{n}}\right\}\,dr\nonumber\\
&=\int_0^1Q_\varrho(e^{-nt^2})\,dt.
\end{align}
For $0<u<1/2$, set $r_u=Q_{\bar\varrho}(2u)$,
$G=\{g:\bar\varrho(f,g)\le r_u\}$, and $p=\pi(G)\ge2u$.
Markov's inequality on $G$ gives
\[
Q_{\varrho_\omega}(u)^2
\le\frac1{p-u}\int_G\varrho_\omega(f,g)^2\,\pi(dg).
\]
Taking expectation and using Cauchy-Schwarz yields
\[
\mathbb E_\omega Q_{\varrho_\omega}(u)
\le\left(\frac1{p-u}\int_G\bar\varrho(f,g)^2\,\pi(dg)\right)^{1/2}
\le\sqrt{\frac p{p-u}}\,r_u
\le\sqrt2\,Q_{\bar\varrho}(2u).
\]
Let $t_0=\sqrt{\log2/n}<1$. Splitting
\eqref{eq: integral quantile representation} at $t_0$ and using
$Q_{\varrho_\omega}\le D$, we obtain
\begin{align*}
\mathbb E_\omega I_{\pi,\varrho_\omega}(f)
&\le Dt_0+\sqrt2\int_{t_0}^1
Q_{\bar\varrho}(2e^{-nt^2})\,dt\\
&\le Dt_0+\sqrt2\int_0^1
Q_{\bar\varrho}(e^{-ns^2})\,ds\\
&=D\sqrt{\frac{\log2}{n}}+\sqrt2\,I_{\pi,\bar\varrho}(f).
\end{align*}
The second inequality uses $s=\sqrt{t^2-t_0^2}$, for which
$2e^{-nt^2}=e^{-ns^2}$ and $dt/ds\le1$.
\hfill$\square$

\paragraph{Proof of Theorem~\ref{thm generic chaining}.}
Condition on $S$ and apply Lemma~\ref{lemma ghost population averaging}
with $\omega=S'$, $D=\sqrt2$, and
$\varrho_\omega=\varrho_{(S,S'),\ell}$. Independence of $S'$ and $S$
gives, for every $f,g\in\cF$,
\[
\mathbb E_{S'}\!\left[
\varrho_{(S,S'),\ell}(f,g)^2\mid S\right]
=(\bP_S+\bP)\bigl(\ell(f;z)-\ell(g;z)\bigr)^2
=\varrho_{(S,\mathbb P),\ell}(f,g)^2.
\]
Consequently, for every $S$ and $f$,
\[
\mathbb E_{S'}\!\left[
I_{\pi,\varrho_{(S,S'),\ell}}(f)\mid S\right]
\le\sqrt2\,I_{\pi,\varrho_{(S,\mathbb P),\ell}}(f)
+\sqrt{\frac{2\log2}{n}}.
\]
Substitute this deterministic comparison into
Lemma~\ref{lemma conditional ghost}. Since
$\log(2\log(2n)/\delta)\ge\log(2\log2)>0$ for $n\ge1$ and
$\delta\in(0,1)$, the additional $n^{-1/2}$ term is absorbed into the
confidence term by enlarging an absolute constant. Hence we finish the proof of
Theorem~\ref{thm generic chaining}.\hfill$\square$

\subsection{Proof of Theorem~\ref{thm lower bound}}
\label{appendix mixed lower bound}
Throughout this proof, $A_\pi(f)$ is the infimum in \eqref{eq: one shot}, and
$C_\pi^\rho(f)$ is the infimum in Theorem~\ref{thm generic chaining} with metric $\rho$;
thus $C_\pi(f)=C_\pi^{\varrho_{(S,\bP),\ell}}(f)$.
All priors are on the displayed loss class. Infima over priors are evaluated samplewise; the priors used for the upper comparisons below are fixed before sampling.

\paragraph{One-shot gap-distance lower bound.}
For the one-shot comparison, take the truncation level equal to the diameter in the gap distance, so every ball has prior mass one. This gives
\[
\sup_f A_\pi(f)
\le\sup_{f,g}|(\bP-\bPn)(\ell(f;z)-\ell(g;z))|
\le2\sup_f|(\bP-\bPn)\ell(f;z)|,
\]
proving \eqref{eq gap objective oscillation} for every sample and prior.
\paragraph{Chaining integral lower bound.}
Lemma~\ref{lemma lower R G}, followed by subtracting the Gaussian contribution of the means, gives
\begin{align}\label{eq: empirical lower}
\bE\sup_f|(\bP-\bPn)\ell(f;z)|
\ge\frac{c_1}{\sqrt{\log n}}
\left\{\bE_{S,g}\sup_f\frac1n\sum_{i=1}^n g_i\ell(f;z_i)
-\sqrt{\frac{2}{\pi n}}\sup_f\bP\ell(f;z)\right\}.
\end{align}
Conditional on $S$, the process $n^{-1}\sum_i g_i\ell(f;z_i)$ has canonical metric
$\varrho_{n,\ell}/\sqrt n$. Lemma~\ref{lemma lower bound} therefore yields
\begin{equation}\label{eq: Gaussian lower}
\bE_{S,g}\sup_f\frac1n\sum_{i=1}^n g_i\ell(f;z_i)
\ge\frac{c_2}{\sqrt n}\bE\inf_\pi\sup_f
\int_0^1\sqrt{\log\frac1{\pi(B_{\varrho_{n,\ell}}(f,\varepsilon))}}\,d\varepsilon.
\end{equation}
Combining these inequalities proves the empirical lower bound. Complementation makes the absolute and one-sided suprema equal.

\paragraph{A family on which both bounds are sharp.}
Let $\bP$ be uniform on $\{1,\ldots,d\}$, where $2\le d\le n$, and take all losses
$\ell(f_A;j)=\mathbf1_{\{j\in A\}}$, $A\subseteq\{1,\ldots,d\}$.
Let $U$ be uniform on these $2^d$ hypotheses, $N_j=\sum_i\mathbf1_{\{z_i=j\}}$, and $\lambda=n/d$.
Put $a_j=1/d-N_j/n$ and $Q^2=\sum_j a_j^2$. Then
\begin{align*}
\sup_f|(\bP-\bPn)\ell(f;z)|&=\sup_f(\bP-\bPn)\ell(f;z)=\frac12\sum_j|a_j|,\\
\bE Q^2&=\frac{1-1/d}{n},\qquad
\bE\left(\sum_j|a_j|\right)^2\le\frac dn.
\end{align*}
For $Y=N_1-\lambda$, we have $\bE Y^2\ge\lambda/2$ and
$\bE Y^4\le3\lambda^2+\lambda\le4\lambda^2$.
H\"older's inequality gives
$\bE|Y|\ge(\bE Y^2)^{3/2}/(\bE Y^4)^{1/2}$; hence
\begin{equation}\label{eq cube gap lower}
\frac1{8\sqrt2}\sqrt{\frac dn}
\le \bE\sup_f|(\bP-\bPn)\ell(f;z)|=\frac d{2n}\bE|Y|
\le\frac12\sqrt{\frac dn}.
\end{equation}

The untruncated comparison follows directly from Gaussian widths.
For weights $w_j\ge0$, the metric
$\rho_w(f_A,f_B)^2=\sum_{j\in A\triangle B}w_j$
is induced by $X_A=\sum_{j\in A}\sqrt{w_j}g_j$, and
$\bE_g\sup_A X_A=(2\pi)^{-1/2}\sum_j\sqrt{w_j}$.
The optimized majorizing-measure integral therefore lies between constant multiples of $\sum_j\sqrt{w_j}$.
The empirical and mixed weights are $N_j/n$ and $N_j/n+1/d$, respectively.
The mixed sum lies between $\sqrt d$ and $\sqrt{2d}$.
Since $\bE N_j^2\le2\lambda^2$, Paley-Zygmund gives
$\bP\{N_j\ge\lambda/2\}\ge1/8$. Jensen's inequality supplies the upper bound, proving
\begin{equation}\label{eq cube integral equivalence}
c\sqrt d\le\bE\inf_\pi\sup_f\int_0^{\sqrt2}
\sqrt{\log\frac1{\pi(B_\rho(f,\varepsilon))}}\,d\varepsilon\le C\sqrt d
\end{equation}
for either $\rho=\varrho_{n,\ell}$ or $\rho=\varrho_{(S,\bP),\ell}$.

\emph{The individually truncated chaining objective.}
The following ball argument also applies to the rare-atom construction below.
Suppose $k$ coordinates of $w$ are at least $b>0$ and set $r=\sqrt{bk}/2$.
A ball of radius $r$ differs from its center on at most $k/4$ of those coordinates, so
Hoeffding's inequality gives $U(B_{\rho_w}(f,r))\le e^{-k/8}$ for every $f$.
For any prior $\pi$, symmetry gives
\[
\bE_{f\sim U}\pi(B_{\rho_w}(f,r))
=\bE_{g\sim\pi}U(B_{\rho_w}(g,r))\le e^{-k/8}.
\]
Thus a single center $f$ has ball mass at most $e^{-k/8}$. At this center, for every $\alpha$,
the chaining objective is at least $\alpha+(r-\alpha)_+\sqrt{k/(8n)}$. Whenever $r\le\sqrt2$, this proves
\begin{equation}\label{eq cube ball truncation}
\inf_\pi\sup_f C_\pi^{\rho_w}(f)\ge\frac{\sqrt{bk}}2\min\left\{1,\sqrt{\frac{k}{8n}}\right\}.
\end{equation}
For the mixed metric, take $k=d$, $b=1/d$.
For the empirical metric, take $K=\#\{j:N_j\ge\lambda/2\}$, $k=K$, $b=1/(2d)$;
then $\inf_\pi\sup_f C_\pi^{\varrho_{n,\ell}}(f)\ge K/(8\sqrt{dn})$ and $\bE K\ge d/8$.
The case $K=0$ gives the trivial lower bound zero.
Conversely, singleton mass $2^{-d}$ and $\alpha=0$ give
$\sup_f C_U^\rho(f)\le\sqrt{2d\log2/n}$ for either metric.
Together with $\varrho_{n,\ell}\le\varrho_{(S,\bP),\ell}$, this gives
\begin{align}\label{eq cube truncated equivalence}
c\sqrt{\frac dn}
\le\bE\inf_\pi\sup_f C_\pi^{\varrho_{n,\ell}}(f)
\le\bE\inf_\pi\sup_f C_\pi(f)
\le\bE\sup_f C_U(f)\le C\sqrt{\frac dn}.
\end{align}

\emph{The gap-distance one-shot objective.}
Let $f_*$ maximize $(\bP-\bPn)\ell(f;z)$. Under $U$, its gap distance to a random hypothesis has the distribution
$\sum_j|a_j|B_j$, where the $B_j$ are independent Bernoulli variables of mean $1/2$.
If $Q>0$, Hoeffding's inequality gives
\[
U(B_{\varrho^{\mathrm{gap}}_{(S,\bP),\ell}}(f_*,\varepsilon))
\le\exp\left(-\frac{(\sum_j|a_j|)^2}{8Q^2}\right)
\quad\text{for }\varepsilon\le\frac14\sum_j|a_j|.
\]
Splitting the optimization at $\frac14\sum_j|a_j|$ yields
\[
\sup_f A_U(f)\ge A_U(f_*)\ge\frac{\sum_j|a_j|}{4(1+\sqrt n Q)}.
\]
This also holds when $Q=0$. Cauchy-Schwarz and the moment bounds above now give
\[
\bE\sup_f A_U(f)
\ge\frac{(\bE\sum_j|a_j|)^2}
{4\{\bE\sum_j|a_j|+\sqrt n\,\bE[Q\sum_j|a_j|]\}}
\ge c\sqrt{\frac dn}.
\]
The reverse bound follows from singleton mass: $\sup_f A_U(f)\le\sqrt{d\log2/n}$.
This proves sharpness of both complexity terms under the same sample-independent prior.

\paragraph{Rare atoms separate the two bounds.}
Give a background atom probability $1-p$ and each of $d\ge8n$ other atoms probability $p/d$.
Retain all subset indicators on the rare atoms and set every loss to zero at the background atom.
Let $K$ count rare observations, so $\bE K=np$, and keep the uniform prior $U$.
The indicator of the unobserved rare atoms has gap at least $7p/8$, while
$\sup_f|(\bP-\bPn)\ell(f;z)|\le p+K/n$. Thus
$7p/8\le\bE\sup_f|(\bP-\bPn)\ell(f;z)|\le2p$.
The mixed weights are $N_j/n+p/d$; \eqref{eq cube ball truncation}, with $k=d$ and $b=p/d$,
gives $\inf_\pi\sup_f C_\pi(f)\ge\sqrt p/2$.
Choosing the truncation level equal to the mixed diameter gives
$\sup_f C_U(f)\le\sqrt{p+K/n}$, whose expectation is at most $\sqrt{2p}$.

The diameter in the gap distance is at most $p+K/n$, so the same radius choice as in
\eqref{eq gap objective oscillation} gives $\bE\sup_f A_U(f)\le2p$.
For the reverse bound, let $f_*$ indicate the unobserved rare atoms.
A gap ball around $f_*$ of radius at most $p/8$ contains only hypotheses with gap at least $3p/4$,
which must include at least $3d/4$ unobserved rare atoms, since observed coordinates have negative gap coefficients $p/d-N_j/n$.
Its $U$-mass is at most $e^{-d/8}$ by Hoeffding's inequality.
Thus $A_U(f_*)\ge\min\{p/8,\sqrt{d/(8n)}\}=p/8$.
This proves \eqref{eq rare chaining separation} and shows that truncation does not restore a universal mixed-metric lower bound.

\paragraph{Scope of tightness.}
The one-shot comparison \eqref{eq gap objective oscillation} depends on the unknown population gaps; it is not an observable guarantee.
Neither objective must match the gap at each hypothesis. In the first construction,
$(\bP-\bPn)\ell(f_\emptyset;z)=0$, while $C_U(f_\emptyset)>0$ for every sample.
For $d=2$,
\[
A_U(f_\emptyset)=\min\left\{\left|\frac{N_1}{n}-\frac12\right|,\sqrt{\frac{\log2}{n}}\right\},
\]
which is positive whenever the counts are unequal.
Finally, truncation and nontruncation are not universally equivalent for the mixed metric:
for the same indicator class with $n$ fixed and unrestricted $d$, the optimized untruncated mixed integral divided by $\sqrt n$
is at least $c\sqrt{d/n}$, whereas every truncated objective is at most $\sqrt2$.
The sharpness assertions concern the complexity terms. At fixed confidence, the expected full upper bounds have the same rate as $\bE\sup_f|(\bP-\bPn)\ell(f;z)|$ on the first family when $d\ge\log(2\log(2n)/\delta)$.
In the second family, taking $p=n^{-1/4}$ and $d=8n$ makes the confidence remainder negligible: the expected one-shot bound and deviation have order $n^{-1/4}$, while the expected chaining bound has order $n^{-1/8}$.\hfill$\square$

\subsection{Background on Gaussian and Empirical Processes}

 It is now well understood that the supremum of Gaussian process can be tightly characterized by the majorizing measure integral  via matching upper and lower bounds up to absolute constants \citep{fernique1975regularite, talagrand1987regularity}; the goal of this section is to extend this characterization to (1)  bounded empirical processes  and (2) a truncated form of integral.

\paragraph{Background on Gaussian Processes.}
We begin by recalling several key results from a series of seminal papers
by  Talagrand, Fernique, and others, which introduced the majorizing‐measure formulation of the generic chaining framework \citep{fernique1975regularite, talagrand1987regularity}. Note that generic chaining have several equivament formulations \citep{talagrand2014upper}, and the one closest to our purpose is through majorizing measure.

A \emph{centered Gaussian random variable} \(X\) is a real-valued measurable function on the outcome space such that the law of \(X\) has density
\[
(2\pi\sigma^2)^{-1/2}\,\exp\!\Bigl(-\tfrac{x^2}{2\sigma^2}\Bigr).
\]
The law of \(X\) is thus determined by \(\sigma=(\mathbb{E}[X^2])^{1/2}\).  If \(\sigma=1\), \(X\) is called \emph{standard normal}.

A \emph{Gaussian process} is a family \(\{X_t\}_{t\in T}\) of random variables indexed by some set \(T\), such that every finite linear combination
\(\sum_{j=1}^k \alpha_j X_{t_j}\)
is Gaussian.  On the index set $T$, consider the semi-metric $\varrho$ given by
\begin{align}\label{eq: Gaussian process metric}
   \varrho(u,v)=\sqrt{\bE[(X_u-X_v)^2]}.
\end{align} Gaussian processes are thus a very rigid class of stochastic processes, with exceptionally nice properties that have been fully developed in the literature.

\cite{fernique1975regularite} proved the following integral upper bound.
\begin{lemma}[Upper Bound of Gaussian Processes via Majorizing Measure, \cite{fernique1975regularite}]\label{lemma Gaussian Processes Upper Bound}
   Given a Gaussian process $(X_t)_{t\in T}$ with its metric $\varrho$ defined by \eqref{eq: Gaussian process metric}, we have
\begin{align*}
    \bE\left[\sup_{t\in T} X_t\right]\leq C\inf_{\pi\in\Delta(T)}\sup_{t\in T} \int_{0}^\infty \sqrt{\log\frac{1}{\pi(B_{\varrho}(t,\varepsilon))}}d\varepsilon,
\end{align*}
where $C>0$ is an absolute constant.
\end{lemma}

A prior $\pi$ that makes the right hand side in Lemma \ref{lemma Gaussian Processes Upper Bound} finite is called a {\it majorizing measure}. Fernique conjectured as early as 1974 that the existence of majorizing measures might characterize the boundedness of Gaussian processes. He proved a number of important
partial results, and his determination eventually motivated the Talagrand to attack the
problem in 1987.  \cite{talagrand1987regularity} proved that the integral  in Lemma \ref{lemma Gaussian Processes Upper Bound} is tight up to absolute constants. The pointwise integral on the right-hand side of Lemma~\ref{lemma Gaussian Processes Upper Bound} is known as the Fernique-Talagrand (majorizing measure) integral; see \citet[Section~6.2]{talagrand2014upper} for a standard treatment and \citet{chu2023majorizing} for a recent reference. A useful variance-adaptive refinement, whose integration range depends on the pointwise standard deviation $\sigma(t)$, is developed in \citet{xu2026pointwise}.

\begin{lemma}[Lower Bound of Gaussian Processes via Majorzing Measure, \cite{talagrand1987regularity}]\label{lemma lower bound} Given a Gaussian process $(X_t)_{t\in T}$ with its metric $\varrho$ defined by \eqref{eq: Gaussian process metric}, we have 
\begin{align*}
    \bE\left[ \sup_{t\in T} X_t\right]\geq c\inf_{\pi\in\Delta(T)}\sup_{t\in T}\int_0^\infty\sqrt{\log\frac{1}{\pi(B_{\varrho}(t,\varepsilon))}}d\varepsilon,
\end{align*}
where $c>0$ is an absolute constant.
\end{lemma}

Thus, the globally optimized Fernique-Talagrand integral, with the infimum over measures and the supremum over index points, characterizes the expected supremum of a Gaussian process up to universal constants.

\paragraph{Background on Empirical Processes.}

We now give several results on upper and lower bounding empirical process by Rademacher and Gaussian complexities \cite{gine1984some, bartlett2002rademacher}.

\begin{definition}[Rademacher and Gaussian complexities]\label{def Rademacher Gaussian complexities} For a function class $\cF$ that consists of mappings from $\cZ$ to $\bR$, define the Rademacher complexity of $\cF$ as 
\begin{align*}
R_n(\cF):=\bE_{z,\xi}\left[\sup_{f\in\cF} \frac{1}{n}\sum_{i=1}^n \xi_i f(z_i)\right],
\end{align*}
    where $\{\xi_i\}_{i=1}^n$ are i.i.d. Rademacher variables; and define the Gaussian complexity of $\cF$ as 
    \begin{align*}
G_n(\cF):=\bE_{z,g}\left[\sup_{f\in\cF} \frac{1}{n}\sum_{i=1}^n g_i f(z_i)\right],
    \end{align*}
    where $\{g_i\}_{i=1}^n$ are i.i.d. standard Gaussian variables.
\end{definition}

It is well-known that Rademacher and Gaussian complexities are upper bounds of empirical processes (see, e.g., Lemma 7.4 in \cite{van2014probability}):
\begin{lemma}[Upper Bounds with Rademacher and Gaussian Complexities]\label{lemma upper R G}
    For any function class $\cF$ that consists of mappings from $\cZ$ to $\bR$, we have
    \begin{align*}
        \bE\left[ \sup_{f\in\cF}(\bP-\bPn) f(z)\right]\leq 2R_n(\cF)\leq \sqrt{2\pi}G_n(\cF),
    \end{align*}
    where $R_n(\cF)$ and $G_n(\cF)$ are (expected) Rademacher and Gaussian complexities defined in Definition \ref{def Rademacher Gaussian complexities}.
\end{lemma}

The following lemma strengthens Theorem~3 of \citet{block2021majorizing}
in the batch setting by allowing pointwise optimization of the truncation level inside the class-wide supremum.
More broadly, this strengthening introduces a new type of truncated chaining or covering integrals for bounding suprema of empirical processes, with the truncation level optimized \emph{separately} at each index. It also serves as a supporting step in the proof of Theorem~\ref{thm ghost}.
\begin{lemma}[Integral Bound with Separately Optimized Truncation Level]\label{lemma truncated integral}
Let $\cF$ be a nonempty separable class of functions from $\cZ$ to $[0,1]$,
and fix $z_1,\ldots,z_n\in\cZ$. Define the empirical semi-metric
\[
\varrho_n(f_1,f_2)\ :=\sqrt{\frac1n\sum_{i=1}^n\left(f_1(z_i)-f_2(z_i)\right)^2}.
\]
There exists an absolute constant $C>0$ such that
\[
\mathbb{E}_\xi\left[\sup_{f\in\cF}\frac{1}{n}\sum_{i=1}^n \xi_i f(z_i)\right]
\ \le\
C\,\inf_{\pi\in\Delta(\cF)}\sup_{f\in\cF}\inf_{0\le\alpha\le1}
\left\{\alpha+\frac{1}{\sqrt{n}}
\int_{\alpha}^{1}\sqrt{\log\frac{1}{\pi\bigl(B_{\varrho_n}(f,\varepsilon)\bigr)}}\,d\varepsilon
\right\},
\]
where $\{\xi_i\}_{i=1}^n$ are independent Rademacher variables.
\end{lemma}

\paragraph{Proof.}
Write $J_\pi(f,\alpha)$ for the expression in braces. Theorem~3 of
\citet{block2021majorizing}, applied to trees whose values at each depth
are independent of the path, gives
\[
\mathbb E_\xi\sup_{f\in\cF}\frac1n\sum_{i=1}^n\xi_i f(z_i)
\le C'\inf_{\pi\in\Delta(\cF)}\inf_{0\le\alpha\le1}
\sup_{f\in\cF}J_\pi(f,\alpha),
\]
where we have interchanged only the two infima.
Fix $\pi$ and put $M:=\sup_f\inf_{0\le\alpha\le1}J_\pi(f,\alpha)\le1$.
For $\eta>0$, choose for each $f$ a cutoff $\alpha_f$ such that
$J_\pi(f,\alpha_f)\le M+\eta$.
Nonnegativity of the integral gives $\alpha_f\le M+\eta$.
Hence $a:=\min\{1,M+\eta\}\ge\alpha_f$, and
\[
J_\pi(f,a)\le a+J_\pi(f,\alpha_f)-\alpha_f\le2(M+\eta).
\]
Taking the supremum over $f$ and letting $\eta\downarrow0$ yields
$\inf_\alpha\sup_f J_\pi(f,\alpha)\le2\sup_f\inf_\alpha J_\pi(f,\alpha)$.
The result follows by taking the infimum over $\pi$, with $C=2C'$.
\hfill$\square$

By translation and rescaling, the same bound holds for functions taking
values in an interval of length $D>0$, with the upper endpoint $1$
replaced by $D$ and $0\le\alpha\le D$. In particular, $D=2$ applies
to the symmetrized loss in Theorem~\ref{thm ghost}.

\paragraph{}
The following result illustrate that Gaussian and Rademacher complexities can also be used to lower bounding empirical processes.
\begin{lemma}[Lower Bounds with Rademacher and Gaussian Complexities]\label{lemma lower R G}
Let $n\ge2$ and let $\mathcal F$ be a class of integrable real-valued functions on $\mathcal Z$, with centered class
$\widetilde{\mathcal F}=\{f-\mathbb P f:f\in\mathcal F\}$. Then
\begin{align*}
\mathbb E\sup_{f\in\mathcal F}|(\mathbb P-\mathbb P_n)f|
\ge\frac12 R_n(\widetilde{\mathcal F})
\ge\frac{c}{\sqrt{\log n}}G_n(\widetilde{\mathcal F}),
\end{align*}
where $c>0$ is an absolute constant. If $\widetilde{\mathcal F}$ is symmetric, the absolute value on the left can be omitted.
\end{lemma}
\paragraph{Proof of Lemma~\ref{lemma lower R G}.}
Let $S'$ be an independent copy of $S$. Jensen's inequality and swapping each pair $(Z_i,Z_i')$ when $\xi_i=-1$ give
\begin{align*}
R_n(\widetilde{\mathcal F})
&\le\mathbb E_{S,S',\xi}\sup_{f\in\mathcal F}
\frac1n\sum_{i=1}^n\xi_i(f(Z_i)-f(Z_i'))\\
&=\mathbb E_{S,S'}\sup_{f\in\mathcal F}(\mathbb P_S-\mathbb P_{S'})f
\le2\mathbb E_S\sup_{f\in\mathcal F}|(\mathbb P_S-\mathbb P)f|.
\end{align*}
The Gaussian-Rademacher comparison gives the second inequality; see Problem~7.1 in \citet{van2014probability}. Symmetry of the centered class makes the absolute and one-sided suprema equal.\hfill$\square$

\paragraph{Background on Covering Numbers}

Formally, we give the definition of covering number as follows.

\begin{definition}[Covering Numbers]\label{def covering number}
Let $(\mathcal Y,\varrho)$ be a metric space and let $\mathcal Z\subseteq\mathcal Y$.
For $\varepsilon>0$, a set $\mathcal N\subseteq\mathcal Z$ is an \emph{internal $\varepsilon$–cover} of $\mathcal Z$
if for every $z\in\mathcal Z$ there exists $y\in\mathcal N\subseteq \cZ$ with $\varrho(z,y)\le \varepsilon$.
The (internal) covering number is
\[
\mathsf N(\mathcal Z,\varrho,\varepsilon)
:= \min\{\, m:\ \exists\ \text{internal $\varepsilon$–cover of $\mathcal Z$ with size } m\,\}.
\]
A set $\mathcal N_{\mathrm{ext}}\subseteq\mathcal Y$ (not necessarily inside $\cZ$) is an \emph{external $\varepsilon$–cover} of $\mathcal Z$
if for every $z\in\mathcal Z$ there exists $y\in\mathcal N_{\mathrm{ext}}$ with $\varrho(z,y)\le \varepsilon$.
The external covering number is
\[
\mathsf N_{\mathrm{ext}}(\mathcal Z,\varrho,\varepsilon)
:= \min\{\, m:\ \exists\ \text{external $\varepsilon$–cover of $\mathcal Z$ with size } m\,\}.
\]
Internal covering numbers depend only on the metric induced on $\mathcal Z$, while external covering numbers also depend on
the ambient space $\mathcal Y$. Throughout the paper, ``covering number''
means the internal one unless otherwise stated.
\end{definition}

We now relate the internal and external covering numbers, showing they are equivalent up to a constant factor in the radius—and thus interchangeable for our purposes.

\begin{lemma}[Properties of External Covering Number]\label{lemma int-ext}
For every $\varepsilon>0$ and $\mathcal Z\subseteq\mathcal Y$,
\begin{align}\label{eq: int ext}
\mathsf N_{\mathrm{ext}}(\mathcal Z,\varrho,\varepsilon)
\;\le\;
\mathsf N(\mathcal Z,\varrho,\varepsilon)
\;\le\;
\mathsf N_{\mathrm{ext}}(\mathcal Z,\varrho,\varepsilon/2).
\end{align}
And the external covering number enjoys monotonicity under set inclusion: if $\mathcal Z_1\subseteq\mathcal Z_2$ then
$\mathsf N_{\mathrm{ext}}(\mathcal Z_1,\varrho,\varepsilon)\ \le\ \mathsf N_{\mathrm{ext}}(\mathcal Z_2,\varrho,\varepsilon)$.
\end{lemma}

\paragraph{Proof of Lemma \ref{lemma int-ext}:} 
The left inequality in \eqref{eq: int ext} is immediate since any internal cover is also an external cover. For the right inequality in \eqref{eq: int ext}, let $\{y_1,\dots,y_m\}\subseteq\mathcal Y$ be an external
$(\varepsilon/2)$–cover of $\mathcal Z$.
For each $i$, define the (possibly empty) cell
$V_i:=\{z\in\mathcal Z:\ \varrho(z,y_i)\le \varepsilon/2\}$. By the very definition of external $(\varepsilon/2)$-cover, every $z\in\mathcal Z$ is within distance
$\varepsilon/2$ of some $y_i$; hence
\[
  \bigcup_{i=1}^m V_i \;=\; \mathcal Z .
\] If $V_i\neq\emptyset$,
pick a representative $z_i\in V_i$.
Then for any $z\in V_i$,
\[
\varrho(z,z_i)\ \le\ \varrho(z,y_i)+\varrho(y_i,z_i)\ \le\ \varepsilon/2+\varepsilon/2\ =\ \varepsilon,
\]
so the selected $\{z_i\}\subseteq\mathcal Z$ form an internal $\varepsilon$–cover.
Hence $\mathsf N(\mathcal Z,\varrho,\varepsilon)\le m=\mathsf N_{\mathrm{ext}}(\mathcal Z,\varrho,\varepsilon/2)$.
Lastly, the monotonicity under set inclusion for the external covering number is a straightforward consequence of its definition.

\hfill$\square$

 \paragraph{Basic Concentration Inequalities.} 
We state McDiarmid's inequality, Hoeffding's inequality, and Bernstein's inequality.

\begin{lemma}[McDiarmid's inequality (bounded differences), \cite{mcdiarmid1998concentration}]\label{lemma Mcdiarmid}
Let $Z_1,\ldots,Z_n$ be independent random variables with $Z_i\in \cZ_i$. Let $h:\mathcal Z_1\times\cdots\times\mathcal Z_n\to\mathbb R$ be a measurable
function satisfying the bounded difference property: there are constants $c_1,\ldots,c_n\ge 0$ such that for all $i\in \{1,\cdots,n\}$ and all $Z_1\in\cZ_1,\cdots, Z_n\in\cZ_n$,
\[
  \sup_{Z'_i\in \cZ_i}\bigl|h(Z_1,\cdots, Z_{i-1},Z_i,Z_{i+1},\cdots, Z_n)-h(Z_1,\cdots, Z_{i-1},Z'_i,Z_{i+1},\cdots, Z_n)\bigr|\;\le\; c_i.
\]
Then for every $t\ge 0$,
\begin{align*}
&\Pr\!\bigl(h(Z_1,\cdots,Z_n)-\mathbb E[h(Z_1,\cdots,Z_n)] \ge t\bigr)
  \;\le\; \exp\!\left(-\frac{2t^2}{\sum_{i=1}^n c_i^2}\right).
\end{align*}
\end{lemma}

\begin{lemma}[Hoeffding's inequality, Chapter 2 in \cite{vershynin2018high}]\label{lemma Hoeffding}
Let $Z_1,\cdots,Z_n$ be independent random variables with
$a_i \le Z_i \le b_i$ almost surely.
Then for every $t\geq 0$,
\[
\Pr\!\left(\sum_{i=1}^n Z_i-\mathbb{E}[Z] \ge t\right)
\;\le\;
\exp\!\left(
-\frac{2t^2}{\sum_{i=1}^n (b_i-a_i)^2}
\right).
\]
\end{lemma}

\section{Proofs for Deep Neural Networks and Riemannian Dimension (Section \ref{sec non-perturbative})}\label{appendix DNN}

\subsection{Proof of Lemma \ref{lemma non-perturbative} (Non-Perturbative Feature Expansion)}
We start with the telescoping decomposition presented in the main paper, which serves as a non-perturbative replacement of conventional Taylor expansion, where in each summand the only difference lies in $W_l'$ and $W_l$.
\begin{align*}  
&F_L(W',X)-F_L(W,X)\\=&\sum_{l=1}^L[\underbrace{\sigma_L(W'_L\cdots W'_{l+1}}_{\textup{controlled by} M_{l\rightarrow L}}\underbrace{\sigma_l}_{\textup{by} 1}(W'_l\underbrace{F_{l-1}(W,X)}_{\textup{learned feature}}))-\sigma_L(W'_L\cdots W'_{l+1}\sigma_l(W_l\underbrace{F_{l-1}(W,X)}_{\textup{learned feature}}))].
\end{align*}
Applying Cauchy-Schwarz inequality to the above identity, we have
\begin{align}\label{eq: Cauchy-Schwarz}\nonumber
    &||F_L(W',X)-F_L(W,X)||^2_{\tF}\\\leq&\sum_{l=1}^L L\|\sigma_L(W'_L\cdots W'_{l+1} \sigma_l(W'_lF_{l-1}(W,X)))-\sigma_L(W'_L\cdots W'_{l+1}\sigma_l(W_lF_{l-1}(W,X)))\|_{\tF}^2.
\end{align}
 By the definition of local Lipschitz constant in Section \ref{sec non-perturbative}, for all $W'\in B_{\varrho_n}(W, \varepsilon)$,
\begin{align} \label{eq: local Lipschitz constant step}
    &\|\sigma_L(W'_L\cdots W'_{l+1} \sigma_l(W'_lF_{l-1}(W,X)))-\sigma_L(W'_L\cdots W'_{l+1}\sigma_l(W_lF_{l-1}(W,X)))\|_{\tF}\nonumber\\\leq &M_{l\rightarrow L}[W,\varepsilon]\|\sigma_l(W_l'F_{l-1}(W,X))-\sigma_l(W_lF_{l-1}(W,X))\|_{\tF}.
\end{align}
Because the activation function $\sigma_l$ is $1-$Lipschitz for each column, we have
\begin{align}\label{eq: activation}
    \|\sigma_l(W_l'F_{l-1}(W,X))-\sigma_l(W_lF_{l-1}(W,X))\|_{\tF}\leq \|(W_l'-W_l)F_{t-1}(W,X)\|_{\tF}.
\end{align}
Combining \eqref{eq: Cauchy-Schwarz} \eqref{eq: local Lipschitz constant step} and \eqref{eq: activation}, we prove that
\begin{align*}
    \|F_L(W',X)-F_L(W,X)\|_{\tF}^2\leq \sum_{l=1}^L L\cdot M_{l\rightarrow L}[W,\varepsilon]^2 \cdot \|(W_l'-W_l)F_{l-1}(W,X)\|_{\tF}^2.
\end{align*}
 \hfill $\square$

\subsection{Metric Domination Lemma}
Our non-perturbative expansion facilitates bounding the pointwise dimension of complex geometries via metric comparison. By constructing a simpler, dominating metric (i.e., one that is pointwise larger), we establish that the pointwise dimension of the original geometry is upper bounded by that of this new, more structured geometry. This ``enlargement" for analytical tractability, a concept with roots in comparison geometry and majorization principles, is operationalized in Lemma \ref{lemma simple metric domination}.
\begin{lemma}[Metric Domination Lemma]\label{lemma simple metric domination}
    For two metrics $\varrho_1, \varrho_2$ defined on $\cW$, if $\varrho_1(W',W)\leq \varrho_2(W',W)$ for all $W'\in B_{\varrho_2}(W,\varepsilon)$,  then for any prior $\pi\in\Delta(\cW)$ and any $\varepsilon>0$, we have
    \begin{align*}
        \log\frac{1}{\pi(B_{\varrho_1}(W,\varepsilon))}\le \log\frac{1}{\pi(B_{\varrho_2}(W,\varepsilon))}.
    \end{align*}
\end{lemma}
\paragraph{Proof of Lemma \ref{lemma simple metric domination}:} Because $\varrho_1(W',W)\leq \varrho_2(W',W)$ for all $W'\in B_{\varrho_2}(W,\varepsilon)$, we have that
\begin{align*}
    B_{\varrho_1}(W,\varepsilon)\supseteq B_{\varrho_2}(W,\varepsilon).
\end{align*}
So for any prior $\pi$ on $\bR^p$, monotonicity of measures gives 
\begin{align*}
    \pi(B_{\varrho_1}(W,\varepsilon))\geq \pi(B_{\varrho_2}(W,\varepsilon)),
\end{align*}
this implies
\begin{align*}
\log\frac{1}{\pi(B_{\varrho_1}(W,\varepsilon))}\le \log\frac{1}{\pi(B_{\varrho_2}(W,\varepsilon))}.
\end{align*}
 \hfill$\square$

We then state an extension of the metric domination lemma, which turns pointwise dimension in a high-dimensional space into a lower-dimensional subspace.

\begin{lemma}[Subspace Metric Domination Lemma]\label{lemma subspace metric dom}
   Given a metric $\varrho_1$ defined on $\bR^p$ a subspace $\cV\subseteq \bR^p$, and a metric $\varrho_2$ defined on $\cV$. Define the orthogonal projector to subspace $\cV$ as $\mathcal{P}_\cV(W):=\arg\min_{\tilde{W}\in \cV} \|\tilde{W}-W\|_2$. If there exists $\varepsilon_1\in(0,\varepsilon)$ such that  for every $W'\in  \cV$,  \begin{align}\label{eq: metric domination condition}
 ( \varrho_1(W',W))^2\leq (\varrho_2(W',\mathcal{P}_\cV(W)))^2+\varepsilon_1^2,
    \end{align}  then for any prior $\pi\in \Delta(\cV)$ , we have 
    \begin{align}\label{eq: subspace conclusion}
        \log\frac{1}{\pi(B_{\varrho_1}(W,\varepsilon))}\le \log\frac{1}{\pi(B_{\varrho_2}(\mathcal{P}_\cV(W),\sqrt{\varepsilon^2-\varepsilon_1^2}))}.
  \end{align}
\end{lemma}

\paragraph{Proof of Lemma \ref{lemma subspace metric dom}:} By the condition \eqref{eq: metric domination condition}, we know
   \begin{align*}
 B_{\varrho_1}(W,\varepsilon)\supseteq B_{\varrho_1}(W,\varepsilon)\cap \cV\supseteq B_{\varrho_2}(\mathcal{P}_{\cV}(W),\sqrt{\varepsilon^2-\varepsilon_1^2}),
    \end{align*}
and this gives the desired conclusion \eqref{eq: subspace conclusion} in 
 Lemma \ref{lemma subspace metric dom}. 
 
 \hfill $\square$


\subsection{Pointwise Dimension Bound with Reference Subspace}

\paragraph{Set Up of Reference Effective Subspace}  Consider the weight space $B_{2}(R)\subset \bR^p$ for vectorized weights $W$, where $B_2(R) := \{w \in \mathbb{R}^p : \|w\|_2 \le R\}$. Given any fixed $p\times p$ PSD matrix $G(W)$, order the eigenvalues $\lambda_1(G(W)),\cdots, \lambda_p(G(W))$ nonincreasingly. For notational convenience, we suppress the dependence on \(G(W)\) and write simply \(\lambda_k\) when no confusion can arise. We denote \(\cV_\eff(G(W), R,\varepsilon)\) to be the {\it effective subspace}---the true top-$r_\eff$ eigenspace---of $G(W)$. For notational convenience, we use $r_\eff$ as the abbreviation of $r_\eff(G(W), R, \varepsilon)$, and $\cV$ as an abbreviation of  \(\cV_\eff(G(W), R,\varepsilon)\) when no confusion can arise.

 Assume there is another $r-$dimensional subspace $\bar\cV$. We will show that if \(\bar \cV\) approximates \(\cV\), then using a prior supported on \(\bar \cV\) still yields a valid effective‐dimension bound.  This observation underpins the hierarchical covering argument in Theorem~\ref{thm matrix}.  For a self‐contained introduction to subspaces (collectively known as the Grassmannian) and their frame parameterizations (the Stiefel manifold); see Section~\ref{subsec Grassmannian}, where we translate algebraic and differential-geometric insights into machine learning terminology.  

\paragraph{Motivation of Approximate Effective Subspace.}  We can view the orthogonal projector to a subspace as a matrix (see the definition via the Stiefel parameterization in \eqref{eq: definition projector}), which is consistent with the earlier operator notation characterized by $\ell_2$–distance in Lemma~\ref{lemma subspace metric dom}. Now define the projected metric $\varrho_{G(W)}^{\bar\cV}$ as 
\begin{align*}
    \varrho_{G(W)}^{\bar\cV}(W_1,W_2)&=\sqrt{(\cP_{\bar\cV}(W_1)-\cP_{\bar\cV}(W_2))^\top{G(W)} (\cP_{\bar\cV}(W_1)-\cP_{\bar\cV}(W_2))}\\&= \sqrt{(W_1-W_2)^\top\cP_{\bar\cV}^\top{G(W)} \cP_{\bar\cV}(W_1-W_2)}.
\end{align*}
By the subspace metric domination lemma (Lemma \ref{lemma subspace metric dom}), if $\cP_{\bar\cV}^\top{G(W)} \cP_{\bar\cV}$ approximates ${G(W)}$, we can use prior over $\bar\cV$ to bound the pointwise dimension and achieve dimension reduction.

 We will require the following approximation error condition: 
\begin{align*}
\varrho_{\proj, {G(W)}}(\cV,\bar\cV)=\|{G(W)}^{\frac{1}{2}} (\mathcal{P}_{\cV}-\mathcal{P}_{\bar\cV})\|_{\op}\leq \frac{\sqrt{n}\varepsilon}{4R}. 
\end{align*}

 In Section \ref{sec geometry algebra}, we systematically study the ellipsoidal covering of Grassmannian, and establish that we can {\it always} find $\bar\cV$ that approximates $\cV$ to the desired precision, with an additional covering cost of the Grassmannian bound in the Riemannian Dimension. This generalizes the canonical projection metric between subspaces into ellipsoidal set-up.

\paragraph{Effective Dimension Bound for Approximate Effective Subspace.}
We now present the lemma that establishes effective dimension bound using prior supported on approximate effective subspace $\bar{\cV}$ (not necessarily the true effective subspace $\cV_\eff({G(W)}, R,\varepsilon)$). We state the main result of this subsection (Lemma \ref{lemma approximate subspace} in the main paper).

Consider the weight space $B_{2}(R)\subset \bR^p$ for vectorized weights, and a pointwise ellipsoidal metric defined via PSD $G(W)$.  
Let $\bar\cV\subseteq\bR^p$ be a fixed $r$-dimensional subspace.  
Define the prior $\pi_{\bar{\cV}}=\textup{Unif}\!\bigl(B_{2}(1.58R)\cap\bar{\cV}\bigr)$.  
Then, uniformly over all $(W,\varepsilon)$ such that the top-$r$ eigenspace $\cV$ of $G(W)$ can be approximated by $\bar\cV$ to  precision 
\begin{align}\label{eq: def approximate}
\varrho_{\proj,G(W)}(\cV,\bar\cV)
:=\bigl\|\,G(W)^{1/2}\bigl(\cP_{\cV}-\cP_{\bar\cV}\bigr)\,\bigr\|_{\op}
\;\leq\; \tfrac{\sqrt{n}\varepsilon}{4R},
\end{align}
we have
\begin{align*}
\log \frac{1}{\pi_{\bar{\cV}}(B_{\varrho_G(W)}(W,\sqrt{n}\varepsilon))}
\;\le\;
\frac{1}{2} \sum_{k=1}^{r_{\eff}(G(W), R, \varepsilon)} \log\left( \frac{40 R^2 \lambda_k(G(W))}{n\varepsilon^2} \right)=d_\eff({G(W)}, \sqrt{5}R,\varepsilon).
\end{align*}

\paragraph{Proof of Lemma \ref{lemma approximate subspace}:} Given a fixed PSD matrix ${G(W)}$ with eigenvalues $\lambda_1\geq \cdots \geq \lambda_p$, denote $r_\eff=r_\eff({G(W)}, R, \varepsilon)$, and the projected metric $\varrho^{\bar{\cV}}_{G(W)}$ on $\bar{\cV}$:
\begin{align*}
\varrho^{\bar{\cV}}_{G(W)}(W_1,W_2)=\sqrt{(W_1-W_2)^\top\cP_{\bar\cV}^\top {G(W)} \cP_{\bar\cV}(W_1-W_2)}.
\end{align*}
Since $\cV$ is the top-$r_\eff$ eigenspace of ${G(W)}$, by the elementary property of eigendecomposition we have that 
\begin{align}\label{eq: elementary property eigendecomposition}
    {G(W)}= &\cP_{\cV}^\top{G(W)} \cP_{\cV}+\cP_{\cV_\perp}^\top {G(W)}\cP_{\cV_\perp} 
    \nonumber\\\preceq &\cP_{\cV}^\top {G(W)} \cP_{\cV}+ \lambda_{r_\eff+1} \cdot\cP_{\cV_\perp}^\top \cP_{\cV_\perp},
\end{align}
where $\cV_\perp$ is orthogonal complement of $\cV$. 
It is also straightforward to see 
\begin{align}\label{eq: completing the square for operator}
    \cP_{\cV}^\top {G(W)} \cP_{\cV} \preceq 2\cP_{\bar\cV}^\top  {G(W)} \cP_{\bar\cV}+2(\cP_{\cV}-\cP_{\bar\cV})^{\top} {G(W)} (\cP_{\cV}-\cP_{\bar\cV}).
\end{align}
Combining \eqref{eq: elementary property eigendecomposition} and \eqref{eq: completing the square for operator}, we have the fundamental loewner order inequality
\begin{align}\label{eq: basic order approximate}
    {G(W)}\preceq  2\cP_{\bar\cV}^\top {G(W)} \cP_{\bar\cV}+2(\cP_{\cV}-\cP_{\bar\cV})^{\top} {G(W)} (\cP_{\cV}-\cP_{\bar\cV})+ \lambda_{r_\eff+1} \cdot\cP_{\cV_\perp}^\top \cP_{\cV_\perp} .
\end{align}
In order to apply the subspace metric domination lemma (Lemma \ref{lemma subspace metric dom}), we hope to bound $\|W'-W\|_2^2$ and apply that bound to the two last remainder terms in the right hand side of \eqref{eq: basic order approximate}.

 To bound $\|W'-W\|_{2}^2$, we firstly state the following lemma on the eigenvalue of $\cP_{\bar\cV}^\top {G(W)} \cP_{\bar\cV}$, whose proof is deferred until after the current proof. Here, the metric tensor is a Riemannian-geometric way of describing a pointwise positive semidefinite, matrix-valued function \(G(W)\); see also Appendix~\ref{subsec: proof them matrix}. The corresponding projected metric tensor is given by \(\cP_{\bar\cV}^\top G(W)\cP_{\bar\cV}\).
\begin{lemma}[Eigenvalue Bound for Projected Metric Tensor]\label{lemma eigenvalue bound}
Assume $\cV$ is the top-$r$ eigenspace of  a PSD matrix $\Sigma$ with eigenvalues $\lambda_1\geq \cdots\geq \lambda_p$, then for a $r-$dimensional subspace $\bar\cV$ we have that for $k=1,2,\cdots, r$,
\begin{align*}
  \lambda_k \geq  \lambda_k(\cP_{\bar\cV}^\top \Sigma\cP_{\bar\cV})\geq \lambda_k /2-\|\Sigma^{\frac{1}{2}}(\cP_{\cV}-\cP_{\bar\cV})\|^2_{\op}.
\end{align*}
\end{lemma}For every ${W}'\in B_{\varrho^{\bar \cV}_{{G(W)}}}(\mathcal{P}_{\bar\cV} (W),\sqrt{n}{\varepsilon}/4)$,  we have $\forall k=1,\cdots, r_\eff$,
\begin{align}
    \|W'-\mathcal{P}_{\bar\cV}(W)\|_{2}^2\leq &\frac{(W'-\cP_{\bar\cV}(W))^\top\cP_{\bar\cV}^\top {{G(W)}} \cP_{\bar\cV}(W'-\cP_{\bar\cV}(W))}{\lambda_{r_\eff}(\cP_{\bar\cV}^\top {{G(W)}} \cP_{\bar\cV})}\leq \frac{ n\varepsilon^2}{16\lambda_{r_\eff}(\cP_{\bar\cV}^\top {{G(W)}} \cP_{\bar\cV})}\nonumber\\\leq &\frac{ n\varepsilon^2}{8\lambda_{r_\eff}-16\|{{G(W)}}^{\frac{1}{2}}(\cP_{\cV}-\cP_{\bar\cV})\|^2_{\op}}\leq \frac{1}{3}R^2,\label{eq: norm bound projected space}
\end{align}
where the first inequality holds because if $A$ is a symmetric positive definite matrix, then for all vectors $x$, we have 
$x^\top A x \geq \lambda_{\min}(A) \|x\|_2^2$; the second inequality used the condition of ${W}'\in B_{\varrho^{\bar \cV}_{{G(W)}}}(\mathcal{P}_{\bar\cV} (W),\sqrt{n}{\varepsilon}/4)$; the third inequality uses Lemma \ref{lemma eigenvalue bound}; and the last inequality uses $\lambda_{r_\eff}\geq \frac{n\varepsilon^2}{2R^2}$ (by definition \eqref{eq: eff rank} of effective rank) and the approximation error condition \eqref{eq: def approximate}. On the other hand, we have that $\|W\|_2^2\leq R^2$, so that for every ${W}'\in B_{\varrho^{\bar\cV}_{{G(W)}}}(\mathcal{P}_{\bar\cV} (W),\sqrt{n}{\varepsilon}/4)$ 
\begin{align*}
    \|W'-W\|_2^2= \|W'-\cP_{\bar\cV}(W)\|_2^2+ \|\cP_{\bar\cV_\perp}(W)\|_2^2\leq \frac{4}{3}R^2,
\end{align*}
combined with \eqref{eq: norm bound projected space}.

From the fundamental loewner order inequality \eqref{eq: basic order approximate}, we establish the desired metric domination condition: for all $W'\in B_{\varrho_{{G(W)}}^{\bar\cV}}(\cP_{\bar\cV}(W), \sqrt{n}\varepsilon/4)$ and $W\in B_2(R)$,
\begin{align*}
   &(W'-W)^\top {G(W)} (W'-W)\\
   \leq &(W'-W)^\top(2\cP_{\bar\cV}^\top {{G(W)}}\cP_{\bar\cV})(W'-W)+(2\|{{G(W)}}^{\frac{1}{2}}(\cP_{\cV}-\cP_{\bar\cV})\|_{\op}^2+ \lambda_{r_\eff+1})\|W'-W\|_2^2\\
   \leq &2\varrho_{{G(W)}}^{\bar\cV}(W',\cP_{\bar\cV}(W))^2+\frac{5n\varepsilon^2}{6},
\end{align*}
where the first inequality holds because of the loewner order inequality \eqref{eq: basic order approximate} and the property of operator norm: $x^\top A x \leq \|A\|_{\mathrm{op}} \cdot \|x\|_2^2$ (one could also apply Lemma \ref{lemma eigenvalue bound} to validate $\|\cP^\top_{\cV_\perp}\cP_{\cV_\perp}\|_\op\leq 1$); and the last inequality uses the fact $\lambda_{r_\eff+1}< \frac{n\varepsilon^2}{2R^2}$ (by definition \ref{eq: eff rank} of effective rank) and the approximation error condition \eqref{eq: def approximate}.
Now we can apply the subspace metric domination lemma (Lemma \ref{lemma subspace metric dom}) and obtain: for any $\pi\in \Delta( \bar\cV)$,
\begin{align}\label{eq: subspace linear approx}
 \log\frac{1}{\pi\bigl(B_{\varrho_{G(W)}}(W,\sqrt{n}\varepsilon)\bigr)} 
    \leq   \log\frac{1}{\pi\bigl(B_{\sqrt{2}\varrho^{\bar\cV}_{{G(W)}}}(\cP_{\bar\cV}(W),\sqrt{n}\varepsilon/\sqrt{6})\bigr)} \leq \log\frac{1}{\pi\bigl(B_{\varrho^{\bar\cV}_{{G(W)}}}(\cP_{\bar\cV}(W),\sqrt{n}\varepsilon/4)\bigr)}.
\end{align}
In particular, we choose $\pi$ to be the uniform prior over $\bar \cV$:
\begin{align*}
    \pi_{\bar \cV}=\textup{Unif}(B_{2}(1.58 R)\cap \bar\cV).
\end{align*}
Then we aim to prove that $B_{\varrho^{\bar\cV}_{{G(W)}}}(\cP_{\bar\cV}(W),\sqrt{n}\varepsilon/4)\subseteq \bar\cV\cap B_{2}(1.58 R)$.  This is true because: 1) for every ${W}'\in B_{\varrho^{\bar\cV}_{{G(W)}}}(\mathcal{P}_{\bar\cV} (W),\sqrt{n}{\varepsilon}/4)$, \eqref{eq: norm bound projected space} suggests $\|W'-\mathcal{P}_{\bar\cV}(W)\|_{2}^2\leq \frac{1}{3}R^2$, 
and 2) for every $W\in B_{2}(R)$, we have $\|\cP_{\bar\cV}(W)\|_2\leq\|W\|_{2}\leq R$.
Combining this and the above inequality we have
\begin{align*}
    \|W'\|_2\leq \|W'-\mathcal{P}_{\bar\cV}(W)\|_2+\|\mathcal{P}_{\bar\cV}(W)\|_2\leq (\sqrt{1/3}+1)R< 1.58R.
\end{align*}
This proves that $B_{\varrho^{\bar\cV}_{{G(W)}}}(\mathcal{P}_{\bar\cV}(W),\sqrt{n}{\varepsilon}/4)\subseteq \bar\cV\cap B_{2}(1.58R)$, so we have
\begin{align}\label{eq: volume equality approx}
    \log\frac{1}{\pi_{\bar\cV}(B_{\varrho^{\bar\cV}_{{G(W)}}}(\cP_{\bar\cV}(W), \sqrt{n}\varepsilon/4)}=\log \frac{\textup{Vol}({\bar\cV}\cap B_{2}(1.58 R))}{\textup{Vol}(B_{\varrho^{\bar\cV}_{{G(W)}}}(\mathcal{P}_{\bar\cV}(W),\sqrt{n}\varepsilon/4))}.
\end{align}
By the change–of–variables theorem in multivariate calculus \citep{wiki:change-of-variables}, the linear map $T=G(W)^{\frac{1}{2}}$ implies the volume formula for ellipsoid $E=B_{\varrho^{\bar\cV}_{{G(W)}}}(\mathcal{P}_{\bar\cV}(W),\sqrt{n}{\varepsilon}/4)$ with dimension
$r_{\text{eff}}$, eigenvalues $\{\lambda_k(\cP_{\bar\cV}^\top{G(W)}\cP_{\bar\cV})\}_{k=1}^{r_\eff}$ and radius $\sqrt{n}\varepsilon/4$
\begin{align*}
     \textup{Vol}\bigl(E \bigr)
     \;=\;
     |\det T|^{-1}\;
     \textup{Vol}\bigl(T(E)\bigr)
     \;=\;
(\det G(W))^{-1/2}\;\textup{Vol}\bigl(B_{2}(\sqrt{n}\varepsilon/4)\bigr)
=\left(\prod_{k=1}^{r_\eff}\lambda_k\right)^{-1/2}\textup{Vol}\bigl(B_{2}(\sqrt{n}\varepsilon/4)\bigr),
\end{align*}
Also by the change-of-variable theorem, we have that the volume of $r_\eff-$dimensional isotropic ball $\cV\cap B_{2}(1.58R)$ is 
\begin{align*}
\mathrm{Vol}(\bar\cV\cap B_{2}(1.58R))
\;=\; \left(\frac{1.58R}{\sqrt{n}\varepsilon/4}\right)^{r_\eff}\textup{Vol}(B_2(\sqrt{n}\varepsilon/4)).
\end{align*}
Hence, applying \eqref{eq: subspace linear approx} \eqref{eq: volume equality approx} and combining it with the two above volume equalities, we have
\begin{align*}
&\log\frac{1}{\pi_{\bar\cV}\bigl(B_{\varrho_{G(W)}}(W,\sqrt{n}\varepsilon)\bigr)} \leq 
   \log\frac{1}{\pi_{\bar\cV}(B_{\varrho^{\bar\cV}_{{{G(W)}}}}(\mathcal{P}_{\bar\cV}(W),\sqrt{n}\varepsilon/4))}=\log\frac{\textup{Vol}(\bar\cV\cap B_{2}(1.58R))}{\textup{Vol}(B_{\varrho^{\bar\cV}_{{G(W)}}}(\mathcal{P}_{\bar\cV}(W),\sqrt{n}{\varepsilon}/4))}\\&\leq \frac{1}{2}\log \frac{ (1.58R)^{2r_\eff}\prod_{k=1}^{r_\eff} \lambda_k}{(\sqrt{n}\varepsilon/4)^{2r_\eff}}\leq \frac{1}{2}\sum_{k=1}^{r_\eff} \log \frac{40R^2\lambda_k}{n\varepsilon^2}\\
   &\leq d_\eff({G(W)}, \sqrt{5}R, \varepsilon).
\end{align*}

Finally, since the prior construction $\pi_{\bar\cV}=\textup{Unif}(B_2(1.58R)\cap \bar\cV)$ only depends on $\bar\cV$ rather than $W$ and $\varepsilon$, we have that uniformly over all $(W,\varepsilon)\in B_{2}(R)\times [0,\infty)$ such that $\bar{\cV}$ approximates $\cV_\eff(G(W),R,\varepsilon)$ to the precision \eqref{eq: def approximate},  
\begin{align*}
    \log\frac{1}{\pi_{\bar\cV}(B_{\varrho_{{G(W)}}}(W,\sqrt{n}\varepsilon))}\leq  d_\eff(G(W), \sqrt{5}R, \varepsilon),
\end{align*} 
which is the claimed bound.
 \hfill $\square$

\paragraph{Proof of Lemma \ref{lemma eigenvalue bound}:}  
The Courant–Fischer–Weyl max-min characterization \citep{wiki:min_max_theorem} states that for any
Hermitian (i.e. symmetric for real matrices studying here) matrix,
\begin{align*}
  \lambda_{k}(\Sigma)
  \;=\;
\max_{\substack{\mathcal{S}\subseteq\mathbb{\bR}^{p}\\
                   \dim\mathcal{S}=k}}
      \;
      \min_{\substack{W\in\mathcal{S}\\ W\neq0}}
      \frac{W^{\top}\Sigma W}{\lVert W\rVert_2^{2}},
\end{align*}
and we have that for any $r-$dimensional subspace $\bar\cV$,
\begin{align*}
    \lambda_{k}(\cP_{\bar\cV}^\top {\Sigma} \cP_{\bar\cV})
  \;=\;
  \max_{\substack{\mathcal{S}\subseteq\bar\cV\\
                   \dim\mathcal{S}=k}}
      \;
      \min_{\substack{W\in\mathcal{S}\\ W\neq0}}
      \frac{W^{\top} \cP_{\bar\cV}^\top \Sigma \cP_{\bar\cV} W}{\lVert W\rVert_2^{2}},
\end{align*}
so we have $ \lambda_{k}(\cP_{\bar\cV}^\top \Sigma \cP_{\bar\cV})\leq \lambda_k$ for $k=1,2,\cdots,r$.

 Moreover, by the elementary property of eigendecomposition, since \(\cV\) is the
top-\(r\) eigenspace of \(\Sigma\), we have
\[
\lambda_k
=
\lambda_k(\cP_{\cV}^\top \Sigma \cP_{\cV}),
\qquad k=1,\ldots,r.
\]
Using the completing-square inequality \eqref{eq: completing the square for operator}, we have the Loewner-order bound
\[
\cP_{\cV}^\top \Sigma \cP_{\cV}
\preceq
2\cP_{\bar\cV}^\top \Sigma \cP_{\bar\cV}
+
2(\cP_{\cV}-\cP_{\bar\cV})^\top
\Sigma
(\cP_{\cV}-\cP_{\bar\cV}).
\]
Furthermore,
\[
(\cP_{\cV}-\cP_{\bar\cV})^\top
\Sigma
(\cP_{\cV}-\cP_{\bar\cV})
\preceq
\|\Sigma^{1/2}(\cP_{\cV}-\cP_{\bar\cV})\|_{\op}^2 I .
\]
Therefore,
\[
\cP_{\cV}^\top \Sigma \cP_{\cV}
\preceq
2\cP_{\bar\cV}^\top \Sigma \cP_{\bar\cV}
+
2\|\Sigma^{1/2}(\cP_{\cV}-\cP_{\bar\cV})\|_{\op}^2 I .
\]
By the monotonicity of eigenvalues under the Loewner order, which follows immediately from the classical Courant–Fischer–Weyl min–max characterization \citep{wiki:min_max_theorem}, we have
\[
\lambda_k(\cP_{\cV}^\top \Sigma \cP_{\cV})
\le
2\lambda_k(\cP_{\bar\cV}^\top \Sigma \cP_{\bar\cV})
+
2\|\Sigma^{1/2}(\cP_{\cV}-\cP_{\bar\cV})\|_{\op}^2 .
\]
Since \(\lambda_k=\lambda_k(\cP_{\cV}^\top \Sigma \cP_{\cV})\), rearranging gives
\[
\lambda_k(\cP_{\bar\cV}^\top \Sigma \cP_{\bar\cV})
\ge
\frac{\lambda_k}{2}
-
\|\Sigma^{1/2}(\cP_{\cV}-\cP_{\bar\cV})\|_{\op}^2 .
\]
Combining this lower bound with the upper bound proved above yields the claim.
 
\hfill$\square$

 \subsection{Proof of Riemannian Dimension Bound for DNN (Theorem \ref{thm matrix})}\label{subsec: proof them matrix}

  In the language of Riemannian geometry \citep{jost2008riemannian}, we regard a pointwise PSD, matrix-valued function $G(W)$ as a (possibly degenerate) {\it metric tensor}; such a $G(W)$ endows the parameter space $\mathbb{R}^{\sum_{l=1}^L d_{l-1}d_l}$ with a (semi-)Riemannian manifold structure. The pointwise ellipsoidal metric in \eqref{eq: metric tensor DNN} belongs to the following family of block-decomposable metric tensors.
  \begin{definition}[Metric Tensor of NN-surrogate Type]\label{def metric tensor NN type} A metric tensor $G(W)$ (pointwise PSD matrix-valued function of size $\sum_{l=1}^L d_{l-1}d_l\times \sum_{l=1}^L d_{l-1}d_l$)  is of ``NN-surrogate'' type if  $G(W)$ is in the form
    \begin{align*}
    G(W)=\textup{blockdiag}(A_1(W)\otimes I_{d_1},\cdots, A_l(W)\otimes I_{d_l},\cdots, A_L(W)\otimes I_{d_L})
\end{align*}
where $A_l(W)\in\bR^{d_{l-1}\times d_{l-1}}$.
\end{definition}
By Lemma~\ref{lemma non-perturbative}, the non-perturbative feature expansion gives rise to the metric tensor $G_{\textup{NP}}(W)$ defined in~\eqref{eq: metric tensor DNN}; $G_{\textup{NP}}(W)$ belongs to the “NN-surrogate” class. We first record some elementary decomposition properties for this family of NN-surrogate metric tensors, and then prove Theorem~\ref{thm matrix}.

 \subsubsection{Decomposition Properties of NN-surrogate Metric Tensor}
 The NN-surrogate metric tensor $G(W)$  in Definition \ref{def metric tensor NN type} has decomposition properties described by the next lemma.

 \begin{lemma}[Decomposition Properties of NN-surrogate Metric Tensor]\label{lemma decomposition of manifold} Given a NN-surrogate metric tensor $G(W)$ defined in Definition \ref{def metric tensor NN type},
 for every $W$, we have the following decomposition properties: First, the effective rank and dimension decompose to 
      \begin{align*}
     r_{\eff}(G(W),R,\varepsilon)&=\sum_{l=1}^L d_l\cdot r_{\eff}(A_l(W), R,\varepsilon);\\
      d_{\eff}(G(W),R,\varepsilon)&=\sum_{l=1}^L d_l\cdot d_{\eff}(A_l(W), R,\varepsilon).
 \end{align*}
 
 Second,  denote $\cV_\eff(A_l(W), R, \varepsilon)$ the effective subspace (i.e., the top-$r_\eff(A_l(W),R,\varepsilon)$ eigenspace) of $A_l(W)$. Then  the effective subspace of  $G(W)$ is \begin{align*}
     \cV_\eff(G(W),R,\varepsilon)= \cV_\eff(A_1(W),R,\varepsilon)^{d_1}\times\cdots\times\cV_\eff(A_L(W),R,\varepsilon)^{d_L}.
 \end{align*}
\end{lemma}
\paragraph{Proof of Lemma \ref{lemma decomposition of manifold}. } It is straightforward to see that, first, the effective rank of the fixed matrix $G(W)$ is 
 \begin{align*}
     &r_\eff(G(W),R,\varepsilon)\\=&\max\{k: 2\lambda_k(G(W)) R^2\geq n\varepsilon^2\}\\=&\sum_{l=1}^L \max\{k: 2\lambda_k(A_l(W)\otimes I_{d_l}) R^2\geq n\varepsilon^2\}\\=&\sum_{l=1}^L d_{l}\max\{k: 2\lambda_k(A_l(W)) R^2\geq n\varepsilon^2\}\\
     =&\sum_{l=1}^L d_l\cdot r_\eff(A_l(W),R,\varepsilon);
 \end{align*}
 and the effective dimension of the fixed matrix $G(W)$ is
 \begin{align*}
    & d_\eff(G(W),R,\varepsilon)\\=
    &\frac{1}{2}\sum_{k=1}^{r_\eff(G(W),R,\varepsilon)}\log\left(\frac{8R^2\lambda_k(G(W))}{n\varepsilon^2}\right)\\
    =&\sum_{l=1}^L \frac{1}{2}\sum_{k=1}^{r_\eff(A_l(W)\otimes I_{d_l},R,\varepsilon)}\log\left(\frac{8R^2\lambda_k(A_l(W)\otimes I_{d_l})}{n\varepsilon^2}\right)\\
    = &\sum_{l=1}^L d_l\cdot \frac{1}{2}\sum_{k=1}^{r_\eff(A_l(W),R,\varepsilon)}\log\left(\frac{8R^2\lambda_k(A_l(W))}{n\varepsilon^2}\right)\\
    = &\sum_{l=1}^L d_l\cdot d_\eff(A_l(W), R,\varepsilon).
 \end{align*}
  Second, as the effective subspace of the matrix tensor product $A_l(W)\otimes I_{d_l}$ is subspace tensor product $\cV_\eff(A_l(W),R,\varepsilon)^{d_l}$, the effective subspace for NN-surrogate metric tensor  $G(W)=\textup{blockdiag}(\cdots;A_l(W)\otimes I_{d_l};\cdots)$ is 
 \begin{align*}
    \cV_\eff(G(W),R,\varepsilon):= \cV_\eff(A_1(W),R,\varepsilon)^{ d_1}\times\cdots\times\cV_\eff(A_L(W),R,\varepsilon)^{ d_L}.
 \end{align*}
 \hfill$\square$

 \subsubsection{Proof of Theorem \ref{thm matrix}}\label{subsec proof NN type}

 We firstly prove the following result, which is almost Theorem \ref{thm matrix}, with the only difference being that the radius in the effective dimension depends on the global radius $R$ rather than the pointwise Frobenious norm $\|W\|_{\tF}$. Extending this result to Theorem \ref{thm matrix} can be achieved via a simple application of the ``uniform pointwise convergence'' principle \citep{xu2025towards} illustrated in Lemma \ref{lemma uniformed localized convergence}.

 \begin{lemma} [Riemannian Dimension for NN-surrogate Metric Tensor---Global Radius Version]\label{lemma matrix}
Consider the NN-surrogate metric tensor in Definition \ref{def metric tensor NN type}, and the weight space $B_{\tF}(R)$.  Then we have that the pointwise dimension is bounded by the pointwise Riemannian Dimension as the following: there exists a prior $\pi$ such that uniformly over all $W\in B_{\tF}(R)$,
\begin{align*}
\log\frac{1}{\pi(B_{\varrho_{G(W)}}(W,\sqrt{n}\varepsilon))}\leq \sum_{l=1}^L\Big( \underbrace{d_l\cdot d_{\eff}(A_l(W),CR,\varepsilon)}_{\textup{``must pay'' cost at each }W}+\underbrace{d_{l-1}\cdot d_\eff(A_l(W), CR, \varepsilon)}_{\textup{covering cost of Grassmannian} }+ \underbrace{\log(d_{l-1})}_{\textup{covering cost of $r_\eff\in[d_{l-1}]$}}\Big),
\end{align*}
where  $C>0$ is an absolute constant.
\end{lemma}
\paragraph{Proof of Lemma \ref{lemma matrix}:} The proof has two key steps: 1. Hierarchical covering argument, and 2. Bound  covering Cost of the Grassmannian. A crucial lemma about the ellipsoidal covering of the Grassmannian, which is new even in the pure mathematics context, is deferred to Section \ref{sec geometry algebra}.
 \paragraph{Step 1: Hierarchical Covering.}

 As explained in the main paper, the major difficulty is that the prior measure $\pi_\cV$ it constructed, is defined over the effective subspace $\cV$, which itself encodes information of the point $W$ and $\varepsilon>0$. The goal of our proof is to construct a ``universal'' prior $\pi$ that does not depend on $\cV$. 
 This is achieved via a hierarchical covering argument \eqref{eq: hierarchical covering},  which we make rigorous below.
 
The key idea of hierarchical covering is as follows: Firstly,  for all $W$, we search for subspace $\bar{\cV}$ that approximates the true effective subspace (top-$r_\eff$ eigenspace) $\cV_\eff(G(W),R,\varepsilon)$ to the precision required by \eqref{eq: def approximate}:
\begin{align}\label{eq: full matrix covering condition}
    \|G(W)^{\frac{1}{2}}(\cP_{\cV}-\cP_{\bar\cV})\|_{\op}\leq \frac{\sqrt{n} \varepsilon}{4R},
\end{align}
where $G(W)^{\frac{1}{2}}$ is the unique square root of PSD matrix $G(W)$ (see, e.g,  \citep{wiki:sqrtmatrix}).
Then by Lemma \ref{lemma approximate subspace} (Pointwise Dimension Bound for Nonlinear Manifold with Approximate Effective Subspace), for every $(W,\varepsilon)\in B_2(R)\times [0,\infty)$ such that $\bar{\cV}$ approximates $\cV_\eff(G(W),R,\varepsilon)$ to the precision \eqref{eq: full matrix covering condition}, the prior $\pi_{\bar{\cV}}=\textup{Unif}(B_{2}(1.58R)\cap\bar{\cV})$ satisfies 
\begin{align}\label{eq: partial Riemannian Dimension}
\log \frac{1}{\pi_{\bar{\cV}}(B_{\varrho_{G(W)}}(W,\sqrt{n}\varepsilon))}
\;\le\;d_\eff(G(W), \sqrt{5}R,\varepsilon)=\sum_{l=1}^L d_l\cdot d_\eff(A_l(W), \sqrt{5}R, \varepsilon),
\end{align}
where the first inequality is by Lemma \ref{lemma approximate subspace} (see definition \eqref{eq: effective dimension} of effective dimension); and the last equality is by the decomposition property of NN-surrogate metric tensor (Lemma \ref{lemma decomposition of manifold}). 

Secondly, we put a prior $\mu$ over all possible subspaces $\cV$ and construct the ``universal'' prior 
\begin{align}\label{eq: universal prior origin}
\pi(W)=\sum_{\cV}{\mu(\cV)}\times \pi_{\cV}(W),
\end{align}
which implies that uniformly over all $W\in B_{\tF}(R)$,
\begin{align}\label{eq: prior argument}
    &\log\frac{1}{\pi(B_{\varrho_{G(W)}}(W,\sqrt{n}\varepsilon))}\nonumber\\= & \log \frac{1}{\sum_{\cV}\mu(\cV)\pi_{\cV}(B_{\varrho_{G(W)}}(W,\sqrt{n}\varepsilon))}\nonumber\\
    \leq & \log \frac{1}{\mu(\bar{\cV}: \bar \cV \textup{ satisfies \eqref{eq: full matrix covering condition}})\inf_{\bar \cV \textup{ satisfies \eqref{eq: full matrix covering condition}}}\pi_{\bar \cV}(B_{\varrho_{G(W)}}(W,\sqrt{n}\varepsilon))}\nonumber\\
    \leq & \underbrace{\log \frac{1}{\mu(\bar{\cV}: \bar\cV\textup{ satisfies \eqref{eq: full matrix covering condition}}}}_{\textup{covering cost of the Grassmannian}}+\sum_{l=1}^L d_l\cdot d_{\eff}(A_l(W),\sqrt{5}R,\varepsilon),
\end{align}where the first equality is by definition \eqref{eq: universal prior origin} of the ``universal'' prior $\pi$; the first inequality is straightforward; and the last inequality is by \eqref{eq: partial Riemannian Dimension} (the result of the ``must pay'' part in the hierarchical covering) and the equivalence between $B_2(R)$ and $B_{\tF}(R)$.

The above hierarchical covering argument successfully gives a valid Riemannian Dimension, with the cost of the additional covering cost given by the subspace prior $\mu$. This explains our basic proof idea.
The remaining proof executes this basic proof idea.

\paragraph{Step 2:  Bounding Covering Cost of the Grassmannian.} 
Section \ref{sec geometry algebra} provides a systematic study to the ellipsoidal metric entropy of Grassmannian manifold, which we detail the conclusion below.

Define
\[
  \textup{Gr}(d,r)\;:=\;\bigl\{\text{$r$–dimensional linear subspaces of }\bR^d\bigr\}
\]
as the \emph{Grassmannian manifold}.

  Given a $d\times d$ PSD $\Sigma$, define the anisometric projection metric between two subspaces by (labeled as Definition \ref{def ellipsoidal} in Section \ref{sec geometry algebra})
\begin{align}\label{eq: ellipsoidal projection in thm proof}
    \varrho_{\textup{proj}, \Sigma}(\cV,\bar\cV)=\|\Sigma^{\frac{1}{2}}(\cP_{\cV}-\cP_{\bar\cV})\|_{\op},
\end{align}
where $\Sigma^{\frac{1}{2}}$ is the square root of the PSD matrix $\Sigma$ (see, e.g.,  \citep{wiki:sqrtmatrix}).

Lemma \ref{lemma ellipsoidal Grassmannian} states that (note that we use $\varepsilon_1$ and $C_0$ here instead of $\varepsilon$ and $C$ in the original statement of Lemma \ref{lemma ellipsoidal Grassmannian}), given a Grassmannian $\Gr(d,r)$, for uniform prior $\mu=\textup{Unif}(\Gr(d,r))$, we have that for every $\cV\in \Gr(d,r)$,  every $\varepsilon_1>0$ and PSD matrix $\Sigma\in\bR^{d\times d}$ with eigenvalues $\lambda_1\geq \cdots\lambda_d\geq 0$, we have the pointwise dimension bound
\begin{align}\label{eq: use lemma ellipsoidal}
    &\log\frac{1}{\mu(B_{\varrho_{\textup{proj},{{\Sigma}}}}(\cV, \varepsilon_1))}\leq  \frac{d-r}{2}\sum_{k=1}^r\log \frac{C_0\max\{\lambda_k,\varepsilon_1^2\}}{\varepsilon_1^2}+ \frac{r}{2}\sum_{k=1}^{d-r}\log \frac{C_0\max\{\lambda_k,\varepsilon_1^2\}}{\varepsilon_1^2},
\end{align}
  where $C_0>0$ is an absolute constant.  We will use the result \eqref{eq: use lemma ellipsoidal} and \eqref{eq: prior argument} to prove Theorem \ref{thm matrix}. 

For a particular layer $l$,  $d_{l-1}\times d_{l-1}$ PSD matrix $A_l(W)$, and a fixed  rank $r_{l}$ denote $\textup{Gr}(d_{l-1}, r_l)$ as a Grassmannian (the collection of all $r_l$-dimensional in $\bR^{d_{l-1}}$). By \eqref{eq: use lemma ellipsoidal} we have that there exists a prior $\mu_l$ over $\Gr(d_{l-1},r_l)$ such that for every $(W,\varepsilon_1)$ such that $r_\eff(A_l(W),R,\varepsilon_1)=r_l$,  and  $\lambda_{r_{l}+1}(A_l(W))\leq c\varepsilon_1^2\leq \lambda_{r_{l}}(A_l(W))$ where $c\geq 1$ can be any absolute constants no smaller than $1$ (later we will specialize to $c=8$), 
\begin{align}\label{eq: layer wise covering}
    \log\frac{1}{\mu_l(\bar{\cV}:\varrho_{\textup{proj},A_l(W)}(\cV_\eff(A_l(W),R,\varepsilon),\bar\cV)\leq \varepsilon_1)}\leq \frac{d_{l-1}}{2}\sum_{k=1}^{r_l}\log\frac{C_1\lambda_k(A_l(W))}{\varepsilon_1^2},
\end{align}
where $C_1=c\max\{C_0,1\}\geq 1$ is an absolute constant depending only on the absolute constant $c$ (later we take $c=8$ so $C_1=8\max\{C_0,1\}$ is indeed an absolute constant). \eqref{eq: layer wise covering} is because: 1) all eigenvalues with index at least \(r_{l}+1\) (each no larger than
\(c\,\varepsilon_{1}^{2}\)) contribute only through the second term in
\eqref{eq: use lemma ellipsoidal}. Their cumulative effect is at most
\begin{align*}
    \mathds{1}\{d_{l-1}-r_l>r_l\}\cdot\frac{r_l}{2}\sum_{k=r_l+1}^{d_{l-1}-r_l}\log\frac{C_0c\varepsilon_1^2}{\varepsilon_1^2}=\frac{r_l\max\{d_{l-1}-2r_l,0\}}{2}\log C_0 c\leq \frac{r_l(d_{l-1}-r_l)}{2}\log C_0c
\end{align*}
unaffected to the spectrum, and we absorb this into the absolute constant \(C_{1}\).
And 2) all eigenvalues with index at most $r_l$'s contribution leads to at most
\begin{align*}
    \frac{d_{l-1}-r_l}{2}\sum_{k=1}^{r_l}\log \frac{C_0\lambda_k(A_l(W))}{\varepsilon_1^2}+\frac{r_l}{2}\sum_{k=1}^{\max\{r_l,d_{l-1}-r_l\}}\log \frac{C_0\lambda_k(A_l(W))}{\varepsilon_1^2}\leq \frac{d_{l-1}}{2}\sum_{k=1}^{r_l}\log \frac{\max\{C_0,1\}\lambda_k(A_l(W))}{\varepsilon_1^2}.
\end{align*}Summing up the contributions two parts of the spectrum together, we get the right hand side of \eqref{eq: layer wise covering}.

By the subspace decomposition property in Lemma \ref{lemma decomposition of manifold}, we have that for $\bar \cV=(\cdots, \underbrace{\bar\cV_l, \cdots, \bar\cV_l}_{\textup{repeat $d_l$ times}}, \cdots)$,  
\begin{align}\label{eq: identity proj max}
    &\varrho_{\textup{proj},G(W)}(\cV_\eff(G(W),R,\varepsilon),\bar\cV)\nonumber\\
    =&\varrho_{\textup{proj},G(W)}(\prod_{l=1}^L\cV_\eff(A_l(W),R,\varepsilon)^{d_l},\prod_{l=1}^L\bar\cV_l^{d_l})
    \nonumber\\= &\max_{l}\varrho_{\textup{proj},A_l(W)}(\cV_\eff(A_l(W),R,\varepsilon), \bar\cV_l),
\end{align}
where the first equality is by Lemma \ref{lemma decomposition of manifold}, and the second equality is by the properties of the spectral norm: $\|\textup{blockdiag}( A, B)\|_{\op}=\max\{\|A\|_{\op}, \|B\|_{\op}\}$ and $\| A\otimes I_d\|_{\op}=\|A\|_{\op}$.

Taking $\varepsilon_1= \frac{\sqrt{n}\varepsilon}{4R }$, by definition \eqref{eq: eff rank} on the threshold to determine effective rank,  we obtain $\lambda_{r_{l}+1}(A_l(W)) \leq 8\varepsilon_1^2= n\varepsilon^2/(2R^2)\leq \lambda_{r_l}(A_l(W))$,  thus this particular choice satisfies the required eigenvalue condition to establish \eqref{eq: layer wise covering} with $c=8$. Then for all layers $l=1,\cdots,L$, given a fixed $\{r_1, \cdots, r_L\}$, by \eqref{eq: layer wise covering}, we have that there exists a prior
\begin{align}\label{eq: product prior}
\mu_{\{r_l\}_{l=1}^L}=\mu_1^{ d_1}\otimes\cdots\otimes\mu_L^{d_L}
\end{align}
over the product Grassmannian $\Gr(d_0,r_1)^{d_1}\times\cdots\times \Gr(d_{L-1},r_L)^{d_L}$ such that uniformly over all $W\in B_{\tF}(R)$ such that $r_\eff(A_l(W),R,\varepsilon)=r_l$, $\forall l\in[L]$ (here $[L]$ is the notation of $\{1,2,\cdots,L\}$), the   ``Grassmannian covering cost'' term in \eqref{eq: prior argument} is bounded by
\begin{align}\label{eq: fixed all r_l}
    &\log\frac{1}{\mu(\bar\cV: \bar\cV \textup{ satisfies \eqref{eq: full matrix covering condition}})} \nonumber\\
= &\log\frac{1}{\mu_{\{r_l\}_{l=1}^L}(\{\bar\cV: \varrho_{\textup{proj},G(W)}(\cV_\eff(G(W),R,\varepsilon),\bar{\cV})\leq \frac{\sqrt{n}\varepsilon}{4R}=\varepsilon_1)\}}\nonumber\\
\leq& \log\frac{1}{\mu_{\{r_l\}_{l=1}^L}(\{(\cdots,\underbrace{\bar\cV_l,\cdots,\bar\cV_l}_{d_l\textup{ times }},\cdots):  \varrho_{\textup{proj},A_l(W)}(\cV_\eff(A_l(W),R,\varepsilon),\bar{\cV}_l)\leq  \varepsilon_1, \quad \forall l\in [L]\})}\nonumber\\
=&\sum_{l=1}^L \log\frac{1}{\mu_{\{r_l\},l }(\{\bar\cV_l:  \varrho_{\textup{proj},A_l(W)}(\cV_\eff(A_l(W),R,\varepsilon),\bar{\cV}_l)\leq  \varepsilon_1\})}\nonumber\\
\leq &\sum_{l=1}^L \frac{d_{l-1}}{2}\sum_{k=1}^{r_l}\log\frac{C_1\lambda_k(A_l(W))}{\varepsilon_1^2}\nonumber\\\leq &\sum_{l=1}^L d_{l-1}d_\eff(A_l(W), \sqrt{2C_1}R,\varepsilon),
\end{align}
 where the first inequality follows by restricting the reference subspace
\(\bar{\cV}\) to the repeated-product form
\(\prod_{l=1}^L \bar{\cV}_l^{d_l}\) and using \eqref{eq: identity proj max}: for such subspaces, the global projection error is the maximum of the layerwise projection errors, so requiring every layerwise error to be at most \(\varepsilon_1\) implies the global condition \eqref{eq: full matrix covering condition}. The second equality follows from the layer-wise product structure of the prior in \eqref{eq: product prior}. More explicitly, if
$E_l :=
\left\{
\bar{\cV}_l:
\varrho_{\mathrm{proj},A_l(W)}
\bigl(
\cV_{\eff}(A_l(W),R,\varepsilon),\bar{\cV}_l
\bigr)
\le \varepsilon_1
\right\},$
then the repeated-product event is \(\bigcap_{l=1}^L E_l\), and the product prior gives
$\mu_{\{r_l\}_{l=1}^L}\!\left(\bigcap_{l=1}^L E_l\right)
=
\prod_{l=1}^L \mu_{l,r_l}(E_l)$.
Here \(\mu_l^{d_l}\) in \eqref{eq: product prior} denotes the pushforward of \(\mu_l\) under the repeated-copy map
\(\bar{\cV}_l\mapsto\bar{\cV}_l^{d_l}\), not the \(d_l\)-fold independent product measure. Thus one samples a single reference subspace \(\bar{\cV}_l\sim\mu_l\) at layer \(l\) and repeats this same subspace \(d_l\) times, so no extra \(d_l\) factor appears in the atlas cost. The second inequality is by the layer-wise covering bound \eqref{eq: layer wise covering}; and the last inequality follows from the choice
\(\varepsilon_1=\sqrt n\varepsilon/(4R)\) and the definition of effective dimension in \eqref{eq: effective dimension}.

Note that \eqref{eq: fixed all r_l} is uniformly over all $W\in{B_{\tF}(R)}$ such that $r_\eff(A_l(W),R,\varepsilon)=r_l$, $\forall l\in[L]$, not uniformly over all $W\in{B_{\tF}(R)}$. We would like to extend \eqref{eq: fixed all r_l} to all $W\in{B_{\tF}(R)}$ over uniform prior over possible integer values of $r_l$. 
Now assign uniform prior over $[d_{l-1}]=\{1,\cdots,d_{l-1}\}$ for $r_l$, we obtain the ``universal'' prior $\pi$ (as we have pursued in  in our hierarchical covering argument \eqref{eq: universal prior origin}) defined by
\begin{align}\label{eq: universal prior}
    \mu(\cV)=& \prod_{l=1}^L\underbrace{\textup{Unif}([d_{l-1}])}_{\textup{prior of }r_l}\otimes \underbrace{\mu_{\{r_k\}_{k=1}^L}}_{\textup{prior over product Grassmannian in \eqref{eq: product prior}}},\nonumber\\
    \pi(W)=&\sum_{\cV}\underbrace{\mu(\cV)}_{\textup{prior over subspaces defined above}}\otimes \underbrace{\textup{Unif}( B_{2}(1.58{R})\cap\bar{\cV})}_{\textup{uniform prior constrained in subspace}}.
\end{align}
Then we have that uniformly over all $W\in {B_{\tF}(R)}$,
\begin{align*}
     &\log\frac{1}{\pi(B_{\varrho_{G(W)}}(W,\sqrt{n}\varepsilon))}\\\leq  &\log\frac{1}{\mu(\bar\cV: \bar\cV \textup{ satisfies \eqref{eq: full matrix covering condition}})} + \sum_{l=1}^L d_l\cdot d_\eff(A_l(W),\sqrt{5}R,\varepsilon))\\
     \leq & \sum_{l=1}^L\log d_{l-1}+\log\frac{1}{\mu_{\{r_k\}_{k=1}^L}(\bar\cV: \bar\cV \textup{ satisfies \eqref{eq: full matrix covering condition}})}+ \sum_{l=1}^L d_l\cdot d_\eff(A_l(W),\sqrt{5}R,\varepsilon))\\
     \leq &\sum_{l=1}^L\log d_{l-1}+\sum_{l=1}^L d_{l-1}\cdot d_\eff(A_l(W),\sqrt{2C_1}R,\varepsilon)+ \sum_{l=1}^L d_l\cdot d_\eff(A_l(W),\sqrt{5}R,\varepsilon)),
\end{align*}
where $C_1>0$ is an absolute constant. Here the first inequality is by the hierarchical covering argument \eqref{eq: prior argument}; the second inequality is by the prior construction \eqref{eq: universal prior}; and the third inequality is by the Grassmannian covering bound \eqref{eq: fixed all r_l} for fixed $\{r_k\}_{k=1}^L$. 
This shows that for NN-surrogate metric tensor $G(W)$,  the pointwise dimension is bounded by the Riemannian Dimension as the following: \begin{align*}
  \log\frac{1}{\pi(B_{\varrho_{G(W)}}(W,\sqrt{n}\varepsilon))}\leq   \sum_{l=1}^L (d_l+d_{l-1})\cdot d_\eff(A_l(W),CR, \varepsilon) + \log (d_{l-1}),
\end{align*}
where $C$ is a positive absolute constant.
This finishes the proof of Lemma \ref{lemma matrix} with $R$ in effective dimension being a global upper bound of $\|W\|_{\tF}$.

\hfill$\square$

\paragraph{Proof of Theorem \ref{thm matrix}:} 
Motivated by the ``uniform pointwise convergence'' principle (proposed in \cite{xu2025towards} and illustrated in Lemma \ref{lemma uniformed localized convergence}), we apply a peeling argument to adapt the Riemannian Dimension to $\|W\|_{\tF}$. Given any $R_0\in(0,R]$, we take $R_k=2^k R_0$ for $k=0,1,\cdots \log_2 \lceil R/R_0 \rceil$.  Taking a uniform prior on these $R_k$, and set 
\begin{align*}
\tilde{\pi}=\underbrace{\textup{Unif}(\{R_0,\cdots, 2^{\log_2\lceil R/R_0 \rceil}R_0\})}_{\textup{prior over upper bound $\tilde{R}$ of $\|W\|_{\tF}$}}\otimes \underbrace{\pi_{\tilde R}}_{\textup{prior defined via \eqref{eq: universal prior}}},
\end{align*}
where $\pi_{\tilde R}$ is the prior defined via \eqref{eq: universal prior} in the proof of Lemma \ref{lemma matrix}.  Then for every $W\in{B_{\tF}(R)}$ where $\|W\|_{\tF}> R_0$, denote $k(W)$ to be the integer such that $2^{k(W)}R_0< \|W\|_{\tF}\leq 2^{k(W)+1}R_0$, then 
\begin{align*}
    &\log\frac{1}{\tilde{\pi}(B_{\varrho_{G(W)}}(W,\sqrt{n}\varepsilon))}\\\leq  &\underbrace{\log\log_2\lceil R/R_0 \rceil}_{\textup{density of $2^{k(W)+1}R_0$}}+ \underbrace{\log\frac{1}{{\pi_{2^{k(W)+1}R_0}}(B_{\varrho_{G(W)}}(W,\sqrt{n}\varepsilon))}}_{\textup{$\pi$ is constructed via \eqref{eq: universal prior}, with global radius taken to be $2^{k(W)+1}R_0$}}\\
    \leq &\log\log_2\lceil R/R_0\rceil+ \sum_{l=1}^L ((d_l+d_{l-1})\cdot d_\eff(A_l(W),C_12^{k(W)+1}R_0, \varepsilon) + \log d_{l-1})\\
    \leq & \log\log_2\lceil R/R_0\rceil+ \sum_{l=1}^L ((d_l+d_{l-1})\cdot d_\eff(A_l(W),C_1\cdot 2\|W\|_{\tF}, \varepsilon) + \log d_{l-1}),
\end{align*}
where the first inequality is due to the product construction of $\tilde{\pi}$;   the second inequality is due to Lemma \ref{lemma matrix}, with $C_1>0$ being an absolute constant; and the last inequality uses the fact $\|W\|_{\tF}\leq 2^{k(W)+1}R_0\leq 2\|W\|_{\tF}$, with $C_1>0$. 

The above bound assumes $\|W\|_{\tF}>R_0$. When $\|W\|_{\tF}\leq R_0$,  we directly apply Lemma \ref{lemma matrix} and obtain
\begin{align*}
    &\log\frac{1}{\tilde{\pi}(B_{\varrho_{G(W)}}(W,\sqrt{n}\varepsilon))}\\\leq  &\underbrace{\log \log_2\lceil R/R_0 \rceil}_{\textup{density of $R_0$}}+ \underbrace{\log\frac{1}{{\pi_{R_0}}(B_{\varrho_{G(W)}}(W,\sqrt{n}\varepsilon))}}_{\textup{$\pi$ is constructed via \eqref{eq: universal prior}, with global radius taken to be $R_0$}}\\
    \leq & \log\log_2\lceil R/R_0\rceil+ \sum_{l=1}^L ((d_l+d_{l-1})\cdot d_\eff(A_l(W),C_1\cdot R_0, \varepsilon)+ \log d_{l-1}).
\end{align*}
Combining the two cases discussed above, we conclude that the pointwise dimension for NN-surrogate metric tensor $G(W)$ in Definition \ref{def metric tensor NN type} is bounded by the  Riemmanin Dimension 
\begin{align*}
&\log\frac{1}{\tilde{\pi}(B_{\varrho_{G(W)}}(W,\sqrt{n}\varepsilon))}\leq d_{\textup{R}}(W,\varepsilon)\\&= \sum_{l=1}^L ((d_l+d_{l-1})\cdot d_\eff(A_l(W), C\max\{\|W\|_{\tF},R_0\},\varepsilon)+ \log (d_{l-1}\log_2 \lceil R/R_0\rceil),
\end{align*} 
where $C=2C_1$ is a positive absolute constant.  The term $\log_2\lceil R/R_0\rceil$ is enlarged by an additional factor of \(L\).

Finally, by the sentence below \eqref{eq: metric tensor DNN} (which is a straightforward result from non-perturbative feature expansion for DNN (Lemma \ref{lemma non-perturbative}) and the metric domination lemma (Lemma \ref{lemma simple metric domination})), we know that there exists a prior $\tilde\pi$ such that uniformly over all $W\in B_{\tF}(R)$, 
\begin{align*}
    &\log\frac{1}{\tilde{\pi}(B_{\varrho_n}(f(W,\cdot), \varepsilon))}\leq \log\frac{1}{\tilde{\pi}(B_{\varrho_{G_{\textup{NP}}(W)}}(W,\sqrt{n}\varepsilon))}\\
    &\leq d_{\textup{R}}(W,\varepsilon)= \sum_{l=1}^L ((d_l+d_{l-1})\cdot d_\eff(A_l(W), C\max\{\|W\|_{\tF},R_0\},\varepsilon)+ \log (d_{l-1}\log_2 \lceil R/R_0\rceil),
\end{align*}where $A_l(W)=L M_{l\rightarrow L}^2(W,\varepsilon)\cdot F_{l-1}(W,X)F_{l-1}^\top(W,X)$ when taking  $G(W)$ to be $G_\textup{NP}(W)$ defined in \eqref{eq: metric tensor DNN}.  Taking $R_0=R/2^n$ proves Theorem \ref{thm matrix}.

 \hfill $\square$

\section{Ellipsoidal Covering of the Grassmannian (Lemma \ref{lemma ellipsoidal Grassmannian})}\label{sec geometry algebra}

The central goal of this section is to prove the following result on the ellipsoidal metric entropy of the Grassmannian manifold.  The definition for $\Gr$ (Grassmannian manifold), $\St$ (Stiefel parameterization manifold) are temporarily deferred to Section \ref{subsec Grassmannian}.
\begin{definition}[Ellipsoidal Projection Metric]\label{def ellipsoidal}For two subspaces $\cV,\bar\cV\in\Gr(d,r)$, and a positive semidefinite matrix $\Sigma$, define the ellipsoidal projection metric $\varrho_{\textup{proj}, \Sigma}$ by 
\begin{align*}
    \varrho_{\textup{proj}, \Sigma}(\cV, \bar\cV)=\|\Sigma^{\frac{1}{2}}( \mathcal{P}_{\cV}-{\mathcal{P}_{\bar\cV}})\|_{\op},
\end{align*}
where $\mathcal{P}_\cV$ and $\mathcal{P}_{\bar\cV}$ are orthogonal projectors to subspace $\cV$ and $\bar\cV$, respectively. 
\end{definition}
We view orthogonal projectors as matrices (see the definition via the Stiefel parameterization in \eqref{eq: definition projector}), consistent with the earlier operator notation characterized by $\ell_2$–distance in Lemma~\ref{lemma subspace metric dom}. In the isotropic case $\Sigma=I_d$, the ellipsoidal projection metric reduces to the standard isotropic projection metric
\[
  \varrho_{\proj}(\cV,\bar\cV)=\bigl\|\mathcal P_{\cV}-\mathcal P_{\bar\cV}\bigr\|_{\op}.
\]

We now state our main result in this section (Lemma \ref{lemma ellipsoidal Grassmannian} in the main paper).

 Consider the Grassmannian $\Gr(d,r)$ and the uniform prior $\mu=\textup{Unif}(\Gr(d,r))$, then  for every $\cV\in \Gr(d,r)$,  every $\varepsilon>0$ and every PSD matrix $\Sigma$ with eigenvalues $\lambda_1\geq \cdots\lambda_d\geq 0$, we have
 \begin{align}\label{eq: Grassmannian appendix}
  \log\frac{1}{\mu(B_{\varrho_{\proj,\Sigma}}(\cV,\varepsilon))}\leq 
  \frac{r}{2}\sum_{k=1}^{d-r}\log \frac{C\max\left\{\lambda_k,\varepsilon^2\right\}}{\varepsilon^2}+\frac{d-r}{2}\sum_{k=1}^r \log\frac{C\max\left\{\lambda_k, \varepsilon^2\right\}}{\varepsilon^2},
\end{align}
where $C>0$ is an absolute constant.

Recall that the traditional covering number bound for the Grassmannian manifold states that
\begin{align}\label{eq: classical}
\left(\frac{c}{\varepsilon}\right)^{r(d-r)}\leq \rN(\Gr(d,r),\varrho_{\textup{proj}},\varepsilon)\leq \left(\frac{C}{\varepsilon}\right)^{r(d-r)}.
\end{align}
Here \(\rN(\cF,\varrho,\varepsilon)\) is the standard covering number—
the smallest size of an \(\varepsilon\)-net that covers \(\cF\) under
the metric \(\varrho\); see Definition \ref{def covering number} for details. 
In comparison, Lemma \ref{lemma ellipsoidal Grassmannian} is much more challenging than proving classical isotropic covering number bounds \eqref{eq: classical} because 
\begin{itemize}
    \item 1) we consider  ellipsoidal metric;
    \item 2) we require the prior $\mu$ to be independent with $\Sigma$ and $\varepsilon$.
\end{itemize} We need to firstly understand how such classical results are proved, and then proceed to generalized them. This suggests that deep mathematical insights  are necessary for the purpose to study neural networks generalization, as we will introduce below.
\paragraph{From Pure Mathematics to Machine Learning Language.}  Understanding the classical proof for the Grassmannian and generalizing them to prove Lemma \ref{lemma ellipsoidal Grassmannian} necessitate the a deep dive in to the geometry and algebra of subspaces and Grassmannians. In fact, 
     traditional treatments to study Grassmannian manifold often invoke advanced machinery—ranging from differential geometry \citep{bendokat2024grassmann} and Lie‐group theory \citep{szarek1997metric} to algebraic geometry \citep{devriendt2024two}, and the seminal covering number proof \citep{szarek1997metric}  is particularly stated in Lie-algebra and differential-geometry language. 
     
     Motivated by the subsequent covering number proof \citep{pajor1998metric} that uses relatively more elementary language, we give an exposition that is elementary and entirely self‐contained, relying only on matrix‐analysis and learning‐theoretic techniques familiar from machine learning.  In particular, every “advanced” fact—for example, the group theory of continuous symmetries traditionally handled via Lie groups—is derived by elementary means (explicit matrix parameterizations, principal-angle/cosine-sine representations, and basic spectral arguments) while preserving the high-level geometric intuition.  We hope that this versatile framework—and our novel contributions (e.g., Definition \ref{def ellipsoidal} and Lemma \ref{lemma ellipsoidal Grassmannian}), which are new even in a pure‐mathematics setting—will establish subspaces, the Grassmannian, and their underlying algebraic structures as powerful tools for future machine learning applications.

\paragraph{Effective Rank vs. Full-Spectrum Complexity.}
Consider a covariance matrix 
$\Sigma$
 with eigenvalues 
$\lambda_1\geq\cdots\lambda_d\geq 0$. By Definition \ref{def ellipsoidal}, the ellipsoidal metric satisfies 
\begin{align*}
\varrho_{\textup{proj},\Sigma}(\cV,\bar\cV)\leq \lambda_1^{\frac{1}{2}}\varrho_{\textup{proj}}(\cV,\bar\cV).
\end{align*} If one is willing to accept a coarser complexity scaling, then one could invoke existing Grassmannian covering results under the canonical isotropic metric \eqref{eq: classical} (taking
$\mu=\mathrm{Unif}(\Gr(d,r))$) and obtain
\begin{align}\label{eq: eff rank Grassmannian}
\log\frac{1}{\mu\!\left(B_{\varrho_{\mathrm{proj},\Sigma}}(\cV,\varepsilon)\right)}
\;\le\;
\log\frac{1}{\mu\!\left(B_{\varrho_{\mathrm{proj}}}\!\left(\cV,\varepsilon/\sqrt{\lambda_1}\right)\right)}
\;\le\;
\bigl(d-r_{\mathrm{eff}}(\Sigma,R,\varepsilon)\bigr)\,
r_{\mathrm{eff}}(\Sigma,R,\varepsilon)\,
\log\!\frac{C\lambda_1}{\varepsilon^{2}} .
\end{align} 
However, this makes the \emph{global atlas} cost dominate the \emph{local chart} cost, yielding a suboptimal bound than the full–spectrum effective dimension in \eqref{eq: Grassmannian appendix}.  
The refined analysis in this section—also simplifying and strengthening the isotropic route—establishes the correct structural
principle: the global–atlas cost must be balanced by the local–chart cost.  
Thus, while the effective–rank bound \eqref{eq: eff rank Grassmannian} serves as a useful sanity check, the full–spectrum treatment is
what delivers the sharpened complexities required for our main results.

\subsection{Grassmannian Manifold,  Stiefel Parameterization, and Orthogonal Groups}\label{subsec Grassmannian}

Fix integers $r\leq d$. Define
\[
  \textup{Gr}(d,r)\;:=\;\bigl\{\text{$r$–dimensional linear subspaces of }\bR^d\bigr\}
\]
as the \emph{Grassmann manifold}.  
Write
\[
  \textup{St}(d,r)\;:=\;\bigl\{\,V\in\bR^{d\times r}\;:\;V^{\top}V=I_r\,\bigr\}
\]
for the \emph{Stiefel manifold} of $r$ orthonormal columns in $\bR^d$.
$\textup{St}(d,r)$ is a convenient \emph{parameterization} of that class $\textup{Gr}(d,r)$. 

 If for subspace $\cV\in\Gr(d,r)$ and matrix $V\in \St(d,r)$ we have $\cV=\textup{span}(V)$, then we say $V$ is a {\it parameterization matrix} of $\cV$.  Though such parameterization is not unique, the associated orthogonal projector and projection metric are both unique. Moreover, the anisometric  projection we define in Definition \ref{def ellipsoidal} is also unique. We will prove these shortly.

Write 
\begin{align*}
    O(r):=\{ Q\in\bR^{r\times r}: Q^\top Q=QQ^\top=I_r\}
\end{align*}
to be the {\it orthogonal group}. Optionally, we also state that (in the real setting)
\begin{align}\label{eq: isomorphism}
  \mathrm{Gr}(d,r)\;\cong\;O(d)\bigl/\bigl(O(r)\times O(d-r)\bigr)
  \;\cong\;\mathrm{Gr}\bigl(d,\,d - r\bigr),
\end{align}
where “\(/\)” denotes the {\it quotient} and “\(\cong\)” denotes a canonical {\it isomorphism} (indeed, a {\it diffeomorphism} of smooth manifolds or a {\it homeomorphism} of topological manifolds; see, e.g., Chapter 1.5 in \citep{awodey2010category}).  Moreover, \(\mathrm{Gr}(d,r)\) can be regarded as a standard {\it algebraic variety} \citep{devriendt2024two}.  We do not aim to explain these notions in detail, but merely note that:
\begin{enumerate}
  \item The geometric properties of \(\mathrm{Gr}(d,r)\) coincide with those of \(\mathrm{Gr}(d,d-r)\) under this isomorphism (geometric equivalence).
  \item The number of degrees of freedom of \(\mathrm{Gr}(d,r)\) is
  \begin{align}\label{eq: degree of freedom}
    \underbrace{\frac{d(d-1)}{2}}_{\dim O(d)}
    \;-\;
    \underbrace{\frac{r(r-1)}{2}}_{\dim O(r)}
    \;-\;
    \underbrace{\frac{(d-r)(d-r-1)}{2}}_{\dim O(d-r)}
    \;=\;r(d-r),
  \end{align}
  which also appears as the dimension factor in the precise covering‐number bounds \eqref{eq: classical}.
\end{enumerate}

We now define the orthogonal projector and the projection metric on the Grassmannian manifold.

\paragraph{Definition of Orthogonal Projector.}
For $V\in\textup{St}(d,r)$ and its column-space $\cV=\textup{span}(V)$, define the rank-$r$ orthogonal projector\footnote{By elementary linear algebra, the matrix definition of the orthogonal projector $\mathcal P$ here coincides with the $\ell_2-$projection characterized in Lemma~\ref{lemma subspace metric dom}; thus the notation is consistent.}
\begin{align}\label{eq: definition projector}
  \mathcal{P}_\cV := VV^{\!\top}\;\;\in\bR^{d\times d}.
\end{align}
Then $\mathcal{P}_\cV$ depends \emph{only} on the subspace $\cV$.
Indeed, if $Q\in O(r)$ then $(VQ)(VQ)^{\top}=VQQ^{\top}V^{\top}=VV^{\top}$,
so $V$ and $VQ$ represent the same subspace.
Hence the map 
\[
  \Psi:\;\textup{St}(d,r)\;\longrightarrow\;\textup{Gr}(d,r),
  \quad
  V\;\mapsto\;\textup{span}(V),
\]
is an \emph{$O(r)-$quotient}: two frames give the same subspace
iff they differ by a right orthogonal factor.

\paragraph{Ellipsoidal Projection Metric.} 
Following Definition \ref{def ellipsoidal},  for $\cV,\bar\cV\in\textup{Gr}(d,r)$,
\begin{align}\label{eq: projection metric}
  \varrho_{\textup{proj}, \Sigma}(\cV,\bar\cV) \;:=\;\|\Sigma^{\frac{1}{2}}(\mathcal{P}_\cV-\mathcal{P}_{\bar\cV})\|_{\op},
\end{align}
where $\mathcal{P}_\cV:=VV^{\!\top}$ for \emph{any} $V$ such that $\textup{span}(V)=\cV$  
(similarly $\mathcal{P}_{\bar\cV}$).  Because $\mathcal{P}_\cV$ is unique for each subspace,   
$\varrho_{\textup{proj},\Sigma}$ is well defined (independent of the chosen $V$).  
The metric can be pulled back to $\textup{St}(d,r)$: 
\begin{align}\label{eq: projection metric Stiefel}
  \varrho_{\textup{proj},\Sigma}(V,\bar V)\;:=\;\varrho_{\textup{proj},\Sigma}\!\bigl(\textup{span}(V),\textup{span}(\bar V)\bigr)
  \;=\;\|\Sigma^{\frac{1}{2}}(V V^\top-\bar V{\bar V}^{\top})\|_{\op}.
\end{align}

\subsection{Principal Angles between Subspaces}\label{subsec princial angles}

W study how metrics and angles between  images $\mathcal{V}$ and $\bar\cV$ affect their spectral properties. We introduce principal angles and the cosine–sine (CS) decomposition—standard tools for analyzing subspaces (see, e.g., Chapter 6.4.3 in \citep{golub2013matrix}).

\paragraph{Principle Angles and Cosine-Sine Representation.}
Let $U$ and $\bar{U}$ be two $d\times d$ orthogonal matrix, and $V$ and $\bar V$ be the first $r$ columns of $U$ and $\bar{U}$, respectively. We are interested in studying the metrics and angles between $r-$dimensional subspaces $\cV=\textup{span}(V)$ and $\bar \cV=\textup{span}(\bar V)$. Formally, denote
\[
U,\;\bar U \;\in\; O(d),
\qquad
U = \bigl[V \;\; V_{\perp}\bigr], 
\;\;
\bar U = \bigl[\bar V \;\; \bar V_{\perp}\bigr],
\]
where
\[
V,\bar V \in \mathbb{R}^{\,d\times r}, 
\quad
V^{\top}V = I_{r}, 
\quad
\bar V^{\top}\bar V = I_{r},
\]
and
\[
V_{\perp},\bar V_{\perp}\in\mathbb{R}^{\,d\times(d-r)},
\quad
V_{\perp}^{\top}V_{\perp} = I_{\,d-r}, 
\quad
\bar V_{\perp}^{\top}\bar V_{\perp} = I_{\,d-r}.
\]

Since \(U,\bar U\in O(d)\), their product \(U^{\top}\bar U\) is itself orthogonal.  Writing
\[
U^{\top}\,\bar U
\;=\;
\begin{pmatrix}
V^{\top}\\[4pt]
V_{\perp}^{\top}
\end{pmatrix}
\;\bigl[\bar V \;\;\bar V_{\perp}\bigr]
\;=\;
\begin{pmatrix}
V^{\top}\bar V        & V^{\top}\bar V_{\perp}\\[6pt]
V_{\perp}^{\top}\bar V & V_{\perp}^{\top}\bar V_{\perp}
\end{pmatrix},
\]
define the four blocks
\begin{align}\label{eq: matrix C}
\underbrace{C}_{r\times r} 
\;=\; V^{\top}\bar V,
\quad
\underbrace{C_{\perp}}_{r\times(d-r)} 
\;=\; V^{\top}\bar V_{\perp},
\end{align}
\begin{align}\label{eq: matrix S}
\underbrace{S}_{(d-r)\times r} 
\;=\; V_{\perp}^{\top}\bar V,
\quad
\underbrace{S_{\perp}}_{(d-r)\times(d-r)} 
\;=\; V_{\perp}^{\top}\bar V_{\perp}.
\end{align}
Thus
\[
U^{\top}\bar U
\;=\;
\begin{pmatrix}
C      & C_{\perp}\\[6pt]
S      & S_{\perp}
\end{pmatrix}
\;\in\; O(d).
\]

Now  we introduce principal angles between \(\cV=\mathrm{span}(V)\) and \(\bar \cV=\mathrm{span}(\bar V)\) by writing
\begin{align}\label{eq: cosine representation}
C = V^{\top}\bar V 
\;=\; 
Q_1\,\diag(\cos\theta_{1},\cdots,\cos\theta_{r})\,W_1^{\top},
\quad
Q_1,W_1\in O(r),
\end{align}
where 
\begin{align*}
    0\leq \theta_1\leq\theta_2\leq\cdots\leq \theta_r\leq \pi/2
\end{align*}
are called the principle angles between subspaces $\cV$ and $\bar\cV$; and where $\{\cos\theta_{1},\cdots,\cos\theta_{r}\}$ are the singular values of $C$. Simultaneously, we have that the eigenvalues of $S$, $C_\perp$, $S_\perp$ are (notation $\textup{sepc}$ means spectrum, the set of singular values)
\begin{align}
    &\textup{spec}(S)= \{\sin \theta_1, \cdots, \sin \theta_{\min\{r,d-r\}}, \underbrace{0, \cdots, 0}_{\max\{d-2r,0\}}\}, \nonumber\\ 
    &\textup{spec}(C_\perp)= \{\sin \theta_1, \cdots, \sin \theta_{\min\{r,d-r\}}, \underbrace{0, \cdots, 0}_{\max\{d-2r,0\}}\}\nonumber\\
    &\textup{spec}(S_\perp)= \{\cos \theta_1,\cdots, \cos\theta_{\min\{r,d-r\}}, \underbrace{1, \cdots, 1}_{\max\{d-2r,0\}}\}.\label{eq: cs representation}
\end{align}
The above representation in \eqref{eq: cosine representation} and \eqref{eq: cs representation}  are without loss of generality: if $r\leq d-r$, then all the four spectrum contain all $r$ principal angles; if $r>d-r$, then only first $d-r$ principal angles $\{\theta_k\}_{k=1}^{d-r}$ can be smaller than  $\pi/2$ and $\theta_k=0$ for all $d-r+1\leq k\leq r$.

The cosine–sine representation of the eigenvalues in \eqref{eq: cosine representation} and \eqref{eq: cs representation} motivates our notation  $C$ and $S$ when defining block matrices in \eqref{eq: matrix C} and \eqref{eq: matrix S}. This  representation is an immediate consequence of the classical CS decomposition for orthogonal matrices \citep{paige1994history, golub2013matrix}, and we henceforth regard the resulting eigenvalue characterization as given.

\paragraph{Projection Metric via Principal Angles.}

For subspaces $\cV$ and $\bar\cV$, recall that for orthogonal projectors 
\[
\mathcal{P}_{\cV}=VV^{\top}, 
\quad
\mathcal{P}_{\bar\cV}=\bar{V}{\bar{V}}^{\top}.
\]
It is known that the projection metric defined in \eqref{eq: projection metric} and \eqref{eq: projection metric Stiefel} are equal to $\sin\theta_r$,  sine of the largest principal angle between the two subspaces. Formally,
there is the fact  (see, e.g., the last equation in Section 6.4.3 in \citep{golub2013matrix}) 
\begin{align}\label{eq: projection equal sine}
\varrho_{\proj}=\|\mathcal{P}_\cV - \mathcal{P}_{\bar\cV}\|_{\mathrm{op}}
\;=\;
\max_{1\le k\le r}\,\sin\theta_{k}
\;=\;\sin \theta_{r}.
\end{align}
Here \(\theta_{i}\) is the \(i\)-th principal‐angle between \(\cV\) and \(\bar\cV\), and the spectral norm of the difference of two projectors equals the largest of these sines.

\subsection{Local Charts of the Grassmannian}\label{subsec isotropic}

In differential geometry, a {\it chart} is a single local coordinate map. An {\it atlas} is the whole collection of charts that covers the manifold. We introduce a useful atlas that consists of finite graph charts, which only rely on elementary linear algebra and avoid more advanced Lie algebra and exponential map techniques in \cite{szarek1997metric}.

Choose a reference subspace $\bar\cV \in \Gr(d,r)$ and its parameterization matrix 
$\bar{V}\in\textup{St}(d,r)$. 
Denote $X\in \bR^{(d-r)\times r}$ to be mappings from $r-$dimensional subspace $\bar\cV $ to $(d-r)-$dimensional subspace $\bar{\cV}_\perp$. Every $r$–dimensional subspace close to $\bar \cV$ can be written as the \emph{graph}  
\begin{align}\label{eq: graph parameterization}
  \cV(X)\;:=\;\textup{span}\Bigl\{
        [\bar V \bar V_\perp]\begin{pmatrix} I_r\\ X\end{pmatrix}
        \Bigr\},
  \qquad
  X\;\in\;\bR^{(d-r)\times r},
\end{align}
where \(\cV(X)\) is the subspace spanned by the columns of
\(\,[\bar V\ \bar V_\perp]\begin{pmatrix} I_r\\ X\end{pmatrix}\) (the matrix multiplication).
Given the reference subspace $\bar\cV$, define the local {\it graph chart} from $\bR^{(d-r)\times r}$ to $\Gr(d,r)$ by
\begin{align}\label{eq: graph chart}
  \phi_{\bar \cV}:\;X\;\longmapsto \cV(X)\in\;\textup{Gr}(d,r).
\end{align}
Note that for the  $(d-r)\times r$ zero matrix (denoted as $0$), we have $\phi_{\bar\cV} (0)=\bar\cV$. 

\paragraph{Intuition for the graph chart.} If a subspace \(\cV\) is close to \(\bar\cV\)—specifically, \(\varrho_{\proj}(\cV,\bar\cV)=\sin\theta_r<1\)—then all principal angles between \(\cV\) and \(\bar\cV\) satisfy $\theta_i<\pi/2$. Equivalently, the orthogonal projection \(\mathcal{P}_{\bar\cV}\) restricted to \(\cV\) is a bijection \(\mathcal{P}_{\bar\cV}|{\cV}:\cV\to\bar\cV\). In the orthonormal basis $[\bar V\ \bar V_\perp]$, this means every \(v\in\cV\) can be written uniquely as
\[
v=[\bar V \bar V_\perp]\begin{pmatrix}\bar v\\ X\,\bar v\end{pmatrix},\qquad \begin{pmatrix}
    \bar v\\ 0
\end{pmatrix}\in\textup{span}\left\{\begin{pmatrix}
    I_r\\ 0
\end{pmatrix}\right\},
\]
for a linear map \(X\in\bR^{(d-r)\times r}\). Thus, locally around \(\bar\cV\) (all principal angles $<\pi/2$), every $r-$plane admits—and is uniquely determined by—its graph parameter $X$.  We call $X$ the {\it graph parameterization} of $\cV(X)$ in this image. This is formalized as the following lemma.
\begin{lemma}[Local Bijection of Graph Chart]\label{lemma bijection graph chart}
Fix an orthonormal decomposition \(\mathbb R^d=\bar\cV\oplus\bar\cV_\perp\) with basis
$[\bar V\ \bar V_\perp]$. Then every $r-$dimensional subspace \(\cV\) such that $\varrho_\proj(\cV,\bar\cV)<1$
(i.e., all principal angles $<\pi/2$) can be written uniquely as a graph
\[
\cV=\phi_{\bar \cV}(X)
=\operatorname{span}\!\left\{[\bar V\ \bar V_\perp]\binom{I_r}{X}\right\},
\qquad X\in\mathbb R^{(d-r)\times r}.
\]
    
\end{lemma}
\paragraph{Proof of Lemma \ref{lemma bijection graph chart}:} If \(V\in\St(d,r)\) spans \(\cV\), block it in the $[\bar V\ \bar V_\perp]$ basis: denote
\begin{align*}
\binom{A}{B} := \begin{pmatrix}\bar V^\top\\ \bar V_\perp^\top\end{pmatrix}V
\quad (A\in\mathbb R^{r\times r},\ B\in\mathbb R^{(d-r)\times r}).
\end{align*}
Then by the principal angle representation \eqref{eq: cosine representation}, $A=\bar{V}^\top V$ is invertible iff all principal angles $<\pi/2$, and choosing
\[
\ X = B\,A^{-1}
\]
leads to 
\begin{align*}
    \cV=\textup{span}(V)=\textup{span}\left\{   [\bar V\bar V_\perp] \begin{pmatrix}
     A\\ B
    \end{pmatrix}\right\}=\textup{span}\left\{  [\bar V\bar V_\perp]  \begin{pmatrix}
        I_r\\ X
    \end{pmatrix}\right\},
\end{align*}
where the last equality is because for invertible $A$ one always have $\textup{span}(ZA)=\textup{span}(Z)$ for any matrix $Z$.

We have already shown existence. For uniqueness, assuming there are two different $X_1, X_2$ such that $\phi_{\bar \cV}(X_1)=\phi_{\bar \cV}(X_2)$. Because two bases of the same $r$–dimensional subspace differ by an invertible change of coordinates, so there exists an invertible $r\times r$ matrix $Y$ such that
\begin{align*}
    [\bar V\bar V_\perp]  \begin{pmatrix}
        I_r\\ X_1 
    \end{pmatrix}Y= [\bar V\bar V_\perp]  \begin{pmatrix}
        I_r\\ X_2 
    \end{pmatrix},
\end{align*}
which results in $Y=I_r$ and $X_1=X_2$. Thus the parameterization $X$ of $\cV$ is unique.

\hfill$\square$

\paragraph{Sine-tangent relationship in graph chart.} We will show that there is a sine-tangent relationship between $\varrho_{\textup{proj}}(\cV,\bar\cV)$ and $\|X\|_\op$. To be specific, we have the following lemma.

\begin{lemma}[Sine-Tangent Relationship in Graph Chart]\label{lemma graph chart}Denote $\theta_r$ is the maximal principal angle between the subspaces $\cV(X)$ and $\bar\cV$, defined in \eqref{eq: cosine representation}. For the graph chart \eqref{eq: graph chart}, we have \begin{align*}
    \varrho_{\textup{proj}}(\cV(X),\bar\cV)=\sin\theta_r, \quad \|X\|_\op=\tan\theta_r. 
\end{align*}
The above relationship immediately implies that 
\begin{align*}
\varrho_\textup{proj}(\cV(X),\bar\cV)=\|X\|_\op/\sqrt{1+\|X\|^2_{\op}}.
\end{align*}
\end{lemma}

\paragraph{Proof of Lemma \ref{lemma graph chart}:}
Given the fact $\varrho_{\proj}(\cV(X), \bar{\cV}) = \sin \theta_r$ (which is already shown in \eqref{eq: projection equal sine}), where $\theta_r$ is the largest principal angle between the subspaces $\cV(X)$ and the reference subspace $\bar{\cV}$, we want to show $\|X\|_{\op} = \tan \theta_r $.

\paragraph{Step 1: Setup and Simplification.}
The projection metric is invariant under orthogonal transformations of the ambient space $\bR^d$. We can therefore choose a coordinate system that simplifies the calculations without loss of generality. We choose a basis such that the reference frame $\bar{V}$ and its orthogonal complement $\bar{V}_\perp$ are represented as:
\begin{align}\label{eq: representation identity}
    \bar{V} = \begin{pmatrix} I_r \\ 0 \end{pmatrix} \in \St(d,r), \qquad
    \bar{V}_\perp = \begin{pmatrix} 0 \\ I_{d-r} \end{pmatrix} \in \St(d, d-r).
\end{align}
In this basis, the reference subspace is $\bar{\cV} = \textup{span}(\bar{V})$. The parameterization matrix (orthonormal basis) $V(X)$ for the subspace $\cV(X)$ simplifies to (here $(I_r+X^\top X)^{-1/2}$ normalize $V(X)$ to be an orthogonal matrix):
\begin{align}\label{eq: graph matrix representation}
    V(X) = [\bar{V} \ \bar{V}_\perp] \begin{pmatrix} I_r \\ X \end{pmatrix} (I_r+X^\top X)^{-1/2}
    = I_d \begin{pmatrix} I_r \\ X \end{pmatrix} (I_r+X^\top X)^{-1/2}
    = \begin{pmatrix} I_r \\ X \end{pmatrix} (I_r+X^\top X)^{-1/2},
\end{align}
where the second equality follows from our choice of basis without loss of generality: the reference frame \(\bar V\) and its
complement \(\bar V_\perp\) are represented as block identity matrices as in
\eqref{eq: representation identity}.

\paragraph{Step 2: Projection Metric and Principal Angles.}

A fundamental result in matrix analysis, our equation \eqref{eq: cosine representation}, states that the cosines of the principal angles, $\cos\theta_i$, between two subspaces spanned by orthonormal bases $V$ and $\bar V$ are the singular values of $V^\top \bar{V}$.
In our case, the principal angles between $\cV(X)$ and $\bar{\cV}$ are determined by the singular values of $V(X)^\top \bar V$---which are, equivalently, the singular values of $\bar V ^\top V(X)$.

\paragraph{Step 3: Calculation of $\cos \theta_i$.}
Let's compute the matrix product $\bar{V}^\top V(X)$ using our simplified forms:
\begin{align*}
    \bar{V}^\top V(X) &= \begin{pmatrix} I_r & 0 \end{pmatrix} \left[ \begin{pmatrix} I_r \\ X \end{pmatrix} (I_r+X^\top X)^{-1/2} \right] \\
    &= \left( \begin{pmatrix} I_r & 0 \end{pmatrix} \begin{pmatrix} I_r \\ X \end{pmatrix} \right) (I_r+X^\top X)^{-1/2} \\
    &= I_r \cdot (I_r+X^\top X)^{-1/2} \\
    &= (I_r+X^\top X)^{-1/2}.
\end{align*}
To find the singular values of this matrix, we use the Singular Value Decomposition (SVD) of $X$. Let $X = U\Sigma W^\top$, where $U \in \bR^{(d-r)\times(d-r)}$ and $W \in \bR^{r\times r}$ are orthogonal, and $\Sigma \in \bR^{(d-r)\times r}$ is a rectangular diagonal matrix with the singular values $\lambda_1 \ge \lambda_2 \ge \dots \ge 0$ on its diagonal. The spectral norm is $\|X\|_{\op} = \lambda_1$.

Then, $X^\top X = (U\Sigma W^\top)^\top(U\Sigma W^\top) = W\Sigma^\top U^\top U\Sigma W^\top = W\Sigma_r^2 W^\top$, where $\Sigma_r^2$ is the $r \times r$ diagonal matrix with entries $\lambda_i^2$.
So, the matrix $I_r + X^\top X = W(I_r + \Sigma_r^2)W^\top$. Its inverse square root is: $(I_r + X^\top X)^{-1/2} = W(I_r + \Sigma_r^2)^{-1/2}W^\top$.

The singular values of $\bar{V}^\top V(X)$ are the diagonal entries of $(I_r + \Sigma_r^2)^{-1/2}$, which are:
$ s_i = \frac{1}{\sqrt{1+\lambda_i^2}}$.
These singular values are the values of $\cos \theta_i$. The largest principal angle, $\theta_r$, corresponds to the smallest cosine value. This occurs when the singular value $\lambda_i$ is largest, i.e., for $\lambda_1 = \|X\|_{\op}$. Thus,
\[
    \cos \theta_{r}  = \frac{1}{\sqrt{1+\|X\|_{\op}^2}}.
\]

\paragraph{Step 4: Deriving $\tan \theta_{r}$.}
Using the fundamental trigonometric identity $\sin^2\theta + \cos^2\theta = 1$ and the fact that principal angles lie in $[0,\pi/2)$,  we have:
\[
    \tan\theta_{r} =  \|X\|_{\op}.
\]
We have shown that for graph charts, there is the relationship $\varrho_\textup{proj}(\cV(X),\bar\cV)=\sin\theta_r$ and $\|X\|_{\op}=\tan \theta_r$. This suggests
\begin{align*}
    \varrho_\textup{proj}(\cV(X),\bar\cV)=\frac{\|X\|_{\op}}{\sqrt{1+\|X\|_{\op}^2}}.
\end{align*}

\hfill$\square$

\subsection{Global Atlas of Graph Charts}

For the Grassmannian $\Gr(d,r)$ we have that for all $\varepsilon>0$, we have the coarse covering number bound $\mathrm{N}(\Gr(d,r), \varrho_{\textup{proj}}, \varepsilon)\leq C^{\frac{r(d-r)}{\varepsilon}}$,
where $C>0$ is an absolute constant. This is a coarse bound—its dependence is exponential in $1/\varepsilon$ (hence not rate–optimal; the optimal dependence is polynomial)—and we use it only as a preliminary supporting estimate. This coarse estimate suggests that, a finite $O(e^{r(d-r)})$ number of graph charts are sufficient to cover the entire $\Gr(d,r)$ such that every subspace $\cV\in \Gr(d,r)$ is contained in the image of a graph chart with its graph parameterization $X$ satisfies $\|X\|_\op\leq 1$. From this intuition, we have the following lemma.
\begin{lemma}[Pointwise Dimension Consequence of Finite Global Atlas]\label{lemma finite atlas}
 The uniform prior $\mu=\textup{Unif}(\Gr(d,r))$ satisfies that for every $\cV\in \Gr(d,r)$, every PSD matrix $\Sigma$ and every $\varepsilon>0$,
\begin{align*}
    \log\frac{1}{\mu(B_{\varrho_{\proj,\Sigma}}(\cV,\varepsilon))}\leq C_1 r(d-r)+\sup_{X\in \cX}\log \frac{1}{\textup{Unif}(\bar \cX)\{X'\in\bar\cX:\varrho_{\proj,\Sigma}(\cV(X), \cV(X')\}\leq\varepsilon)},
\end{align*}
where $\cX=\{X\in\bR^{(d-r)r}:\|X\|_\op\leq 1\}$ and $\bar\cX=\{X\in\bR^{(d-r)r}:\|X\|_\op\leq 2\}$ (we make $\bar\cX$ slightly larger than $\cX$ for later technical derivation), $\textup{Unif}(\bar \cX)\{\cdot\}$ is the uniform measure over $\bar \cX$, and $C_1>0$ is an absolute constant.
\end{lemma}

    \paragraph{Proof of Lemma \ref{lemma finite atlas}:} Proposition 6 in \citep{pajor1998metric} prove a coarse covering number bound
\begin{align*}
    \mathrm{N}(\Gr(d,r), \varrho_{\textup{proj}}, \varepsilon)\leq C^{\frac{r(d-r)}{\varepsilon}}
\end{align*}where $C>0$ is an absolute constant; this coarse estimate is exponential rather than polynomial in $\varepsilon$, so it is used only for preliminary supporting purposes. For every \(\cV\in\Gr(d,r)\), by the homogeneity of the Grassmannian (under the action of \(O(d)\)),
the \(\varrho_{\proj}\)-ball \(B_{\proj}(\cV,\varepsilon)\) has volume independent of its center. We
therefore refer to this common value as the volume of an \(\varepsilon\)–\(\varrho_{\proj}\) ball, written as $\textup{Vol}(\textup{$\varepsilon-\varrho_\proj$ ball})$. By the definition of covering number (see Definition~\ref{def covering number} and the subsequent inequality for background), we have that
\begin{align*}
    \mathrm{N}(\Gr(d,r), \varrho_\proj, \varepsilon)\cdot \textup{Vol}(\textup{$\varepsilon-\varrho_\proj$ ball})\geq \textup{Vol}(\Gr(d,r)),
\end{align*}
then for the uniform prior $\nu=\textup{Unif}(\Gr(d,r))$, we have that for every $\bar\cV\in\Gr(d,r)$,
\begin{align*}
    \log\frac{1}{\nu(B_{\varrho_\proj}(\bar\cV,\varepsilon))}=\log\frac{\textup{Vol}(\Gr(d,r))}{\textup{Vol}(\textup{$\varepsilon-\varrho_\proj$  ball})}\leq r(d-r)\frac{\log C}{\varepsilon}.
\end{align*}
Note that $\varrho_\proj$ is not the target metric; our goal is the ellipsoidal metric $\varrho_{\proj,\Sigma}$. Taking $\varepsilon=1/\sqrt{2}$, we obtain: 
\begin{align}\label{eq: finite atlas entropy}
    \log\frac{1}{\nu(B_{\varrho_\proj}(\bar\cV,1/\sqrt{2}))}\leq C_1 r(d-r),
\end{align}
 where $C_1>0$ is an absolute constant.  By Lemma \ref{lemma graph chart}, we have that inside the ball $B_{\varrho_\proj}(\bar\cV,1/\sqrt{2})$, by choosing $\bar\cV$ as the reference subspace, the graph parameterization $X$ of $\cV$ satisfies \begin{align*}  \|X\|_{\op}\leq  1,
\end{align*}
which follows from that if \( \varrho_\textup{proj}(\cV(X),\bar\cV) \le 1/\sqrt{2}\) 
(i.e., \( \sin\theta_r \le 1/\sqrt{2} \)), we have \( \|X\|_\op \le 1 \).
See \eqref{eq: graph parameterization} for the definition of this graph chart parameterization; the existence and uniqueness of the parameterization $X$ is by Lemma \ref{lemma bijection graph chart} (local bijection of graph chart).
Furthermore, again by Lemma \ref{lemma bijection graph chart} and Lemma \ref{lemma graph chart}, $\cX=\{X\in\bR^{(d-r)r}:\|X\|_\op\leq 1\}$ satisfies ($\cong$ means isomorphism/bijection)
\begin{align}\label{eq: bijection cX}
  B_{\varrho_\proj}(\bar\cV,1/\sqrt{2})\cong \cX \subset \bar\cX\cong B_{\varrho_\proj}(\bar\cV,2/\sqrt{5}).
\end{align}
Let
\begin{align*}
 \mu_{\bar\cV}=\textup{Unif}(B_\proj(\bar\cV, 2/\sqrt{5})), \quad \mu(\cV)=\int \nu(\bar\cV)\mu_{\bar\cV}(\cV)d{\bar\cV}=\textup{Unif}(\Gr(d,r)).
\end{align*}
Then we have
\begin{align*}
    \log\frac{1}{\mu(B_{\varrho_{\proj,\Sigma}}(\cV,\varepsilon))}= &\log\frac{1}{\int\nu(\bar\cV)\mu_{\bar\cV}(B_{\varrho_{\proj,\Sigma}}(\cV, \varepsilon))d{\bar\cV}}\\=&\log\frac{1}{\int \nu(\bar\cV)\mu_{\bar\cV}(B_{\varrho_{\proj,\Sigma}}(\cV, \varepsilon)\cap B_\proj(\bar\cV, 2/\sqrt{5}))d\bar\cV}\\\leq &\log\frac{1}{\nu(B_{\varrho_{\proj}}(\cV,1/\sqrt{2}))\underset{\bar\cV\in B_{\varrho_{\proj}}(\cV,1/\sqrt{2})}{\min}\mu_{\bar\cV}(X'\in \bar\cX:\varrho_{\proj,\Sigma}(\cV(X),\cV(X') )\leq \varepsilon)}\\
    \leq &C_1 r(d-r)+\sup_{X\in\cX}\log \frac{1}{\textup{Unif}(\bar\cX)\{X'\in\bar\cX:\varrho_{\proj,\Sigma}(\cV(X), \cV(X'))\leq\varepsilon\}},
\end{align*}
where the first inequality is by restricting $\bar\cV$ to $B_{\varrho_{\proj}}(\cV,1/\sqrt{2})$; and the second inequality is by \eqref{eq: finite atlas entropy} as well as the bijection stated in \eqref{eq: bijection cX} and Lemma \ref{lemma bijection graph chart}. Note that we use different radius here than in $\mu_{\bar\cV}$ to enusre that the set $\bar\cX$ for $X'$, which is inside the uniform distribution in the final bound, to be larger than the domain $\cX$ for $X$ to take sup. This will help later technical derivation.

\hfill$\square$

\subsection{Decomposition and Lipschitz Properties inside Graph Chart}
We apply a non-perturbative analysis to the ellipsoidal projection metric.

\begin{lemma}[Non-Perturbative Decomposition of Projector Difference]\label{lemma decomposition of graph chart}
 Let $X, X' \in \mathbb{R}^{(d-r) \times r}$ be two matrices.
Given any reference subspace $\bar\cV$, consider the graph chart $\phi_{\bar\cV}: X\mapsto \cV(X)$ defined in  \eqref{eq: graph parameterization}.
Then the difference between two projectors $\mathcal{P}_{\cV(X)}$, $\mathcal{P}_{\cV(X')}$ be decomposed as follows: 
\begin{align*} 
&\mathcal{P}_{\cV(X)}-\mathcal{P}_{\cV(X')}\\
= &\mathcal{P}_{{\cV(X)}_\perp}\begin{pmatrix}
        0 \\ I_{d-r}
    \end{pmatrix}(X-X')\begin{pmatrix}
        I_r & 0
    \end{pmatrix}\mathcal{P}_{{\cV(X')}}+\mathcal{P}_{\cV(X)}\begin{pmatrix}
        I_r \\ 0
    \end{pmatrix}(X^\top-{X'}^\top)\begin{pmatrix}
        0 & I_{d-r}
    \end{pmatrix}\mathcal{P}_{{\cV(X')}_{\perp}}.
    \end{align*}

\end{lemma}
\paragraph{Proof of Lemma \ref{lemma decomposition of graph chart}:} The projector is invariant under orthogonal transformations of the ambient space $R^d$. We can therefore choose a coordinate system that simplifies the
calculations without loss of generality. By the matrix representation \eqref{eq: graph matrix representation} (which, without loss of generality,  uses a convenient orthogonal basis specified by \eqref{eq: representation identity}), we denote
\begin{align*}
    A(X)=\begin{pmatrix}
        I_r \\ X
    \end{pmatrix}, \quad M(X)=(I_r+X^\top X)^{-1}, 
\end{align*}and have the following facts: 
\begin{align}
V(X)=&A(X)M(X)^{1/2},\nonumber\\
\mathcal{P}_{\cV(X)}=&A(X)M(X)A(X)^\top=A(X)M(X)\begin{pmatrix}
    I_r & X^\top
\end{pmatrix}\label{eq: matrix representation graph 2}\\
    \mathcal{P}_{\cV(X)}-\mathcal{P}_{\cV(X')}=&A(X)M(X)A(X)^\top-A(X')M(X')A(X')^\top\nonumber\\
    A(X)M(X)=&\mathcal{P}_{\cV(X)}\begin{pmatrix}
        I_r \\ 0
    \end{pmatrix}\label{eq: matrix representation graph 4}\\
    A(X)M(X)X^\top=  &\mathcal{P}_{\cV(X)}\begin{pmatrix}
        0 \\ I_{d-r}
    \end{pmatrix}\label{eq: matrix representation graph 5},
\end{align}
where \eqref{eq: matrix representation graph 4} and \eqref{eq: matrix representation graph 5} are straightforward consequences of \eqref{eq: matrix representation graph 2}.

We begin with a non-perturbative decomposition:
\begin{align}\label{eq: decomposition projector difference 0}
&  \mathcal{P}_{\cV(X)}-\mathcal{P}_{\cV(X')}\nonumber\\
=&A(X)M(X)A(X)^\top - A(X')M(X')A(X')^\top\nonumber\\ = &(A(X)-A(X'))M(X')A(X')^\top+A(X)(M(X)-M(X'))A(X')^\top+ A(X)M(X)(A(X)-A(X'))^\top .
\end{align}

We continue to decompose each term non-perturbatively. First,  
\begin{align}\label{eq: decomposition projector difference 1}
    & (A(X)-A(X'))M(X')A(X')^\top\nonumber\\=&\mat{0 \\ X-X'} M(X')A(X')^\top\nonumber\\=&\begin{pmatrix}
        0 \\ I_{d-r}
    \end{pmatrix} (X-X')M(X')A(X')^\top\nonumber\\=& \begin{pmatrix}
        0 \\ I_{d-r}
    \end{pmatrix} (X-X')\begin{pmatrix}
        I_r & 0
    \end{pmatrix}\mathcal{P}_{\cV(X')},
\end{align}
where the last equality uses the fact \eqref{eq: matrix representation graph 4} and symmetry of $\cP_{\cV(X)}$. 

Second, because we have the non-perturbative decomposition
\begin{align*}
    &M(X)-M(X')\\=&(I_r+{X}^\top X)^{-1}\left((I_r+{X'}^\top X')-(I_r+X^\top X)\right)(I_r+{X'}^\top X')^{-1}\\
    = &(I_r+{X}^\top X)^{-1}\left({X'}^\top X'-X^\top X\right)(I_r+{X'}^\top X')^{-1}\\
    =& (I_r+{X}^\top X)^{-1}\left(X^\top(X'-X)+({X'}^\top-X^\top)X'\right)(I_r+{X'}^\top X')^{-1}\\
    = &M(X)X^\top(X'-X)M(X')+ M(X)({X'}^\top-X^\top)X'M(X'),
\end{align*}
we have
    \begin{align}\label{eq: decompisition projector difference 2}
        &A(X)(M(X)-M(X'))A(X')^\top\nonumber\\=  &A(X)M(X)X^\top(X'-X)M(X')A(X')^\top+A(X)M(X)({X'}^\top-X^\top)X'M(X')A(X')^\top\nonumber\\
        =& -\mathcal{P}_{\cV(X)}\begin{pmatrix}
            0\\I_{d-r}
        \end{pmatrix}(X-X')\begin{pmatrix} I_r & 0 \end{pmatrix}\cP_{{\cV(X')}}-\cP_{\cV(X)}\begin{pmatrix}
            I_r \\ 0
        \end{pmatrix}(X^\top-{X'}^\top)\begin{pmatrix}
            0 & I_{d-r}
        \end{pmatrix}\cP_{\cV(X')},
    \end{align}
where the last equality uses the fact \eqref{eq: matrix representation graph 4} and the fact \eqref{eq: matrix representation graph 5}.

    Third, we have 
    \begin{align}\label{eq: decomposition projector difference 3}
     &A(X)M(X)(A(X)-A(X'))^\top\nonumber\\=   &A(X)M(X)\begin{pmatrix} 0 & X^\top-{X'}^\top \end{pmatrix}\nonumber\\=&\cP_{\cV(X)}\begin{pmatrix}
         I_r\\ 0
     \end{pmatrix} (X^\top-{X'}^\top)\begin{pmatrix}
         0 & I_{d-r}
     \end{pmatrix},
    \end{align}
where the last equality uses the fact \eqref{eq: matrix representation graph 4}.

Substituting \eqref{eq: decomposition projector difference 1}, \eqref{eq: decompisition projector difference 2}, \eqref{eq: decomposition projector difference 3} back into \eqref{eq: decomposition projector difference 0}, we have
\begin{align*}
    &\cP_{\cV(X)}-\cP_{\cV(X')} \\=&\begin{pmatrix}
        0 \\ I_{d-r}
    \end{pmatrix} (X-X')\begin{pmatrix}
        I_r & 0
    \end{pmatrix}\mathcal{P}_{\cV(X')}\\&-\mathcal{P}_{\cV(X)}\begin{pmatrix}
            0\\I_{d-r}
        \end{pmatrix}(X-X')\begin{pmatrix} I_r & 0 \end{pmatrix}\cP_{{\cV(X')}}-\cP_{\cV(X)}\begin{pmatrix}
            I_r \\ 0
        \end{pmatrix}(X^\top-{X'}^\top)\begin{pmatrix}
            0 & I_{d-r}
        \end{pmatrix}\cP_{\cV(X')}\\
        &+\cP_{\cV(X)}\begin{pmatrix}
         I_r\\ 0
     \end{pmatrix} (X^\top-{X'}^\top)\begin{pmatrix}
         0 & I_{d-r}
     \end{pmatrix}\\
     =& \mathcal{P}_{{\cV(X)}_{\perp}}\begin{pmatrix}
            0\\I_{d-r}
        \end{pmatrix}(X-X')\begin{pmatrix} I_r & 0 \end{pmatrix}\cP_{{\cV(X')}}+ \cP_{\cV(X)}\begin{pmatrix}
         I_r\\ 0
     \end{pmatrix} (X^\top-{X'}^\top)\begin{pmatrix}
         0 & I_{d-r}
     \end{pmatrix}\cP_{{\cV(X')}_\perp},
\end{align*}
where the last equality uses $I_d-\cP_{\cV(X)}=\cP_{{\cV(X)}_\perp}$ and $I_d-\cP_{\cV(X')}=\cP_{{\cV(X')}_\perp}$.

\hfill$\square$

Building upon the non-perturbative decomposition in Lemma \ref{lemma decomposition of graph chart}, we have the following Lipschitz property of graph chart.

\begin{lemma}[Lipschitz of Graph Chart]\label{lemma Lipschitz of graph chart}
 Let $X, X' \in \mathbb{R}^{(d-r) \times r}$ be two matrices. Given any reference subspace $\bar\cV$, consider the graph chart defined in  \eqref{eq: graph matrix representation}.
Then the ellipsoidal projection metric is Lipschitz to ellipsoidal spectral metrics as follows: for every rank-$r$ PSD ${\Sigma}\in\bR^{d\times d}$, 
\begin{align*} 
&\varrho_{\textup{proj},\Sigma}(\cV(X),\cV(X'))\\
\\ \le &\left\|{\left(\begin{pmatrix}
        0 & I_{d-r}
    \end{pmatrix}\cP_{{\cV(X)}_\perp}^\top\Sigma \cP_{{\cV(X)}_\perp}\begin{pmatrix}
        0 \\ I_{d-r}
    \end{pmatrix}\right)}^{\frac{1}{2}}(X-X')\right\|_{\op}+ \left\|{\left(\begin{pmatrix}
        I_r & 0
    \end{pmatrix}\cP_{\cV(X)}^\top\Sigma\cP_{\cV(X)}\begin{pmatrix}
        I_r \\ 0
    \end{pmatrix}\right)}^{\frac{1}{2}}(X^\top-{X'}^\top)\right\|_{\op}.
    \end{align*}

\end{lemma}

\paragraph{Proof of Lemma \ref{lemma Lipschitz of graph chart}:} By Lemma \ref{lemma decomposition of graph chart}, we have
\begin{align*}
    &\varrho_{\proj, \Sigma}(\cV(X), \cV(X'))=\left\|\Sigma^{\frac{1}{2}}(\cP_{\cV(X)}-\cP_{\cV(X')})\right\|_\op\\
    = &\left\|\Sigma^{\frac{1}{2}}\mathcal{P}_{{\cV(X)}_{\perp}}\begin{pmatrix}
            0\\I_{d-r}
        \end{pmatrix}(X-X')\begin{pmatrix} I_r & 0 \end{pmatrix}\cP_{{\cV(X')}}+ \Sigma^{\frac{1}{2}}\cP_{\cV(X)}\begin{pmatrix}
         I_r\\ 0
     \end{pmatrix} (X^\top-{X'}^\top)\begin{pmatrix}
         0 & I_{d-r}
     \end{pmatrix}\cP_{{\cV(X')}_\perp}\right\|_\op\\
     \leq &\left\|\Sigma^{\frac{1}{2}}\mathcal{P}_{{\cV(X)}_{\perp}}\begin{pmatrix}
            0\\I_{d-r}
        \end{pmatrix}(X-X')\right\|_\op+\left\|\Sigma^{\frac{1}{2}}\cP_{\cV(X)}\begin{pmatrix}
         I_r\\ 0
     \end{pmatrix} (X^\top-{X'}^\top)\right\|_\op\\
     = &\left\|{\left(\begin{pmatrix}
        0 & I_{d-r}
    \end{pmatrix}\cP_{{\cV(X)}_\perp}^\top\Sigma \cP_{{\cV(X)}_\perp}\begin{pmatrix}
        0 \\ I_{d-r}
    \end{pmatrix}\right)}^{\frac{1}{2}}(X-X')\right\|_{\op}+ \left\|{\left(\begin{pmatrix}
        I_r & 0
    \end{pmatrix}\cP_{\cV(X)}^\top\Sigma\cP_{\cV(X)}\begin{pmatrix}
        I_r \\ 0
    \end{pmatrix}\right)}^{\frac{1}{2}}(X^\top-{X'}^\top)\right\|_{\op}.
\end{align*}
where the inequality follows from the triangle inequality and the facts that the spectral norms of  $\cP_{\cV(X')}$, $\cP_{{\cV(X')}_{\perp}}$, and the two block–identity matrices are all at most $1$ (the fact that spectral norms of projectors are at most $1$ can be proved via the first inequality in Lemma \ref{lemma eigenvalue bound}); and the last equality is because for any matrices $A$, $B$ we have
\begin{align*}
    \|\Sigma^{\frac{1}{2}}AB\|_\op=\sqrt{\|B^\top A^\top\Sigma A B\|_\op}=\|(A^\top \Sigma A)^{\frac{1}{2}}B\|_\op.
\end{align*}
\hfill$\square$

We continue to present the following lemma, which implies that the projectors and the block-identity matrices in Lemma \ref{lemma Lipschitz of graph chart} only reduces the effective dimensions of the ellipsoidal map, and does not increase the eigenvalues (up to absolute constants).

\begin{lemma}[Spectral Domination under Contractions]\label{lemma spectral-compression}
Let $\Sigma\succeq 0$ be a $d\times d$ PSD matrix with ordered eigenvalues
$\lambda_1(\Sigma)\ge\cdots\ge\lambda_d(\Sigma)$.
Let $A\in\mathbb{R}^{d\times m}$ for some $m\leq d$ and write $s:=\|A\|_{\op}$.
Denote by $\mu_1\ge\cdots\ge\mu_m$  the eigenvalues of $A^\top\Sigma A$.
Then, for every $k=1,\dots,m$,
\[
\mu_m\ \le\ s^2\,\lambda_m(\Sigma).
\]
\end{lemma}

\paragraph{Proof of Lemma \ref{lemma spectral-compression}:}

By the Courant–Fischer–Weyl max-min characterization (see, e.g., \citep{wiki:min_max_theorem}), we have \begin{align*}
\lambda_k(A^\top\Sigma A)\;=&\;\min_{\substack{S\subset\mathbb{R}^d\\ \dim S=d-k+1}}\ \sup\{\|A^\top\Sigma^{\frac{1}{2}}x\|^2_2:\ x\in S,\ \|x\|_2=1\}\\\leq &\;s^2\cdot \min_{\substack{S\subset\mathbb{R}^d\\ \dim S=d-k+1}}\ \sup\{\|\Sigma^{1/2}x\|_2:\ x\in S,\ \|x\|_2=1\}\\
= &s^2\lambda_k(\Sigma).
\end{align*}
\hfill$\square$

\subsection{Proof of the Main Result}\label{subsec grassmannian main result}
From Lemma \ref{lemma finite atlas}, to cover $\textup{Gr}(d,r)$ it suffices to cover the unit ball of
$(d-r)\times r$ matrices under the ellipsoidal spectral metric. We are now ready to prove Lemma \ref{lemma ellipsoidal Grassmannian}, the main result for ellipsoidal Grassmannian covering.

\paragraph{Proof of Lemma \ref{lemma ellipsoidal Grassmannian}:} We present the proof in multiple parts.

\paragraph{Part 1: Applying Lemma \ref{lemma finite atlas}.}
Define $\cX=\left\{X\in \bR^{(d-r)\times r}:\|X\|_{\op}\le 1 \right\}$ and further $\bar\cX=\left\{X\in \bR^{(d-r)\times r}:\|X\|_{\op}\le 2 \right\}$. By Lemma \ref{lemma finite atlas} (Pointwise Dimension Consequence of Finite Global Atlas), for $\mu=\textup{Unif}(\Gr(d,r))$, we have that for all $\cV\in\Gr(d,r)$ and all $\varepsilon>0$,
\begin{align}
  \log\frac{1}{\mu(B_{\varrho_{\proj,\Sigma}}(\cV,\varepsilon))}\leq C_1 r(d-r)+\sup_{X\in\cX}\log \frac{1}{\textup{Unif}(\bar\cX)\{X'\in\bar\cX:\varrho_{\proj,\Sigma}(\cV(X), \cV(X'))\leq\varepsilon\}}\label{eq: covering number 1},
\end{align}
where $C_1>0$ is an absolute constant.

 Define the $(d-r)\times (d-r)$ positive definite matrices $H_1(X)$ and the $r\times r$ positive definite matrix $H_2(X)$ as the following
\begin{align*}
    H_1(X)&=\begin{pmatrix}
        0 & I_{d-r}
    \end{pmatrix}\cP_{{\cV(X)}_\perp}^\top\Sigma \cP_{{\cV(X)}_\perp}\begin{pmatrix}
        0 \\ I_{d-r}
    \end{pmatrix},\\
    H_2(X)&=\begin{pmatrix}
        I_r & 0
    \end{pmatrix}\cP_{\cV(X)}^\top\Sigma\cP_{\cV(X)}\begin{pmatrix}
        I_r \\ 0
    \end{pmatrix}.
\end{align*}
By Lemma \ref{lemma Lipschitz of graph chart}  (Lipschitz of Graph Chart), 
we have that 
\begin{align*}
    \varrho_{\proj,\Sigma}(\cV(X),\cV(X'))\leq \|H_1(X)^{\frac{1}{2}}(X'-X)\|_{\op}+\|H_2(X)^{\frac{1}{2}}(X'-X)^\top\|_{\op}.
\end{align*}

\paragraph{Part 2: Volumetric Arguments.}
We analyze the log density complexity in \eqref{eq: covering number 1} via volumetric arguments. 

\paragraph{A technical step: ball inclusion via thresholding}
In order to compute the log density complexity with the uniform prior, one needs the operator norm ball to be included in the support of the prior. Given a PSD matrix $H\in \bR^{m\times m}$ and an eigenvalue threshold $\alpha$, assume its eigendecomposition is $H=U \diag(\beta_1,\cdots, \beta_m) U^\top$, define the thresholding function $T_\alpha$ by
\begin{align*}
    T_\alpha(H)=U\diag(\max\{\beta_1,\alpha\},\cdots, \max\{\beta_m,\alpha\})U^\top.
\end{align*}Clearly this function only increases the metric.
 We further define the following two ellipsoidal metrics: 
\begin{align*}
    \varrho^2_1(X,X')=&\|({X'}-X)^\top \bar H_1(X) (X'-X)\|_\op, \quad \bar{H}_1(X)=T_{\varepsilon^2}\left(H_1(X)\right)
    \\
\varrho^2_2(X,X')=&\|(X'-X)\bar H_2(X)(X-X')^\top\|_\op, \quad \bar{H}_2(X)=T_{\varepsilon^2}\left(H_2(X)\right)
\end{align*}
We note that the two balls $B_{\varrho_1}(X, \varepsilon)$, $B_{\varrho_2}(X, \varepsilon)$ are contained in $\bar \cX$, as we have applied the thresholding function to ensure this inclusion. For example, for the first ball, from
\begin{align*}
X'-X= \left(\bar H_1(X)\right)^{-1/2} \underbrace{\left(\bar H_1(X)\right)^{\frac{1}{2}}(X'-X)}_{\textup{spectral norm  $\leq\varepsilon$ for $X'\in B_{\varrho_1}(X,\varepsilon)$}},
\end{align*}
we have (by using the \(\varepsilon\) estimate from the second underbraced term above, and combining it
with the thresholding guarantee \(\lambda_{\min}(\bar H_1(X))\ge \varepsilon^{2}\))
\begin{align*}
    \|X'-X\|_{\op}\leq \lambda_{\min}(\bar H_1(X))^{-1/2} \cdot \varepsilon\leq 1,
\end{align*}
which resulting in $\|X'\|_{\op}\leq \|X'-X\|_{\op}+\|X\|_{\op}\leq 2$ and thus $B_{\varrho_1}(X, \varepsilon)\subseteq \bar \cX$. Similarly, we can show $B_{\varrho_2}(X, \varepsilon)\subseteq \bar \cX$. this gives us the auxiliary ball-inclusion result:
\begin{align}\label{eq: ball inclusion}
    B_{\varrho_1+\varrho_2}(X, \varepsilon)\subseteq B_{\varrho_1}(X, \varepsilon)\cap B_{\varrho_2}(X, \varepsilon)\subseteq  B_{\varrho_1}(X, \varepsilon)\cup B_{\varrho_2}(X, \varepsilon)\subseteq \bar \cX.
\end{align}

Now we are ready to proceed with the main part of the proof. By Lemma \ref{lemma Lipschitz of graph chart}  (Lipschitz of Graph Chart) and the fact that threholding only increases the spectral norm,  the ellipsoidal projection metric is bounded by $\varrho_1+\varrho_2$, so for any $X\in\cX$,
\begin{align}\label{eq: covering number 2}
&\log \frac{1}{\textup{Unif}(\bar\cX)\{X'\in\bar\cX:\varrho_{\proj,\Sigma}(\cV(X), \cV(X'))\leq\varepsilon\}}\nonumber\\
\leq &\log \frac{1}{\textup{Unif}(\bar\cX)\{X'\in\bar\cX: \varrho_1(X,X')+\varrho_2(X,X')\leq \varepsilon\}}\nonumber\\
= &\log \frac{1}{\textup{Unif}(\bar\cX)\{B_{\varrho_1+\varrho_2}(X,\varepsilon)\}}\\
=&\frac{\textup{Vol}(\bar\cX)}{\textup{Vol}(B_{\varrho_1+\varrho_2}(X,\varepsilon))},
\end{align}
where the first equality uses the ball-inclusion result \eqref{eq: ball inclusion}.

\paragraph{Covering number in normed vector space.} Definition~\ref{def covering number} is stated for a general metric space. In the special case of a normed vector space (see, e.g., \cite{wiki:normed_vector_space}), however, the covering number admits a more explicit characterization via standard volume-ratio arguments, up to absolute-constant factors in the radius. This simplification arises from the strong homogeneity of the metric induced by the norm, in particular the absolute homogeneity property
\begin{align}\varrho(\lambda y,\lambda z)=|\lambda|\,\varrho(y,z), \qquad \forall \lambda\in\mathbb{R}.
\end{align}

Classical volume‐ratio  arguments  give the following results on the covering number of balls in general normed vector space $\mathcal{Y}$. For a \(p\)-dimensional normed vector space  equipped with the metric associated to its norm \(\lVert\cdot\rVert\), we denote by \(B(y, R)\) the ball in \(\mathcal{Y}\) centered at $y\in\mathcal{Y}$ with radius $R$, and by \(\textup{N}(\cZ, \|\cdot\|,\varepsilon)\) the covering number of a subset $\cZ \subseteq\mathcal Y$.

Proposition 4.2.10 in \cite{vershynin2018high} (the proof is elementary and clearly holds true for general metric in a normed vector space) states that for $\cZ\subseteq \mathcal{Y}$ and  general metric $\|\cdot\|$, we  have that for any $y\in \mathcal{Y}$,
\begin{align*}
   \frac{\textup{Vol}(\cZ)}{\textup{Vol}(B(y,\varepsilon))} \leq \textup{N}(\cZ, \|\cdot\|,\varepsilon)\leq \frac{\textup{Vol}(\cZ+B(y,\frac{\varepsilon}{2}))}{\textup{Vol}(B(y,\frac{\varepsilon}{2}))},
\end{align*}
where the set $\mathcal{A}+\mathcal{B}:=\{a+b: a\in\mathcal{A}, b\in\mathcal{B}\}$. When $\cZ$ is convex and $B(y,\varepsilon)\subseteq \cZ$, we further have
\begin{align}\label{eq: volume inequality}
   \frac{\textup{Vol}(\cZ)}{\textup{Vol}(B(y,\varepsilon))} \leq \textup{N}(\cZ, \|\cdot\|,\varepsilon)\leq \frac{\textup{Vol}(\cZ+B(y,\frac{\varepsilon}{2}))}{\textup{Vol}(B(y,\frac{\varepsilon}{2}))}\leq \frac{\textup{Vol}(\frac{3}{2}\cZ)}{\textup{Vol}(B(y,\frac{\varepsilon}{2}))}=3^p\frac{\textup{Vol}(\cZ)}{\textup{Vol}(B(y,\varepsilon))},
\end{align}
where $\lambda \mathcal{A}:=\{\lambda a:a\in\mathcal{A}\}$ for $\lambda>0$. Lastly, when the normed space $\mathcal{Y}$ is $p-$dimensional, for every \(\varepsilon \in (0,R]\), setting $\cZ=B(0,R)$ turns the above inequality \eqref{eq: volume inequality} into the optimal covering number bound
\begin{align}\label{eq: covering normed space}
\left(\frac{R}{\varepsilon}\right)^{p}
\;\le\;
\rN\!\bigl(B(0,R),\,\lVert\cdot\rVert,\,\varepsilon\bigr)
\;\le\;
\left(\frac{3R}{\varepsilon}\right)^{p}.
\end{align}
Note that this result is for general normed space, not only for the $\ell_2$ norm in Euclidean space (see, e.g., display (1) in \cite{pajor1998metric}; see also \cite{milman1986asymptotic, pisier1999volume}).

\paragraph{A technical step: lifting to product space.} Consider the product space $\bR^{(d-r)\times r}\times \bR^{(d-r)\times r}$ (of dimension $2\times(d-r)\times r$). Given any $(d-r)\times (d-r)$ positive definite matrix $H_1$ and $r\times r$ positive definite matrix $H_2$, define the modified spectral norm by 
\begin{align*}
    \|(X_1,X_2)-(X_1',X_2')\|_{\op, H_1, H_2}:=\|H_1^{\frac{1}{2}}(X_1-X_1')\|_\op+\|H_2^{\frac{1}{2}}(X_2^\top-{X_2'}^\top)\|_\op.
\end{align*}
Consider the constrained set 
\begin{align*}
\mathcal{S}:=\{(X_1,X_2)\in\bR^{(d-r)\times r}\times \bR^{(d-r)\times r}: X_1=X_2\}=\{(X,X):X\in \bR^{(d-r)\times r}\},
\end{align*} 
which is a normed space with dimension $(d-r)\times r$ (isomorphic to $\bR^{(d-r)\times r}$), equipped with the modifed spectral norm 
\begin{align*}
\|(X,X)-(X',X')\|_{\op, H_1, H_2}=\|H_1^{\frac{1}{2}}(X-X')\|_\op+\|H_2^{\frac{1}{2}}(X^\top-{X'}^\top)\|_\op.
\end{align*}
 Denote $B^{\mathcal{S}}_{\op, H_1, H_2}((X,X),R)=\{(X',X')\in \mathcal{S}: \|(X',X')-(X,X)\|_{\op, H_1, H_2}\leq R\}$ (the ball constrained in $\mathcal{S}$). Because there is a bijective, distance-preserving (isometric) map between $B_{\varrho_1+\varrho_2}(X,\varepsilon)$ and $B^{\mathcal{S}}_{\op, \bar H_1(X), \bar H_2(X)}((X,X), \varepsilon)$, 
 and likewise $B^{\mathcal{S}}_{\op, I_{d-r}, I_r}((0,0),4)$ and $\bar\cX$ (here $0$ denotes the $(d-r)\times r$ $0$ matrix), we obtain 
 \begin{align}\label{eq: covering number 3}
     \frac{\textup{Vol}(\bar\cX)}{\textup{Vol}(B_{\varrho_1+\varrho_2}(X,\varepsilon))}=  \frac{\textup{Vol}(B^{\mathcal{S}}_{\op, I_{d-r}, I_r}((0,0),4))}{\textup{Vol}(B^{\mathcal{S}}_{\op,\bar H_1(X), \bar H_2(X) }((X,X),\varepsilon))},
 \end{align}
 where the volume on $\mathcal{S}$ is defined via the surface area measure. 
\eqref{eq: covering number 3} is exactly the objective we need to bound in \eqref{eq: covering number 2}. 

Given $\varepsilon>0$, by the property \eqref{eq: volume inequality} of covering number, we have that for every  $X\in \cX$ and $\varepsilon>0$,
\begin{align}\label{eq: covering number 4}
    \frac{\textup{Vol}(B^{\mathcal{S}}_{\op, I_{d-r}, I_{r}}((0,0),4))}{\textup{Vol}(B^{\mathcal{S}}_{\op, \bar H_1(X), \bar H_2(X)}((X,X), \varepsilon))}
\;\le\;
\rN\!\bigl(B^{\mathcal{S}}_{\op, I_{d-r}, I_{r}}((0,0),4),\,\lVert\cdot\rVert_{\op, \bar H_1(X), \bar H_2(X)},\,\varepsilon\bigr).
\end{align}
\paragraph{Remark on why lifting to product space double the degree of freedom.} We now lift the $\mathcal{S}-$constrained ball $B^{\mathcal{S}}_{\op, I_{d-r}, I_{r}}((0,0),4)$ to the product space $\bar\cX\times\bar\cX$, using the covering number of the lifted product space to bound the covering number of the original space, in order to obtain an upper bound on \eqref{eq: covering number 4} and \eqref{eq: covering number 3}. This is the reason why our final bound will scale (in the isotropic case) in the order $O((d-r)r\log\frac{1}{\varepsilon^2})=O(2(d-r)r\log\frac{1}{\varepsilon})$ rather than the classical optimal order $\Theta((d-r)r\log\frac{1}{\varepsilon})$---the lifting to product space increase the number of freedom by a multiplicative factor of 2. Nevertheless, such difference is negligible in our theory.

For every $(X_1, X_2)\in \bR^{(d-r)\times r}\times \bR^{(d-r)\times r}$, every $(d-r)\times(d-r)$ matrix $H_1\succ 0$, and every $r\times r$ matrix $H_2\succ 0$, and radius $R$, denote $B_{\op, H_1, H_2}((X_1,X_2),R)$ to be the unconstrained ball in $\bR^{(d-r)\times r}\times \bR^{(d-r)\times r}$:
\begin{align*}
    B_{\op, H_1, H_2}((X_1,X_2),R):=\{(X_1',X_2')\in \bR^{(d-r)\times r}\times \bR^{(d-r)\times r}: \|(X_1,X_2)-(X_1'-X_2')\|_{\op, H_1, H_2}\leq R\}.
\end{align*}
Lifting to the product space can only increase the external covering number (monotonicity under set inclusion), and the external covering number is equivalent to the internal covering number up to a constant factor in the radius. To be specific, by Lemma \ref{lemma int-ext}, we have
\begin{align}\label{eq: first equality lift}
    &\rN\!\bigl(B^{\mathcal{S}}_{\op, I_{d-r}, I_{r}}((0,0),4),\,\lVert\cdot\rVert_{\op, \bar H_1(X), \bar H_2(X)},\,\varepsilon\bigr)\nonumber\\\leq  &\rN_{\textup{ext}}\!\bigl(B^{\mathcal{S}}_{\op, I_{d-r}, I_{r}}((0,0),4),\,\lVert\cdot\rVert_{\op, \bar H_1(X), \bar H_2(X)},\,\varepsilon/2\bigr)\nonumber\\\leq &\rN_{\textup{ext}}\!\bigl(B_{\op, I_{d-r}, I_{r}}((0,0),4),\,\lVert\cdot\rVert_{\op, \bar H_1(X), \bar H_2(X)},\,\varepsilon/2\bigr)\nonumber\\
    \leq &\rN\!\bigl(B_{\op, I_{d-r}, I_{r}}((0,0),4),\,\lVert\cdot\rVert_{\op, \bar H_1(X), \bar H_2(X)},\,\varepsilon/2\bigr).
\end{align}
For every $X\in\cX$, the ball-inclusion argument \eqref{eq: ball inclusion} is strong enough to imply that the unconstrained ball $B_{\op, \bar H_1(X), \bar H_2(X)}((X,X), \varepsilon)\subseteq \bR^{(d-r)\times r}\times \bR^{(d-r)\times r}$ is also included in the lifted ball $B_{\op, I_{d-r}, I_{r}}((0,0),4)$, which gives that
\begin{align*}
   B_{\op, \bar H_1(X), \bar H_2(X)}((X,X), \varepsilon/2)\subset B_{\op, \bar H_1(X), \bar H_2(X)}((X,X), \varepsilon)\subseteq B_{\op, I_{d-r}, I_{r}}((0,0),4).
\end{align*}
This satisfies the inclusion condition required to establish \eqref{eq: volume inequality}, and we have 
\begin{align}\label{eq: result lift}
    \rN\!\bigl(B_{\op, I_{d-r}, I_{r}}((0,0),4),\,\lVert\cdot\rVert_{\op, \bar H_1(X), \bar H_2(X)},\,\varepsilon/2\bigr)\leq &3^{2(d-r)r}\frac{\textup{Vol}(B_{\op, I_{d-r}, I_{r}}((0,0),4))}{\textup{Vol}(B_{\op, \bar H_1(X), \bar H_2(X)}((X,X),\varepsilon/2))}\nonumber\\
    =&6^{2(d-r)r}\frac{\textup{Vol}(B_{\op, I_{d-r}, I_{r}}((0,0),4))}{\textup{Vol}(B_{\op, \bar H_1(X), \bar H_2(X)}((X,X),\varepsilon))}.
\end{align}

\paragraph{Part 3: Applying Change of Variable and Calculating the Jacobian Determinant.}
Applying the standard change of variables
\[
Y_1=\bar H_1(X)^{1/2} X_1,\qquad
Y_2=X_2\,\bar H_2(X)^{1/2},
\]
the map on vectorized variables is
\[
\operatorname{vec}(Y_1)=(I_r\otimes \bar H_1(X)^{1/2})\,\operatorname{vec}(X_1),\qquad
\operatorname{vec}(Y_2)=(\bar H_2(X)^{\top 1/2}\otimes I_{d-r})\,\operatorname{vec}(X_2),
\]
and the total Jacobian is 
\begin{align*}
  J(X)=  \begin{pmatrix}
I_r\otimes \bar H_1(X)^{1/2} & 0 \\
0 & \bar H_2(X)^{\top 1/2}\otimes I_{d-r}).
\end{pmatrix}
\end{align*}
The two block–diagonal Jacobian determinants are
\begin{align*}
\Bigl|\det\bigl(I_r\otimes \bar H_1(X)^{1/2}\bigr)\Bigr|
&= \bigl(\det \bar H_1(X)^{1/2}\bigr)^{\,r}
 = \det\!\bigl(\bar H_1(X)\bigr)^{\,r/2},\\
\Bigl|\det\bigl(\bar H_2(X)^{\top 1/2}\otimes I_{d-r}\bigr)\Bigr|
&= \bigl(\det \bar H_2(X)^{1/2}\bigr)^{\,d-r}
 = \det\!\bigl(\bar H_2(X)\bigr)^{\,(d-r)/2}.
\end{align*}
Multiplying the two factors, the total Jacobian of the linear change of variables is
\[
\;
\det(J(X))
= \det\!\bigl(\bar H_1(X)\bigr)^{\,r/2}\,
  \det\!\bigl(\bar H_2(X)\bigr)^{\,(d-r)/2}.
\;
\]
(We used $\det(B^\top)=\det(B)$ and that $\bar H_1(X),\bar H_2(X)\succ0$, so determinants are positive.) By the change of variable formula in integration (see, e.g., \cite{wiki:change-of-variables}), we have
  \begin{align*}
  &\operatorname{Vol}\!\big(B_{\op, \bar H_1(X), \bar H_2(X)}((X,X),\varepsilon)\big)
  \\\;=\;
  &\operatorname{Vol}\!\big(B_{\op, I_{d-1}, I_r}((X,X),\varepsilon)\big)\,(\det (J(X)))^{-1}
  \\\;=\;
  &\operatorname{Vol}\!\big(B_{\op, I_{d-1}, I_r}((X,X),\varepsilon)\big)\,
  \prod_{k=1}^{d-r}\lambda_k(\bar H_1(X))^{-r/2}\prod_{k=1}^{r}\lambda_k(\bar H_2(X))^{-(d-r)/2},
\end{align*}which implies 
\begin{align}\label{eq: result ratio}
    \frac{\textup{Vol}(B_{\op, I_{d-r}, I_{r}}((0,0),4))}{\textup{Vol}(B_{\op, \bar H_1(X), \bar H_2(X)}((X,X),\varepsilon))}= \prod_{k=1}^{d-r}\lambda_k(\bar H_1(X))^{r/2}\prod_{k=1}^{r}\lambda_k(\bar H_2(X))^{(d-r)/2}\frac{\textup{Vol}(B_{\op, I_{d-r}, I_{r}}((0,0),4))}{\textup{Vol}(B_{\op, I_{d-r}, I_r}((X,X),\varepsilon))}.
\end{align}

\paragraph{Part 4: Proving the Final Bound.} 
 For all $X\in\cX$ and $\varepsilon\leq 1$, we have that $B_{\op, I_{d-r}, I_r}((X,X),\varepsilon)\subseteq B_{\op, I_{d-r}, I_{r}}((0,0),4)$ and thus by \eqref{eq: volume inequality} and \eqref{eq: covering normed space}, we have
\begin{align}\label{eq: ratio bound}
    \frac{\textup{Vol}(B_{\op, I_{d-r}, I_{r}}((0,0),4))}{\textup{Vol}(B_{\op, I_{d-r}, I_r}((X,X),\varepsilon))}\leq {\left(\frac{12}{\varepsilon}\right)}^{2(d-r)r}.
\end{align}
Combining the above inequality \eqref{eq: ratio bound} with \eqref{eq: result lift} and \eqref{eq: result ratio}, we have
\begin{align}\label{eq: result sum log}
   &\log \rN\!\bigl(B_{\op, I_{d-r}, I_{r}}((0,0),4),\,\lVert\cdot\rVert_{\op, \bar H_1(X), \bar H_2(X)},\,\varepsilon/2\bigr)\nonumber\\\leq &2(d-r)r \log \frac{72}{\varepsilon}+ \frac{r}{2}\sum_{k=1}^{d-r}\log \lambda_k(\bar H_1(X))+\frac{d-r}{2}\sum_{k=1}^r \log\lambda_k(\bar H_2(X))\nonumber\\
   = &\frac{r}{2}\sum_{k=1}^{d-r}\log \frac{72^2\lambda_k(\bar H_1(X))}{\varepsilon^2}+\frac{d-r}{2}\sum_{k=1}^r \log\frac{72^2\lambda_k(\bar H_2(X))}{\varepsilon^2}.
\end{align}
Combing the above inequality \eqref{eq: result sum log} with  \eqref{eq: covering number 3}, \eqref{eq: covering number 4} and \eqref{eq: first equality lift}, we have that for all $X\in\cX$,
\begin{align}\label{eq: original ratio sum log bound}
    \log\frac{\textup{Vol}(\bar\cX)}{\textup{Vol}(B_{\varrho_1+\varrho_2}(X,\varepsilon)}\leq \frac{r}{2}\sum_{k=1}^{d-r}\log \frac{72^2\lambda_k(\bar H_1(X))}{\varepsilon^2}+\frac{d-r}{2}\sum_{k=1}^r \log\frac{72^2\lambda_k(\bar H_2(X))}{\varepsilon^2}.
\end{align}
Finally, combine the above inequality \eqref{eq: original ratio sum log bound} with \eqref{eq: covering number 1} and \eqref{eq: covering number 2}, we prove that for $\mu=\textup{Unif}(\Gr(d,r))$, we have that for all $\cV\in\Gr(d,r)$ and all $\varepsilon>0$,
\begin{align}\label{eq: second final bound}
  \log\frac{1}{\mu(B_{\varrho_{\proj,\Sigma}}(\cV,\varepsilon))}\leq &C_1 r(d-r)+\frac{r}{2}\sum_{k=1}^{d-r}\log \frac{72^2\lambda_k(\bar H_1(X))}{\varepsilon^2}+\frac{d-r}{2}\sum_{k=1}^r \log\frac{72^2\lambda_k(\bar H_2(X))}{\varepsilon^2}\nonumber\\
  = &\frac{r}{2}\sum_{k=1}^{d-r}\log \frac{C\lambda_k(\bar H_1(X))}{\varepsilon^2}+\frac{d-r}{2}\sum_{k=1}^r \log\frac{C\lambda_k(\bar H_2(X))}{\varepsilon^2},
\end{align}
where $C>0$ is an absolute constant.

We end the proof by applying Lemma \ref{lemma spectral-compression} and Lemma \ref{lemma eigenvalue bound}: since 
\begin{align*}
    \lambda_k(H_1(X))&\leq \lambda_k(\cP_{{\cV(X)}_\perp}^\top\Sigma \cP_{{\cV(X)}_\perp})\leq \lambda_k, \quad k=1,\cdots, d-r;\\
    \lambda_k(H_2(X))&\leq \lambda_k(\cP_{\cV(X)}^\top\Sigma \cP_{\cV(X)})\leq \lambda_k, \quad k=1, \cdots, r,
\end{align*}
we have 
\begin{align*}
    \lambda_k(\bar H_1(X))&\leq \max\{\lambda_k,\varepsilon^2\}, \quad k=1,\cdots, d-r;\\
    \lambda_k(\bar H_2(X))&\leq \max\{\lambda_k, \varepsilon^2\}, \quad k=1, \cdots, r.
\end{align*}
Substituting this bound to \eqref{eq: second final bound}, we prove that for $\mu=\textup{Unif}(\Gr(d,r))$, we have that for all $\cV\in\Gr(d,r)$ and all $\varepsilon>0$,
\begin{align*}
  \log\frac{1}{\mu(B_{\varrho_{\proj,\Sigma}}(\cV,\varepsilon))}\leq 
  \frac{r}{2}\sum_{k=1}^{d-r}\log \frac{C\max\left\{\lambda_k,\varepsilon^2\right\}}{\varepsilon^2}+\frac{d-r}{2}\sum_{k=1}^r \log\frac{C\max\left\{\lambda_k, \varepsilon^2\right\}}{\varepsilon^2},
\end{align*}
where $C>0$ is an absolute constant.

\hfill$\square$

\section{Proofs for Generalization Bounds and Comparison (Section \ref{sec generalization DNN})}\label{appendix generalization}

\subsection{Proof of Theorem~\ref{thm dnn}}
Let $\pi$ be the data-independent hierarchical prior of Theorem~\ref{thm matrix}. For fixed $S,W$ and scale $u$, define
\[
G_{\mathrm{NP}}^{S,\mathbb P}(W,u)
=\operatorname{blockdiag}\!\left(\ldots,
L\bar M_{l\to L}^2(W,u)\Gamma_{l-1}^{S,\mathbb P}(W)\otimes I_{d_l},\ldots\right).
\]
The non-perturbative expansion on the same bounded parameter charts applies in the mixed $L_2(\mathbb P_S+\mathbb P)$ norm. Its quadratic terms satisfy
\[
\mathbb P_S\|\Delta_lf_{l-1}(W,X)\|_2^2+
\mathbb P\|\Delta_lf_{l-1}(W,X)\|_2^2
=\frac1n\operatorname{Tr}\!\left(\Delta_l\Gamma_{l-1}^{S,\mathbb P}(W)\Delta_l^\top\right).
\]
The local chart and metric-domination argument used in Theorem~\ref{thm matrix}, followed by the $\beta$-Lipschitz loss comparison, therefore gives
\[
\log\frac1{\pi(B_{\varrho_{(S,\mathbb P),\ell}}(f(W,\cdot),\eta))}
\le d_{\mathrm R}^{S,\mathbb P}(W,c\eta/\beta).
\]
Here $\varrho_{(S,\mathbb P),\ell}$ is the mixed empirical-population loss metric in Theorem~\ref{thm generic chaining}. The geometric calculation applies to every finite PSD feature matrix; it does not require a finite-column realization. In particular, it applies directly to $\Gamma^{S,\mathbb P}$. Equivalently, with ghost-independent common transfer bounds, the quadratic tensor is the conditional mean of its ghost counterpart. This observation concerns the quadratic form only: no expectation is passed through an effective rank or a truncated spectral sum.

Substitution into Theorem~\ref{thm generic chaining}, followed by $\eta=\beta\varepsilon$ and the same absolute-constant endpoint rescaling as below, proves the integral bound. Expanding each block by the definition of $d_{\mathrm{eff}}$ gives \eqref{eq: R D DNN}. This proves Theorem~\ref{thm dnn}. \hfill$\square$

\subsection{Proof of Theorem~\ref{thm empirical dnn feature iso}}
Work on the event in Definition~\ref{def feature isomorphism}. Put $K_j=n\Sigma_j^{\mathbb P}(W)$, $P=P_j^S(W,\varepsilon)$, $Q=I-P$, and $\Delta_l=U_l-W_l$. Positivity of $K_j$ and the two assumed subspace inequalities give
\[
\operatorname{Tr}(\Delta_lK_j\Delta_l^\top)
\le2\kappa\operatorname{Tr}(\Delta_lP\Gamma_j^S(W)P\Delta_l^\top)
+2b_{\mathrm{sub}}\vartheta_j(W,\varepsilon)\|\Delta_l\|_{\tF}^2.
\]
Thus the finite-scale domination argument proved in detail in Appendix~\ref{subsec empirical ghost ellipsoid} applies with $\Gamma_j^{S'}$ replaced by $K_j$ and the empirical-ghost loss metric replaced by $\varrho_{(S,\mathbb P),\ell}$. On the same bounded local support, $\sum_l\|\Delta_l\|_{\tF}^2\le CR_W^2$, while
\[
L\bar M_{l\to L}^2(W,\varepsilon)\vartheta_{l-1}(W,\varepsilon)
=\frac{n\varepsilon^2}{2R_W^2}.
\]
The inactive contribution is therefore at most $Cn\varepsilon^2$. Choosing $c_{\mathrm{sub}}$ sufficiently small gives the same finite-scale ball inclusion and hence
\[
\log\frac1{\pi(B_{\varrho_{(S,\mathbb P),\ell}}(f(W,\cdot),\eta))}
\le d_{\mathrm R}^{S}(W,c_{\mathrm{sub}}\eta/\beta).
\]
Apply Theorem~\ref{thm generic chaining} and use a union bound with the population-subspace event. This proves \eqref{eq empirical feature iso bound main}. No ghost exceptional event is used in this argument. \hfill$\square$

The population condition and the conditional ghost condition below are alternative sufficient conditions. With positive ghost failure probability, averaging the latter need not imply the former, because the feature covariance on the exceptional event need not be controlled. The conditional ghost statement retains its original $\zeta$ loss term.

\subsection{Ghost formulation and proof of the DNN bound}\label{subsec dnn ghost}
For the proofs and refinements below, retain the original conditional ghost formulation. Write $F_l^T(W)=F_l(W,X_T)$ and $\Gamma_l^T(W)=F_l^T(W)F_l^T(W)^\top$ for $T\in\{S,S'\}$, and
\[
\Gamma_l^{S,S'}(W)=F_l^S(W)F_l^S(W)^\top+F_l^{S'}(W)F_l^{S'}(W)^\top.
\]
Use common outer transfer bounds valid for the observed and ghost samples and their concatenation. They may be chosen to be the uniform bounds over inputs used in the main theorem. The following result needs no population feature moment assumption.

\begin{lemma}[Generalization Bound for DNN; mixed empirical-ghost form]\label{lemma dnn ghost}
Let the loss $\ell(f(W,x),y)$ be bounded in $[0,1]$ and $\beta$-Lipschitz with respect to $f(W,x)$.  For every $\delta\in(0,1)$, with probability at least $1-\delta$ over the observed sample $S$, uniformly over all $W\in B_{\tF}(R)$,
\begin{align*}
&(\bP-\bPn)\ell(f(W,x),y)\nonumber\\
&\le C_1\left(\beta\,\mathbb E_{S'}\left[\inf_{0\le\alpha\le 1/\beta}\left\{\alpha+\frac{1}{\sqrt n}\int_\alpha^{1/\beta}\sqrt{d_{\textup{R}}^{S,S'}(W,c\varepsilon)}\,d\varepsilon\right\}\Bigm|S\right]
+\sqrt{\frac{\log\frac{2\log(2n)}{\delta}}{n}}\right),
\end{align*}
where $c,C_1>0$ are absolute constants and
\begin{align}\label{eq: R D DNN ghost}
d_{\textup{R}}^{S,S'}(W,\varepsilon)=\frac{1}{2}\sum_{l=1}^L \bigg(&(d_l+d_{l-1}) \sum_{k=1}^{r^{S,S'}_{\eff}[W, l]}\underbrace{\log\frac{8C_2^2\lambda_k(\Gamma_{l-1}^{S,S'}(W))}{n\varepsilon^2}}_{\textup{spectrum of inner layers }1{:}l\!-\!1}\nonumber\\
&+(d_l+d_{l-1}) r^{S,S'}_{\eff}[W, l]\cdot\underbrace{\log\Big(\bar{M}^2_{l\rightarrow L}(W,\varepsilon)  L\max\{||W||_{\tF}^2,R^2/4^n\}\Big)}_{\textup{spectrum of outer layers }l+1:L}+2\log(d_{l-1}n)\bigg),
\end{align}
 with
\[
r^{S,S'}_{\eff}[W,l]
:=r_{\eff}\!\left(L\bar{M}^2_{l\rightarrow L}(W,\varepsilon)\Gamma_{l-1}^{S,S'}(W), C_2\max\{||W||_{\tF}, R/2^n\}, \varepsilon\right),
\]
and $C_2>0$ is an absolute constant.
\end{lemma}
\paragraph{Proof of Lemma~\ref{lemma dnn ghost}.}
The proof consists of two steps. First, we combine Lemma~\ref{lemma conditional ghost} with the Riemannian Dimension obtained from Theorem~\ref{thm matrix}, applied to the mixed sample; this yields an integral bound in terms of the mixed empirical-ghost Riemannian Dimension. Second, we expand this mixed Riemannian Dimension into the layerwise expression displayed in the theorem.

\paragraph{Step 1: Obtaining the Integral Bound on Generalization Gap.}
Let $\pi$ be the data-independent hierarchical prior from Theorem~\ref{thm matrix}.  Lemma~\ref{lemma conditional ghost} gives, with probability at least $1-\delta$ over the observed sample $S$, uniformly over all $W\in B_{\tF}(R)$,
\begin{align}\label{eq mixed theorem one in dnn proof}\nonumber
&(\bP-\bPn)\ell(f(W,x),y)
\\\le &C_0\left(\mathbb E_{S'}\left[\inf_{0\le\alpha\le \sqrt2}\left\{\alpha+\frac1{\sqrt n}\int_\alpha^{\sqrt2}
\sqrt{\log\frac1{\pi(B_{\varrho_{(S,S'),\ell}}(f(W,\cdot),\eta))}}\,d\eta\right\}\Bigm|S\right]
+\sqrt{\frac{\log(2\log(2n)/\delta)}{n}}\right),
\end{align}
where \(C_0>0\) is the absolute constant in Lemma~\ref{lemma conditional ghost}.
We now compare the mixed empirical-ghost loss metric with the mixed
non-perturbative NN-surrogate metric.   As presented in \eqref{eq: metric tensor DNN}, we construct the mixed metric tensor for the mixed sample $(S,S')$, 
\begin{align*}
G_{\textup{NP}}^{S,S'}(W,u)
      :=\textup{blockdiag}\left(\cdots, L \bar{M}^2_{l\rightarrow L}(W,u)\cdot \Gamma_{l-1}^{S,S'}(W)\otimes  I_{d_{l}},\cdots\right),\\
       \varrho_{G_{\mathrm{NP}}^{S,S'}(W,u)}(W,W')^2=\textup{vec}(W'-W)^\top G_{\textup{NP}}^{S,S'}(W,u)\textup{vec}(W'-W),
\end{align*}
where $G_{\textup{NP}}^{S,S'}$ is obtained by replacing each empirical feature Gram matrix by the mixed feature Gram matrix $\Gamma_{l-1}^{S,S'}(W)$, and \(\varrho_{G_{\mathrm{NP}}^{S,S'}(W,u)}\) is the corresponding
NN-surrogate metric. By Lipschitz property of the loss function we have
\[
\varrho_{(S,S'),\ell}(f(U,\cdot),f(W,\cdot))
\le
\beta\,\left(
(\bP_S+\bP_{S'})
\|f(U,x)-f(W,x)\|_2^2
\right)^{1/2}
=
\frac\beta{\sqrt n}
\|F_L^{S,S'}(U)-F_L^{S,S'}(W)\|_{\tF},
\]
where $F_l^{S,S'}(W)$ is defined as
\[
F_l^{S,S'}(W):=
\bigl[F_l^S(W),F_l^{S'}(W)\bigr].\]
The non-perturbative expansion used in Lemma~\ref{lemma non-perturbative} gives
\[
\|F_L^{S,S'}(U)-F_L^{S,S'}(W)\|_{\tF}^2
\le  \varrho_{G_{\textup{NP}}^{S,S'}(W,u)}(U,W)^2.
\]
Combining the above two inequalities and taking $u= \eta/\beta$ we have \begin{align*}
  \varrho_{(S,S'),\ell}(f(U,\cdot),f(W,\cdot))
\le
\frac{\beta}{\sqrt n}
\varrho_{G_{\mathrm{NP}}^{S,S'}(W,\eta/\beta)}(U,W).  
\end{align*}
In particular, we have the ball inclusion
\[
B_{\varrho_{G_{\mathrm{NP}}^{S,S'}(W,\eta/\beta)}}
\left(W,\frac{\sqrt n\,\eta}{\beta}\right)
\subseteq
B_{\varrho_{(S,S'),\ell}}
(f(W,\cdot),\eta).
\]
By the metric domination lemma (Lemma \ref{lemma simple metric domination}), we have the pointwise dimension bound: for every $W\in{B_{\tF}(R)}$,
\begin{align*}
  \log\frac1{\pi(B_{\varrho_{(S,S'),\ell}}(f(W,\cdot),\eta))}\leq \log\frac{1}{\pi(B_{\varrho_{G_{\textup{NP}}^{S,S'}(W,\eta/\beta)}}(W,\sqrt{n}\eta/\beta))}.
\end{align*}
Applying Theorem~\ref{thm matrix} (Riemannian Dimension Bound for DNN)  to this mixed metric, we have that  there exists a prior $\pi$ such that uniformly over every $W\in B_{\tF}(R)$,
  \[
\log\frac1{\pi(B_{\varrho_{(S,S'),\ell}}(f(W,\cdot),\eta))}
\le d_{\textup{R}}^{S,S'}(W,c\eta/\beta)
\]
with absolute-constant changes, where \(c>0\) is an absolute constant. 
Substituting this into \eqref{eq mixed theorem one in dnn proof} and changing variables $\eta=\beta\varepsilon$ gives
\begin{align*}
&(\bP-\bPn)\ell(f(W,x),y)\nonumber\\
&\le C_0\left(\beta\,\mathbb E_{S'}\left[\inf_{0\le\alpha\le \sqrt{2}/\beta}\left\{\alpha+\frac{1}{\sqrt n}\int_\alpha^{\sqrt{2}/\beta}\sqrt{d_{\textup{R}}^{S,S'}(W,c\varepsilon)}\,d\varepsilon\right\}\Bigm|S\right]
+\sqrt{\frac{\log(2\log(2n)/\delta)}{n}}\right).
\end{align*}
The substitution $\varepsilon=\sqrt2 u$, $\alpha=\sqrt2 a$ changes the upper endpoint to $1/\beta$; absorb the resulting factors of $\sqrt2$ into the absolute constants $C_1$ and $c$. Thus,
\begin{align}\label{eq: integral bound dnn}
&(\bP-\bPn)\ell(f(W,x),y)\nonumber\\
&\le C_1\left(\beta\,\mathbb E_{S'}\left[\inf_{0\le\alpha\le 1/\beta}\left\{\alpha+\frac{1}{\sqrt n}\int_\alpha^{1/\beta}\sqrt{d_{\textup{R}}^{S,S'}(W,c\varepsilon)}\,d\varepsilon\right\}\Bigm|S\right]
+\sqrt{\frac{\log(2\log(2n)/\delta)}{n}}\right).
\end{align}
This proves the integral part of Lemma~\ref{lemma dnn ghost}.

\paragraph{Step 2: Obtaining the Expression of Riemannian Dimension.}
It remains to expand the mixed Riemannian Dimension $d_{\textup{R}}^{S,S'}$ by Theorem~\ref{thm matrix}.  By definition of the mixed feature Gram matrix,
\begin{align}
 d_{\textup{R}}^{S,S'}(W,\varepsilon)=&\sum_{l=1}^L \Big( (d_l+d_{l-1})\cdot d_\eff(L\bar{M}_{l\rightarrow L}^2(W,\varepsilon)\cdot \Gamma_{l-1}^{S,S'}(W), C_2\max\{\|W\|_{\tF}, R/2^n\},\varepsilon)\nonumber\\
&+\log(d_{l-1}n)\Big), \label{eq: d R Thm DNN mixed}
\end{align}
where $R=\sup_{\cW}\|W\|_{\tF}$ and $C_2$ is an absolute constant.  The effective dimension is
\begin{align}
&d_\eff(L\bar{M}_{l\rightarrow L}^2(W,\varepsilon)\cdot \Gamma_{l-1}^{S,S'}(W),C_2 \max\{\|W\|_{\tF},R/2^n\}, \varepsilon)\nonumber\\
&\qquad=\frac{1}{2}\sum_{k=1}^{r^{S,S'}_\eff[W,l]}\log\frac{8C_2^2\max\{\|W\|_{\tF}^2,R^2/4^n\}L\bar{M}_{l\rightarrow L}^2(W,\varepsilon)\lambda_k(\Gamma_{l-1}^{S,S'}(W))}{n\varepsilon^2},\label{eq: d eff step mixed}
\end{align}
where $r^{S,S'}_\eff[W,l]$ abbreviates $r_\eff(L\bar{M}_{l\rightarrow L}^2(W,\varepsilon)\Gamma_{l-1}^{S,S'}(W), C_2\max\{\|W\|_{\tF},R/2^n\}, \varepsilon)$.  Combining \eqref{eq: d R Thm DNN mixed} and \eqref{eq: d eff step mixed} gives the displayed expression \eqref{eq: R D DNN ghost}.  Combining the integral upper bound \eqref{eq: integral bound dnn} with this expression concludes the proof of Lemma~\ref{lemma dnn ghost}. 
\hfill$\square$

\subsection{Empirical-Ghost Subspace Isomorphism for the Pointwise Ellipsoidal Metric}\label{subsec empirical ghost ellipsoid}
\begin{definition}[Conditional Ghost Subspace Isomorphism]\label{def ghost feature isomorphism}
Fix constants $\kappa\ge1$, $b_{\mathrm{sub}}\ge1$, and a failure level $\zeta\in[0,1]$.  We say that the observed sample $S$ satisfies the conditional layer-wise finite-resolution subspace isomorphism if, conditionally on $S$, with probability at least $1-\zeta$ over an independent ghost sample $S'$, the following two estimates hold simultaneously for every feature layer $j=0,\ldots,L-1$, every center $W\in B_{\tF}(R)$, and every scale $\varepsilon\in(0,1/\beta]$:
\begin{align}
P_j^S(W,\varepsilon)\Gamma_j^{S'}(W)P_j^S(W,\varepsilon)
&\preceq
\kappa\,P_j^S(W,\varepsilon)\Gamma_j^S(W)P_j^S(W,\varepsilon),
\label{eq ghost feature isomorphism active}\\
\big\|Q_j^S(W,\varepsilon)\Gamma_j^{S'}(W)Q_j^S(W,\varepsilon)\big\|_{\op}
&\le b_{\mathrm{sub}}\,\vartheta_j(W,\varepsilon).
\label{eq ghost feature isomorphism inactive}
\end{align}
\end{definition}

\begin{lemma}[Fully Empirical DNN Bound under Subspace Isomorphism]\label{lemma empirical dnn ghost feature iso}
Assume the setting of Lemma~\ref{lemma dnn ghost}.  Suppose that, with probability at least $1-\delta_{\mathrm{iso}}$ over $S$, the conditional layer-wise finite-resolution subspace isomorphism in Definition~\ref{def ghost feature isomorphism} holds.  Then there is a constant $c_{\mathrm{sub}}>0$, depending only on $(\kappa,b_{\mathrm{sub}})$, such that for every $\delta\in(0,1)$, with probability at least $1-\delta-\delta_{\mathrm{iso}}$ over $S$, uniformly over all $W\in B_{\tF}(R)$,
\begin{align}\label{eq empirical ghost feature iso bound}
&(\bP-\bPn)\ell(f(W,x),y)\nonumber\\
&\le C\Bigg(\beta\inf_{0\le\alpha\le 1/\beta}\left\{\alpha+\frac1{\sqrt n}\int_\alpha^{1/\beta}\sqrt{d_{\textup{R}}^S(W,c_{\mathrm{sub}}\varepsilon)}\,d\varepsilon\right\}
+\sqrt{\frac{\log(2\log(2n)/\delta)}{n}}+\zeta\Bigg),
\end{align}
where $C>0$ is an absolute constant.  In particular, if \(\kappa,b_{\mathrm{sub}}=O(1)\), then 
\(c_{\mathrm{sub}}\) is an absolute constant. If moreover
\(\zeta\le \sqrt{\log(2\log(2n)/\delta)}/n\), the last term is absorbed into the displayed concentration term and the leading complexity is the clean observed-sample Riemannian Dimension $d_{\textup{R}}^S$.
\end{lemma}

This subsection proves Lemma~\ref{lemma empirical dnn ghost feature iso}.  The proof uses only the finite-resolution subspace isomorphism in Definition~\ref{def ghost feature isomorphism}.  Its role is to show that the mixed empirical-ghost metric in Lemma~\ref{lemma conditional ghost} is dominated, at the finite scale used in the pointwise dimension, by the clean observed-sample ellipsoidal metric.  Unlike a full Loewner isomorphism, the argument does not insert a fixed isotropic ridge into the observed spectrum.

For a sample $T$ of size $n$, write $X_T$ for its input matrix and
\[
F_{l-1}^T(W):=F_{l-1}(W,X_T),\qquad
A_{l,T}(W,u):=L\bar{M}_{l\to L}^2(W,u)F_{l-1}^T(W)F_{l-1}^T(W)^\top .
\]
The empirical non-perturbative metric tensor is
\[
G_{\textup{NP}}^T(W,u):=\operatorname{blockdiag}\bigl(\ldots,A_{l,T}(W,u)\otimes I_{d_l},\ldots\bigr).
\]
The block-diagonal form is essential: the local-chart contribution is $d_l$ copies of the ellipsoid generated by $A_{l,T}(W,u)$, while the Grassmannian atlas cost is governed by the same spectrum and contributes the $d_{l-1}$ factor.

\paragraph{Subspace isomorphism converts the mixed metric to the empirical ellipsoid.}
We now prove the fully empirical statement under Definition~\ref{def ghost feature isomorphism}.  For $j=0,\ldots,L-1$, let $P_j^S(W,\varepsilon)$ and $Q_j^S(W,\varepsilon)$ be the active and inactive projectors below \eqref{eq subspace iso threshold main}.  The next lemma is the precise point at which the subspace isomorphism is used.

\begin{lemma}[Subspace Isomorphism Implies Finite-Scale Metric Domination]\label{lemma feature iso metric domination}
Assume that the event in Definition~\ref{def ghost feature isomorphism} holds for a pair $(S,S')$.  Then there is a constant $c_{\mathrm{sub}}>0$, depending only on $(\kappa,b_{\mathrm{sub}})$, such that the following holds.  For every center $W\in B_{\tF}(R)$ and every scale $\eta\in(0,1]$, the empirical ellipsoidal ball at radius $c_{\mathrm{sub}}\sqrt n\eta/\beta$, restricted to the same Euclidean shell used in the proof of Theorem~\ref{thm matrix}, is contained in the mixed loss ball:
\begin{align}\label{eq feature iso metric domination}
B_{\varrho_{G_{\textup{NP}}^S(W,c_{\mathrm{sub}}\eta/\beta)}}\!\left(W,c_{\mathrm{sub}}\sqrt n\eta/\beta\right)
\subseteq
B_{\varrho_{(S,S'),\ell}}\bigl(f(W,\cdot),\eta\bigr).
\end{align}
\end{lemma}

\paragraph{Proof of Lemma~\ref{lemma feature iso metric domination}.}
Recall that the proof of
Theorem~\ref{thm matrix} uses a dyadic peeling over the Frobenius norm scale. For the
dyadic component containing the center \(W\), the corresponding local support
has Euclidean radius comparable to
\[
R_W:=C_2\max\{\|W\|_{\tF},R/2^n\}.
\]
Throughout this lemma we work on this same local support. Put $\varepsilon:=c_{\mathrm{sub}}\eta/\beta$ and write $\Delta_l:=U_l-W_l$.
It suffices to show that any
\[
U\in 
B_{\varrho_{G_{\textup{NP}}^S(W,\varepsilon)}}(W,\sqrt n\varepsilon)
\]
inside the same Euclidean shell used in the proof of Theorem~\ref{thm matrix} satisfies
\[
\varrho_{(S,S'),\ell}(f(U,\cdot),f(W,\cdot))\le \eta .
\] 
Hence, for every
\(U\) in the relevant local support,
\begin{align}\label{eq local shell norm proof}
\sum_{l=1}^L\|\Delta_l\|_{\tF}^2
=
\|U-W\|_{\tF}^2
\le C R_W^2 .
\end{align}

By the non-perturbative expansion in Lemma~\ref{lemma non-perturbative}, applied to $S$ and to $S'$ on the same local-Lipschitz event,
\begin{align*}
\|F_L(U,X_S)-F_L(W,X_S)\|_{\tF}^2
&\le \sum_{l=1}^L L\bar{M}_{l\to L}^2(W,\varepsilon)\|\Delta_lF_{l-1}^{S}(W)\|_{\tF}^2,\\
\|F_L(U,X_{S'})-F_L(W,X_{S'})\|_{\tF}^2
&\le \sum_{l=1}^L L\bar{M}_{l\to L}^2(W,\varepsilon)\|\Delta_lF_{l-1}^{S'}(W)\|_{\tF}^2.
\end{align*}
Fix the summand \(l\), set \(j=l-1\), and write
\[
P=P_j^S(W,\varepsilon),\qquad Q=Q_j^S(W,\varepsilon)=I-P.
\]
Since \(P,Q\) are spectral projectors of
\[
\Gamma_j^S(W)=F_j^S(W)F_j^S(W)^\top,
\]
the observed covariance has no \(P\)--\(Q\) cross term:
\[
P\Gamma_j^S(W)Q=0,\qquad Q\Gamma_j^S(W)P=0.
\]
Therefore,
\[
\Gamma_j^S(W)=P\Gamma_j^S(W)P+Q\Gamma_j^S(W)Q,
\]
and hence
\[
\begin{aligned}
\|\Delta_lF_j^S(W)\|_{\tF}^2
 =
\operatorname{Tr}\bigl(\Delta_l \Gamma_j^S(W) \Delta_l^\top\bigr) 
 =
\operatorname{Tr}\bigl(\Delta_lP\Gamma_j^S(W)P\Delta_l^\top\bigr)
+
\operatorname{Tr}\bigl(\Delta_lQ\Gamma_j^S(W)Q\Delta_l^\top\bigr).
\end{aligned}
\]
For the ghost term, \(P,Q\) are not spectral projectors of \(\Gamma_j^{S'}(W)\), so cross terms may appear. Since \(\Gamma_j^{S'}(W)\succeq0\), for any row vector \(a\),
\[
a\Gamma_j^{S'}(W)a^\top
\le
2(aP)\Gamma_j^{S'}(W)(aP)^\top
+
2(aQ)\Gamma_j^{S'}(W)(aQ)^\top .
\]
Applying this row-wise to the rows of \(\Delta_l\) and summing gives
\[
\operatorname{Tr}\bigl(\Delta_l\Gamma_j^{S'}(W)\Delta_l^\top\bigr)
\le
2\operatorname{Tr}\bigl(\Delta_lP\Gamma_j^{S'}(W)P\Delta_l^\top\bigr)
+2\operatorname{Tr}\bigl(\Delta_lQ\Gamma_j^{S'}(W)Q\Delta_l^\top\bigr).
\]
By Definition~\ref{def ghost feature isomorphism}, on the active subspace,
\[
P\Gamma_j^{S'}(W)P
\preceq
\kappa P\Gamma_j^S(W)P,
\]
while on the inactive subspace,
\[
Q\Gamma_j^{S'}(W)Q
\preceq
b_{\mathrm{sub}}\vartheta_j(W,\varepsilon)Q.
\]
Consequently,
\begin{align*}
\|\Delta_lF_j^{S'}(W)\|_{\tF}^2
=
\operatorname{Tr}\bigl(\Delta_l\Gamma_j^{S'}(W)\Delta_l^\top\bigr)
&\le
2\kappa\operatorname{Tr}\bigl(\Delta_lP\Gamma_j^S(W)P\Delta_l^\top\bigr)
+2b_{\mathrm{sub}}\vartheta_j(W,\varepsilon)\|\Delta_lQ\|_{\tF}^2\\
&\le
2\kappa\operatorname{Tr}\bigl(\Delta_lP\Gamma_j^S(W)P\Delta_l^\top\bigr)
+2b_{\mathrm{sub}}\vartheta_j(W,\varepsilon)\|\Delta_l\|_{\tF}^2.
\end{align*}
Moreover, by the definition of \(Q\) as the inactive spectral projector of \(\Gamma_j^S(W)\),
\[
Q\Gamma_j^S(W)Q\preceq\vartheta_j(W,\varepsilon)Q.
\]
Combining the observed decomposition with the preceding ghost estimate gives
\[
\|\Delta_lF_j^S(W)\|_{\tF}^2+\|\Delta_lF_j^{S'}(W)\|_{\tF}^2
\le
C_\kappa\|\Delta_lF_j^S(W)\|_{\tF}^2
+
C_b\vartheta_j(W,\varepsilon)\|\Delta_l\|_{\tF}^2.
\]
Substituting this estimate into the two non-perturbative expansions and summing over \(l\), we obtain
\begin{align*}
&\|F_L(U,X_S)-F_L(W,X_S)\|_{\tF}^2+
\|F_L(U,X_{S'})-F_L(W,X_{S'})\|_{\tF}^2\\
&\qquad\le
C_\kappa\sum_{l=1}^L
L\bar M_{l\to L}^2(W,\varepsilon)
\|\Delta_lF_{l-1}^S(W)\|_{\tF}^2 
+C_b\sum_{l=1}^L
L\bar M_{l\to L}^2(W,\varepsilon)
\vartheta_{l-1}(W,\varepsilon)
\|\Delta_l\|_{\tF}^2.
\end{align*}
The first sum is exactly the observed empirical ellipsoidal energy,
\[
\sum_{l=1}^L
L\bar M_{l\to L}^2(W,\varepsilon)
\|\Delta_lF_{l-1}^S(W)\|_{\tF}^2
=
\varrho_{G_{\textup{NP}}^S(W,\varepsilon)}(U,W)^2.
\]
For the second sum, using the definition of the finite resolution in \eqref{eq subspace iso threshold main},
\[
\vartheta_{l-1}(W,\varepsilon)
=
\frac{n\varepsilon^2}
{2L\bar M_{l\to L}^2(W,\varepsilon)R_W^2},
\]
we have
\[
L\bar M_{l\to L}^2(W,\varepsilon)\vartheta_{l-1}(W,\varepsilon)
=
\frac{n\varepsilon^2}{2R_W^2}.
\]
Therefore, by \eqref{eq local shell norm proof},
\[
\sum_{l=1}^L
L\bar M_{l\to L}^2(W,\varepsilon)
\vartheta_{l-1}(W,\varepsilon)\|\Delta_l\|_{\tF}^2
\le
C n\varepsilon^2.
\]
Thus
\begin{align*}
&\|F_L(U,X_S)-F_L(W,X_S)\|_{\tF}^2+
\|F_L(U,X_{S'})-F_L(W,X_{S'})\|_{\tF}^2\\
&\qquad\le
C_{\kappa,b}\,\varrho_{G_{\textup{NP}}^S(W,\varepsilon)}(U,W)^2
+C_{\kappa,b}n\varepsilon^2.
\end{align*}

If \(U\) belongs to the empirical ellipsoidal ball on the left-hand side of \eqref{eq feature iso metric domination}, then
\[
\varrho_{G_{\textup{NP}}^S(W,\varepsilon)}(U,W)
\le
\sqrt n\varepsilon.
\]
Hence the right-hand side above is bounded by \(C_{\kappa,b}n\varepsilon^2\). Since the loss is \(\beta\)-Lipschitz,
\begin{align*}
\varrho_{(S,S'),\ell}(f(U,\cdot),f(W,\cdot))^2
&\le
\frac{\beta^2}{n}
\Bigl(
\|F_L(U,X_S)-F_L(W,X_S)\|_{\tF}^2+
\|F_L(U,X_{S'})-F_L(W,X_{S'})\|_{\tF}^2
\Bigr)\\
&\le
C_{\kappa,b}\beta^2\varepsilon^2.
\end{align*}
Since \(\varepsilon=c_{\mathrm{sub}}\eta/\beta\), this becomes
\[
\varrho_{(S,S'),\ell}(f(U,\cdot),f(W,\cdot))^2
\le
C_{\kappa,b}c_{\mathrm{sub}}^2\eta^2.
\]
Choosing \(c_{\mathrm{sub}}>0\) sufficiently small, depending only on \((\kappa,b_{\mathrm{sub}})\), gives
\[
\varrho_{(S,S'),\ell}(f(U,\cdot),f(W,\cdot))\le \eta.
\]
This proves \eqref{eq feature iso metric domination}. \hfill$\square$

\begin{lemma}[Empirical Pointwise Dimension under Subspace Isomorphism]\label{lemma empirical pointwise dimension feature iso}
Assume that the event in Definition~\ref{def ghost feature isomorphism} holds for the pair $(S,S')$. Then the hierarchical prior $\pi$ from Theorem~\ref{thm matrix} satisfies, uniformly over $W\in B_{\tF}(R)$ and $\eta>0$,
\begin{align}\label{eq empirical pointwise dimension feature iso}
\log\frac{1}{\pi\bigl(B_{\varrho_{(S,S'),\ell}}(f(W,\cdot),\eta)\bigr)}
\le
d_{\textup{R}}^S\!\left(W,\frac{c_{\mathrm{sub}}\eta}{\beta}\right).
\end{align}
\end{lemma}

\paragraph{Proof of Lemma~\ref{lemma empirical pointwise dimension feature iso}.}
Lemma~\ref{lemma feature iso metric domination} gives the finite-scale ball inclusion required to compare pointwise dimensions.  By monotonicity of pointwise dimension under metric domination, Lemma~\ref{lemma simple metric domination},
\begin{align*}
\log\frac{1}{\pi\bigl(B_{\varrho_{(S,S'),\ell}}(f(W,\cdot),\eta)\bigr)}
\le
\log\frac{1}{\pi\bigl(B_{\varrho_{G_{\textup{NP}}^S(W,c_{\mathrm{sub}}\eta/\beta)}}(W,c_{\mathrm{sub}}\sqrt n\eta/\beta)\bigr)}.
\end{align*}
The tensor $G_{\textup{NP}}^S$ has exactly the same NN-surrogate block-diagonal form treated in Theorem~\ref{thm matrix}, with the feature Gram equal to the observed matrix $\Gamma_{l-1}^S(W)$.  Thus the hierarchical covering theorem applies directly to this observed empirical metric tensor. 
Applying Theorem~\ref{thm matrix} with 
\(\varepsilon=c_{\mathrm{sub}}\eta/\beta\) gives
\[
\log
\frac{1}{
\pi\!\left(
B_{\varrho_{G_{\mathrm{NP}}^S(W,c_{\mathrm{sub}}\eta/\beta)}}
\left(W,c_{\mathrm{sub}}\sqrt n\,\eta/\beta\right)
\right)}
\le
d_{\mathrm R}^S\!\left(W,\frac{c_{\mathrm{sub}}\eta}{\beta}\right).
\]
Combining this with the preceding display proves \eqref{eq empirical pointwise dimension feature iso}. \hfill$\square$

\paragraph{Proof of Lemma~\ref{lemma empirical dnn ghost feature iso}.}
Let $\Psi_{S,S'}(W)$ denote the generic-chaining functional inside Lemma~\ref{lemma conditional ghost},
\[
\Psi_{S,S'}(W):=
\inf_{0\le\alpha\le \sqrt2}\left\{\alpha+\frac1{\sqrt n}\int_\alpha^{\sqrt2}\sqrt{\log\frac1{\pi(B_{\varrho_{(S,S'),\ell}}(f(W,\cdot),\eta))}}\,d\eta\right\}.
\]
Since the pointwise dimension is nonincreasing in the radius \(\eta\), a change of variables \(\eta=\sqrt2 t\), followed by taking the infimum over \(\alpha\), allows us to replace the upper endpoint \(\sqrt2\) by \(1\) at the cost of an absolute constant:
\[
\Psi_{S,S'}(W)\le C
\inf_{0\le\alpha\le 1}\left\{\alpha+\frac1{\sqrt n}\int_\alpha^{1}\sqrt{\log\frac1{\pi(B_{\varrho_{(S,S'),\ell}}(f(W,\cdot),\eta))}}\,d\eta\right\},
\]
where the constant is enlarged harmlessly.
Conditionally on an observed sample \(S\) satisfying Definition~\ref{def ghost feature isomorphism}, let
\(\mathcal E_{\mathrm{iso}}(S,S')\) denote the event over the ghost sample \(S'\) on which the two
estimates in Definition~\ref{def ghost feature isomorphism} hold. By definition,
\[
\mathbb P_{S'}(\mathcal E_{\mathrm{iso}}(S,S')\mid S)\ge 1-\zeta .
\]
On \(\mathcal E_{\mathrm{iso}}(S,S')\), Lemma~\ref{lemma empirical pointwise dimension feature iso}
and the change of variables \(\eta=\beta\varepsilon\) imply
\[
\Psi_{S,S'}(W)
\le
C\beta
\inf_{0\le\alpha\le 1/\beta}
\left\{
\alpha+
\frac1{\sqrt n}
\int_\alpha^{1/\beta}
\sqrt{d_R^S(W,c_{\mathrm{sub}}\varepsilon)}\,d\varepsilon
\right\}.
\]
On the complement \(\mathcal E_{\mathrm{iso}}(S,S')^c\), the trivial choice
\(\alpha=\sqrt2\) in the original definition of \(\Psi_{S,S'}(W)\) gives \(\Psi_{S,S'}(W)\le\sqrt2\) for every \(W\). Hence,
conditionally on such an \(S\),
\begin{align}\label{eq conditional expectation feature iso proof}\nonumber
\mathbb E_{S'}[\Psi_{S,S'}(W)\mid S]&=
\mathbb E_{S'}[\Psi_{S,S'}(W)\mathbf 1_{\mathcal E_{\mathrm{iso}}}\mid S]
+
\mathbb E_{S'}[\Psi_{S,S'}(W)\mathbf 1_{\mathcal E_{\mathrm{iso}}^c}\mid S] \\\nonumber&\le
\mathbb E_{S'}[\Psi_{S,S'}(W)\mathbf 1_{\mathcal E_{\mathrm{iso}}}\mid S]
+
\sqrt2\,
\mathbb P_{S'}(\mathcal E_{\mathrm{iso}}^c\mid S) \\&\le C\beta\inf_{0\le\alpha\le 1/\beta}\left\{\alpha+\frac1{\sqrt n}\int_\alpha^{1/\beta}\sqrt{d_{\textup{R}}^S(W,c_{\mathrm{sub}}\varepsilon)}\,d\varepsilon\right\}+C\zeta,
\end{align}
where the last inequality absorbs \(\sqrt2\) into the absolute constant $C$.
Substituting \eqref{eq conditional expectation feature iso proof} into Lemma~\ref{lemma conditional ghost} proves \eqref{eq empirical ghost feature iso bound}.  The probability is at least $1-\delta-\delta_{\mathrm{iso}}$ by a union bound between the generic-chaining event and the event that Definition~\ref{def ghost feature isomorphism} holds for $S$. \hfill$\square$

\subsection{How Feature Regularity Can Imply the Subspace Isomorphism}\label{subsec feature regularity}
The two-sample verification below is retained for the conditional ghost result. The same relative covariance certificate restricted to $T=S$ also implies the population condition in Definition~\ref{def feature isomorphism}: rearrange it as $(1-\gamma)\Sigma_j\preceq\widehat\Sigma_{j,S}+\omega_jI$, then restrict to the observed active and inactive subspaces. This gives constants $\kappa,b_{\mathrm{sub}}\le(1+c_0)/(1-\gamma)$ at the stated resolution.

Definition~\ref{def ghost feature isomorphism} is a finite-resolution, center-uniform condition on the learned feature class.  It is weaker than full Loewner domination because it only asks for multiplicative transfer on the empirical active subspace and below-resolution leakage on its orthogonal complement.  Let
\[
\widehat\Sigma_{j,T}(W):=\frac1nF_j^T(W)F_j^T(W)^\top,
\qquad
\Sigma_j(W):=\mathbb{P}\bigl[f_j(W,x)f_j(W,x)^\top\bigr],
\qquad
\bar\vartheta_j(W,\varepsilon):=\frac{\vartheta_j(W,\varepsilon)}{n}.
\]
A convenient verification certificate for Definition~\ref{def ghost feature isomorphism} is the following regularized relative covariance estimate: for some $\gamma\in(0,1/4)$ and some $\omega_j(W,\varepsilon)$ satisfying $\omega_j(W,\varepsilon)\le c_0\bar\vartheta_j(W,\varepsilon)$ with a sufficiently small absolute constant $c_0$, both samples $T\in\{S,S'\}$ obey, uniformly over all $j,W,\varepsilon$ and all $v\in\mathbb R^{d_j}$,
\begin{align}\label{eq relative covariance certificate main}
\left|v^\top\big(\widehat\Sigma_{j,T}(W)-\Sigma_j(W)\big)v\right|
\le
\gamma\,v^\top\Sigma_j(W)v+\omega_j(W,\varepsilon)\|v\|_2^2.
\end{align}
Indeed, \eqref{eq relative covariance certificate main} implies Definition~\ref{def ghost feature isomorphism}. 
Applying \eqref{eq relative covariance certificate main} with \(T=S'\) gives
\[
v^\top\widehat\Sigma_{j,S'}(W)v
\le
(1+\gamma)v^\top\Sigma_j(W)v
+
\omega_j(W,\varepsilon)\|v\|_2^2 .
\]
On the other hand, applying \eqref{eq relative covariance certificate main} with \(T=S\) gives
\[
v^\top\widehat\Sigma_{j,S}(W)v
\ge
(1-\gamma)v^\top\Sigma_j(W)v
-
\omega_j(W,\varepsilon)\|v\|_2^2.
\]
Combining the two inequalities yields
\begin{align}\label{eq: open display}
v^\top\widehat\Sigma_{j,S'}(W)v
\le
\frac{1+\gamma}{1-\gamma}
v^\top\widehat\Sigma_{j,S}(W)v
+
\left(1+\frac{1+\gamma}{1-\gamma}\right)
\omega_j(W,\varepsilon)\|v\|_2^2 .
\end{align}
For $v$ in the active empirical subspace, $v^\top\widehat\Sigma_{j,S}(W)v\ge\bar\vartheta_j(W,\varepsilon)\|v\|_2^2$. Since \(\omega_j(W,\varepsilon)\le c_0\bar\vartheta_j(W,\varepsilon)\), the preceding inequality \eqref{eq: open display} gives
\[
v^\top\widehat\Sigma_{j,S'}(W)v
\le \kappa v^\top\widehat\Sigma_{j,S}(W)v, \quad \kappa:= \frac{1+\gamma}{1-\gamma}
+
\left(1+\frac{1+\gamma}{1-\gamma}\right)c_0.
\]
For $v$ in the inactive empirical subspace, $v^\top\widehat\Sigma_{j,S}(W)v\le\bar\vartheta_j(W,\varepsilon)\|v\|_2^2$, so the same inequality \eqref{eq: open display} gives
\[
v^\top\widehat\Sigma_{j,S'}(W)v\le b_{\mathrm{sub}}\bar\vartheta_j(W,\varepsilon)\|v\|_2^2, \quad b_{\mathrm{sub}}:=\frac{1+\gamma}{1-\gamma}
+
\left(1+\frac{1+\gamma}{1-\gamma}\right)c_0.
\]
Thus Definition~\ref{def ghost feature isomorphism} holds with with constants that may be taken as
\[
\kappa=b_{\mathrm{sub}}:=\frac{1+\gamma}{1-\gamma}
+
\left(1+\frac{1+\gamma}{1-\gamma}\right)c_0.
\]
Since \(\gamma<1/4\) and \(c_0\) is an absolute constant chosen sufficiently small, \(\kappa=b_{\mathrm{sub}}=O(1)\).
Since
\[
\Gamma_j^T(W)=n\widehat\Sigma_{j,T}(W),
\qquad
\vartheta_j(W,\varepsilon)=n\bar\vartheta_j(W,\varepsilon),
\]
multiplying by \(n\) gives
\eqref{eq ghost feature isomorphism active}--\eqref{eq ghost feature isomorphism inactive}.

Thus, sub-Gaussian or small-ball assumptions at the subspace level
(see, e.g., \cite{mendelson2015learning,mendelson2021extending})
should be used to prove the relative certificate
\eqref{eq relative covariance certificate main}, rather than a fixed additive-ridge theorem.
The relevant empirical process is the scalar quadratic class
\begin{align}
\label{eq: quadratic class}
\mathcal Q_j
=
\Bigl\{
x\mapsto \langle v,f_j(W,x)\rangle^2:
W\in B_{\tF}(R),\ v\in S^{d_j-1}
\Bigr\}.
\end{align}
Consequently, the directional part of the verification scales with the vector-sphere
dimension \(d_j\), rather than with the matrix dimension \(d_j^2\); the remaining cost is
the complexity of the learned feature class as \(W\) varies. This vector-size saving is
crucial for proving the isomorphism event, while the theorem above keeps the final
generalization bound pointwise and compressed, since no below-resolution leakage is
inserted into \(d_{\mathrm R}^S\). Concretely, the open question in Section~\ref{subsec empirical ghost ellipsoid} can be reframed in the following special case:

\begin{center}
\begin{minipage}{0.94\linewidth}
\itshape
    Can \eqref{eq relative covariance certificate main} be proved for the two-layer quadratic
class \eqref{eq: quadratic class} under sub-Gaussian or small-ball feature regularity assumptions at the
subspace level?
\end{minipage}
\end{center}
We hope the discussion above provides a concrete route toward resolving this open question.

\subsection{Proof for Regularized ERM in Section \ref{subsec implicit bias}}\label{appendix implicit bias}
\begin{lemma}[Excess Risk Bound for Regularized ERM]\label{lemma regularized ERM}Assume we have high-probability pointwise generalization bound in  the form of \eqref{eq: generalization gap}, and the loss $\ell(f;z)$ is uniformly bounded by $[0,1]$. 
    Then for the regularized ERM 
    \begin{align*}
    \hat{f} = \operatorname{argmin}_f \left\{\mathbb{P}_n \ell(f;z) + C\sqrt{\frac{d(f) + \log (2/\delta)}{n}}\right\},
    \end{align*}
    we have the excess risk bound against the population risk minimizer $f^\star:=\arg\min_{\cF} \mathbb{P}\ell(f;z)$: with probability at least $1-\delta$,
    \begin{align*}
    \mathbb{P}\ell(\hat{f};z)-\mathbb{P}\ell(f^\star;z)\leq &\inf_{f \in \mathcal{F}}\left\{\mathbb{P}_n\ell(f;z)+C\sqrt{\frac{d (f)+ \log (2/\delta)}{n}}\right\}-\mathbb{P}\ell(f^\star;z)\\
    \leq &(C+\sqrt{1/2})\sqrt{\frac{d (f^\star)+ \log (2/\delta)}{n}}.
\end{align*}
\end{lemma}
    
\paragraph{Proof of Lemma \ref{lemma regularized ERM}:} By \eqref{eq: generalization gap}, for every $\delta\in(0,1)$, take $\delta_1=\delta_2=\delta/2$, we have that with probability at least $1-\delta_1-\delta_2=1-\delta$, we have
\begin{align*}
    \mathbb{P}\ell(\hat{f};z)\leq &\inf_{f\in \mathcal{F}}\left\{ \mathbb{P}_n \ell(f;z)+C\sqrt{\frac{d (f)+ \log (1/\delta_1)}{n}}\right\}\\
    \leq &\mathbb{P}_n\ell(f^\star;z)+C\sqrt{\frac{d(f^\star)+\log(1/\delta_1)}{n}}\\
    \leq & \mathbb{P}\ell(f^\star;z)+\sqrt{\frac{\log({1}/{\delta_2})}{2n}}+C\sqrt{\frac{d(f^\star)+\log(1/\delta_1)}{n}}\\
    = & \mathbb{P}\ell(f^\star;z)+\sqrt{\frac{\log({2}/{\delta})}{2n}}+C\sqrt{\frac{d(f^\star)+\log(2/\delta)}{n}}\\
    \leq &\mathbb{P}\ell(f^\star;z)+(C+\sqrt{1/2})\sqrt{\frac{d(f^\star)+\log(2/\delta)}{n}},
 \end{align*}
 where the first inequality uses the bound of the form \eqref{eq: generalization gap}; the second inequality uses definition of $\hat{f}$; and the third inequality is an application of the Hoeffding's inequality (Lemma \ref{lemma Hoeffding}) at $f^\star$; the equality is by $\delta_1=\delta_2=\delta/2$; and the last inequality follows from the monotonicity of the square root function.
 Thus we have that the excess risk is bounded by
\begin{align*}
    \mathbb{P}\ell(\hat{f};z)-\mathbb{P}\ell(f^\star;z)\leq &\inf_{f \in \mathcal{F}}\left\{\mathbb{P}_n\ell(f;z)+C\sqrt{\frac{d (f)+ \log (2/\delta)}{n}}\right\}-\mathbb{P}\ell(f^\star;z)\\
    \leq &(C+\sqrt{1/2})\sqrt{\frac{d (f^\star)+ \log (2/\delta)}{n}}.
\end{align*}

\hfill$\square$

\subsection{Improvement over Norm Bounds in Section \ref{subsec comparison}}

\subsubsection{Exponential Improvement to a Norm Bound and Comparison}\label{subsec norm bound}
This subsection records a norm-bound relaxation that follows directly from the mixed empirical-population bound in Theorem~\ref{thm dnn}. It does not use the subspace-isomorphism condition in Definition~\ref{def feature isomorphism}, which is needed only for replacing the population feature contribution by observed feature information.

Recall that
\[
\Gamma_l^{S,\mathbb P}(W)=F_l^S(W)F_l^S(W)^\top+n\mathbb E\big[f_l(W,X)f_l(W,X)^\top\big].
\]
Invoking the elementary bound \(\log x\le \log(1+x)\le x\) for \(x>0\), the effective-dimension part in Theorem~\ref{thm dnn} satisfies, for every layer \(l\),
\begin{align}\label{eq:mixed-rank-free-eig}
&\sum_{k=1}^{r_{\eff}^{S,\mathbb P}[W,l]}
\log\!\left(
\frac{8C_2^2\lambda_k(\Gamma_{l-1}^{S,\mathbb P}(W))
R_n(W)^2L\bar M_{l\to L}^2(W,\varepsilon)}{n\varepsilon^2}
\right)\nonumber\\
&\qquad\le
\frac{8C_2^2\operatorname{Tr}(\Gamma_{l-1}^{S,\mathbb P}(W))
R_n(W)^2L\bar M_{l\to L}^2(W,\varepsilon)}{n\varepsilon^2},
\end{align}
where \(R_n(W):=\max\{\|W\|_{\tF},R/2^n\}\). Here
\[
\operatorname{Tr}(\Gamma_{l-1}^{S,\mathbb P}(W))
=\|F_{l-1}^S(W)\|_{\tF}^2+n\mathbb E\|f_{l-1}(W,X)\|_2^2.
\]
This observation gives the following corollary.

\begin{corollary}[Mixed Empirical-Population Rank-Free and Spectral-Norm Relaxation]\label{coro worst case}
Assume the hypotheses of Theorem~\ref{thm dnn}. Define
\(
\overline M_l(W):=\sup_{0<\varepsilon\le \max\{1,c\}/\beta}\bar M_{l\to L}(W,\varepsilon),
\)
where \(c\) is the radius constant in Theorem~\ref{thm dnn}, and let
\[
\alpha_n:=\min\{n^{-5},(2\beta)^{-1}\},\quad
\Lambda_n:=1+\log\frac1{\alpha_n\beta}.
\]
There exists an absolute constant \(C>0\) such that, with probability at least \(1-\delta\) over \(S\), uniformly over \(W\in B_{\tF}(R)\),
\begin{align}\label{eq: norm}
&(\mathbb P-\mathbb P_n)\ell(f(W,x),y)\nonumber\\
&\le C\Bigg(
\frac{\beta\Lambda_n R_n(W)}{n}
\sqrt{\sum_{l=1}^L L(d_l+d_{l-1})
\operatorname{Tr}(\Gamma_{l-1}^{S,\mathbb P}(W))\overline M_l(W)^2}
\nonumber\\
&\hspace{2cm}+\beta\alpha_n
+\sqrt{\frac{\sum_{l=1}^L\log(d_{l-1}n)+\log\frac{2\log(2n)}{\delta}}{n}}
\Bigg).
\end{align}
The bound is rank-free and contains no \(r_{\eff}\) or \(d_{\eff}\).

If, in addition, the activations are $1$-Lipschitz and satisfy \(\sigma_l(0)=0\), then
\begin{align}\label{eq: spectral relaxation}
\operatorname{Tr}(\Gamma_{l-1}^{S,\mathbb P}(W))^{1/2}
\le
\left(\prod_{i<l}\|W_i\|_{\op}\right)
\left(\|X_S\|_{\tF}^2+n\mathbb E\|X\|_2^2\right)^{1/2}.
\end{align}
If the common outer transfer bounds on the relevant weight charts also satisfy
\[
\overline M_l(W)\le C_0\prod_{i>l}\|W_i\|_{\op}
\]
for an absolute constant \(C_0\), then
\begin{align}\label{eq: spectral}
&(\mathbb P-\mathbb P_n)\ell(f(W,x),y)\nonumber\\
&\le\widetilde O\!\Bigg(
\frac{\beta R_n(W)}{n}
\left(\|X_S\|_{\tF}^2+n\mathbb E\|X\|_2^2\right)^{1/2}
\sqrt{\sum_{l=1}^{L}L(d_l+d_{l-1})\prod_{i\ne l}\|W_i\|_{\op}^2}
\nonumber\\
&\hspace{2cm}+\beta\alpha_n
+\sqrt{\frac{\sum_{l=1}^L\log(d_{l-1}n)+\log\frac{2\log(2n)}{\delta}}{n}}
\Bigg),
\end{align}
where \(\widetilde O\) hides absolute constants and \(\Lambda_n\). The outer-chart condition is additional: the spectral product at the center alone need not dominate transfer constants throughout a chart.

If the inputs are almost surely bounded, \(\|X\|_2\le B_x\), then
\begin{align}\label{eq:mixed-input-norm-bound}
\left(\|X_S\|_{\tF}^2+n\mathbb E\|X\|_2^2\right)^{1/2}
\le \left(\|X_S\|_{\tF}^2+nB_x^2\right)^{1/2}
\le\sqrt{2n}\,B_x.
\end{align}
Hence \eqref{eq: spectral} becomes the familiar \(n^{-1/2}\)-order spectral-norm bound
\begin{align}\label{eq: spectral bounded input}
&(\mathbb P-\mathbb P_n)\ell(f(W,x),y)\nonumber\\
&\le\widetilde O\!\Bigg(
\frac{\beta B_x R_n(W)}{\sqrt n}
\sqrt{\sum_{l=1}^{L}L(d_l+d_{l-1})\prod_{i\ne l}\|W_i\|_{\op}^2}
\nonumber\\
&\hspace{2cm}+\beta\alpha_n
+\sqrt{\frac{\sum_{l=1}^L\log(d_{l-1}n)+\log\frac{2\log(2n)}{\delta}}{n}}
\Bigg).
\end{align}
\end{corollary}

\paragraph{Discussion of Corollary \ref{coro worst case}:}
We proceed in three  paragraphs of discussion.  
First, we show that the Riemannian Dimension bound in
Theorem~\ref{thm dnn} can be \emph{exponentially} tighter than the
spectrally normalized bound in~\eqref{eq: spectral}.  
Second, we offer a metric–tensor interpretation that clarifies the
source of this improvement.
Finally, we position \eqref{eq: spectral} relative to the most
representative spectrally normalized  bounds (SNB) in the existing literature. 
 Besides, the additional normalization \(\sigma_l(0)=0\) is used only for the spectral-norm relaxation \eqref{eq: spectral relaxation}.
It is standard in spectrally normalized norm-bound analyses; for example,
\citet{bartlett2017spectrally}.

\paragraph{I: Why the improvement can be exponential.}
The Riemannian-Dimension bound retains logarithmic ellipsoidal volumes, whereas \(\log(1+x)\le x\) replaces them by linear trace bounds. For a concrete comparison, take a scalar one-layer network with identity activation, \(W=R=\beta=1\), deterministic input \(X=e^t\), and \(n=\lceil e^{4t}\rceil\), for \(t\ge1\). The loss may be any bounded $1$-Lipschitz loss, and the outer transfer constant is exactly one. Then \(\Gamma_0^{S,\mathbb P}=2ne^{2t}\), and
\[
d_{\mathrm R}^{S,\mathbb P}(W,c\varepsilon)
=O\!\left(t+\log\frac1\varepsilon\right),
\qquad 0<\varepsilon\le1.
\]
At fixed confidence, choosing \(\alpha=0\) in Theorem~\ref{thm dnn} gives \(O(\sqrt t\,e^{-2t})\), whereas the leading spectral term in \eqref{eq: spectral} is \(\widetilde\Theta(e^{-t})\). Thus the displayed bounds can differ exponentially in \(t\), even when both vanish. This compares the bounds, not the actual generalization gap. Feature compression can give an additional improvement: replacing \(\operatorname{Tr}(\Gamma_{l-1}^{S,\mathbb P})^{1/2}\) by the right-hand side of \eqref{eq: spectral relaxation} discards the learned feature spectrum. Both improvements are already available in Theorem~\ref{thm dnn}, without the subspace-isomorphism condition.

\paragraph{II: Metric tensor interpretation.}
Under the outer-chart condition in Corollary~\ref{coro worst case}, the spectral-norm relaxation replaces the mixed ellipsoidal metric tensor \(G_{\mathrm{NP}}^{S,\mathbb P}(W)\), up to an absolute factor, by
\[
G_{\mathrm{SNB}}^{S,\mathbb P}(W)
=\operatorname{blockdiag}\!\left(
\ldots,
L\big(\|X_S\|_{\tF}^2+n\mathbb E\|X\|_2^2\big)
\prod_{i\ne l}\|W_i\|_{\op}^2\,I_{d_ld_{l-1}},
\ldots
\right).
\]
This tensor is isotropic inside each layer and forgets the spectrum of the learned feature Gram. The Riemannian-Dimension bound keeps the anisotropic tensor
\(L\bar M_{l\to L}^2\Gamma_{l-1}^{S,\mathbb P}(W)\otimes I_{d_l}\), which is why it adapts to low rank and spectral decay in the learned representations.

\paragraph{III: Relation to existing spectrally normalized bounds.}
The bound in~\eqref{eq: spectral} is structurally close to the classical
SNB results of \cite{bartlett2017spectrally} and
\cite{neyshabur2017pac}; the three bounds differ only in the \emph{global
ball} used to constraint the hypothesis class.

\begin{itemize}
\item[\textbf{(a)}]%
Our bound~\eqref{eq: spectral} controls \emph{all} layers simultaneously
via the global Frobenius norm~$\lVert W\rVert_{\tF}$, hence the factor
$R_n(W)$ in the numerator, which differs from $\lVert W\rVert_{\tF}$ only below the discretization scale $R/2^n$.

\item[\textbf{(b)}]%
\cite{neyshabur2017pac} bounds each layer $l$ separately by its Frobenius
norm $\|W_l\|_\tF$.  Strengthening their argument with Dudley’s entropy integral
(one-shot optimization in the original paper) gives
\begin{equation}\label{eq: F norm SNB}
  (\mathbb{P}-\mathbb{P}_{n})\,
  \ell\bigl(f(W,x),y\bigr)
  \;\le\;
  \tilde{O}\!\Bigl(
    \frac{
      \beta\,\lVert X\rVert_{\tF}\;
      \sqrt{
        \sum_{l=1}^{L}
          L^{2}(d_{l}+d_{l-1})\,
          \lVert W_{l}\rVert_{\tF}^{2}\,
          \prod_{i\neq l}\lVert W_{i}\rVert_{\op}^{2}
      }}{n}
    +\sqrt{\frac{\log\frac{1}{\delta}}{n}}
  \Bigr).
\end{equation}
Neither \eqref{eq: spectral} nor \eqref{eq: F norm SNB} strictly
dominates the other, since factors of the form \((\sum_{l}a_{l})(\sum_{l}b_{l})\) in \eqref{eq: spectral} \emph{vs.} factors of the form
\(L\sum_{l}a_{l}b_{l}\) in \eqref{eq: F norm SNB}
can swap their relative order.

\item[\textbf{(c)}]%
\cite{bartlett2017spectrally} replaces each Frobenius norm by the
\(\lVert\,\cdot\,\rVert_{2,1}\) norm, obtaining the tighter
\begin{equation}\label{eq: 2,1 norm SNB}
  (\mathbb{P}-\mathbb{P}_{n})\,
  \ell\bigl(f(W,x),y\bigr)
  \;\le\;
  \tilde{O}\!\Bigl(
    \frac{\beta\,\lVert X\rVert_{\tF}\,
      \bigl (\,
        \sum_{l}\lVert W_{l}\rVert_{2,1}^{2/3}\,
        \sum_{l}\bigl(\prod_{i\neq l}\lVert W_{i}\rVert_{\op}\bigr)^{2/3}
      \bigr )^{3/2}}{n}
    +\sqrt{\frac{\log\frac{1}{\delta}}{n}}
  \Bigr),
\end{equation}
which improves on \eqref{eq: spectral} and \eqref{eq: F norm SNB}
thanks to the sharper \(2,1\) norm.  Extending our Riemannian‐dimension
analysis to the \(2,1\) norm setting is an interesting direction for future
work.

\item[\textbf{(d)}]%
Size‐independent SNB bounds (pioneered by \cite{golowich2020size}) remove
all depth/width dependence at the price of a worse scaling in~\(n\);
incorporating their technique is left for future research.

\item[\textbf{(e)}]  \citet{pinto2025generalization} impose explicit per-layer rank constraints on the weight matrices, thereby
replacing the width factors in \eqref{eq: F norm SNB} with the corresponding ranks while leaving the product of spectral norms unchanged.
Their bound includes an additional \(C^{L}\) factor, which is subsequently removed by \citet{ledent2025generalization}. Moreover,  \citet{ledent2025generalization} seek to bridge the spectral–norm and parameter–count regimes by leveraging the Schatten–$p$ framework of \citet{golowich2020size}, which interpolates between the product-of-spectral-norm regime ($p\to\infty$) and layerwise low-rank scalings ($p\to 0$). In the extreme $p\to 0$ limit, a representative consequence (Theorem~E.8 of \citealp{ledent2025generalization}) yields
\begin{align*}
(\mathbb{P}-\mathbb{P}_n)\ell\bigl(f(W,x),y\bigr)
\le
\widetilde{O}\left(
\frac{\sup_{i}\|x_i\|_2^2}{\sqrt{n}}
\sqrt{\sum_{l=1}^{L} L(d_l+d_{l-1})\operatorname{rank}(W_l)}
\right).
\end{align*}
Notably, the explicit dependence on the \emph{ranks of the weight matrices}—rather than on spectrum-aware or \emph{feature}-rank quantities—renders this result structurally similar to VC-dimension bounds (indeed, the proof proceeds via uniform covering numbers, and packing/VC dimensions for matrices are known to adapt to explicit rank constraints \citep{srebro2004generalization}). As the authors acknowledge, this is a principal limitation: empirical evidence suggests that deep networks exhibit low rank in their \emph{features} rather than their weights, a phenomenon this bound does not capture. 
\end{itemize}

In any case, \eqref{eq: spectral} is a representative SNB bound, and the key message in this subsection is that our Riemannian-Dimension result in Theorem~\ref{thm dnn} can be \emph{exponentially} sharper than~\eqref{eq: spectral}.

\subsubsection{Proof of Corollary \ref{coro worst case}}\label{subsec proof coro}
We work directly with the empirical-population metric throughout. Theorem~\ref{thm dnn} gives, with probability at least \(1-\delta\), uniformly over \(W\in B_{\tF}(R)\),
\begin{align}\label{eq: upper chaining bound}
&(\bP-\bPn)\ell(f(W,x),y)\nonumber\\
&\le C_1\left(\beta\inf_{0\le\alpha\le1/\beta}
\left\{\alpha+\frac1{\sqrt n}\int_\alpha^{1/\beta}
\sqrt{d_{\mathrm R}^{S,\mathbb P}(W,c\varepsilon)}\,d\varepsilon\right\}
+\sqrt{\frac{\log\frac{2\log(2n)}{\delta}}{n}}\right).
\end{align}
The supremum defining \(\overline M_l\) covers all scales \(c\varepsilon\) used here; the fixed factor \(c\) is absorbed into the absolute constants below.

\paragraph{Step 1: The Trace Relaxation.}
Summing \eqref{eq:mixed-rank-free-eig} over layers gives
\begin{align}\label{eq: d R upper coro}
d_{\mathrm R}^{S,\mathbb P}(W,c\varepsilon)
&\le\frac{C R_n(W)^2}{n\varepsilon^2}
\sum_{l=1}^L L(d_l+d_{l-1})
\operatorname{Tr}(\Gamma_{l-1}^{S,\mathbb P}(W))\overline M_l(W)^2
+\sum_{l=1}^L\log(d_{l-1}n).
\end{align}
Here we used the identity
\begin{align}\label{eq: property Frobenius}
\sum_k\lambda_k(\Gamma_{l-1}^{S,\mathbb P}(W))
=\operatorname{Tr}(\Gamma_{l-1}^{S,\mathbb P}(W))
=\|F_{l-1}^S(W)\|_{\tF}^2+n\mathbb E\|f_{l-1}(W,X)\|_2^2.
\end{align}
Thus, for \(0<\alpha\le1/\beta\),
\begin{align}\label{eq mixed dominant integral bound}
&\frac1{\sqrt n}\int_\alpha^{1/\beta}
\sqrt{d_{\mathrm R}^{S,\mathbb P}(W,c\varepsilon)}\,d\varepsilon\nonumber\\
&\le\frac{C R_n(W)}{n}
\sqrt{\sum_{l=1}^L L(d_l+d_{l-1})
\operatorname{Tr}(\Gamma_{l-1}^{S,\mathbb P}(W))\overline M_l(W)^2}
\log\frac1{\alpha\beta}
+\frac1\beta\sqrt{\frac{\sum_{l=1}^L\log(d_{l-1}n)}{n}}.
\end{align}
Choose \(\alpha=\alpha_n\) in \eqref{eq: upper chaining bound} and use \(\log(1/(\alpha_n\beta))\le\Lambda_n\). This proves \eqref{eq: norm}. The truncation contributes only
\begin{align}\label{eq: hide 1}
\beta\alpha_n=\min\{\beta n^{-5},1/2\}.
\end{align}
In particular, there is no need to integrate the trace bound down to zero.

\paragraph{Step 2: The Spectral-Norm Relaxation.}
For every input \(x\), the activation assumptions give
\[
\|f_{l-1}(W,x)\|_2
\le\left(\prod_{i<l}\|W_i\|_{\op}\right)\|x\|_2.
\]
Squaring and integrating against \(n(\mathbb P_S+\mathbb P)\) proves \eqref{eq: spectral relaxation}. Under the stated common outer-chart bound, substituting this inequality and \(\overline M_l(W)\le C_0\prod_{i>l}\|W_i\|_{\op}\) into \eqref{eq: norm} proves \eqref{eq: spectral}. The bound is simultaneous in \(W\) on the same event. Replacing \(R_n(W)\) by \(\|W\|_{\tF}\), as in the main-text display, incurs at most the additional term
\begin{align}\label{eq: hide 2}
\frac{C\beta\Lambda_n R}{n2^n}
\left(\|X_S\|_{\tF}^2+n\mathbb E\|X\|_2^2\right)^{1/2}
\sqrt{\sum_{l=1}^{L}L(d_l+d_{l-1})\prod_{i\ne l}\|W_i\|_{\op}^2}.
\end{align}
Finally, \eqref{eq:mixed-input-norm-bound} yields \eqref{eq: spectral bounded input}. No subspace-isomorphism assumption or ghost-sample averaging is used. \hfill$\square$

\newpage

\bibliography{references.bib}

\end{document}

%% file: math_commands.tex
\usepackage{amsmath,amsfonts,bm}

\def\1{\bm{1}}

\DeclareMathAlphabet{\mathsfit}{\encodingdefault}{\sfdefault}{m}{sl}
\SetMathAlphabet{\mathsfit}{bold}{\encodingdefault}{\sfdefault}{bx}{n}
\newcommand{\tens}[1]{\bm{\mathsfit{#1}}}

\def\tF{{\tens{F}}}

\newcommand{\E}{\mathbb{E}}



%% file: style.tex
\usepackage[linktocpage=true, colorlinks=true, linkcolor=red, citecolor=blue,urlcolor=black]{hyperref}
\usepackage{url}
\usepackage{fullpage}
\usepackage[authoryear]{natbib}   
\usepackage{amsmath}
\usepackage{mathtools}
\usepackage{dsfont}
\usepackage{booktabs}       

\numberwithin{equation}{section}

\newcommand{\diag}{\operatorname{\textup{diag}}}

\newtheorem{lemma}{Lemma}
\newtheorem{theorem}{Theorem}

\newtheorem{corollary}{Corollary}
\newtheorem{definition}{Definition}

\newtheorem{example}{Example}

\newcommand{\lsim}{\raisebox{-0.13cm}{~\shortstack{$<$ \\[-0.07cm]
      $\sim$}}~}

\newcommand{\cH}{\mathcal{H}}
\newcommand{\cZ}{\mathcal{Z}}
\newcommand{\bE}{\mathbb{E}}
\newcommand{\bP}{\mathbb{P}}
\newcommand{\bPn}{\mathbb{P}_n}
\newcommand{\bR}{\mathbb{R}}

\newcommand{\cF}{\mathcal{F}}

\newcommand{\cW}{\mathcal{W}}
\newcommand{\cV}{\mathcal{V}}
\newcommand{\cX}{\mathcal{X}}

\newcommand{\op}{\textup{op}}

\newcommand{\eff}{\textup{eff}}

\newcommand{\St}{\textup{St}}
\newcommand{\Gr}{\textup{Gr}}
\newcommand{\proj}{\textup{proj}}
\newcommand{\mat}[1]{\begin{pmatrix}#1\end{pmatrix}}
\newcommand{\rN}{\mathrm{N}}
\newcommand{\cP}{\mathcal{P}}
